\documentclass[sigconf]{acmart}
\acmJournal{TOG}

\AtBeginDocument{%
  \providecommand\BibTeX{{%
    \normalfont B\kern-0.5em{\scshape i\kern-0.25em b}\kern-0.8em\TeX}}}

\copyrightyear{2022}
\acmYear{2022}
\setcopyright{iw3c2w3}
\acmConference[SA '22 Conference Papers]{SIGGRAPH Asia 2022 Conference Papers}{December 6--9, 2022}{Daegu, Republic of Korea}
\acmBooktitle{SIGGRAPH Asia 2022 Conference Papers (SA '22 Conference Papers), December 6--9, 2022, Daegu, Republic of Korea}\acmDOI{10.1145/3550469.3555417}
\acmISBN{978-1-4503-9470-3/22/12}

\acmSubmissionID{488}

\usepackage{soul}
\usepackage[ruled, vlined, linesnumbered]{algorithm2e}
\usepackage{booktabs}
\usepackage{subcaption}
\usepackage{overpic}
\usepackage{mathtools}
\usepackage{transparent}
\usepackage{multirow}
\usepackage{tcolorbox}

\makeatletter
\let\@authorsaddresses\@empty
\makeatother

\begin{document}

\title{VIINTER: View Interpolation with Implicit Neural Representations of Images}

\author{Brandon Yushan Feng}
\email{yfeng97@umd.edu}
\affiliation{%
  \institution{University of Maryland, College Park}
  \country{}
}

\author{Susmija Jabbireddy}
\email{jsreddy@umd.edu}
\affiliation{%
  \institution{University of Maryland, College Park}
  \country{}
}

\author{Amitabh Varshney}
\email{varshney@umd.edu}
\affiliation{%
  \institution{University of Maryland, College Park}
  \country{}
}

\renewcommand{\shortauthors}{Feng et al.}

\begin{abstract}
  We present VIINTER, a method for view interpolation by interpolating the implicit neural representation (INR) of the captured images.
  We leverage the learned code vector associated with each image and interpolate between these codes to achieve viewpoint transitions.
  We propose several techniques that significantly enhance the interpolation quality.
  VIINTER signifies a new way to achieve view interpolation without constructing 3D structure, estimating camera poses, or computing pixel correspondence.
  We validate the effectiveness of VIINTER on several multi-view scenes with different types of camera layout and scene composition.
  As the development of INR of images (as opposed to surface or volume) has centered around tasks like image fitting and super-resolution, with VIINTER, we show its capability for view interpolation and offer a promising outlook on using INR for image manipulation tasks.
\end{abstract}

\ccsdesc[500]{Computing methodologies~Image manipulation}
\ccsdesc[500]{Computing methodologies~Image processing}
\ccsdesc[300]{Computing methodologies~Image-based rendering}
\ccsdesc[500]{Computing methodologies~Neural networks}

\keywords{implicit neural representation, coordinate network, view synthesis}

\begin{teaserfigure}
  \includegraphics[width=\linewidth, trim={40 0 0 0}, clip]{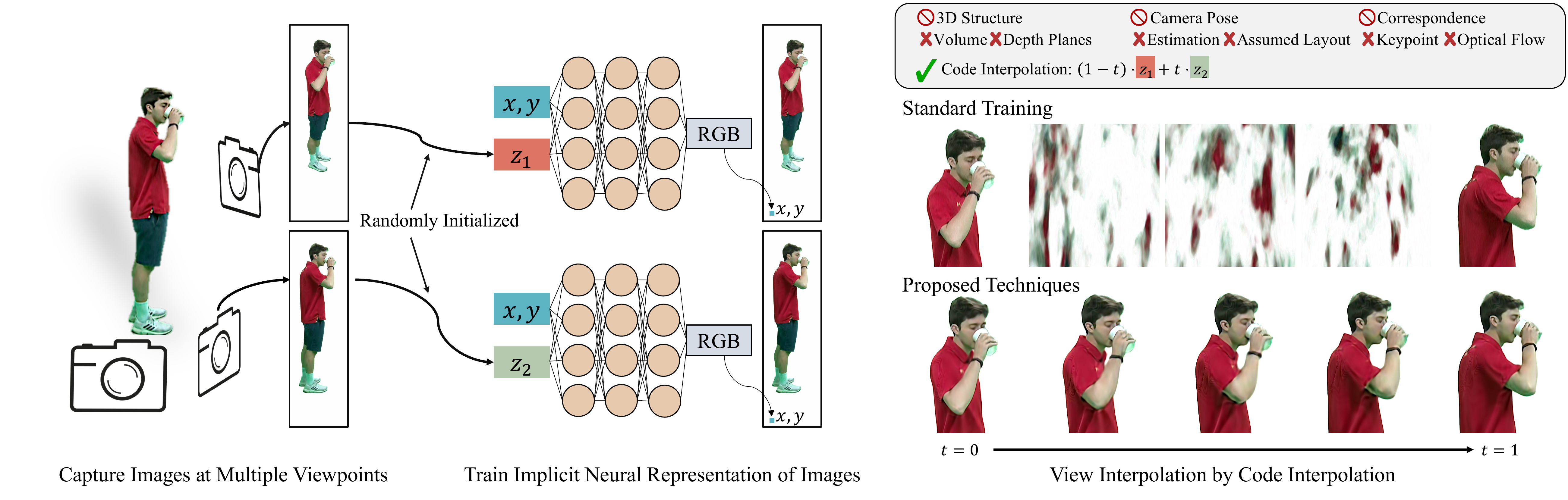}
  \vspace{-25pt}
  \caption{We propose a new method for view interpolation through implicit neural representations (INR) of images. After each image is randomly assigned a code vector $z$, the codes are then jointly trained with the neural network to produce the RGB color given coordinate $(x, y)$. With standard training, the INR fails to decode coherent images from new codes interpolated by two trained codes, but our method enables smooth transition between two known viewpoints. Contrary to common methods for view interpolation, our method does not use 3D structure, camera poses, or pixel correspondence during training.}
  \label{fig:teaser}
\end{teaserfigure}

\maketitle

\newcommand{\etal}{{\em et al.} }
\newcommand{\eg}{{\em e.g.} }
\newcommand{\degree}{\ensuremath{^{\circ}} }
\newcommand{\ceil}[1]{{\lceil #1 \rceil}}
\newcommand{\floor}[1]{{\lfloor #1 \rfloor}}

\newcommand{\Log}[1]{\log\left(#1\right)}
\newcommand{\NormTwo}[1]{\left\lVert#1\right\rVert_{2}}
\newcommand{\Norm}[1]{\left\lVert#1\right\rVert}
\newcommand{\mathbfit}[1]{\textbf{\textit{#1}}}
\newcommand{\Origin}{\mathbfit{O}}

\newcommand{\FigTwoWidth}{21mm}
\newcommand{\FigTwoSubfig}[2]{	\begin{subfigure}{29mm}
	\centering
	\includegraphics[width=29mm, trim={320 310 280 320}, clip]{#1}\vspace{-5pt}
    $\underbracket[0pt][1mm]{\hspace{10pt}}_%
    {\hspace{-63pt}\substack{\vspace{-18pt}\\ \colorbox{white}{\tiny #2}}}$
	\end{subfigure}
}

\newcommand{\FigThreeSubfigWidth}{28mm}
\newcommand{\FigThreeSubfig}[2]{	\begin{subfigure}{\FigThreeSubfigWidth}
	\centering
	\includegraphics[width=\FigThreeSubfigWidth, height=24.5mm, trim={210 1170 30 240}, clip]{#1}
    $\underbracket[0pt][1mm]{\hspace{10pt}}_%
    {\hspace{-63pt}\substack{\vspace{-28pt}\\ \colorbox{white}{\tiny #2}}}$
	\end{subfigure}
}

\newcommand{\FigFourSubfigWidth}{24.5mm}
\newcommand{\FigFourSubfig}[2]{	\begin{subfigure}{\FigFourSubfigWidth}
	\centering
	\includegraphics[width=\FigFourSubfigWidth, trim={50 575 50 200}, clip]{#1}
    $\underbracket[0pt][1mm]{\hspace{10pt}}_%
    {\hspace{-63pt}\substack{\vspace{-28pt}\\ \colorbox{white}{\tiny #2}}}$
	\end{subfigure}
}

\newcommand{\FigFiveSubfig}[1]{	\begin{subfigure}{29mm}
	\centering
	\includegraphics[width=29mm, height=29mm]{#1}
	\end{subfigure}
}

\newcommand{\FigFiveSubfigWidth}{35mm}
\newcommand{\FigFiveSubfigHeight}{28mm}
\newcommand{\FigFiveSubfigCaption}[1]{	    \begin{subfigure}{\FigFiveSubfigWidth}
	\centering
    $\underbracket[0pt][1mm]{\hspace{3cm}}_%
    {\substack{\vspace{-6pt}\\ \colorbox{white}{#1}}}$
    \vspace{-5pt}
	\end{subfigure}
}
\newcommand{\FigFiveSubfigA}[1]{	\begin{subfigure}{\FigFiveSubfigWidth}
	\centering
	\includegraphics[width=\FigFiveSubfigWidth, trim={320 430 200 310}, clip]{#1}
	\end{subfigure}
}
\newcommand{\FigFiveSubfigB}[1]{	\begin{subfigure}{\FigFiveSubfigWidth}
	\centering
	\includegraphics[width=\FigFiveSubfigWidth, trim={930 725 120 290}, clip]{#1}
	\end{subfigure}
}
\newcommand{\FigFiveSubfigC}[1]{	\begin{subfigure}{\FigFiveSubfigWidth}
	\centering
	\includegraphics[width=\FigFiveSubfigWidth, trim={350 280 180 480}, clip]{#1}
	\end{subfigure}
}

\newcommand{\FigSixSubfigWidth}{35mm}
\newcommand{\FigSixSubfigHeight}{28mm}
\newcommand{\FigSixSubfigCaption}[1]{	    \begin{subfigure}{\FigSixSubfigWidth}
	\centering
    $\underbracket[0pt][1mm]{\hspace{3cm}}_%
    {\substack{\vspace{-6pt}\\ \colorbox{white}{#1}}}$
    \vspace{-5pt}
	\end{subfigure}
}
\newcommand{\FigSixSubfigA}[1]{	\begin{subfigure}{\FigSixSubfigWidth}
	\centering
	\includegraphics[width=\FigSixSubfigWidth, trim={170 100 200 150}, clip]{#1}
	\end{subfigure}
}
\newcommand{\FigSixSubfigB}[1]{	\begin{subfigure}{\FigSixSubfigWidth}
	\centering
	\includegraphics[width=\FigSixSubfigWidth, trim={0 0 0 0}, clip]{#1}
	\end{subfigure}
}

\newcommand{\FigSevenSubfigWidth}{35mm}
\newcommand{\FigSevenSubfigHeight}{28mm}
\newcommand{\FigSevenSubfigCaption}[1]{	    \begin{subfigure}{\FigSevenSubfigWidth}
	\centering
    $\underbracket[0pt][1mm]{\hspace{3cm}}_%
    {\substack{\vspace{-6pt}\\ \colorbox{white}{#1}}}$
    \vspace{-5pt}
	\end{subfigure}
}
\newcommand{\FigSevenSubfigA}[1]{	\begin{subfigure}{\FigSevenSubfigWidth}
	\centering
	\includegraphics[width=\FigSevenSubfigWidth, trim={0 400 0 230}, clip]{#1}
	\end{subfigure}
}
\newcommand{\FigSevenSubfigB}[1]{	\begin{subfigure}{\FigSevenSubfigWidth}
	\centering
	\includegraphics[width=\FigSevenSubfigWidth, trim={0 100 0 990}, clip]{#1}
	\end{subfigure}
}

\newcommand{\FigLimitationSubfigWidth}{27mm}
\newcommand{\FigLimitationSubfig}[2]{	\begin{subfigure}{\FigLimitationSubfigWidth}
	\centering
	\includegraphics[width=\FigLimitationSubfigWidth, trim={0 0 0 0}, clip]{#1}
    $\underbracket[0pt][1mm]{\hspace{10pt}}_%
    {\hspace{-63pt}\substack{\vspace{-28pt}\\ \colorbox{white}{\tiny #2}}}$
	\end{subfigure}
}

\newcommand*{\colorboxed}{}
\def\colorboxed#1#{%
  \colorboxedAux{#1}%
}
\newcommand*{\colorboxedAux}[3]{%
  \begingroup
    \colorlet{cb@saved}{.}%
    \color#1{#2}%
    \boxed{%
      \color{cb@saved}%
      #3%
    }%
  \endgroup
}

\definecolor{mycolor}{rgb}{1.0, 0.1, 0.15}

\newtcbox{\mybox}{on line, standard jigsaw,
  colframe=mycolor,colback=mycolor!10!white, opacityback=0.0,
  boxrule=1.5pt,arc=0pt,boxsep=0pt,left=6pt,right=6pt,top=6pt,bottom=6pt}

\newcommand{\SuppLFSubfig}[2]{	\begin{subfigure}{\FigFiveSubfigWidth}
	\centering
	\includegraphics[width=\FigFiveSubfigWidth, trim=#1, clip]{#2}
	\end{subfigure}
}
\newcommand{\SuppSixSubfigWidth}{28mm}
\newcommand{\SuppSixSubfig}[2]{	\begin{subfigure}{\SuppSixSubfigWidth}
	\centering
	\includegraphics[width=\SuppSixSubfigWidth, trim=#1, clip]{#2}
	\end{subfigure}
}

\newcommand{\SuppHoloSubfig}[2]{	\begin{subfigure}{25mm}
	\centering
	\includegraphics[width=25mm, trim=#1, clip]{#2}
	\end{subfigure}
}

\newcommand{\SuppThreeSubfigWidth}{35mm}
\newcommand{\SuppThreeSubfig}[2]{	\begin{subfigure}{\SuppThreeSubfigWidth}
	\centering
	\includegraphics[width=\SuppThreeSubfigWidth, trim=#1, clip]{#2}
	\end{subfigure}
}

\newcommand{\SuppResolutionCompareSubfigWidth}{28mm}
\newcommand{\SuppResolutionCompareSubfig}[1]{	\begin{subfigure}{\SuppResolutionCompareSubfigWidth}
	\centering
	\includegraphics[width=\SuppResolutionCompareSubfigWidth, trim={120 1200 50 200}, clip]{#1}
	\end{subfigure}
}

\section{Introduction}
Neural networks have become a prevalent component in various computational systems over the past decade. 
For the graphics and vision community, they have been an effective tool in tasks involving visual data, such as recognition, segmentation, and 3D reconstruction.
In these classic tasks, neural networks are often deployed as a feature extractor from the input visual signal (\eg image), but more recently, coordinate network has emerged as a new concept.
Instead of extracting features from the signal, the network takes in a coordinate and produces the signal value at that coordinate.
Such a network learns a continuous function that maps signal coordinates to values, and it is often referred to as an implicit neural representation (INR) of the signal.
INR has led to remarkable success in representing visual signals such as images, videos, signed distance fields, and radiance fields.

In scenarios where only 2D images are available, INR has found two prominent applications.
One of them is image fitting, where INRs are trained to produce the color of each known image pixel.
Along this line, much progress has been made to improve the accuracy and speed of fitting INR on images, as well as its ability for compression and super-resolution.
The other prominent application is reconstructing 3D scenes from 2D images. Here, INRs produce the attribute values (\eg radiance and opacity) at each spatial coordinate, which are then differentiably rendered into pixels.
In this case, the INRs are optimized such that these rendered pixels reproduce the known image pixels.
Once sufficiently trained, these INRs can synthesize plausible novel views outside the training set.

On both fitting and view synthesis, INRs achieve impressive visual results that closely resembled the original 2D images.
However, it also appears that the development of INRs has gone into two orthogonal directions.
On one hand, the quality of fitting images with INRs is improved by incorporating traditional signal processing techniques like multi-scale subsampling and filtering.
On the other hand, the quality of view synthesis is improved by augmenting INRs with well-established 3D graphics techniques, such as spatial subdivision, parametric modeling, and level-set methods.

Although the exciting advancements towards these two directions are rapidly pushing the state of the art, we like to explore a different direction and ask a new question: Given multiple 2D image views of a 3D scene, can we use the INR of those 2D images alone to do view synthesis without any 3D reconstruction, pose, or correspondence?
In this paper, driven by this question, we present an initial exploration towards view interpolation with INR of images (VIINTER).
With randomly initialized INR weights and code vectors for individual images, we modify the standard INR training process such that the trained INR can both faithfully reproduce the given images and synthesize plausible novel views when we interpolate between those learned image codes.

It is nontrivial to obtain sensible novel views through code interpolation with standard training of INR.
We experiment on a range of changes to the training of INR and provide details in Section~\ref{Section:Method}.
We present further evaluation results on different types of multi-view scenes in Section~\ref{Section:Experiments}.
Our work takes an important early step toward revealing new potential of INR of images, and we summarize our main contributions as the followings:
\begin{itemize}

\item We present a novel approach to view interpolation by interpolating INRs trained to fit 2D images without any knowledge of 3D structure, pose, or correspondence.

\item We introduce several modifications to the common process of training image-fitting INRs, which significantly improve the view interpolation quality.

\item We show that the proposed non-3D approach achieves smooth and photorealistic interpolation across several scenes with a variety of viewpoint layout and scene content.

\end{itemize}
\section{Related Work}
 \label{Section:Related}

In this section, we review recent work on implicit neural representation, as well as prior techniques for view interpolation.

\subsection{Implicit Neural Representations.}
Following seminal works~\cite{chen2019learning, park2019deepsdf, mescheder2019occupancy} showing successful applications of neural network to encode 3D shapes, many methods have been introduced to solve various vision and graphics tasks using INRs of 3D shapes.
These INRs usually use the multilayer perceptron (MLP) architecture to encode geometric information of a 3D shape by learning the mapping from a given 3D spatial point and a scalar value denoting either the signed distance or occupancy.
\subsubsection{3D Reconstruction.}
As differentiable rendering becomes more practical, researchers have succeeded in training INRs to learn, not just fit, the geometry and appearance of a 3D scene based on 2D image observations.
The most prominent works is Neural Radiance Fields (NeRF)~\cite{Mildenhall2020NeRF}, which learns an INR of the view-dependent radiance volume inside a 3D scene and naturally enables view synthesis.
The success of NeRF sparked an enthusiastic trend of improving INRs for highly photorealistic view synthesis in terms of their training speed, rendering speed, and rendering quality.
A wide range of techniques have been studied and incorporated to 3D INRs, including spatial subdivision or octree~\cite{liu2020neural, yu2021plenoctrees}, parametric modeling with human body shape prior~\cite{liu2021neural, peng2021animatable}, level-set methods for more accuracy geometry~\cite{wang2021neus, bergman2021fast}, caching and distillation for faster rendering`\cite{yu2021plenoctrees, hedman2021baking}, camera pose refinement~\cite{lin2021barf, meng2021gnerf, wang2021nerf}, and lighting and camera variation during capture to better extract physical attributes~\cite{bi2020deep, zhang2021physg}.
{Convolutional neural networks~\cite{eslami2018neural,tatarchenko2016multi, dosovitskiy2016learning, Bemana2020xfields} have also been trained to take camera pose as input and produce 2D renderings of simple 3D scenes}.

\begin{figure*}[!ht]
    \FigTwoSubfig{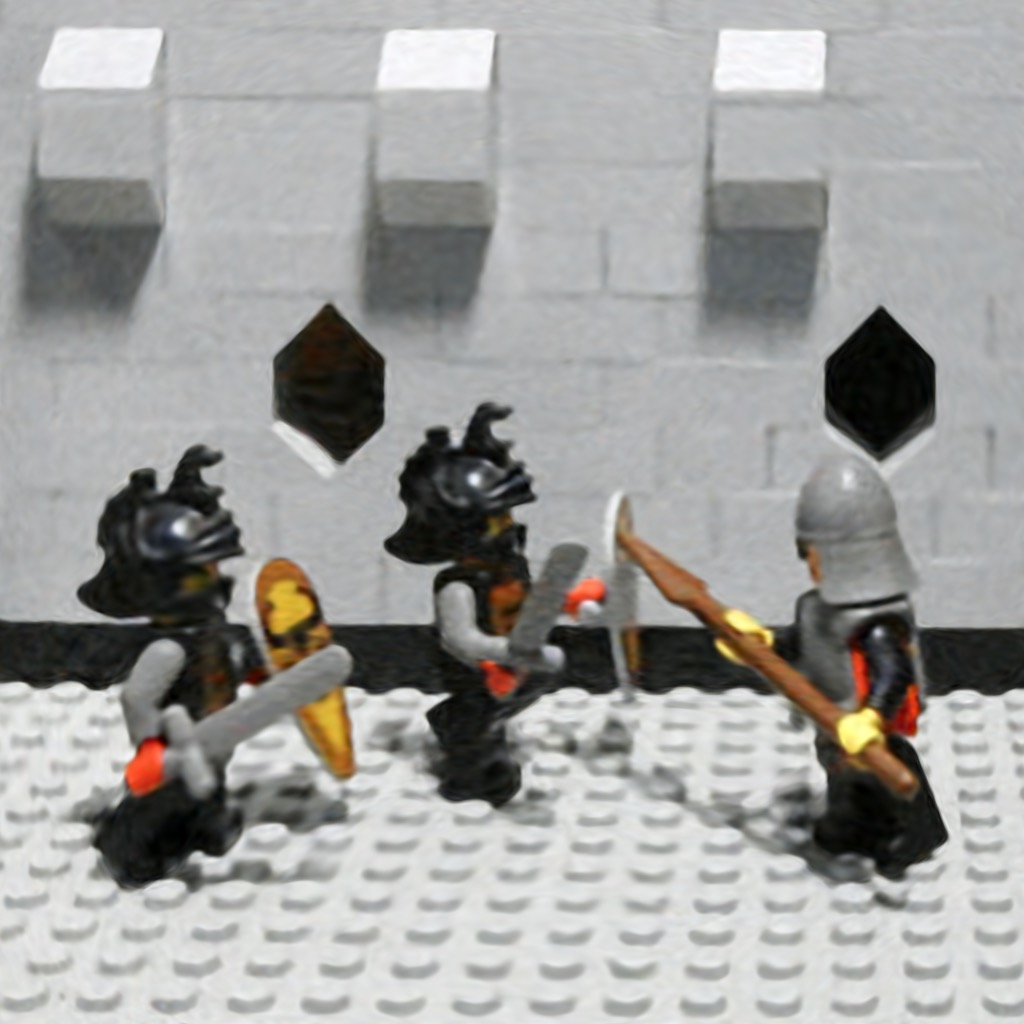}{$t=0$}%
    ~
    \FigTwoSubfig{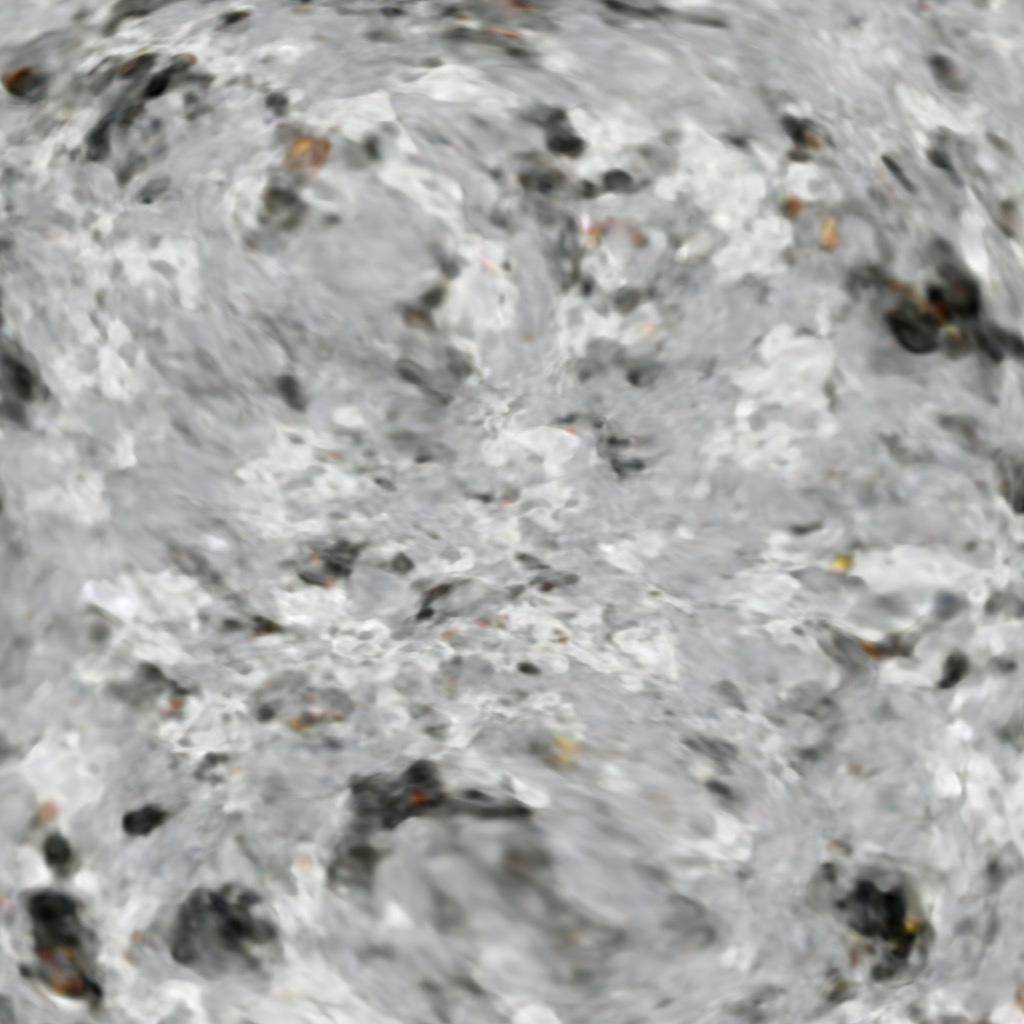}{$t=0.5$}%
    ~
    \FigTwoSubfig{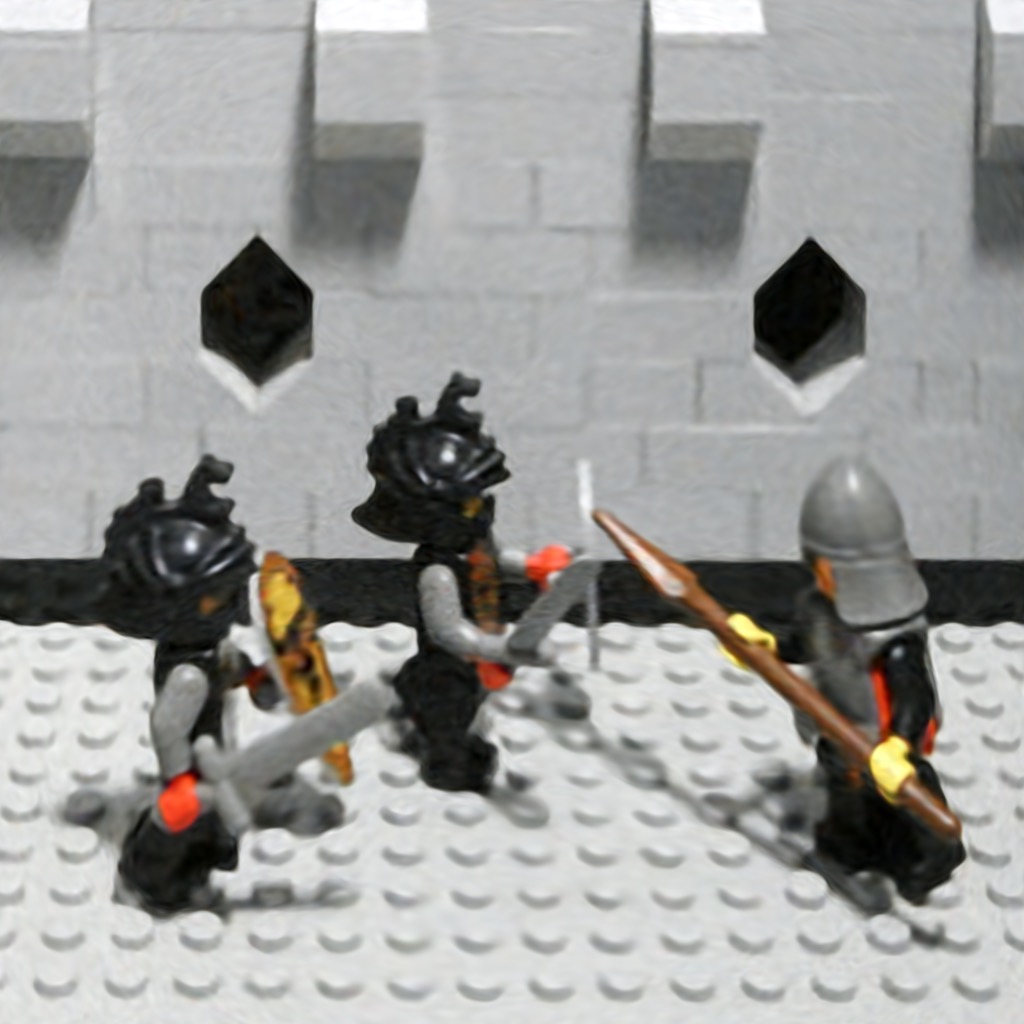}{$t=1$}%
    ~
    \FigTwoSubfig{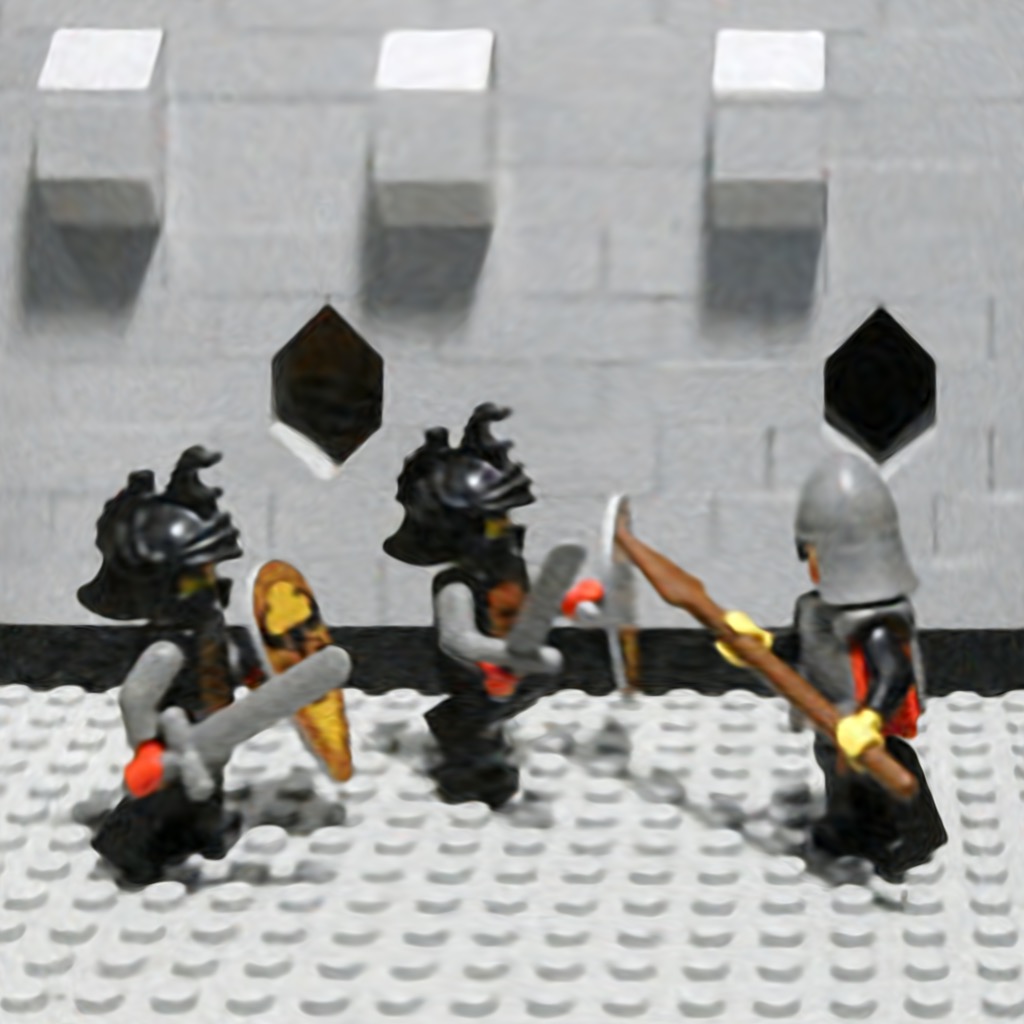}{$t=0$}%
    ~
    \FigTwoSubfig{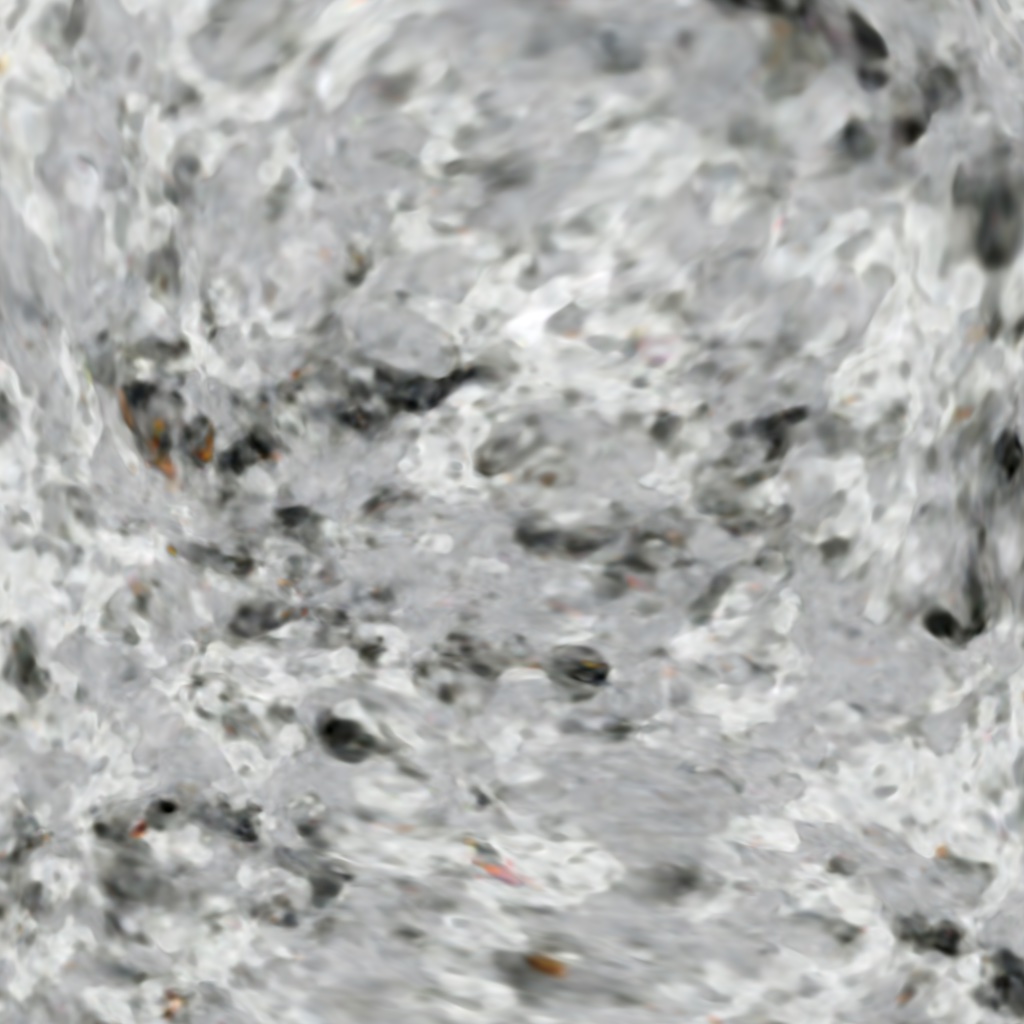}{$t=0.5$}%
    ~
    \FigTwoSubfig{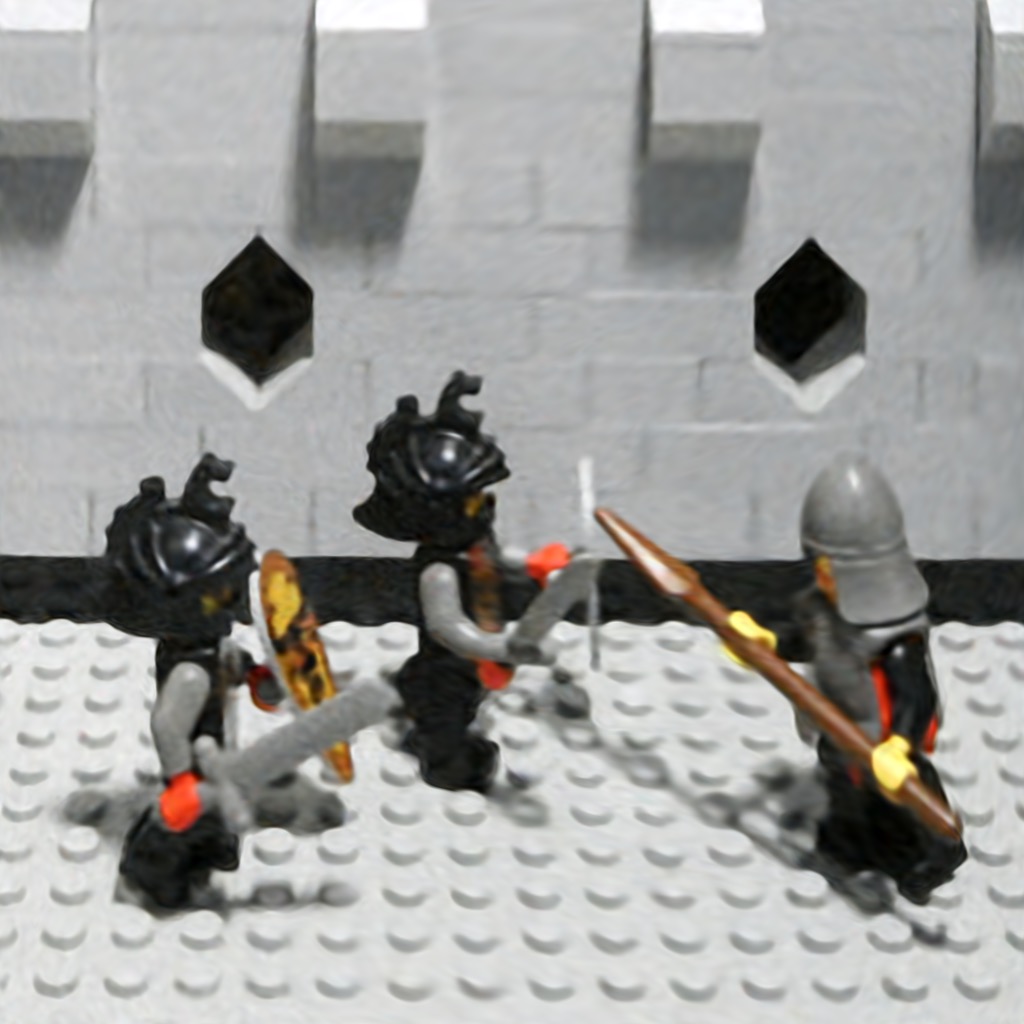}{$t=1$}%
    
    \vspace{-8pt}\hspace{0.1cm}$\underbracket[0.5pt][3pt]{\hspace{8.1cm}}_%
    {\substack{\vspace{-10pt}\\ \quad \colorbox{white}{No Control}}}$    \vspace{0pt}
    \hspace{0.8cm}$\underbracket[0.5pt][3pt]{\hspace{8.1cm}}_%
    {\substack{\vspace{-10pt}\\ \colorbox{white}{$\infty$-norm}}}$
    \vspace{0pt}

    \FigTwoSubfig{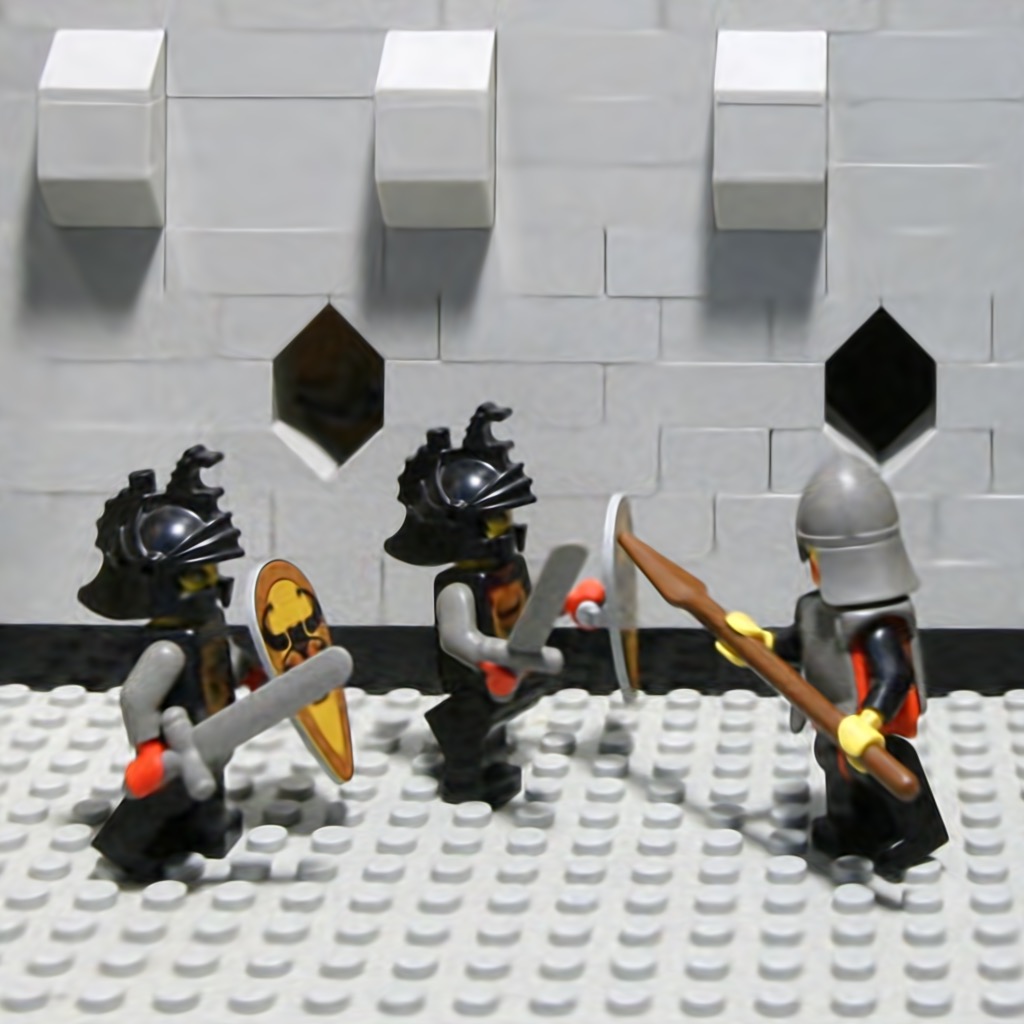}{$t=0$}%
    ~
    \FigTwoSubfig{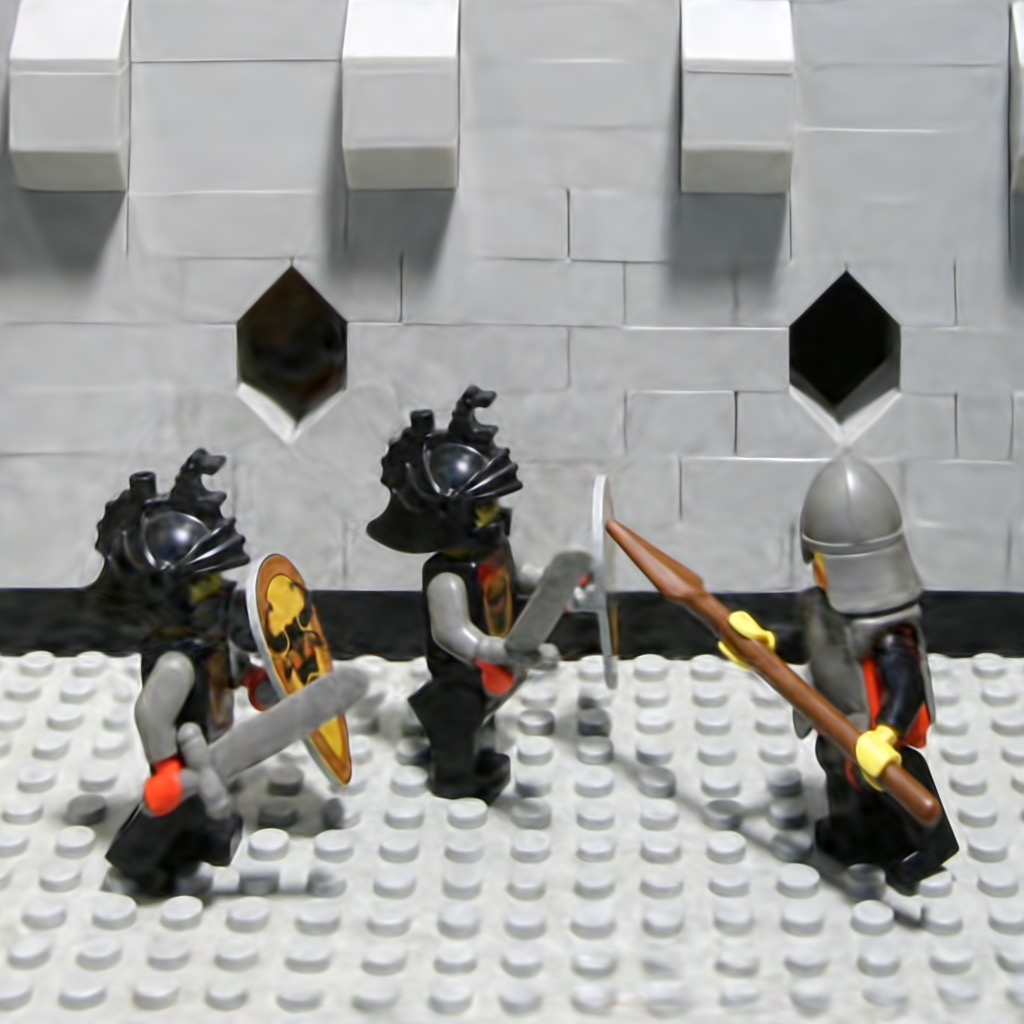}{$t=0.5$}%
    ~
    \FigTwoSubfig{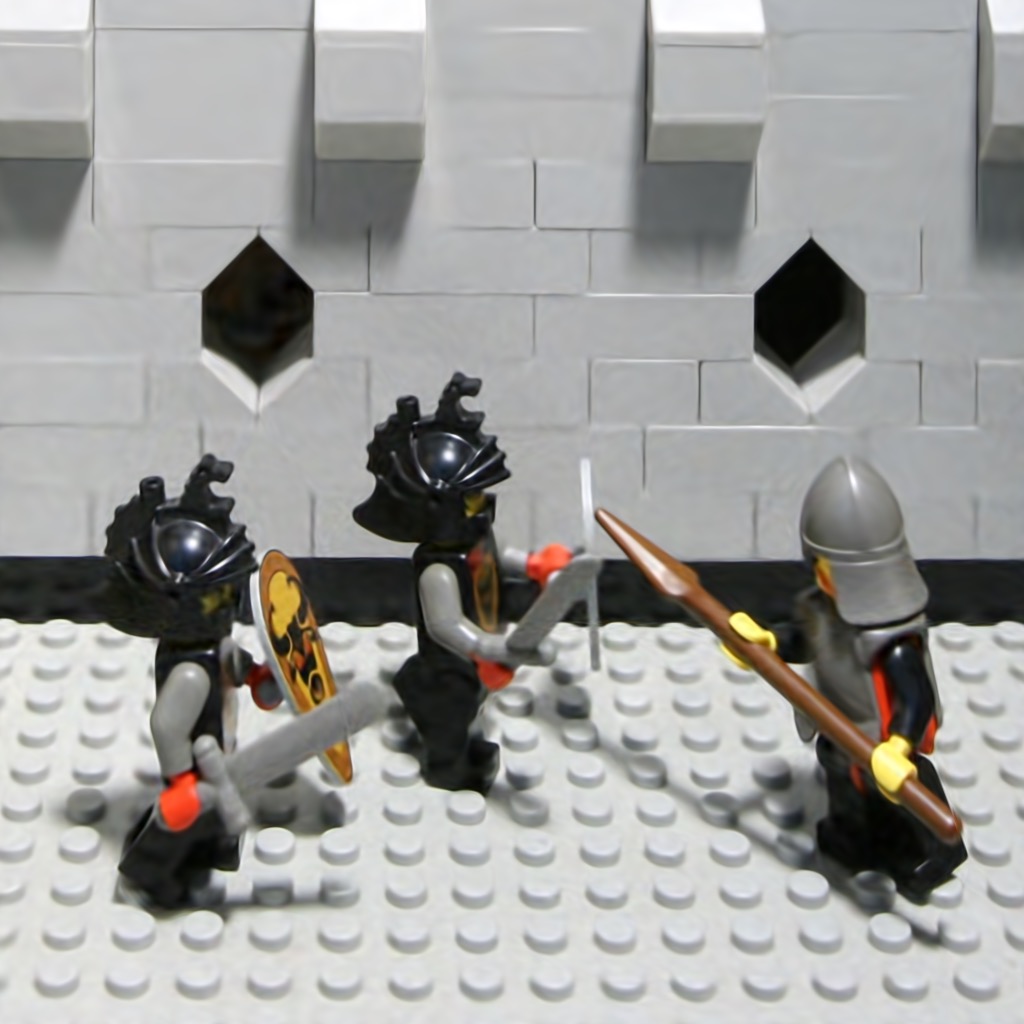}{$t=1$}%
    ~
    \FigTwoSubfig{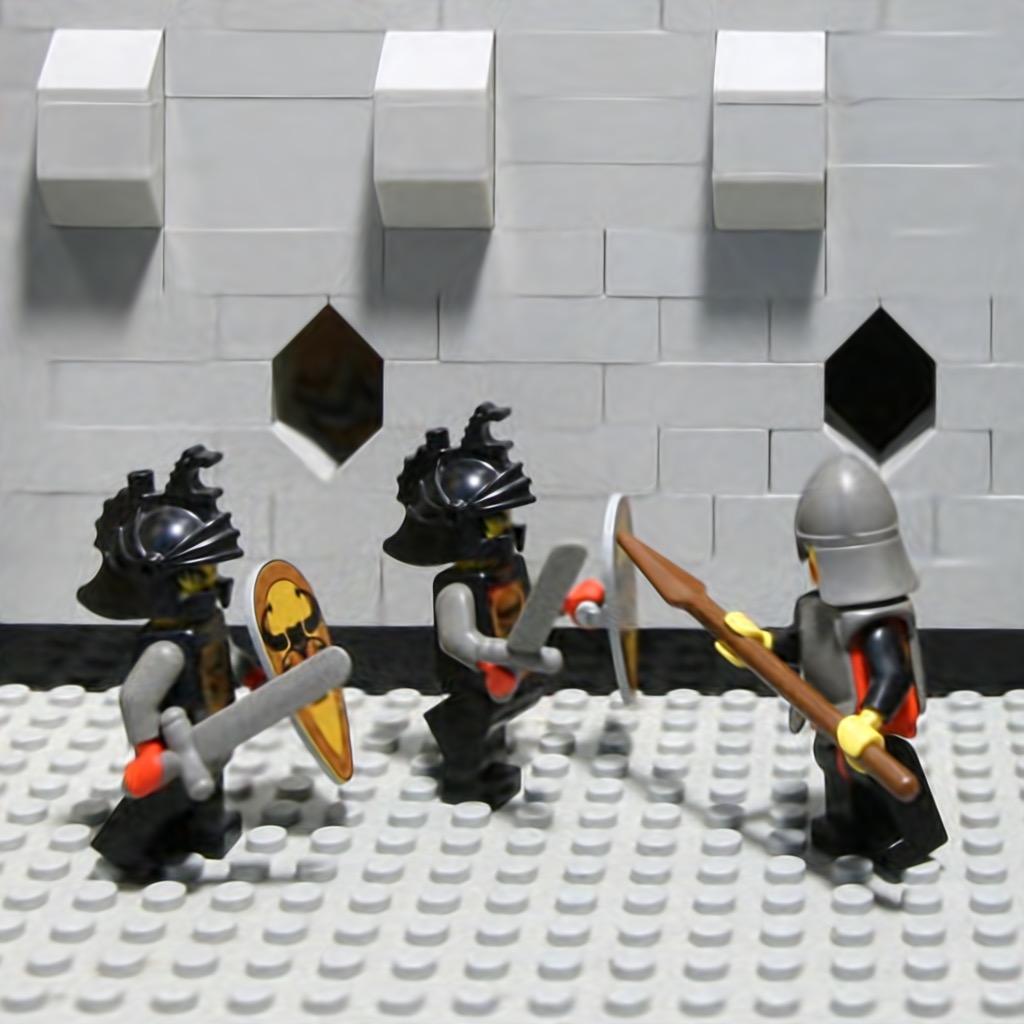}{$t=0$}%
    ~
    \FigTwoSubfig{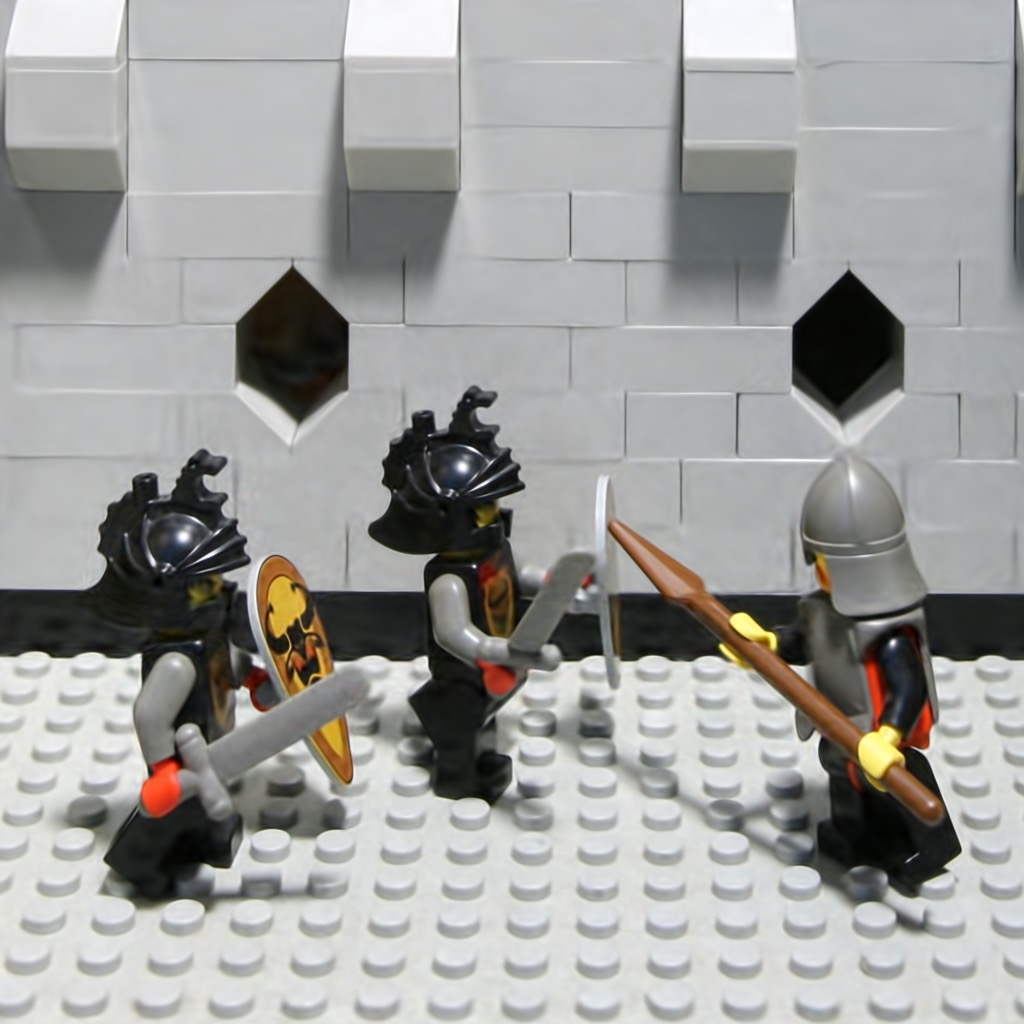}{$t=0.5$}%
    ~
    \FigTwoSubfig{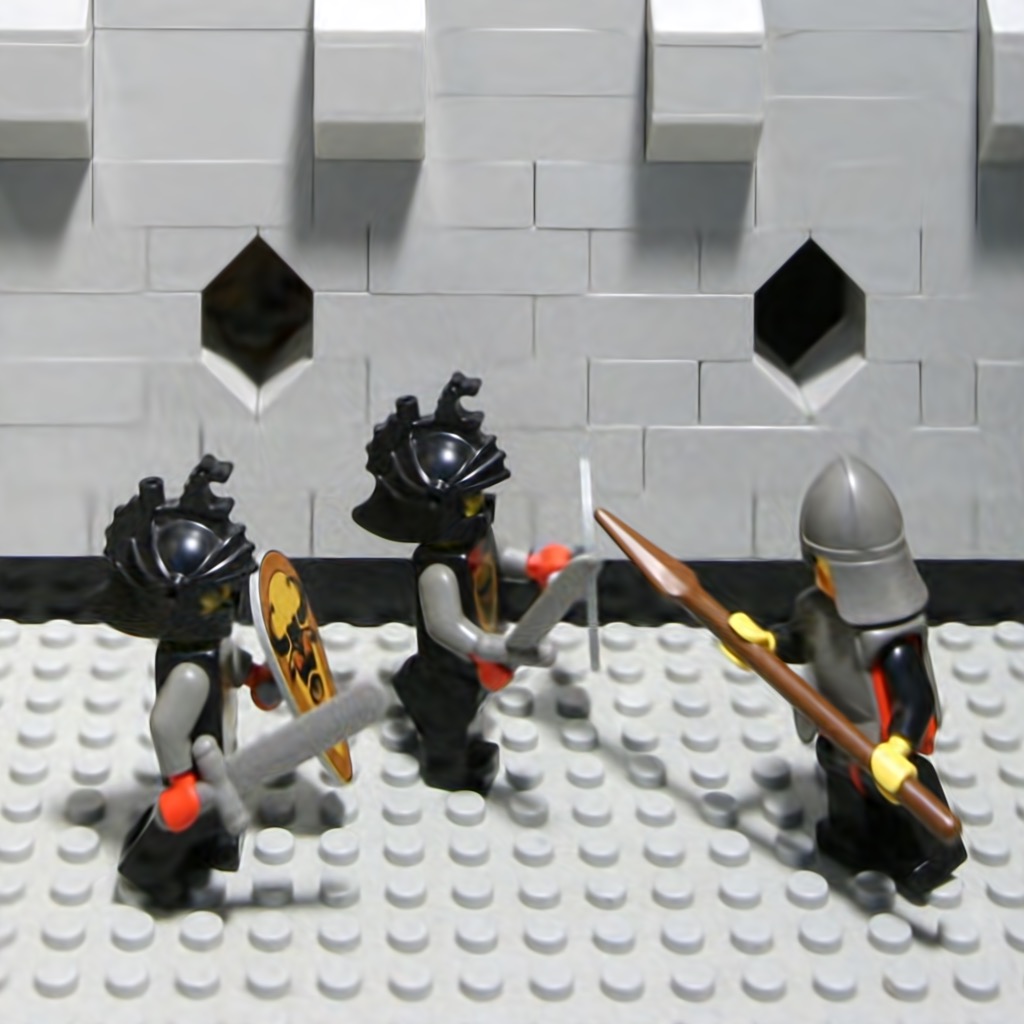}{$t=1$}%
    
    \vspace{-8pt}\hspace{0.1cm}$\underbracket[0.5pt][3pt]{\hspace{8.1cm}}_%
    {\substack{\vspace{-10pt}\\ \colorbox{white}{$2$-norm}}}$    \vspace{-7pt}
    \hspace{0.8cm}$\underbracket[0.5pt][3pt]{\hspace{8.1cm}}_%
    {\substack{\vspace{-10pt}\\ \colorbox{white}{$1$-norm}}}$
    
    $\underbracket[0pt][1mm]{\hspace{15pt}}_%
    {\hspace{-2.2cm}\substack{\vspace{-163pt}\\ {\Large \mybox{\quad \quad}}}}$%
    $\underbracket[0pt][1mm]{\hspace{15pt}}_%
    {\hspace{7.1cm}\substack{\vspace{-163pt}\\ {\Large \mybox{\quad \quad}}}}$
    \vspace{-10pt}
	\caption{{\bf Effect of Controlling $\|z\|_{p} = 1$ with Different $p$.} For each condition, we show the INR-produced images given $z_{i}$ (left), $0.5 z_{i} + 0.5 z_{j}$ (center), and $z_{j}$ (right). ``No Control'' trains INR $\mathcal{F}$ and all codes for known images without controlling their scales, showing proper reconstruction at known views (left and right) but complete failure in interpolation (center). ``$\infty$-norm'' scales each $z$ with its maximum norm, but still does not interpolate well. ``$2$-norm'' significantly improve interpolation and reconstructs known views better, but ``$1$-norm'' is much better at interpolation (see red boxes). }
	\label{fig:NormCompare}
    \vspace{-5pt}
\end{figure*}

\subsubsection{Image Fitting.}
Images are arguably the most dominant form of visual data, and many efforts on INRs are to make them fit 2D images as accurately and quickly as possible~\cite{tancik2021learned, muller2022instant}. 
ACORN~\cite{martel2021acorn} applies spatial subdivision to more efficiently train INR of a single gigapixel image (with 1 billion pixels).
Various signal processing techniques are also shown useful in making INRs fit images more accurately, such as image pyramids~\cite{saragadam2022miner}, sinusoidal and Fourier basis functions~\cite{sitzmann2020implicit, tancik2020fourier}, and multiplicative filtering with Fourier or Gabor wavelet basis functions~\cite{fathony2020multiplicative, huang2021textrm, lindell2021bacon}.
Many of these techniques are applicable for other signals like 3D MRI data or 3D signed distance fields, which are beyond the scope of this paper.

While most of these methods focus on training a single network as INR of a single image, a single network may also serve as INR of multiple images.
For consecutive images at a fixed viewpoint, an extra time dimension is added to the coordinate input~\cite{sitzmann2020implicit}.
For structured 4D light fields where the views lie on a 2D plane, those 2D coordinates can be re-parameterized as input~\cite{feng2021signet, attal2021learning}.
A single network can further serve as a generalizable INR of arbitrary images, by concatenating the code with the 2D pixel coordinate as input to the INR~\cite{mehta2021modulated}.
An alternative approach is to modulate network activation based on the code~\cite{mehta2021modulated, dupont2022data}, which has success in fitting arbitrary image patches.
We find the simple code concatenation is sufficient for our problem, and it has been successfully used to train an INR of light rays from different scenes~\cite{sitzmann2021light, feng2022}.
Unlike modulation, it avoids the cost of additionally training an encoder and a modulator network, allowing us to study of INR with its most basic form.
\vspace{-10pt}

\subsection{Image-based Rendering.}
The early approaches of image-based rendering (IBR) achieve novel view synthesis through explicitly blending relevant pixels from known images~\cite{gortler1996lumigraph, levoy1996light, debevec1996modeling}.
The visual quality of IBR is heavily dependent on the strategy of deciding the blending weights of images, and researchers have developed a line of techniques improving blending weights selection, such as ray-space proximity~\cite{levoy1996light, chai2000plenoptic}, proxy geometry~\cite{heigl1999plenoptic, buehler2001unstructured, debevec1996modeling}, optical flow~\cite{chen1993view, du2018montage4d}, soft blending~\cite{penner2017soft, riegler2020free}, and neural-network-assisted blending~\cite{thies2019deferred, wang2021ibrnet, mildenhall2019local, rombach2021geometry}.
These techniques often require an approximate 3D structure (proxy geometry or depth) of the scene so that pixels can be re-projected to the novel view.
For methods that do not involve 3D re-projection~\cite{ng2005light, levoy1996light}, many still assume the knowledge of the 3D camera locations and orientations of each image and leverage the spatial relationship among the cameras to decide the blending weights.
In contrast, we explore a different and more challenging problem setting which does not involve 3D reconstruction nor the knowledge of 3D locations and camera orientations.
Our problem setup is similar to prior work on image morphing~\cite{wolberg1998image, chen1993view, seitz1996view, liao2014automating}, but we achieve the morphing effect without finding pixel-wise correspondences between images. 

\begin{figure*}[!ht]
    \FigThreeSubfig{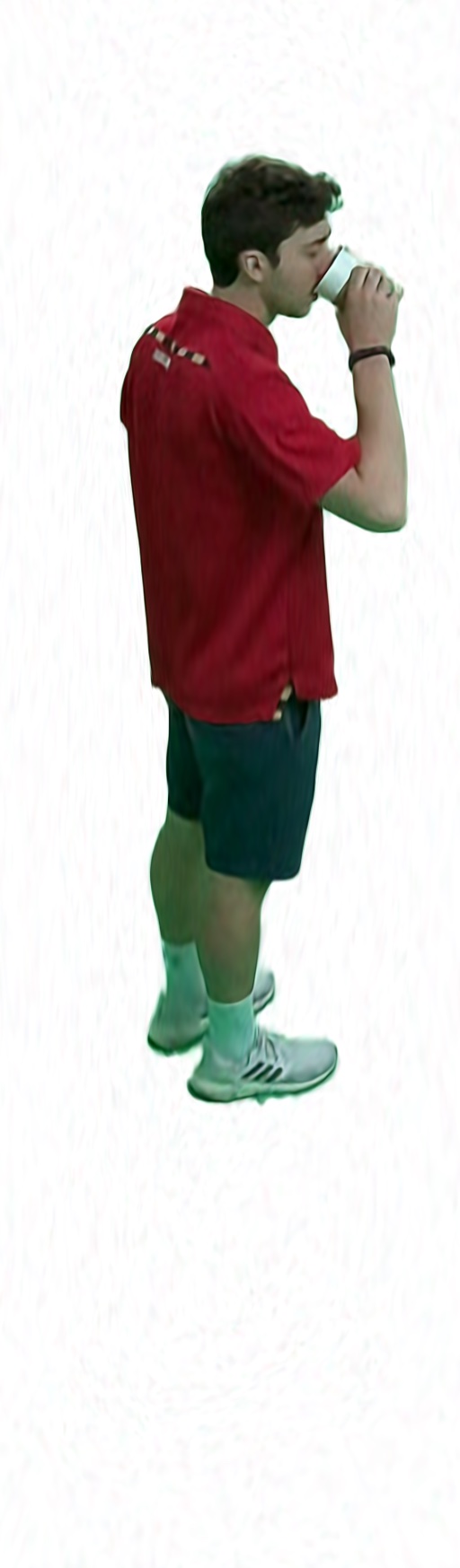}{$t=0$}%
    ~
    \FigThreeSubfig{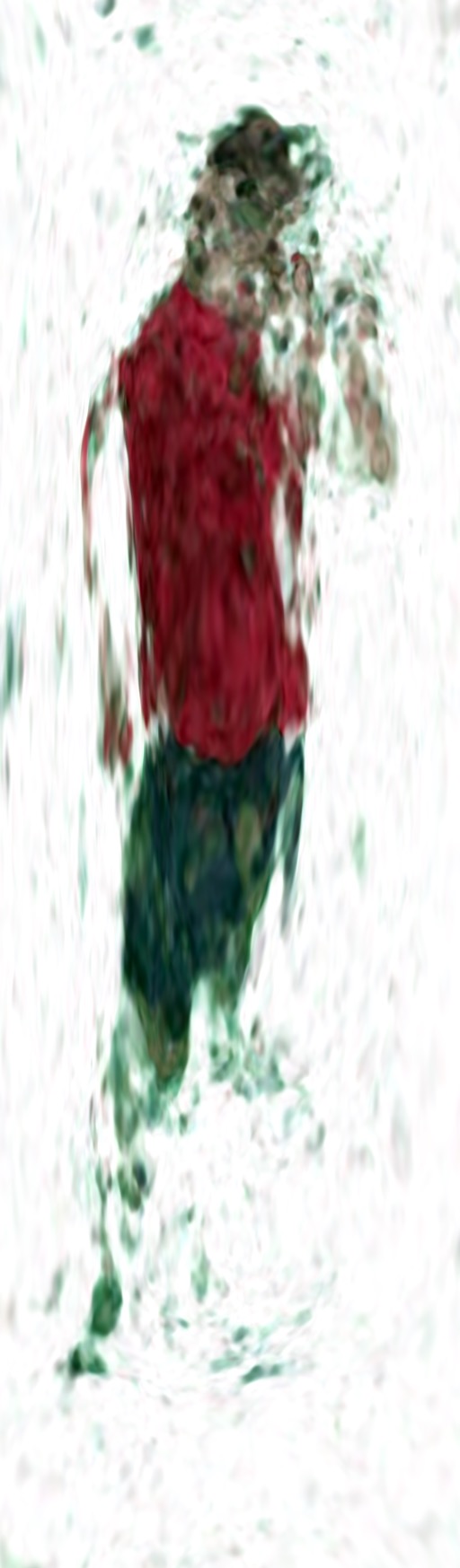}{$t=0.5$}%
    ~
    \FigThreeSubfig{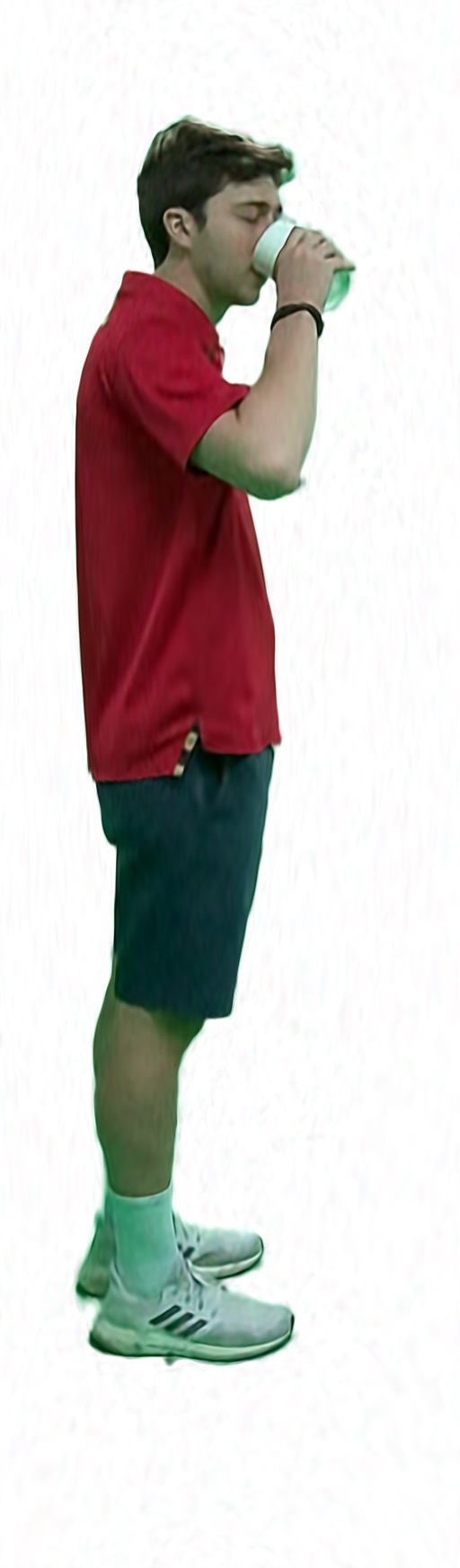}{$t=1$}%
    ~
    \FigThreeSubfig{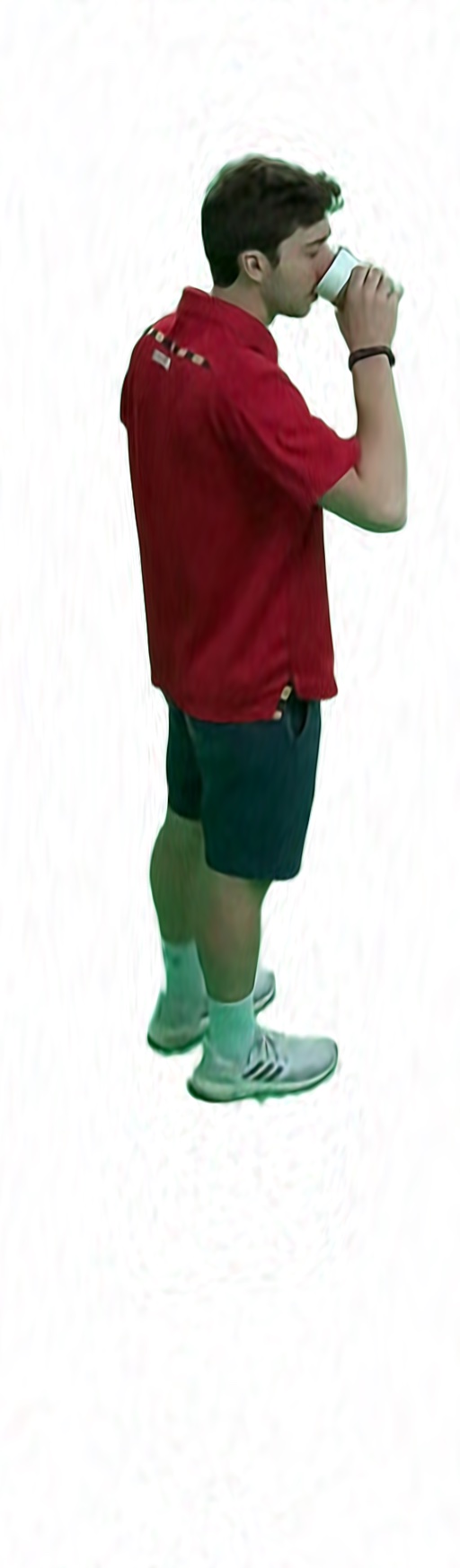}{$t=0$}%
    ~
    \FigThreeSubfig{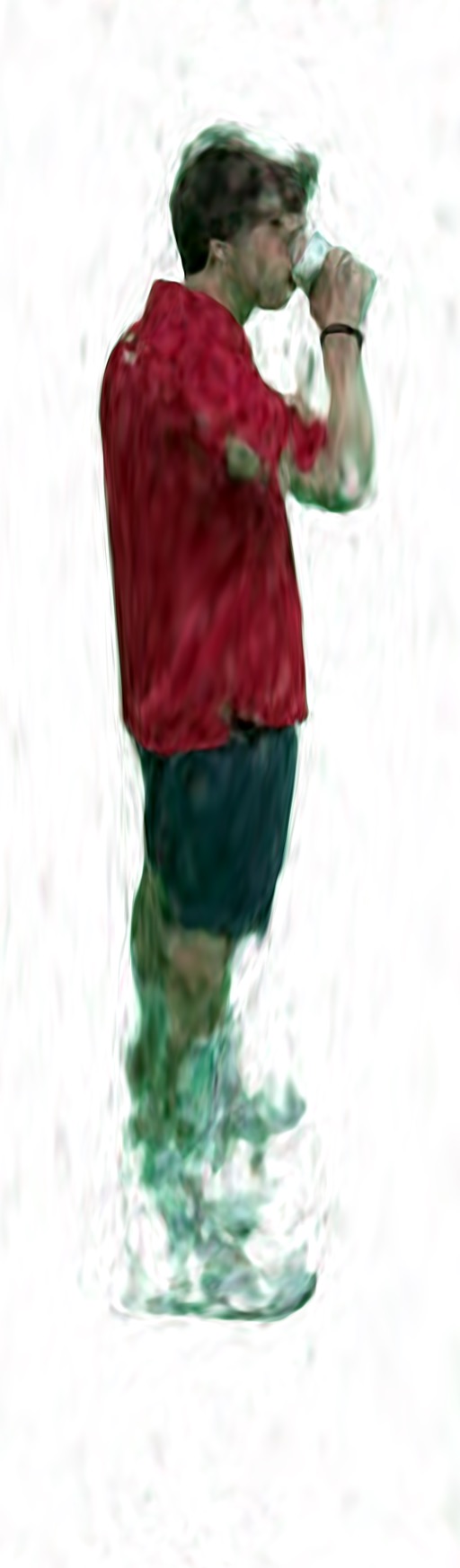}{$t=0.5$}%
    ~
    \FigThreeSubfig{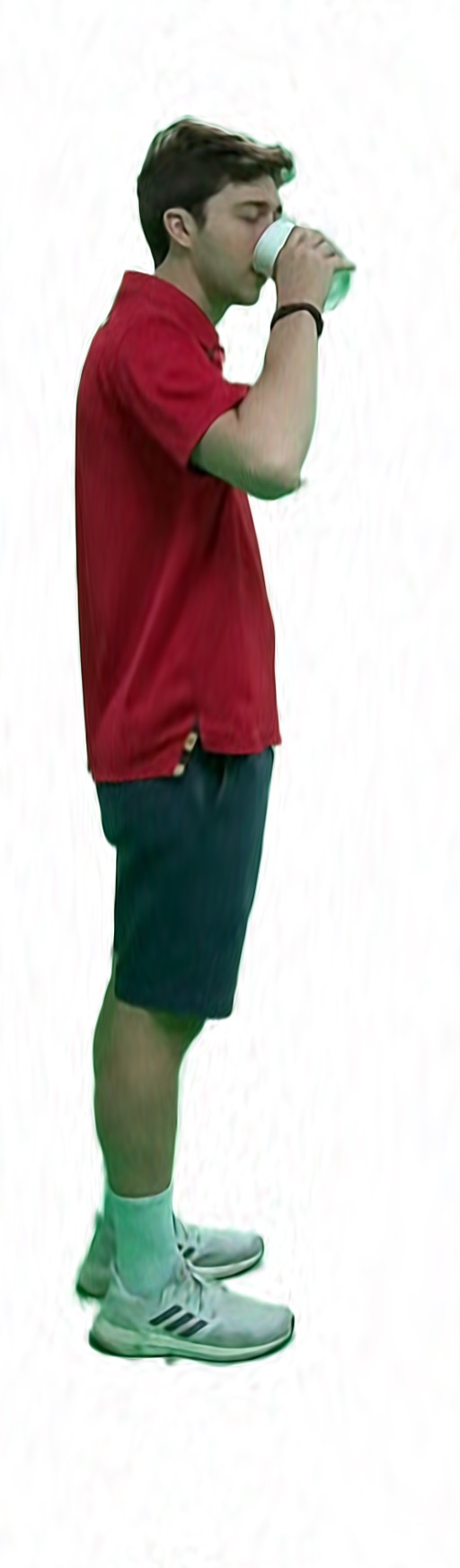}{$t=1$}%
    
    \vspace{-12pt}\hspace{0.1cm}$\underbracket[0.5pt][3pt]{\hspace{8.1cm}}_%
    {\substack{\vspace{-10pt}\\ \colorbox{white}{$M=16$}}}$    \vspace{0pt}
    \hspace{0.8cm}$\underbracket[0.5pt][3pt]{\hspace{8cm}}_%
    {\substack{\vspace{-10pt}\\ \colorbox{white}{$M=32$}}}$
    \vspace{0pt}

    \FigThreeSubfig{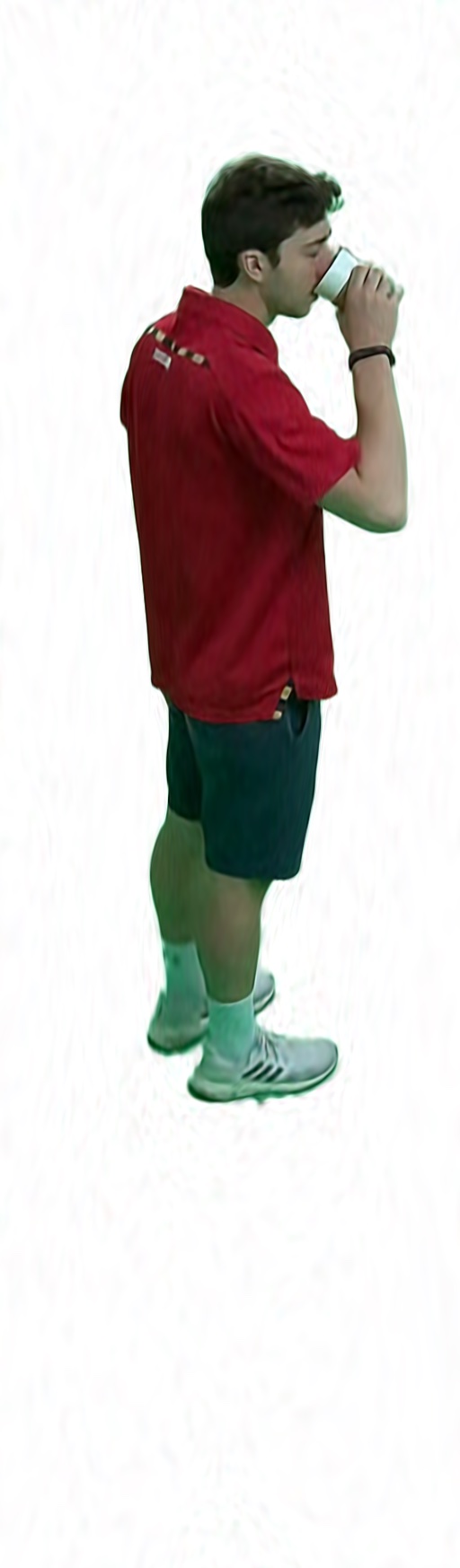}{$t=0$}%
    ~
    \FigThreeSubfig{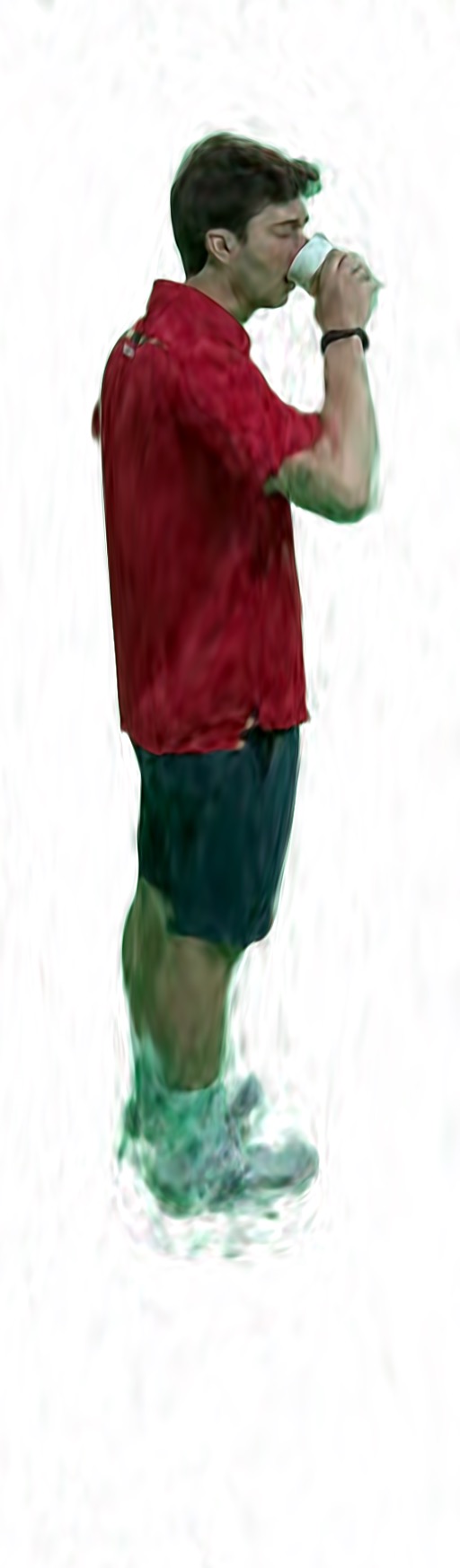}{$t=0.5$}%
    ~
    \FigThreeSubfig{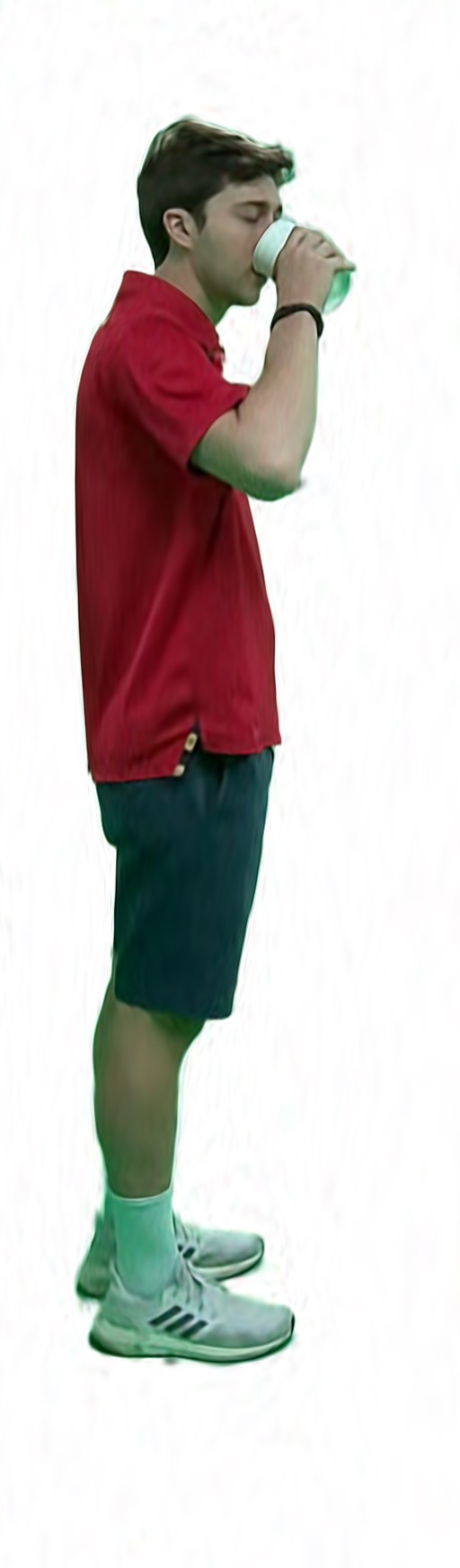}{$t=1$}%
    ~
    \FigThreeSubfig{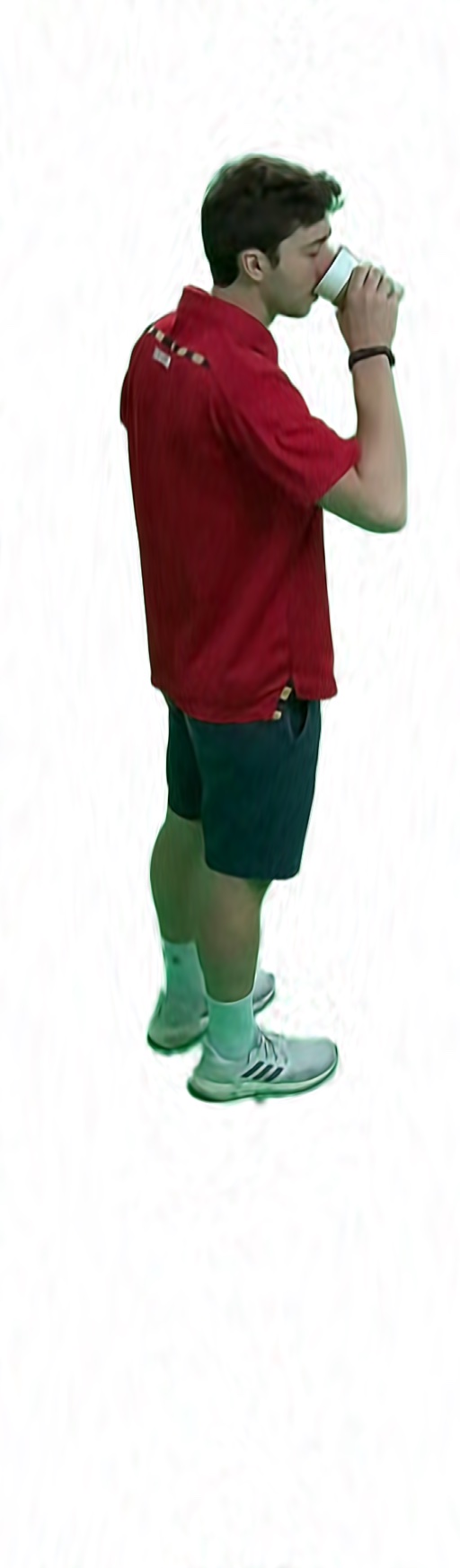}{$t=0$}%
    ~
    \FigThreeSubfig{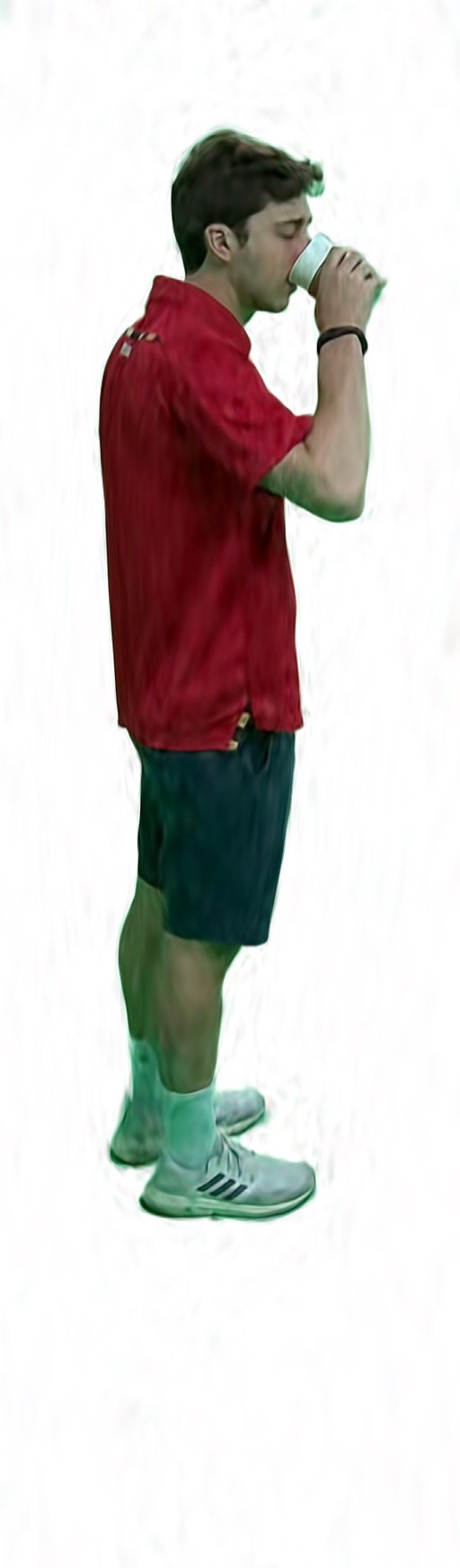}{$t=0.5$}%
    ~
    \FigThreeSubfig{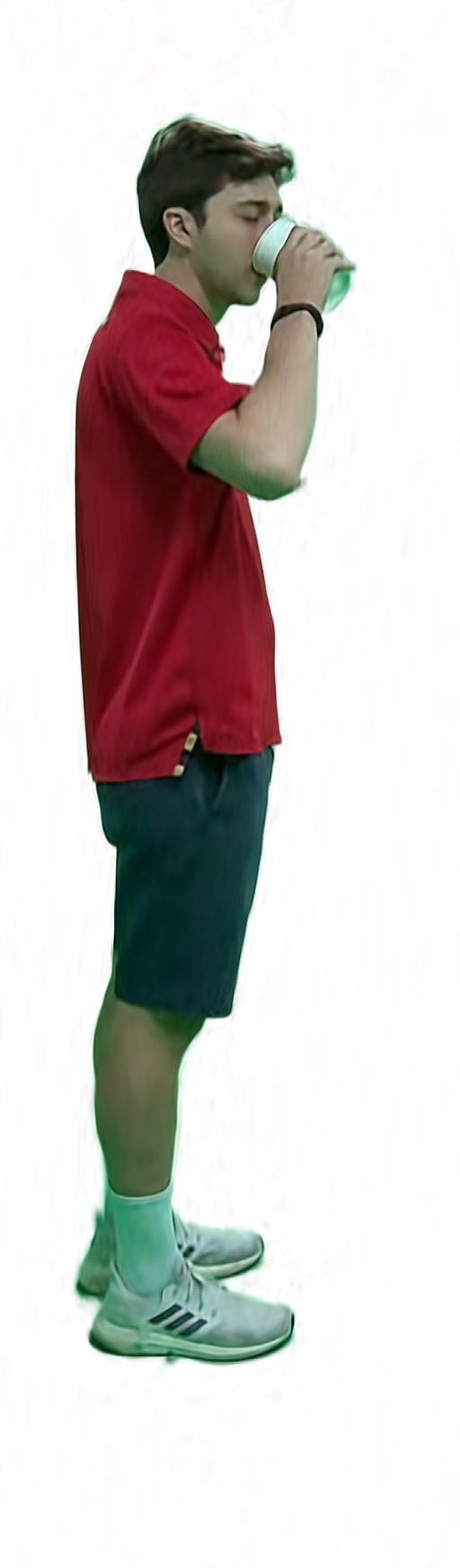}{$t=1$}%
    
    \vspace{-12pt}\hspace{0.1cm}$\underbracket[0.5pt][3pt]{\hspace{8.1cm}}_%
    {\substack{\vspace{-10pt}\\ \colorbox{white}{$M=64$}}}$    \vspace{-7pt}
    \hspace{0.8cm}$\underbracket[0.5pt][3pt]{\hspace{8cm}}_%
    {\substack{\vspace{-10pt}\\ \colorbox{white}{$M=128$}}}$
    \vspace{-7pt}

	\caption{{\bf Effect of Code Length $M$.} When trained with shorter code vectors, the INR can still produce good results at known views ($t=0, 1$) where the code $z$s are well-trained to reconstruct the pixel color. However, the output given interpolated codes ($t=0.5$) rapidly decreases as the code length decreases. }
	\label{fig:CodeCompare}
\end{figure*}

\section{Method} \label{Section:Method}
We provide details on the INR parametrization adopted in our study, and we introduce the proposed modifications to INR training.

\begin{figure}[!ht]
    \FigFourSubfig{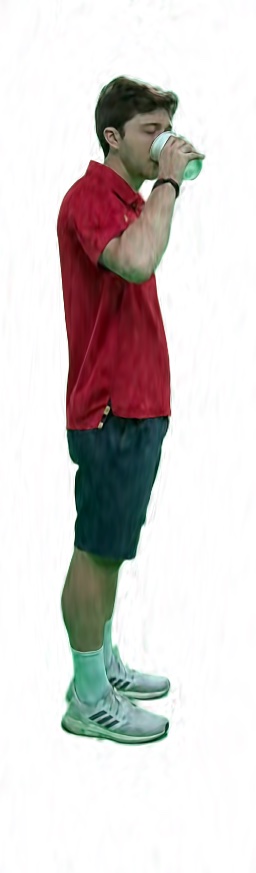}{$t=0.25$}%
    ~
    \FigFourSubfig{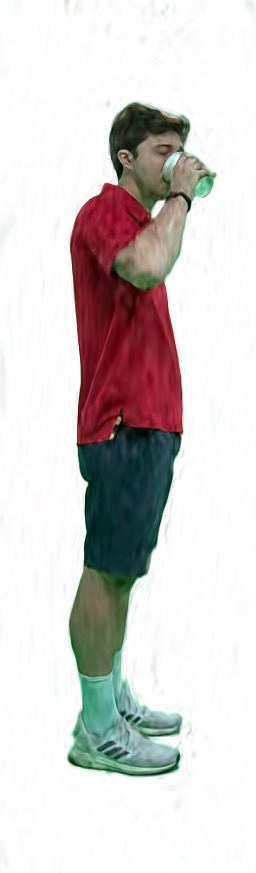}{$t=0.5$}%
    ~
    \FigFourSubfig{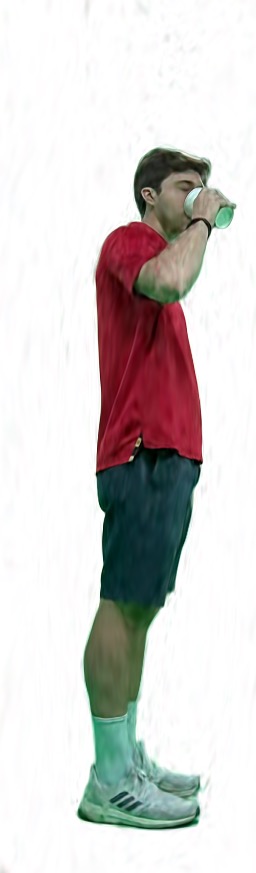}{$t=0.75$}%

    \vspace{-14pt}\hspace{0cm}$\underbracket[0.5pt][3pt]{\hspace{8cm}}_%
    {\substack{\vspace{-3mm}\\ \colorbox{white}{No $L_{Inter}$}}}$
    \vspace{0pt}
    \FigFourSubfig{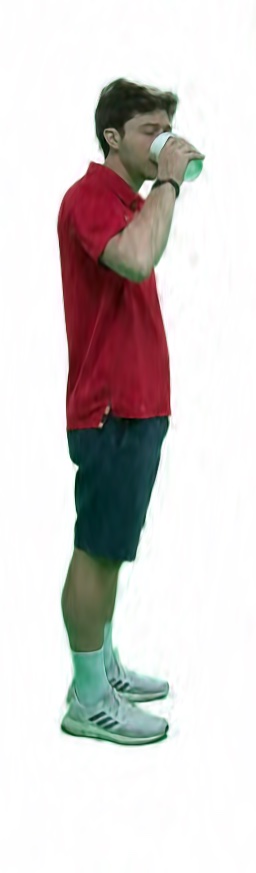}{$t=0.25$}%
    ~
    \FigFourSubfig{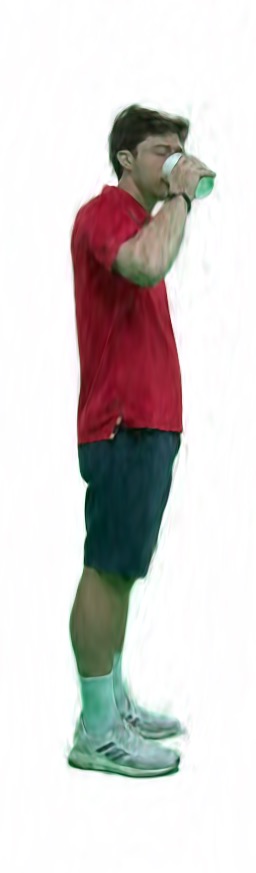}{$t=0.5$}%
    ~
    \FigFourSubfig{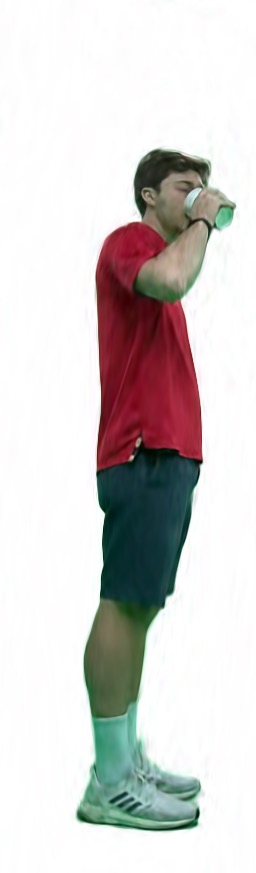}{$t=0.75$}%

    \vspace{-14pt}\hspace{0cm}$\underbracket[0.5pt][3pt]{\hspace{8cm}}_%
    {\substack{\vspace{-3mm}\\ \colorbox{white}{VGGNet}}}$
    \vspace{0pt}
    \FigFourSubfig{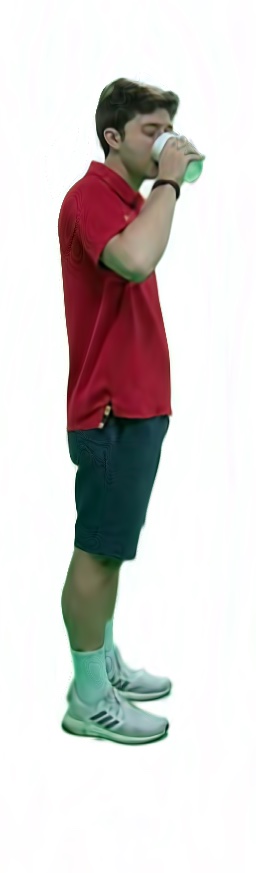}{$t=0.25$}%
    ~
    \FigFourSubfig{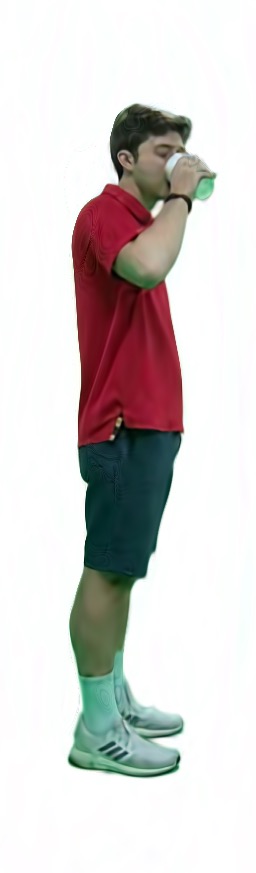}{$t=0.5$}%
    ~
    \FigFourSubfig{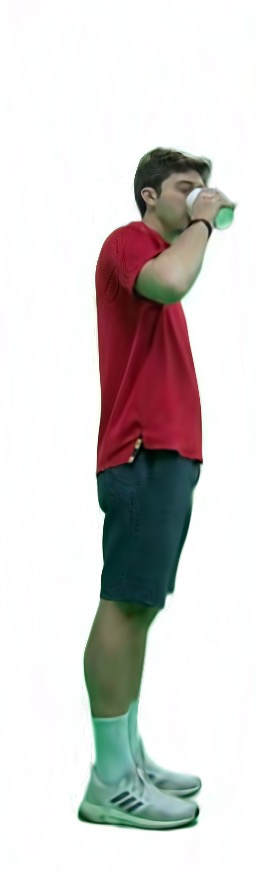}{$t=0.75$}%
    
    \vspace{-14pt}\hspace{0cm}$\underbracket[0.5pt][3pt]{\hspace{8cm}}_%
    {\substack{\vspace{-3mm}\\ \colorbox{white}{CLIP}}}$
    \vspace{-15pt}
	\caption{{{\bf Effect of $L_{Inter}$ Loss.}} Without considering $L_{Inter}$, the interpolation contains visible artifact. With $L_{Inter}$ computed based on the common VGGNet-based perceptual features, the results are worsened by over-smoothing artifacts. We propose using CLIP-extracted features to compute $L_{Inter}$, which significantly reduces the artifacts during interpolation. }
	\label{fig:CLIPCompare}
	\vspace{-15pt}
\end{figure}

\subsection{INR for Image Fitting}
Let $\mathcal{F}$ denote the INR of images.
In the case of a single image, for all pixels $p$ of the image, the INR $\mathcal{F}$ defines
\begin{equation}
\mathcal{F}(p_{x}, p_{y}) = p_{c},
\end{equation}
{where $(p_{x}, p_{y})$ denotes the coordinate of the pixel $p$}, with {$p_x \in \mathbb{R}$} and {$p_y \in \mathbb{R}$}. {$p_c \in \mathbb{R}^3$} denotes the value (often the RGB vector) associated with the pixel $p$.
In itself, the INR formulation is invariant to different numeric ranges of $(p_{x}, p_{y})$ or $p_{c}$, and for simplicity we rescale the pixel coordinates and values to be within [0, 1].

\begin{figure*}[!ht]
    \FigFiveSubfigA{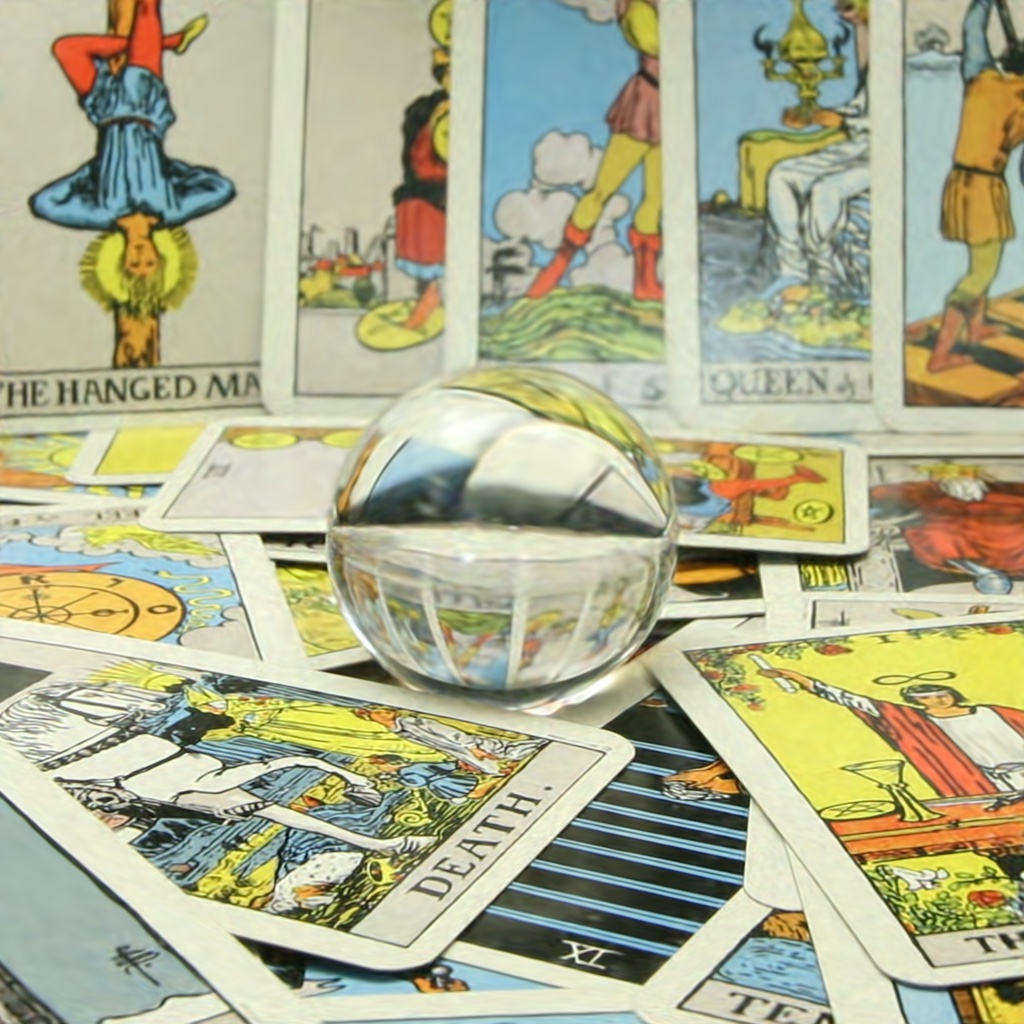}%
    ~
    \FigFiveSubfigA{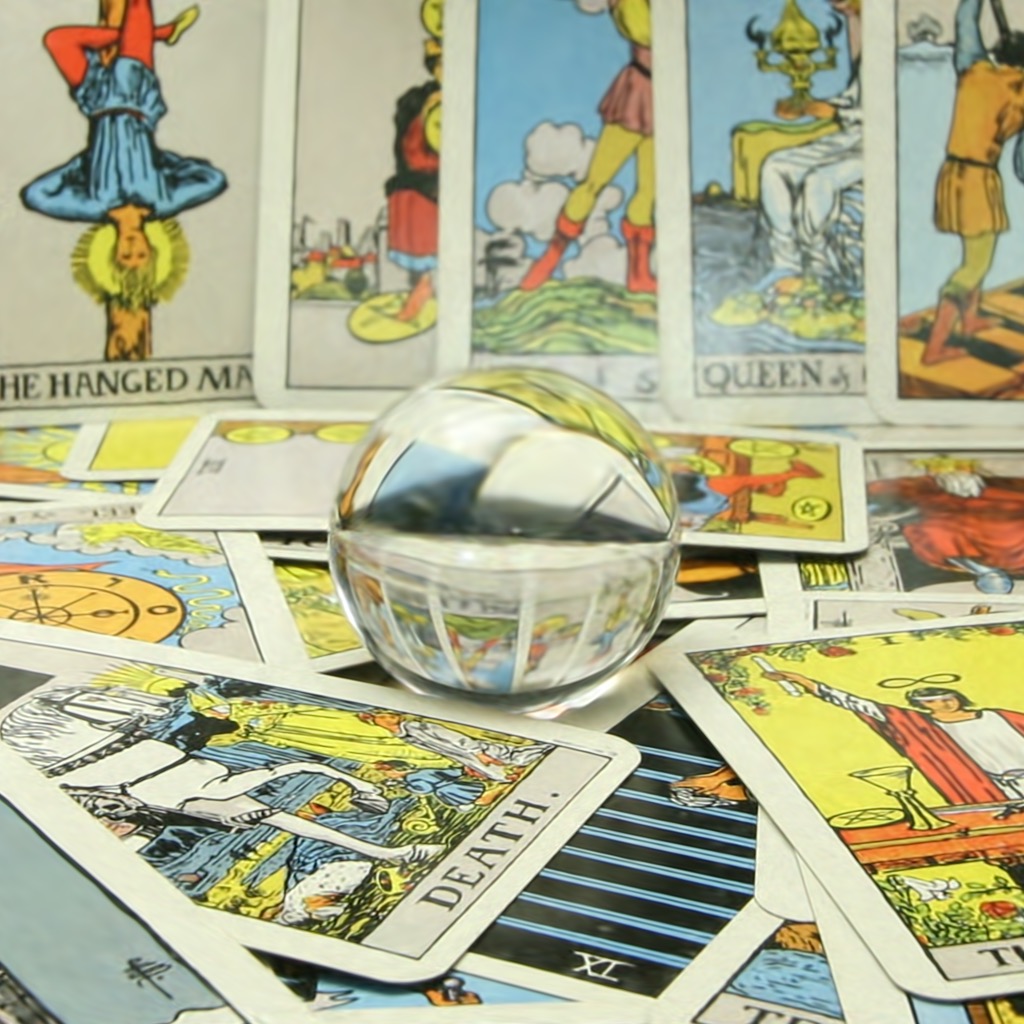}%
    ~
    \FigFiveSubfigA{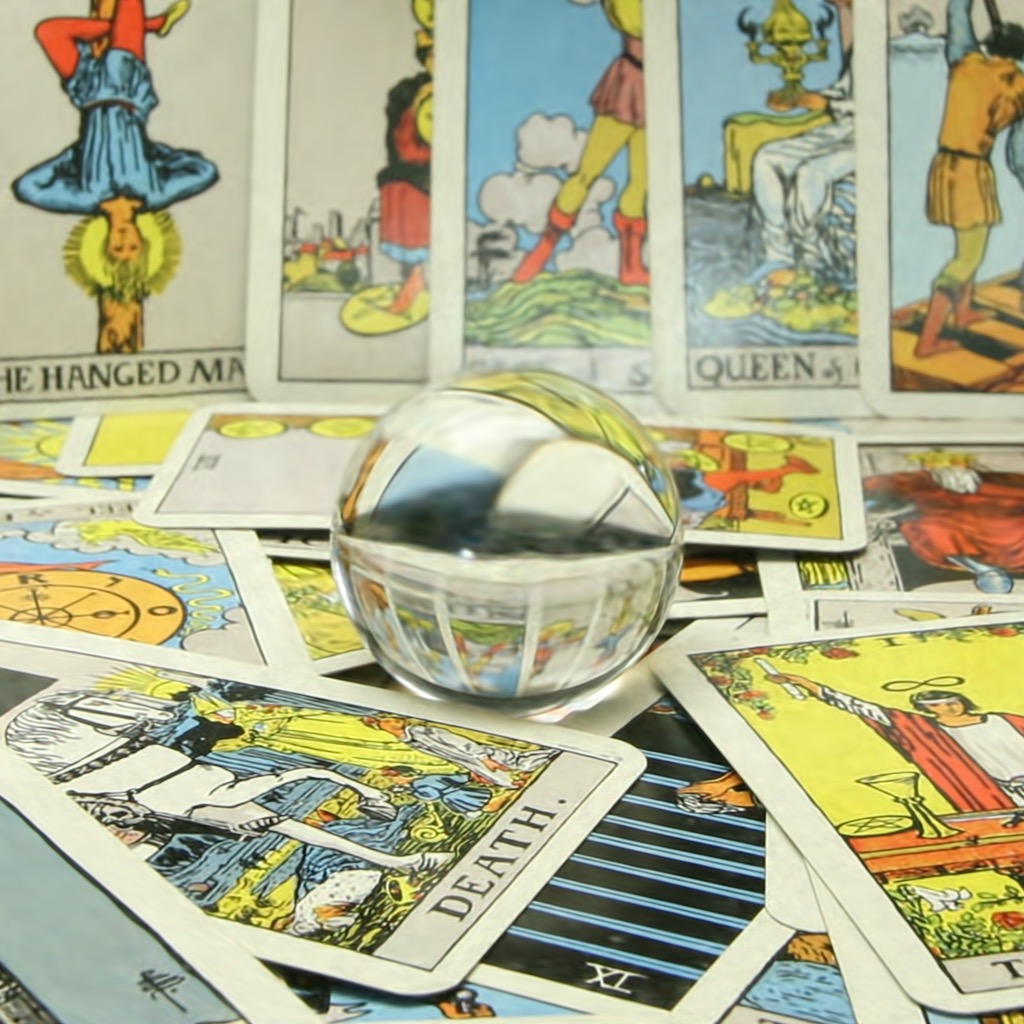}%
    ~
    \FigFiveSubfigA{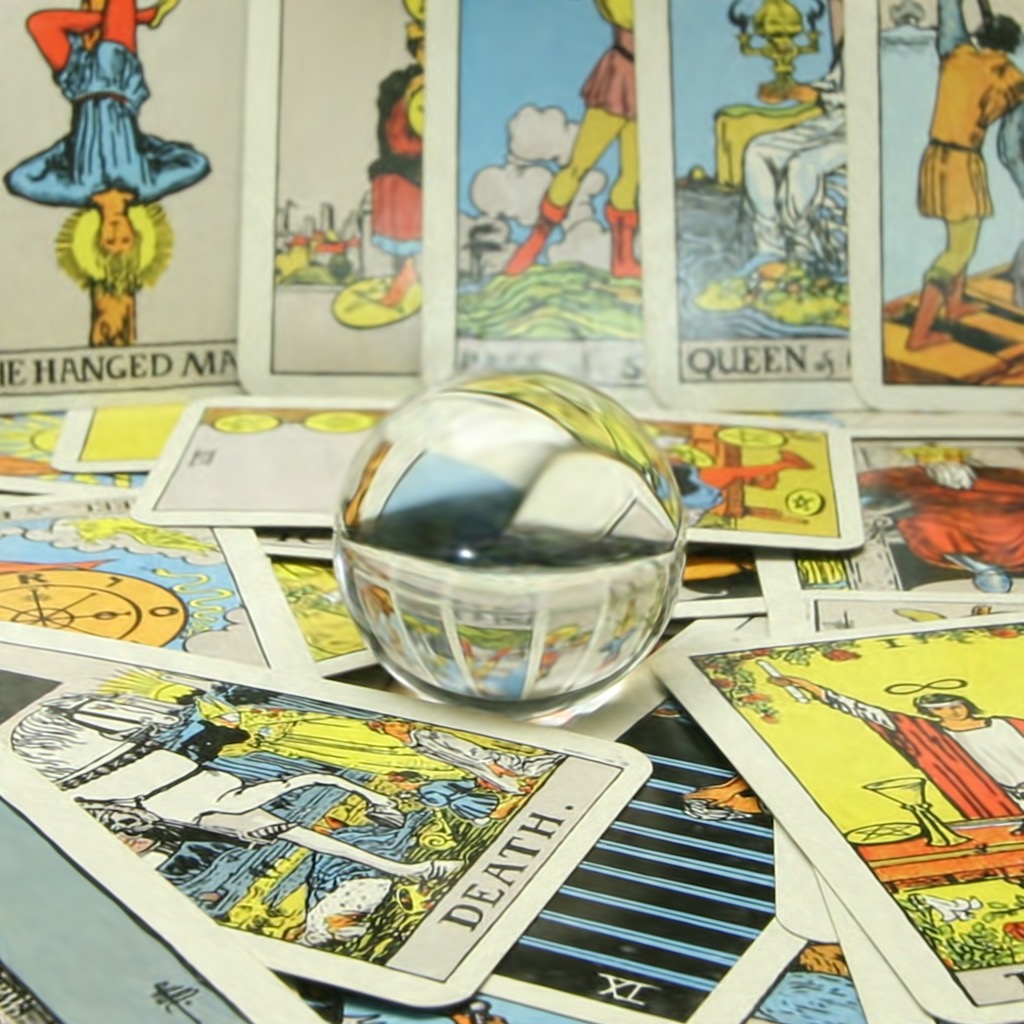}%
    ~
    \FigFiveSubfigA{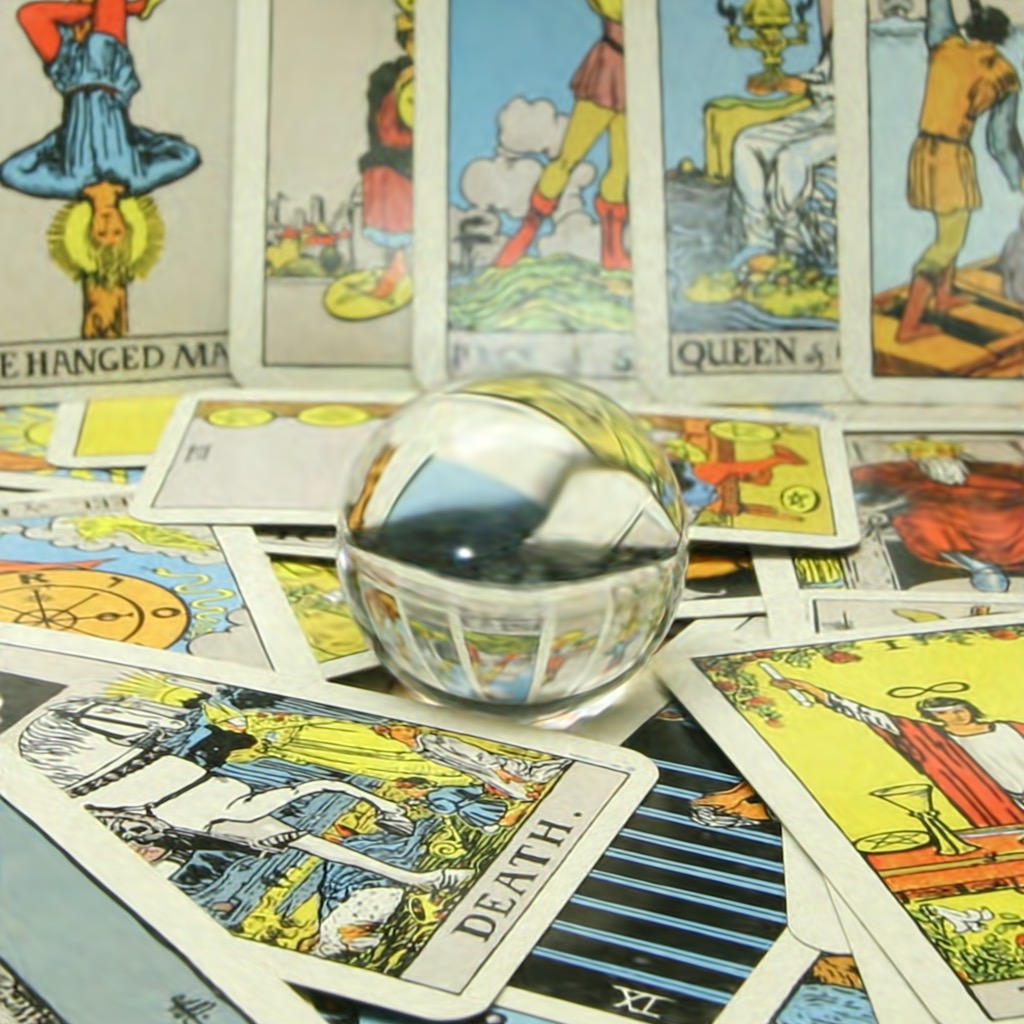}%

    \FigFiveSubfigB{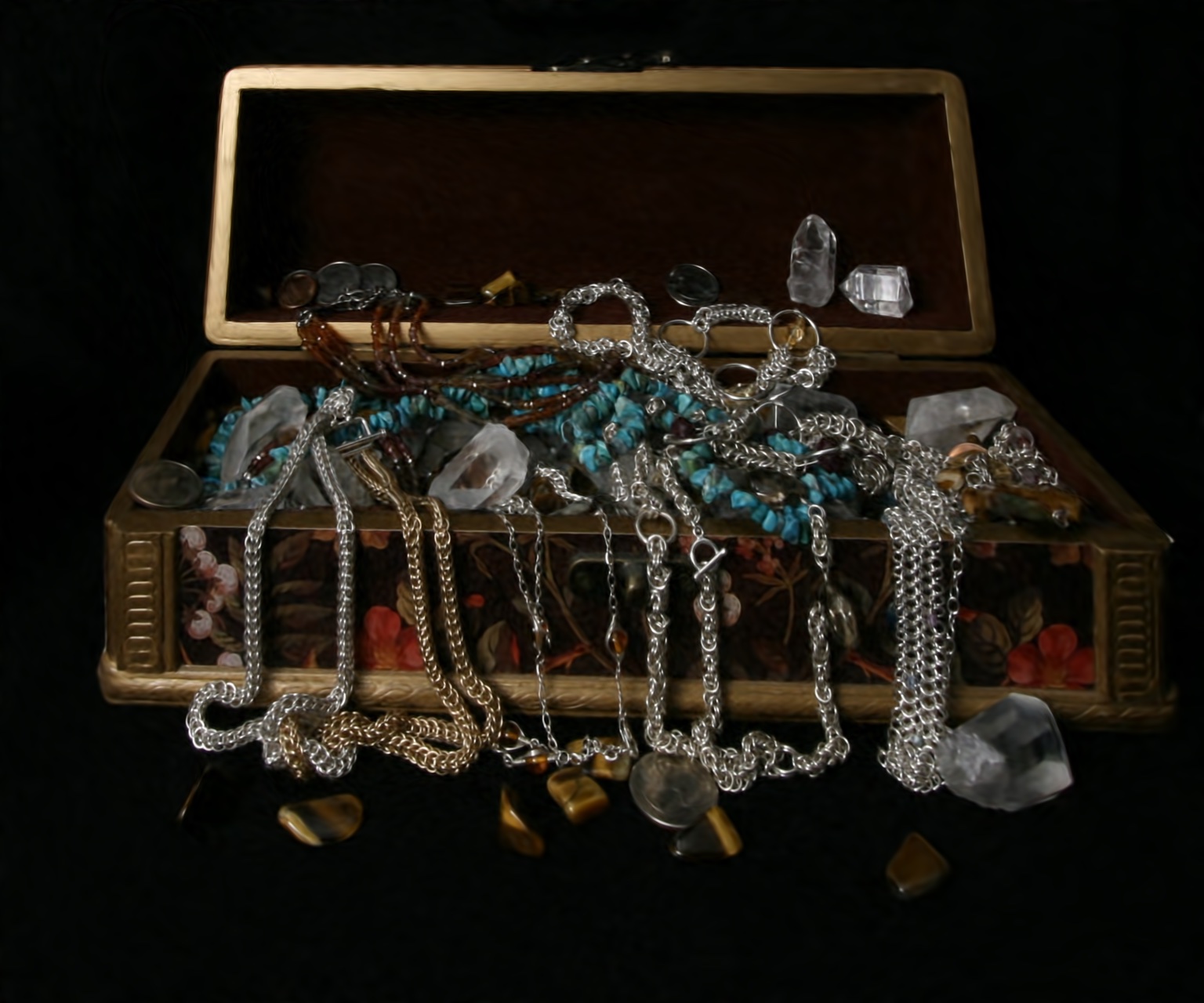}%
    ~
    \FigFiveSubfigB{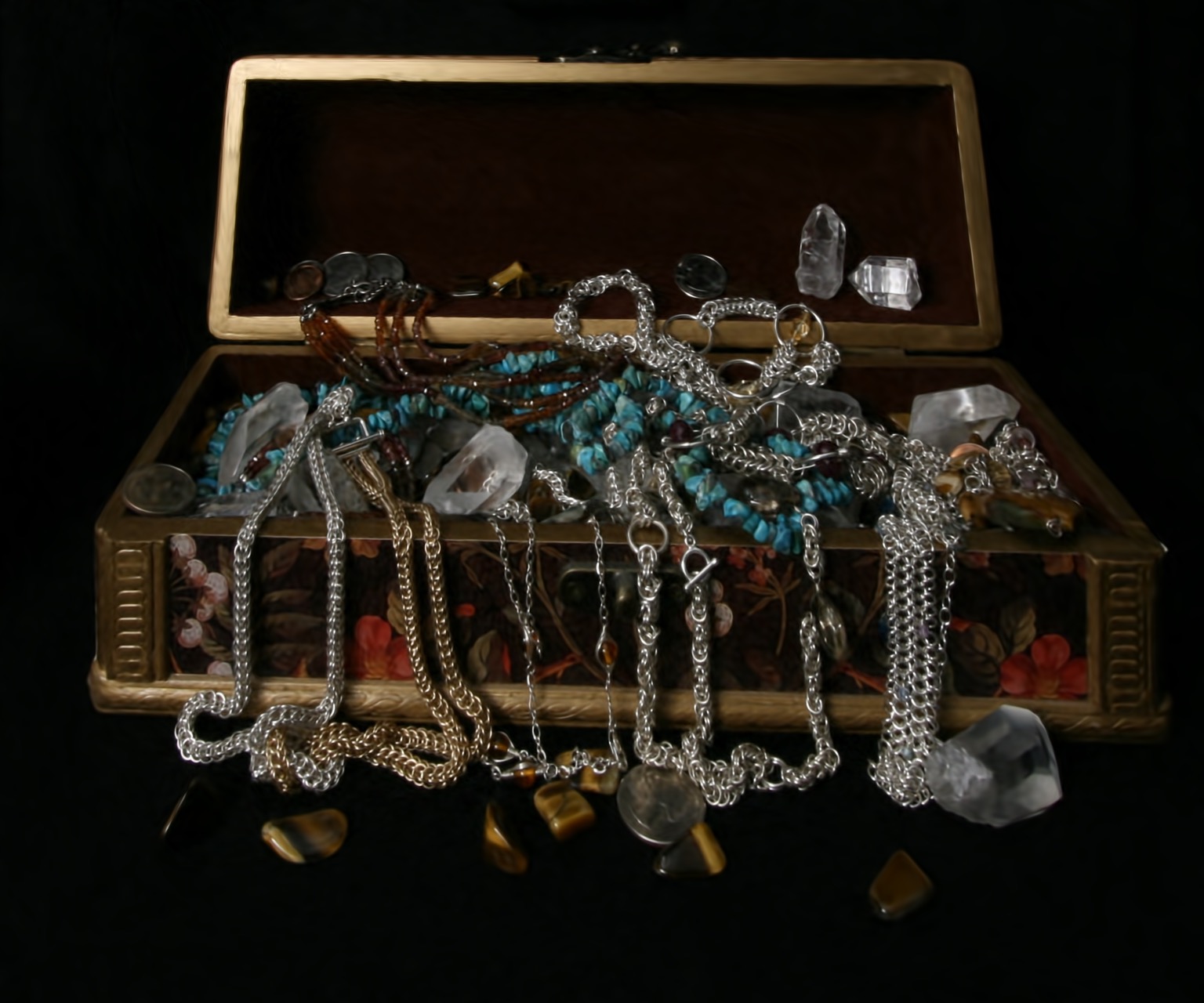}%
    ~
    \FigFiveSubfigB{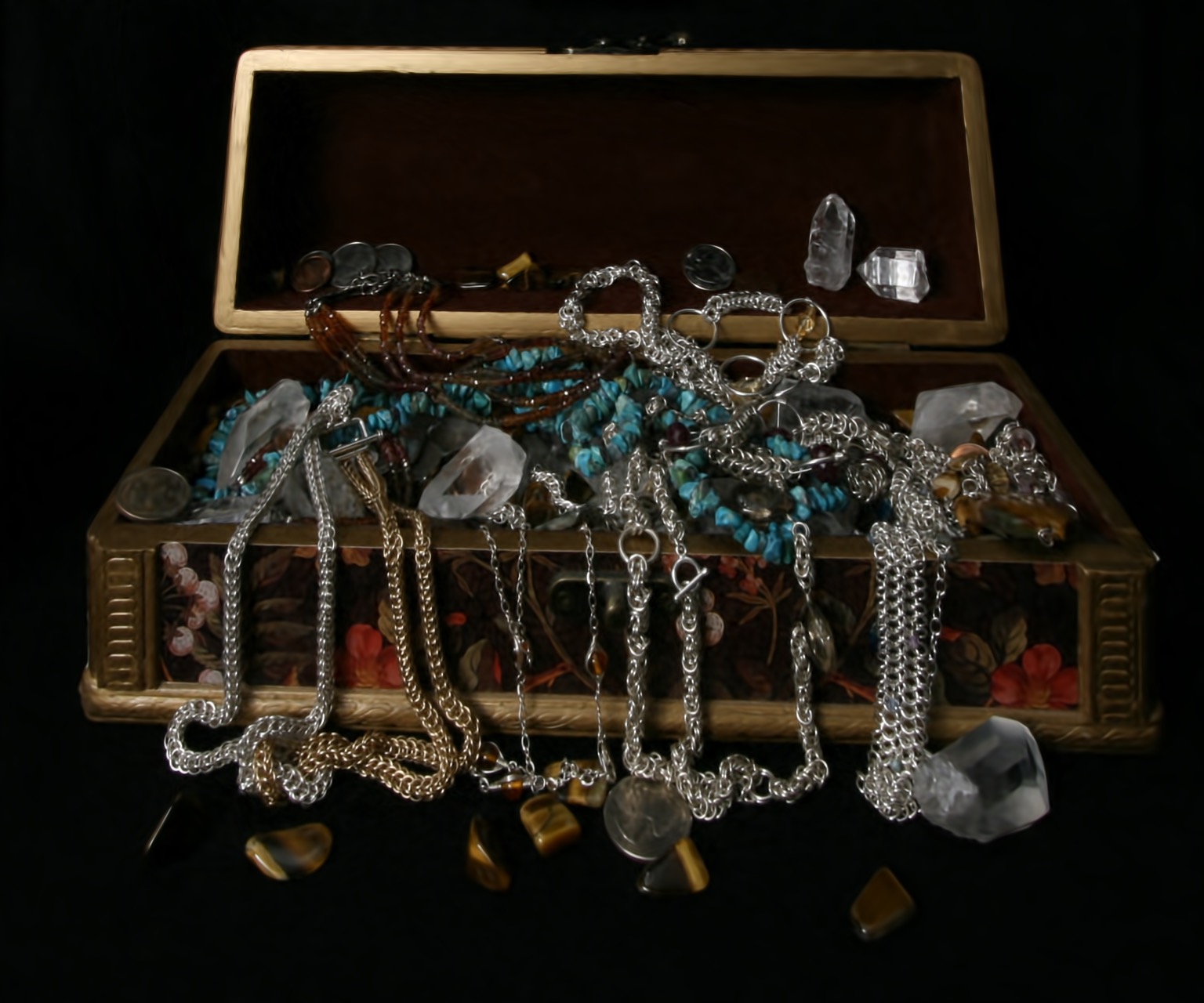}%
    ~
    \FigFiveSubfigB{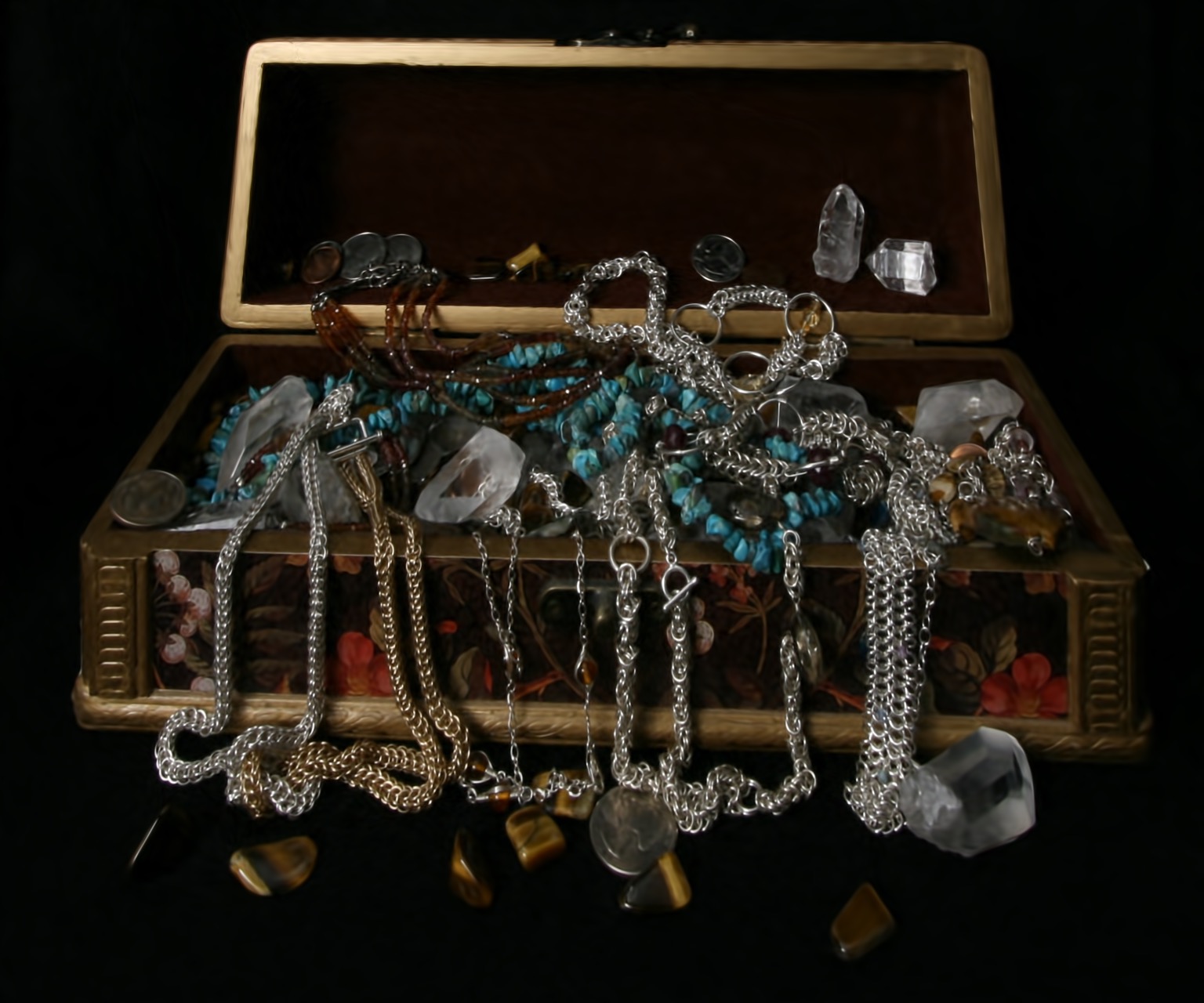}%
    ~
    \FigFiveSubfigB{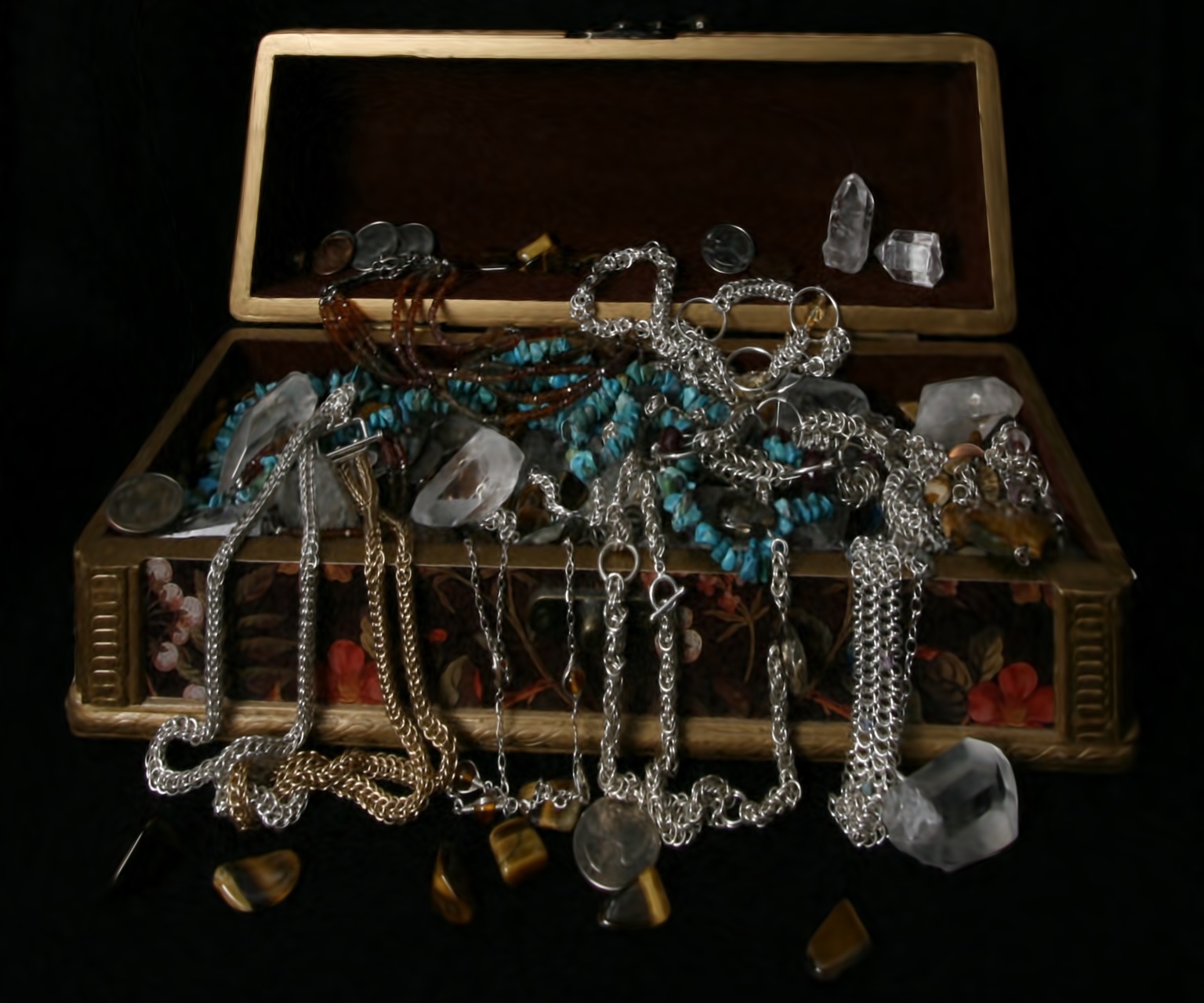}%

    \FigFiveSubfigC{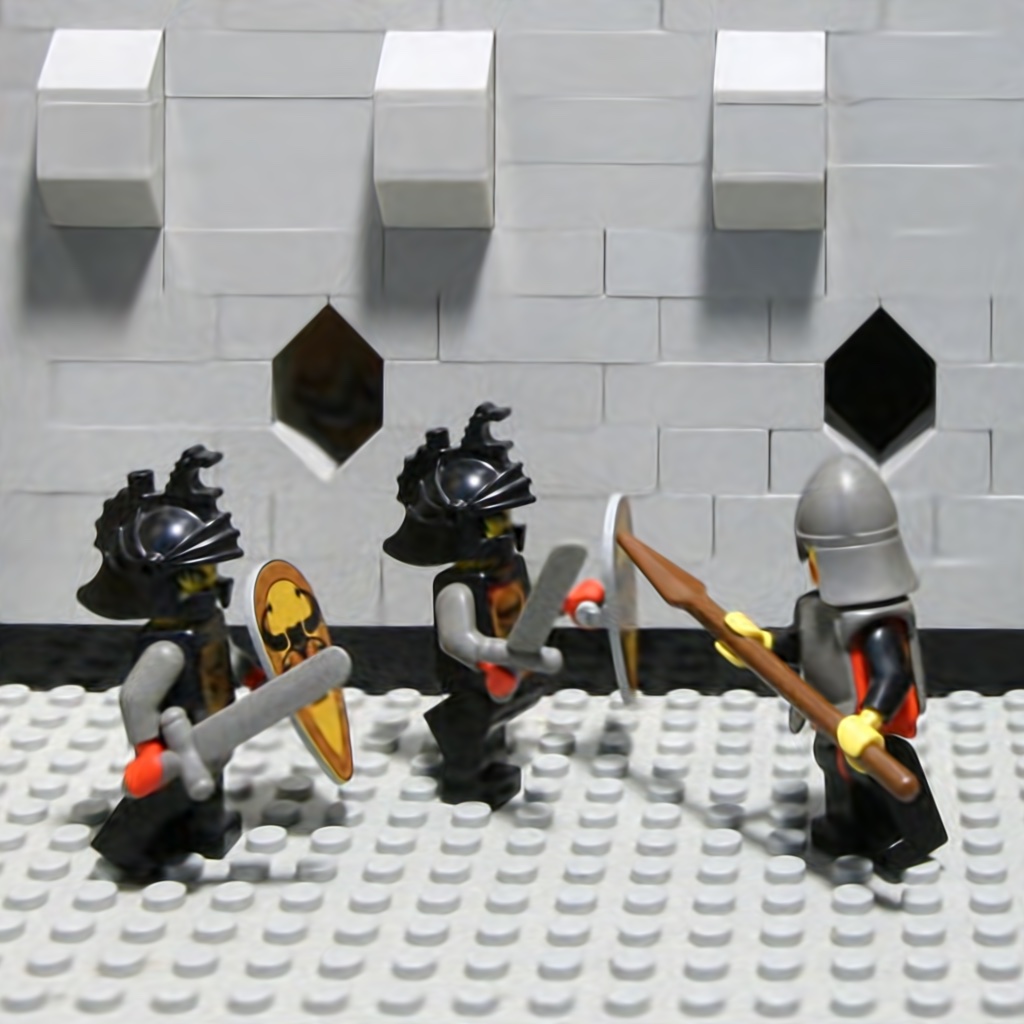}%
    ~
    \FigFiveSubfigC{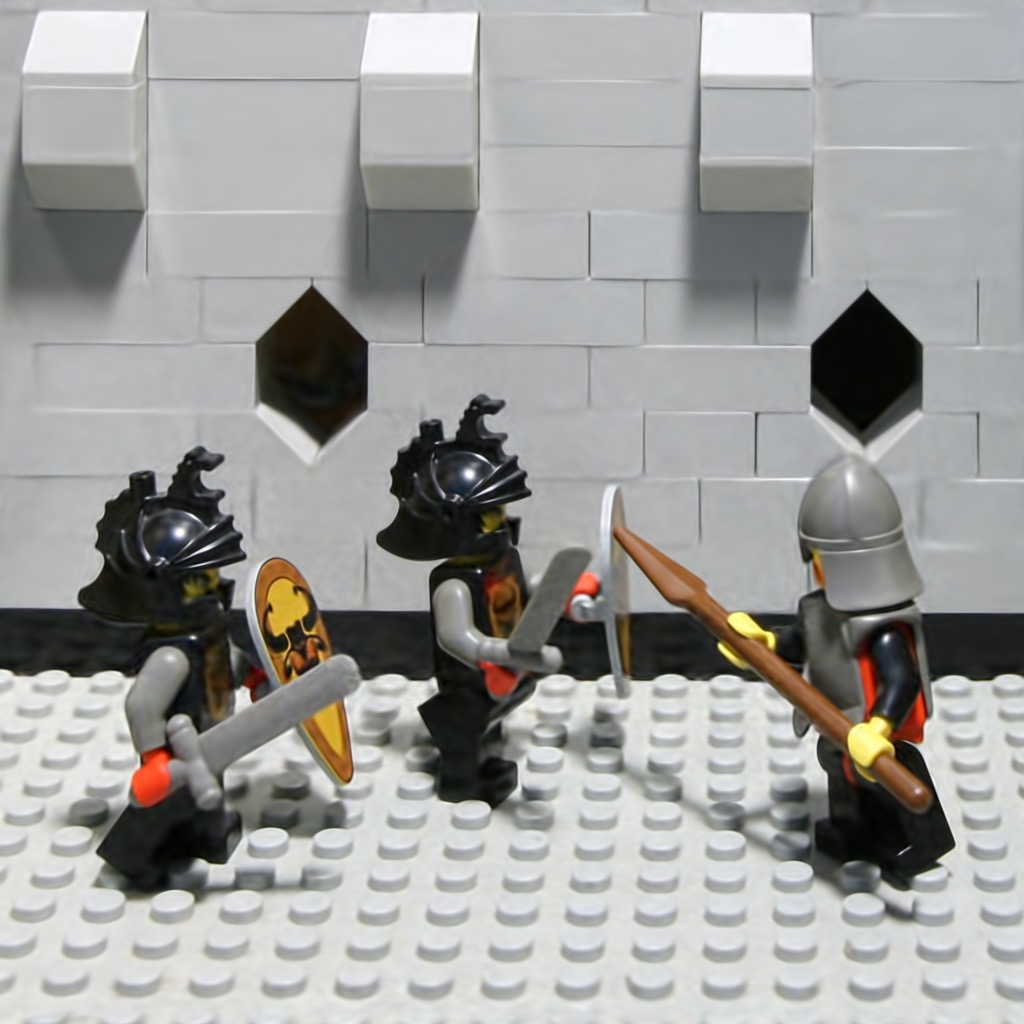}%
    ~
    \FigFiveSubfigC{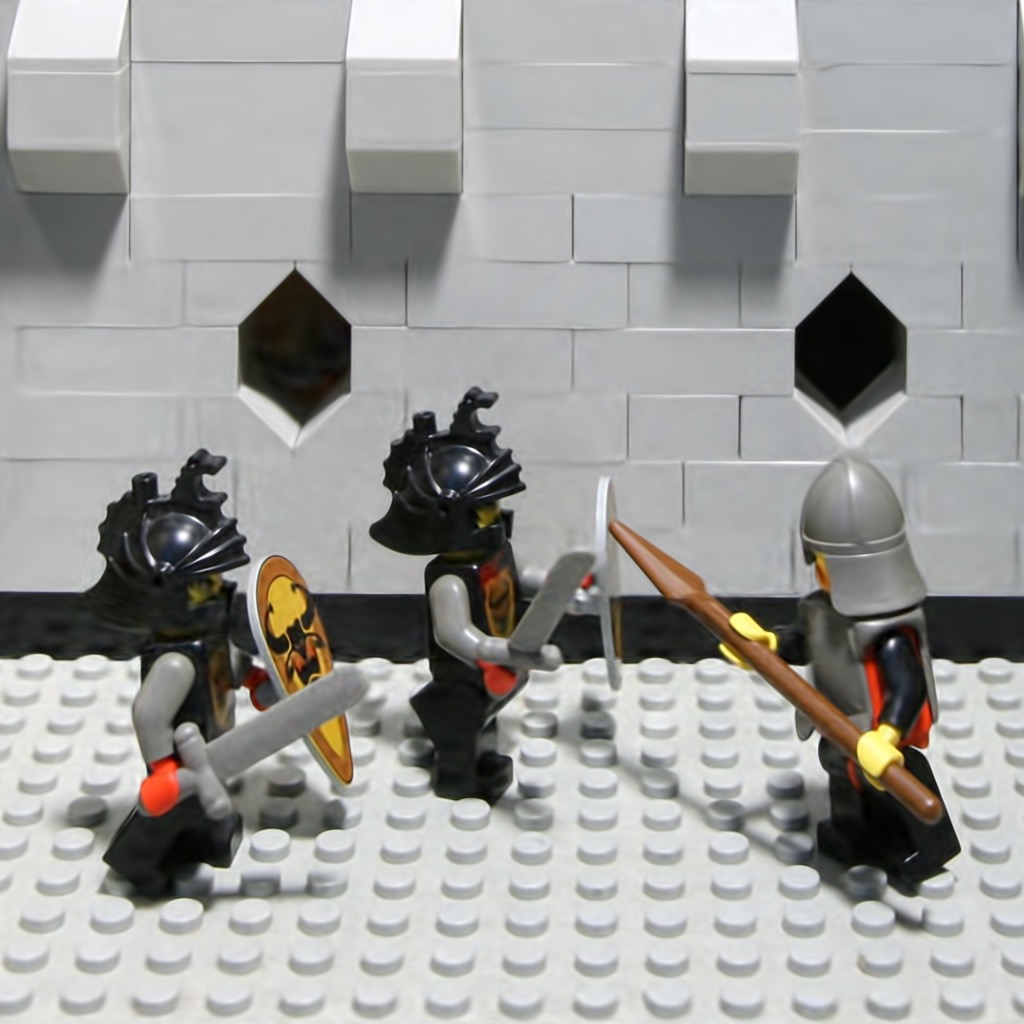}%
    ~
    \FigFiveSubfigC{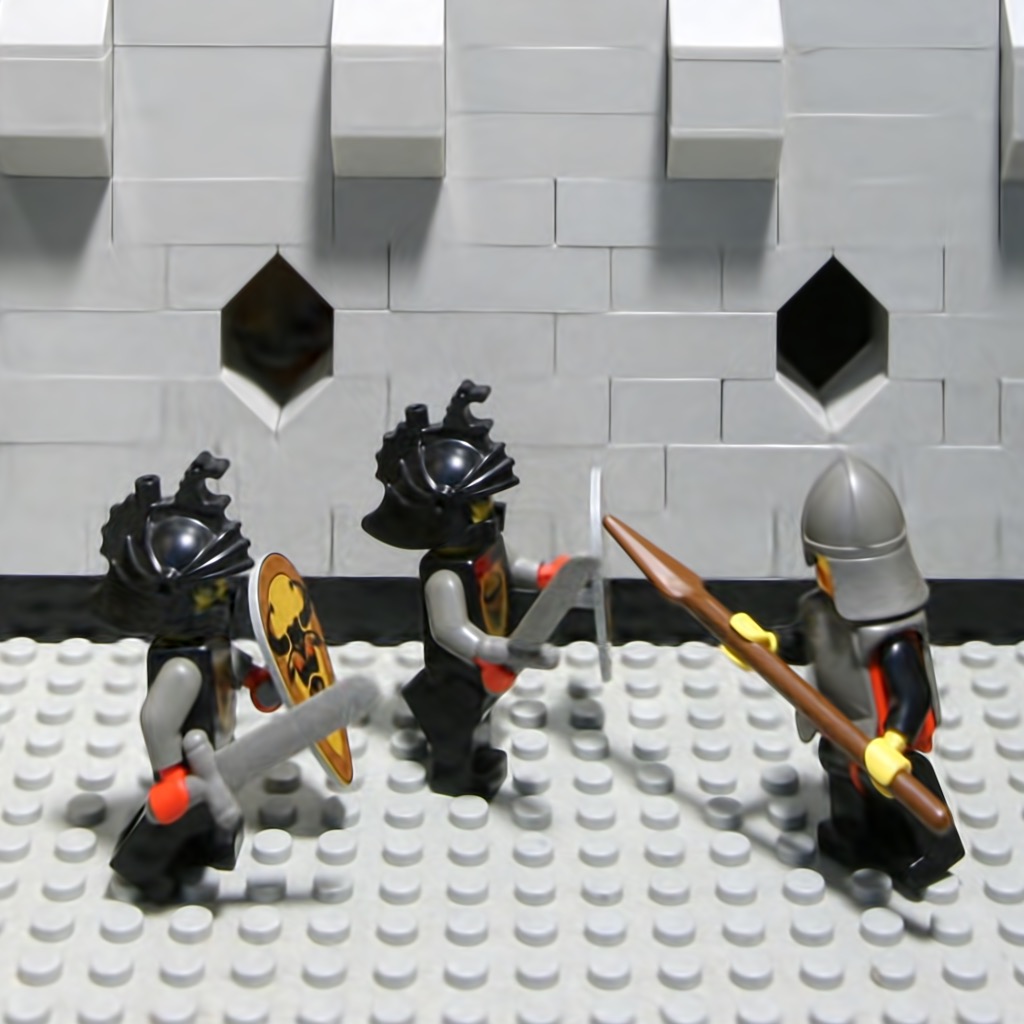}%
    ~
    \FigFiveSubfigC{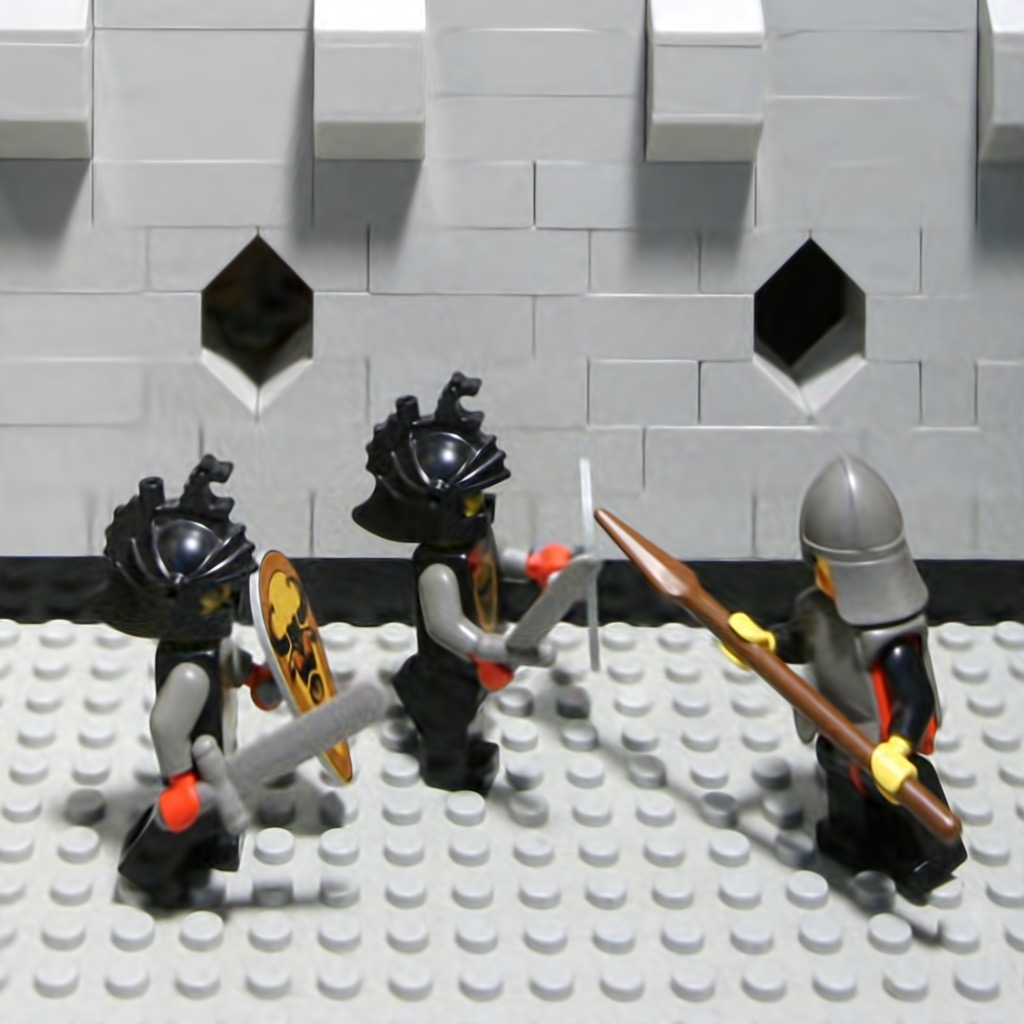}%

    \FigFiveSubfigCaption{$t=0$}%
    ~
    \FigFiveSubfigCaption{$t=0.25$}%
    ~
    \FigFiveSubfigCaption{$t=0.5$}%
    ~
    \FigFiveSubfigCaption{$t=0.75$}%
    ~
    \FigFiveSubfigCaption{$t=1$}%
    $\underbracket[0pt][1mm]{\hspace{15pt}}_%
    {\hspace{-13.0cm}\substack{\vspace{-125pt}\\ {\Large \mybox{\quad \quad}}}}$%
    $\underbracket[0pt][1mm]{\hspace{15pt}}_%
    {\hspace{-13.4cm}\substack{\vspace{-177pt}\\ {\Large \mybox{\quad \quad}}}}$%
    $\underbracket[0pt][1mm]{\hspace{10pt}}_%
    {\hspace{-16.4cm}\substack{\vspace{-285pt}\\ {\large \mybox{\quad \quad}}}}$
    \vspace{-25pt}
	\caption{{\bf Stanford 4D Light Fields Results.} We interpolate learned codes of views $i$ and $j$ as $(1 - t)\cdot z_{i} + t\cdot z_{j}$. The INR preserves the image details from the known views and smoothly transitions between them, such as bright speckles on the ball (row 1) and view-dependent reflections (rows 2 and 3). Images are zoomed in for easier evaluations.}
	\label{fig:4DLFMorph}
\end{figure*}

We adopt the conventional MLP architecture to parameterize $\mathcal{F}$ as a chain of fully connected layers, with activation function usually set as a ReLU or sinusoidal function.
Various embedding functions of the input coordinate $(x, y)$ have been proposed, but in this work we apply no embedding and use sinusoidal activation~\cite{sitzmann2020implicit}, which are sufficient for fitting single 2D images.

The primary training objective of INR $\mathcal{F}$ {for single 2D images} is to minimize the reconstruction error between the predicted $p_{c}$ and ground truth $p_{c}^{GT}$ across all known pixels in a single image, namely
\begin{equation}
L_{{SingleRecon}} = \sum_{p} \| p_{c} - p_{c}^{GT} \|^{2}.
\end{equation}

\subsection{Extension to Multiple Images}
Our goal is to use a single network $\mathcal{F}$ as the INR for multiple images from the same scene.
Prior methods assume the camera layout (for planar light fields~\cite{feng2021signet}) or known camera poses in the pipeline (for general light fields~\cite{sitzmann2021light, attal2021learning}), but we are interested in pushing the limit to where the camera pose of each image is unknown.

In our 3D-agnostic setup which does not consider camera poses, we assign a randomly initialized vector $z \in \mathbb{R}^{M}$ for each image, which serves as its identity code.
We then modify the INR setup so that the operation on each pixel coordinate $(p_{x}, p_{y})$ is now conditional on the code $z$ of length $M$.
In practice, we concatenate $z$ with $(p_{x}, p_{y})$ to form the input vector to the network.
Formally, with $N$ images and $n = 1,..., N$,
\begin{equation}
    \mathcal{F}(p_{x}, p_{y} \mid z_{n}) = p_{c|n},
\end{equation}
{where $p_{c|n}$ stands for the predicted value of pixel $p$ in image $I_n$}.

The training loss function can be easily modified as
\begin{equation}
L_{Recon} =  \sum_{n} \sum_{p} \| p_{c|n} - p_{c|n}^{GT} \|^{2},
\end{equation}
such that the INR $\mathcal{F}$ fits the pixels among all $N$ images. 

While it is easy to optimize $\mathcal{F}$ and $\mathrm{Z}_{N} = \{z_{n}\}_{n=1}^{N}$ to reach a low $L_{Recon}$ across all known pixels, it remains unclear how the learned $\mathcal{F}$ would perform given a novel $z \not \in \mathrm{Z}_{N}$.
Of course, it would be too demanding to expect $\mathcal{F}$ to always produce sensible results for any random $z \in \mathbb{R}^{M}$.
Nonetheless, we believe it is fair to inquire $\mathcal{F}$ under a more relaxed setting:
With $z_{i}, z_{j} \in \mathrm{Z}_{N}$, can $\mathcal{F}$ produce sensible results given a novel $z_{Inter} = \alpha z_{i} + \beta z_{j}$?
In other words, as $\mathcal{F}$ is trained to produce good results with $z_{i}$ and $z_{j}$, what would it produce with a weighted combination of $z_{i}$ and $z_{j}$?

\subsection{Direct Regularization}
The reason we are interested in the above problem is that, if $\mathcal{F}$ would produce good results on weighted combinations of known $z_{i}, z_{j} \in \mathrm{Z}_{N}$, it could produce a smooth transition from image $I_i$ to image $I_j$.
Effectively, it could achieve view interpolation without using any correspondence point or 3D information.
\begin{table}[th!]
\caption{Top: Effect of varying $p$ for code rescaling. Quantitative results reaffirm that rescaling based on $1$-norm achieves the best quality. Bottom: Effect of varying the length of code $M$. Results indicate that the code length cannot be arbitrary, as a small $M$ can be detrimental to the interpolation quality. Setting $M$ too large is also not helpful, as the quality appears to peak at $M=128$. {More details about the experimental setting are in Sec.~\ref{Section:Experiments}.}}
\begin{tabular}{ccccccc}
\toprule
$p$-norm &  ($M$ = 128)  & No & $\infty$ & $2$ & 1.5 & $1$ \\
\midrule
\multirow{2}{*}{Known} & SSIM & 0.901    & 0.903   & 0.951     & 0.955   &  0.962 \\
&  PSNR & 27.49   & 27.57   & 31.63     & 31.92   &  33.38 \\
\multirow{2}{*}{Novel} & SSIM & 0.595    & 0.583   & 0.937     & 0.952   &  0.958 \\
& PSNR & 11.02    & 11.04   & 29.15     & 31.25   &  32.39 \\
\bottomrule
\end{tabular}
\begin{tabular}{cccccccc}
\toprule
$M$  & ($p$ = 1) & 16 & 32 & 64 & 128 & 256 & 512\\
\midrule
\multirow{2}{*}{Known} & SSIM &  0.953  & 0.958  & 0.961   & 0.962   &  0.961  & 0.960 \\
& PSNR & 32.64  & 33.03     & 33.37   & 33.38  & 33.16  & 32.95\\
\multirow{2}{*}{Novel} & SSIM &  0.891  & 0.951  & 0.958   & 0.958   &  0.958  & 0.956 \\
& PSNR & 25.27  & 31.37     & 32.56   & 32.49  & 32.39  & 32.10\\
\bottomrule
\end{tabular}
\label{table:table_1}
\vspace{-20pt}
\end{table}

In this paper, we select the interpolation weights $\alpha, \beta$ with the simple linear interpolation, inducing $\alpha = 1-t, \beta = t$ with $0 \leq t \leq 1$.
Unfortunately, as shown in Fig.~\ref{fig:NormCompare}, linearly interpolating between the two learned codes fails to let $\mathcal{F}$ produce any meaningful result.

The initial failure is not really a surprise.
The codes in $\mathrm{Z}_{N}$ are optimized only towards minimizing $L_{Recon}$.
It would make sense for them to end up with different scales so that $\mathcal{F}$ can better distinguish them and reduce $L_{Recon}$.
Thus, interpolating between them would likely produce a noise vector which $\mathcal{F}$ cannot meaningfully decode.

To address this issue, we directly regularize the $\mathrm{Z}_{N}$ during training.
In particular, we prevent the codes from having different scales by explicitly enforcing the learnable code $z$ as unit $p$-norm.
For our method we select $p = 1$ and enforce
\begin{equation}
z =  \frac{z}{\|z\|_{1}}, \forall z \in \mathrm{Z}_{N}
\vspace{-3pt}
\end{equation}
in addition to training $\mathcal{F}$ and $\mathrm{Z}_{N}$ based on $L_{Recon}$.

Although a more instinctive option to many people is to rescale $z$ based on its Euclidean norm (sum of squares), this Euclidean norm is only a special case for the general $p$-norm when $p=2$, or $\|z\|_{2}$.
We investigate the effect of varying $p$ while computing the norm of $z$, and we present the results in Table~\ref{table:table_1} and in Fig.~~\ref{fig:NormCompare}.
With rescaling based on $1$-norm, the interpolation is more natural and stable than with $2$-norm.
Moreover, as indicated by Table~\ref{table:table_1}, $1$-norm even leads to more accurate reconstruction at the known image views, which is the original task of fitting INR of images.

\subsection{Indirect Regularization}
Although the unit norm constraint significantly improves visual quality, artifacts are observable as shown in Fig.~\ref{fig:CLIPCompare}.
This shortcoming does not come as a shock, because there must be a limit as to how smoothly $\mathcal{F}$ can interpolate, given how little prior knowledge it has about the content.
After all, $\mathcal{F}$ is only trained on a small set of images, and, unlike powerful generative networks, it does not possess domain knowledge (\eg faces, cars) from a huge dataset.
\begin{figure*}[!ht]
    \FigSixSubfigA{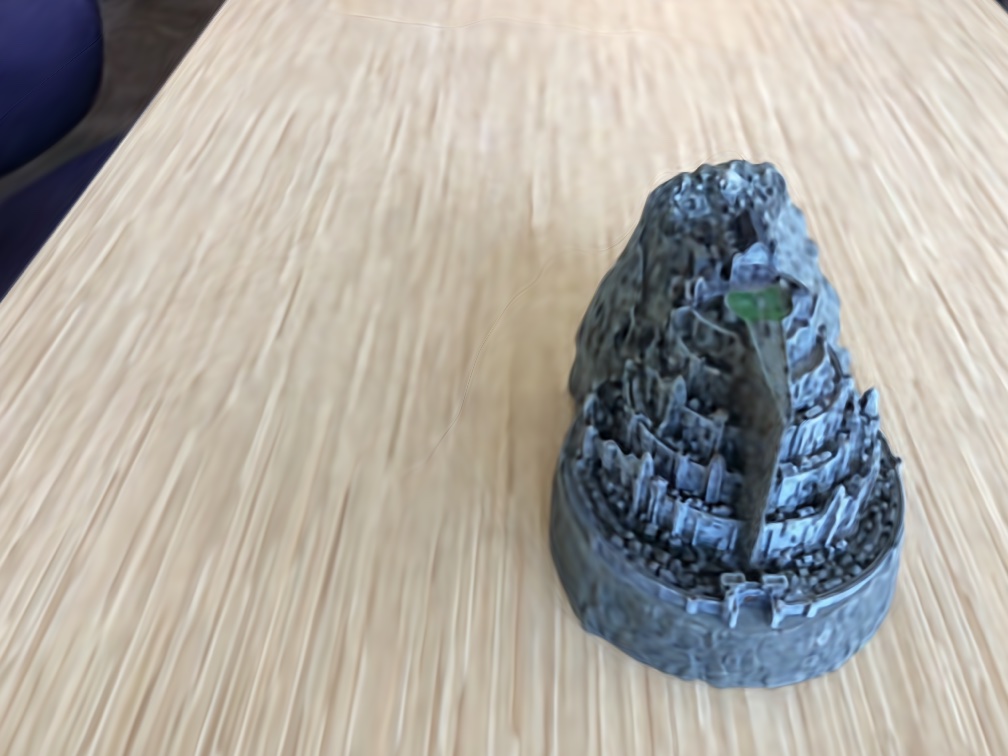}%
    ~
    \FigSixSubfigA{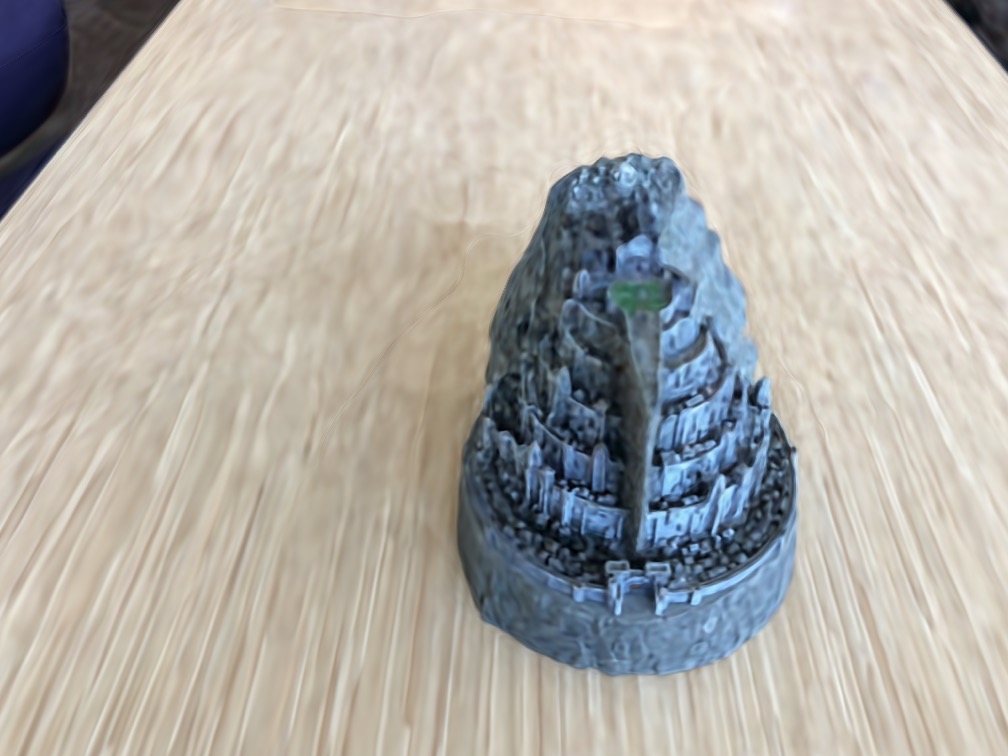}%
    ~
    \FigSixSubfigA{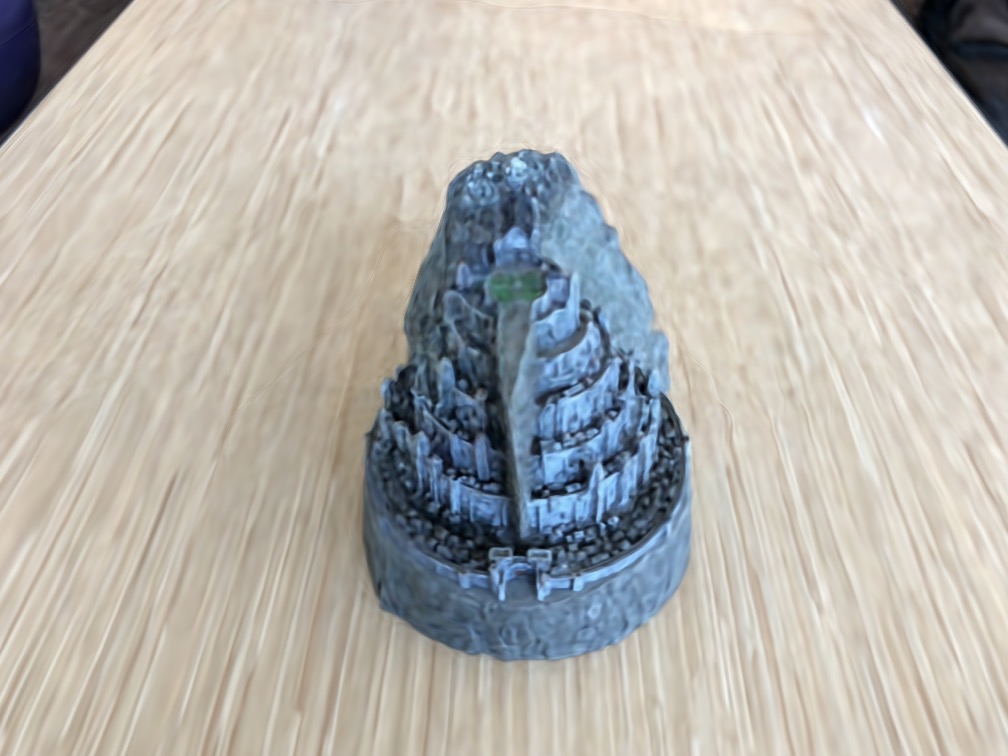}%
    ~
    \FigSixSubfigA{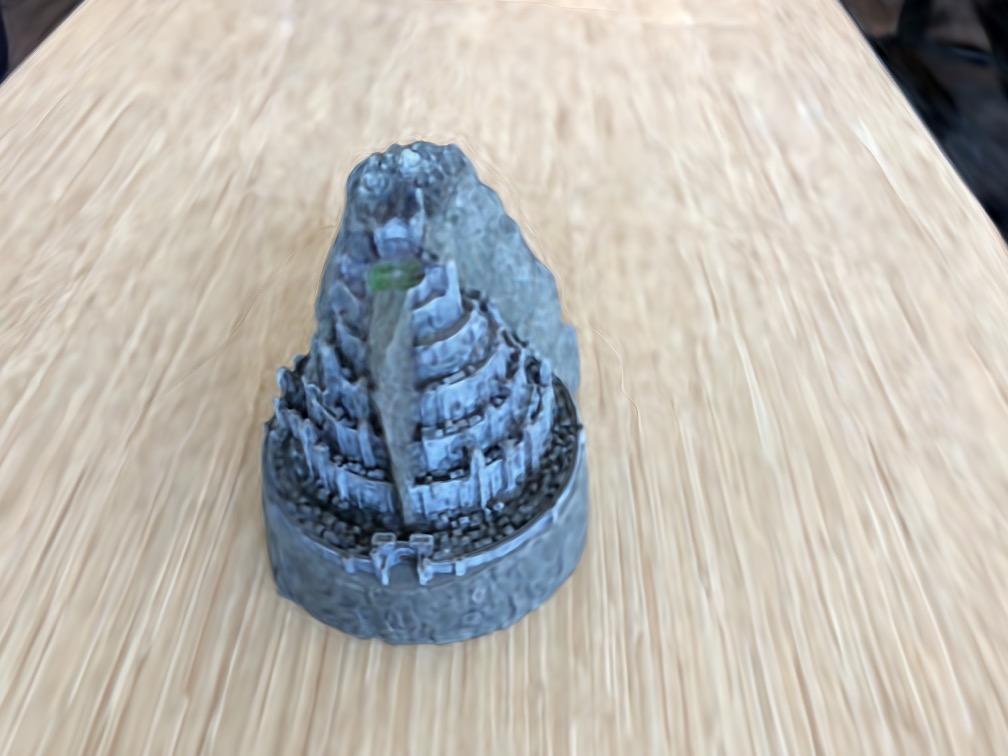}%
    ~
    \FigSixSubfigA{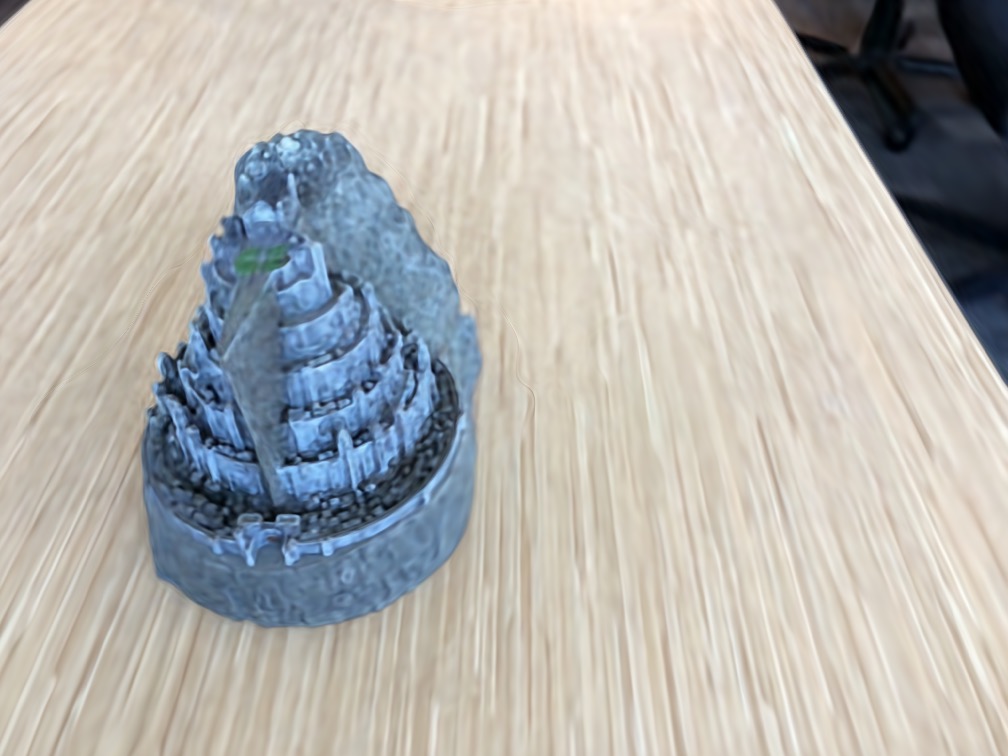}%

    \FigSixSubfigB{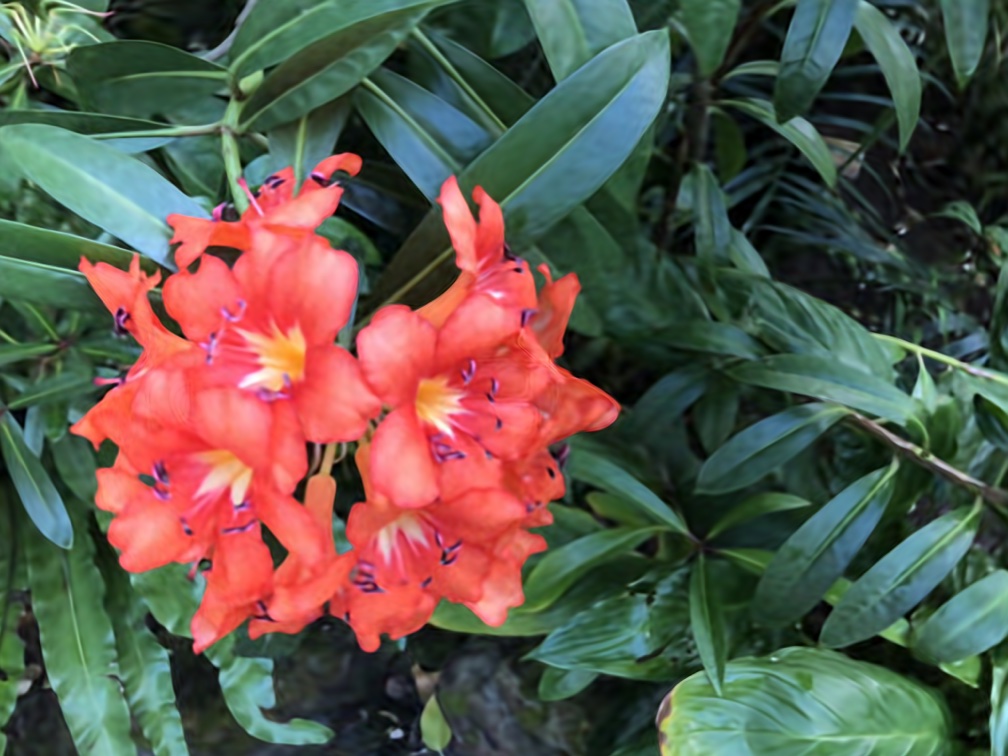}%
    ~
    \FigSixSubfigB{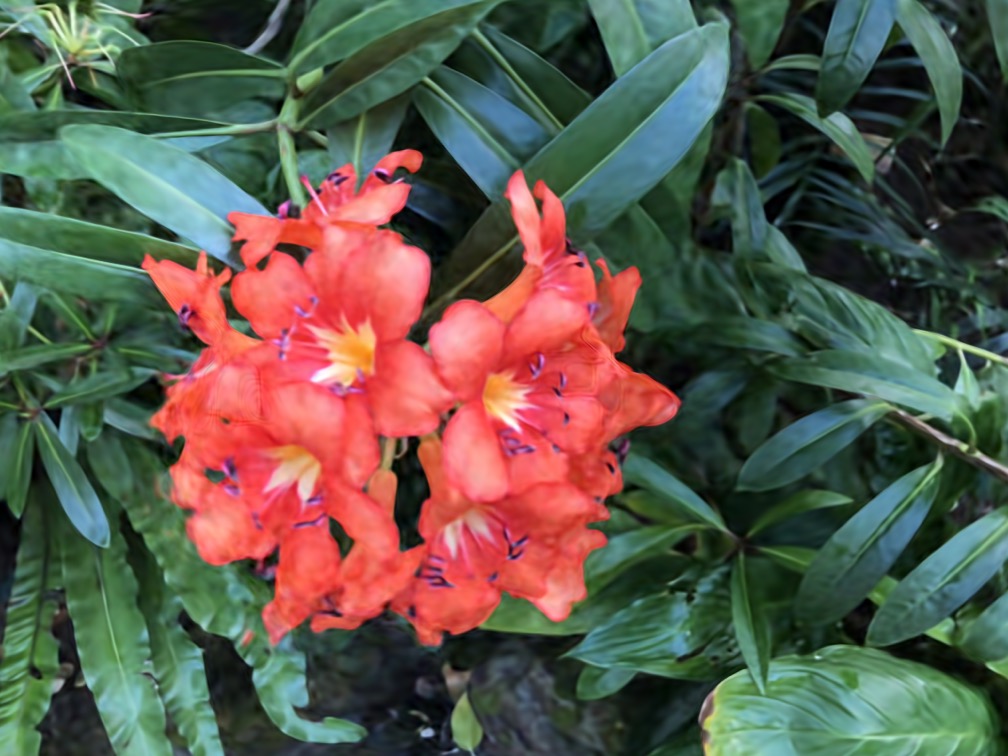}%
    ~
    \FigSixSubfigB{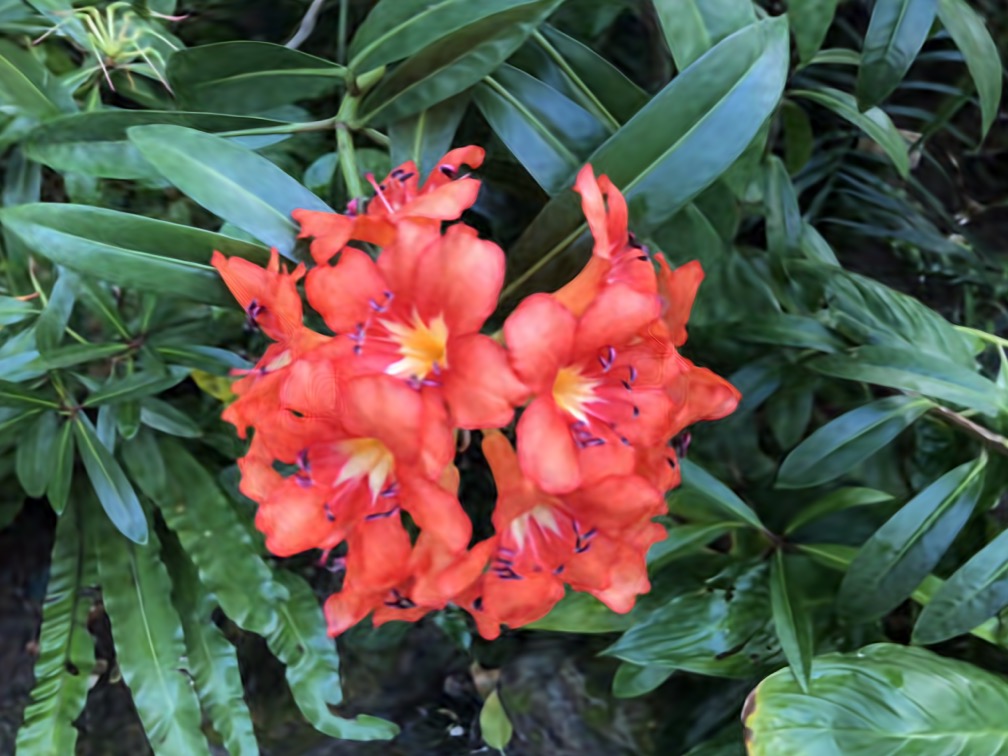}%
    ~
    \FigSixSubfigB{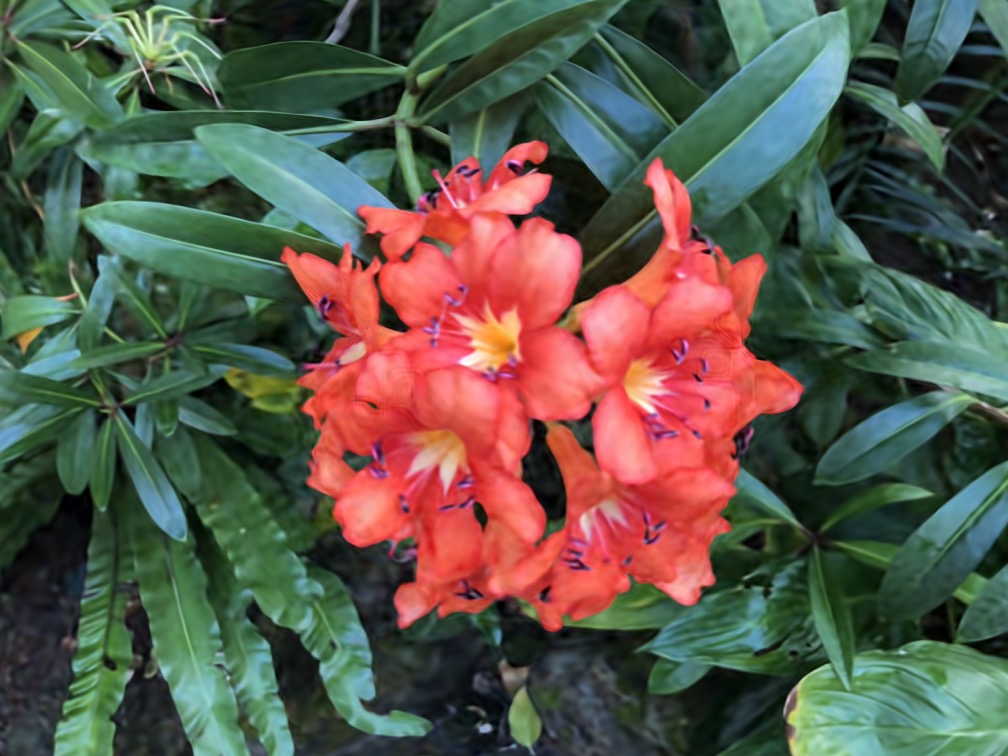}%
    ~
    \FigSixSubfigB{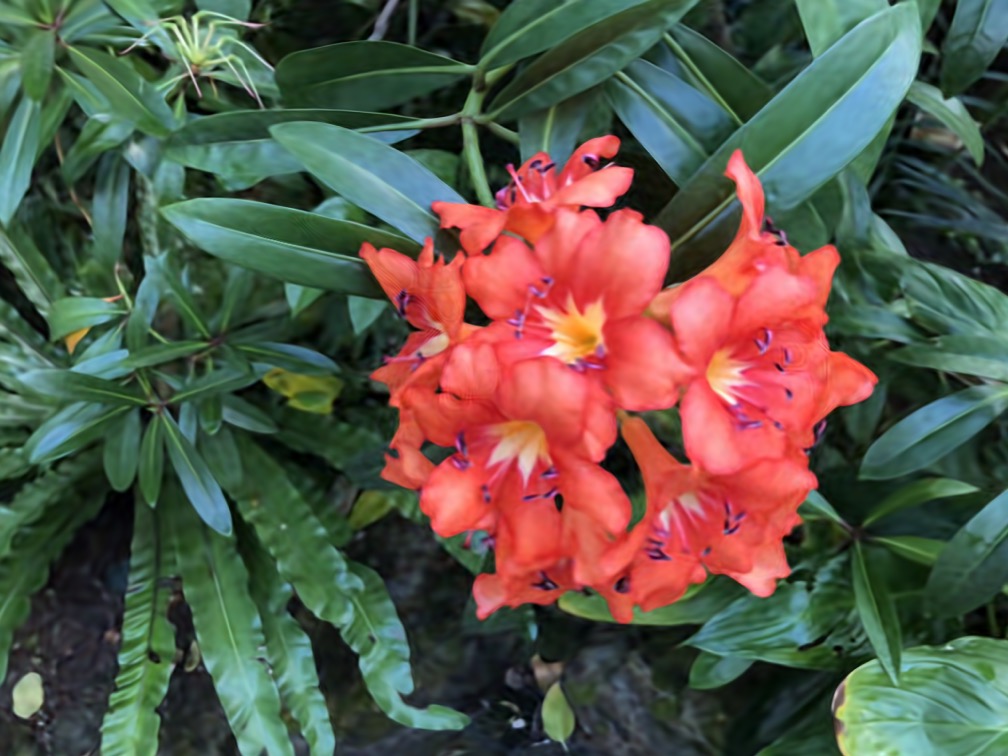}%

    \FigSevenSubfigB{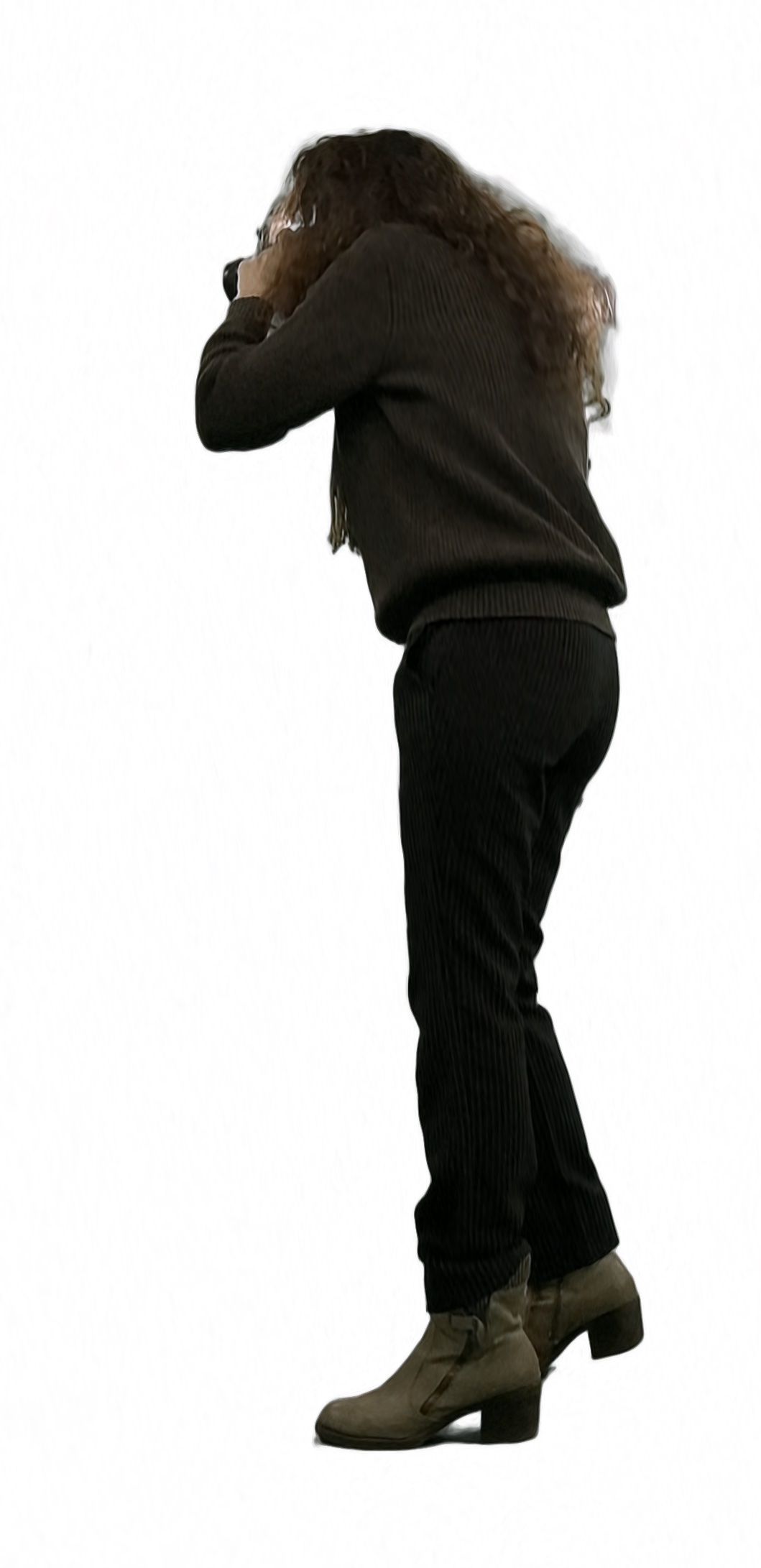}%
    ~
    \FigSevenSubfigB{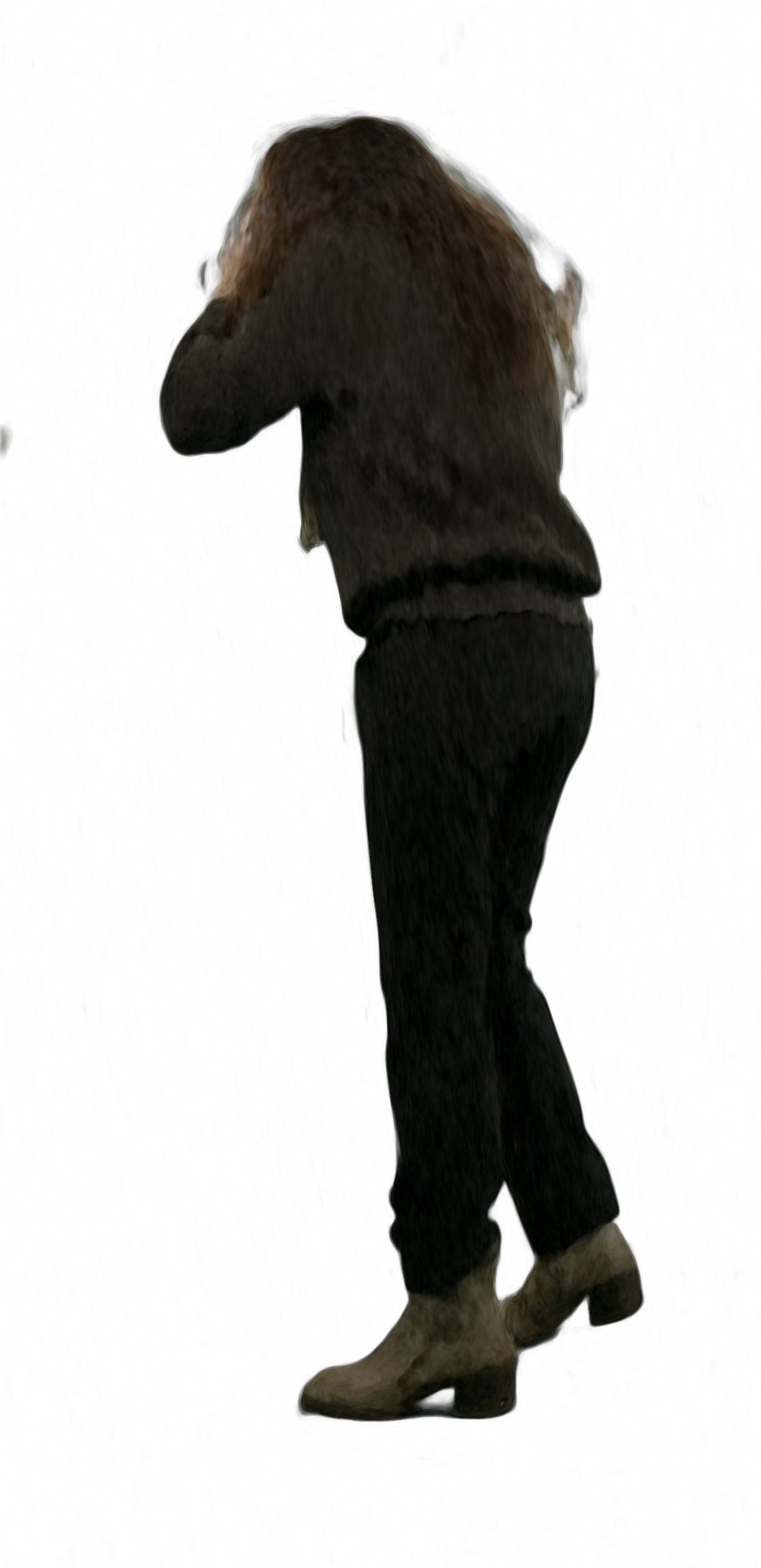}%
    ~
    \FigSevenSubfigB{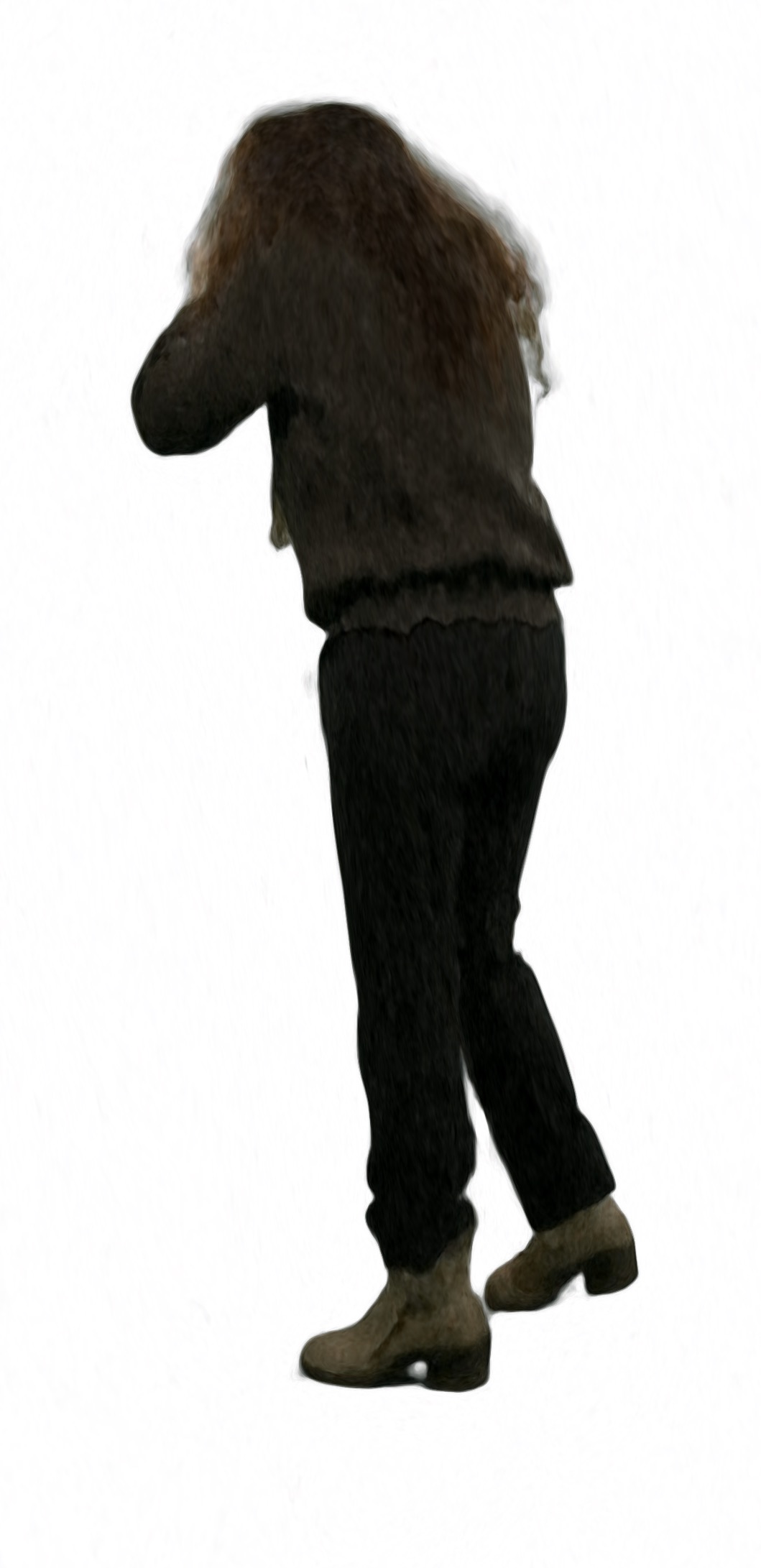}%
    ~
    \FigSevenSubfigB{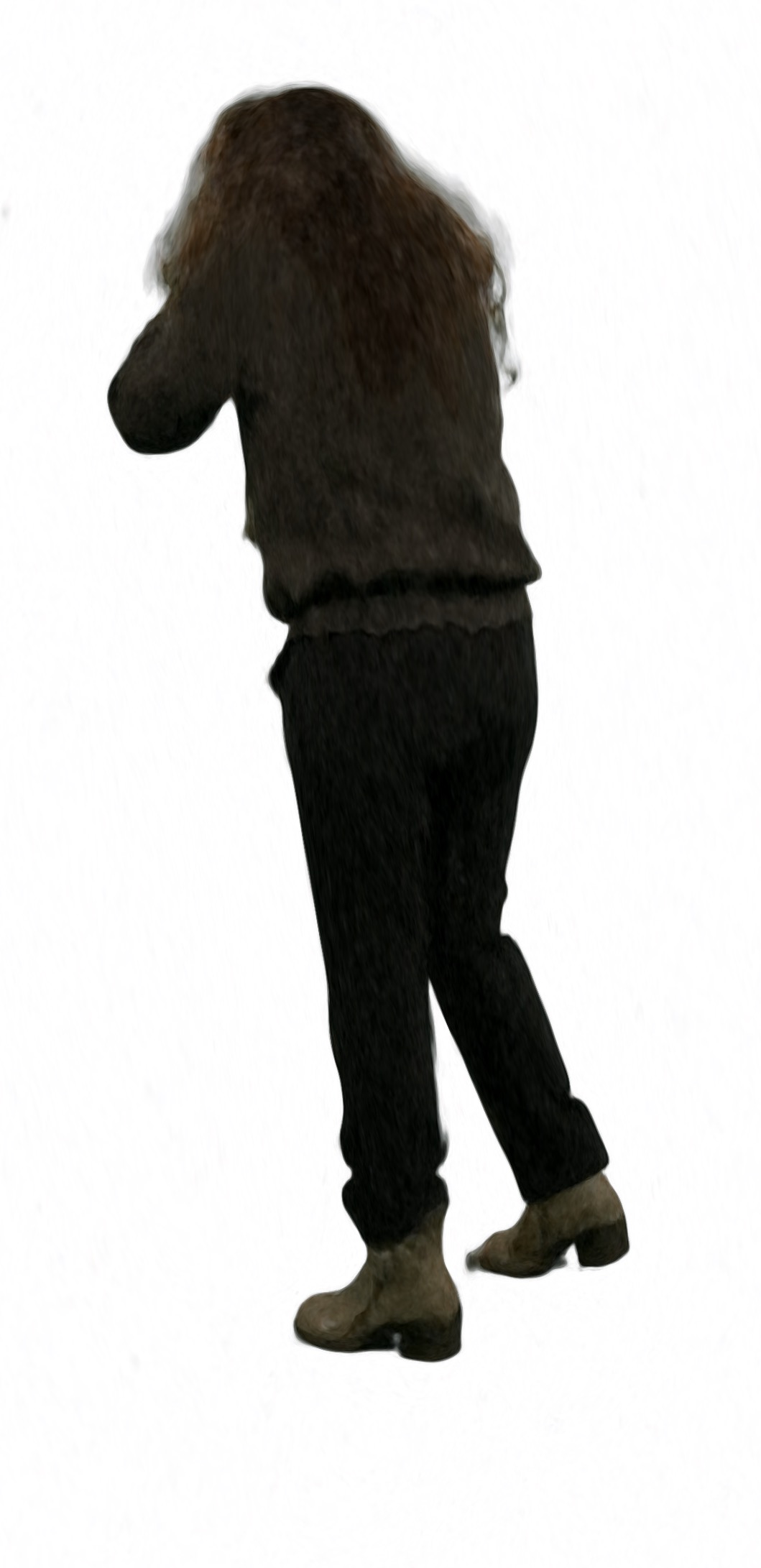}%
    ~
    \FigSevenSubfigB{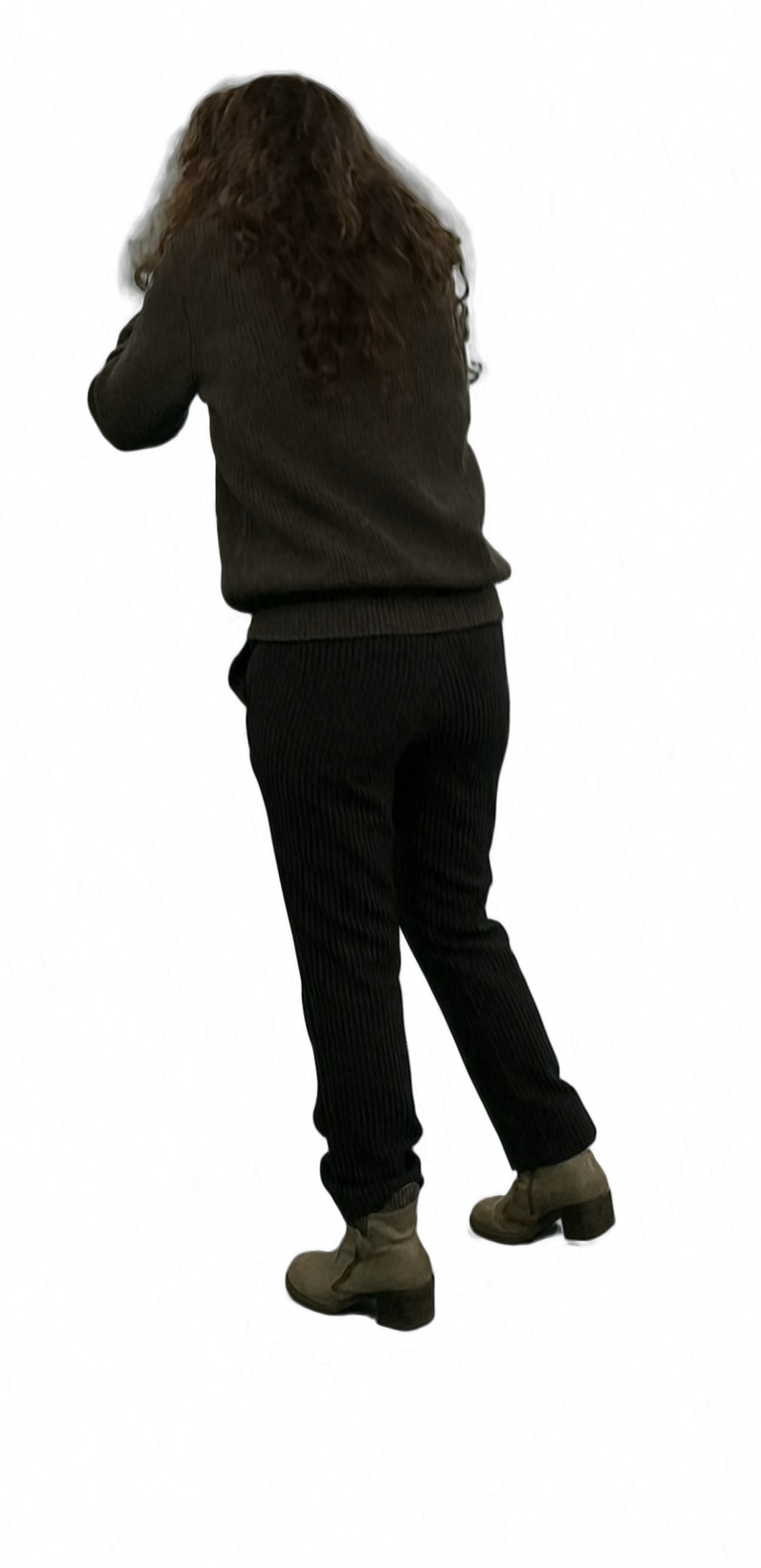}%

    \FigSevenSubfigA{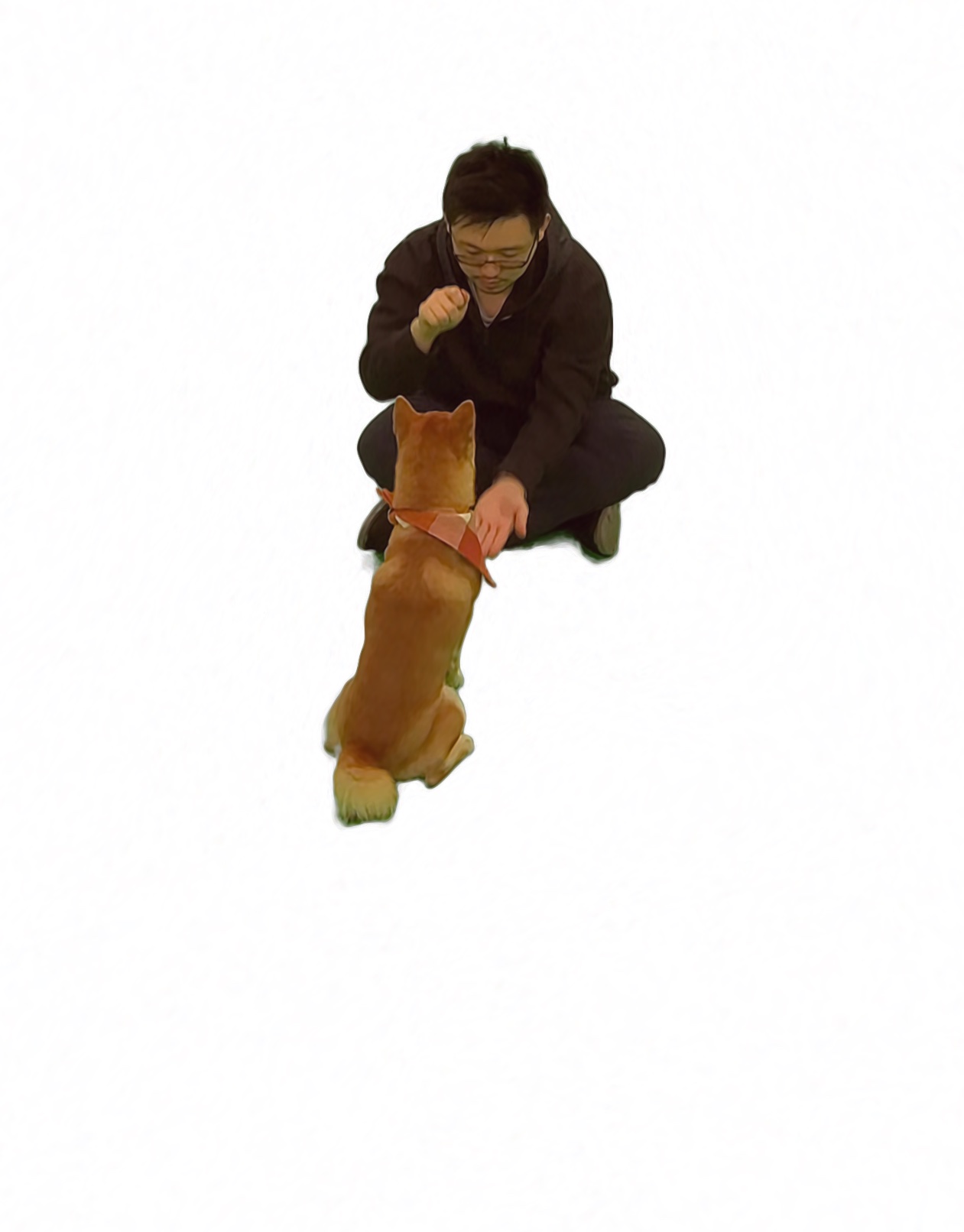}%
    ~
    \FigSevenSubfigA{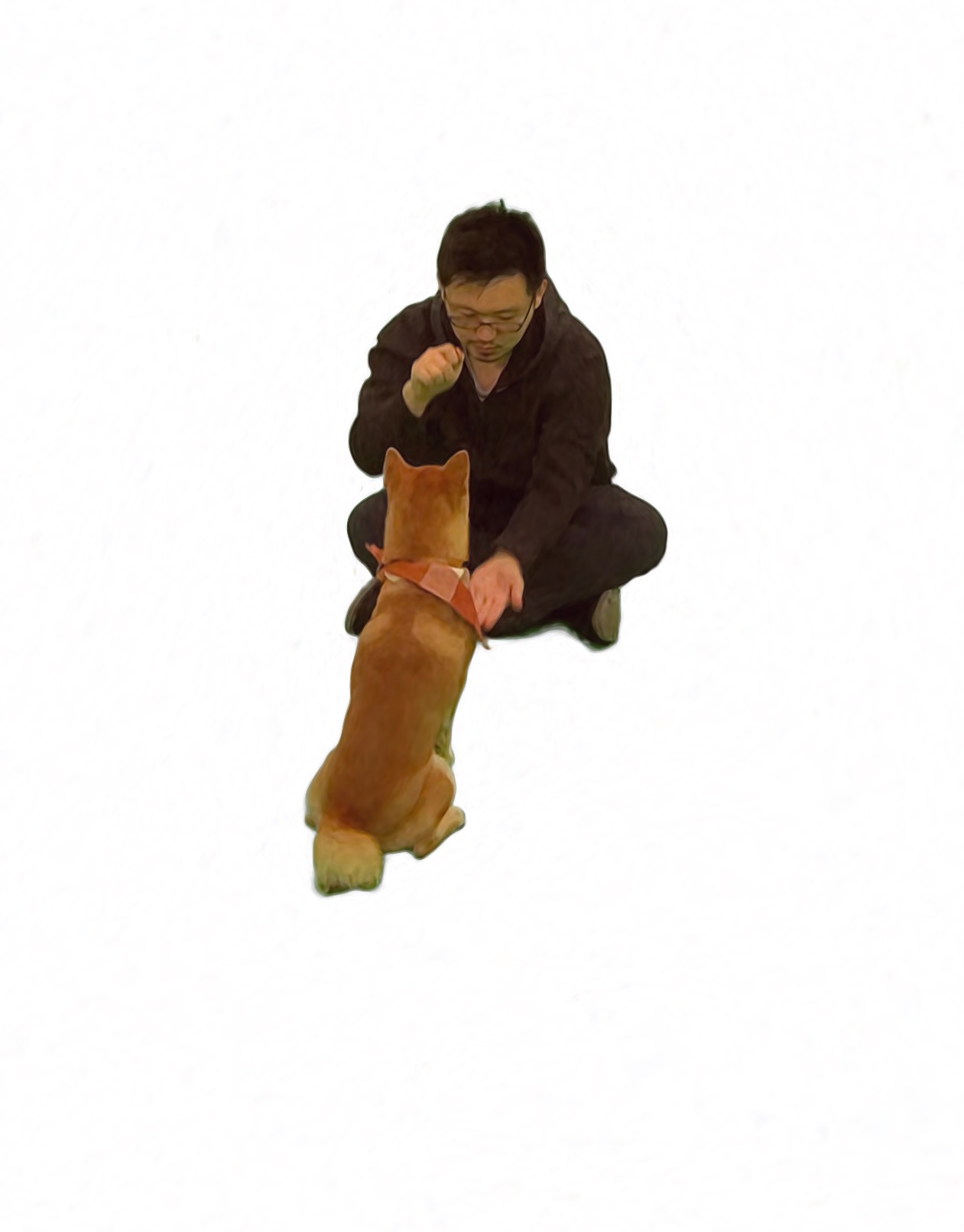}%
    ~
    \FigSevenSubfigA{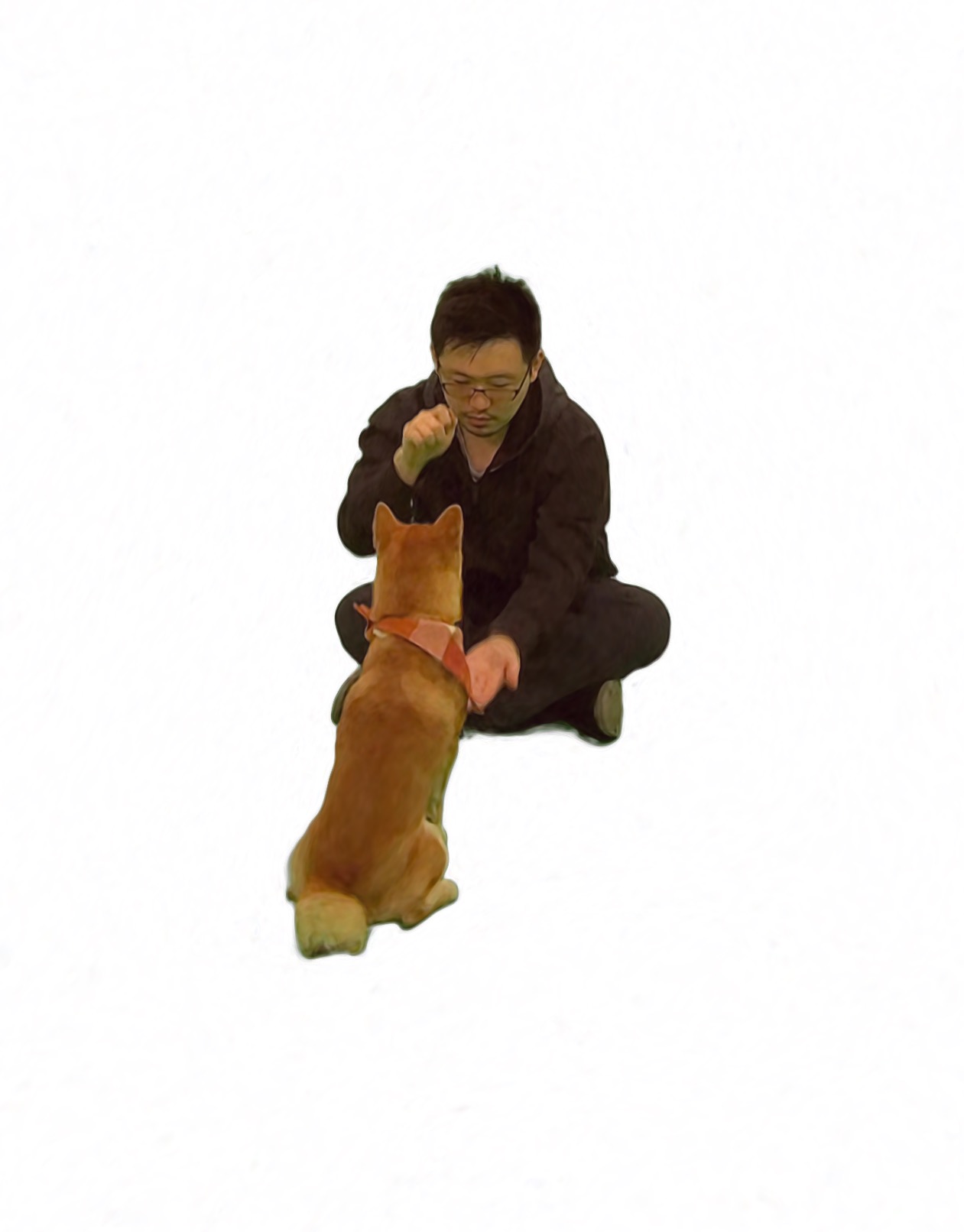}%
    ~
    \FigSevenSubfigA{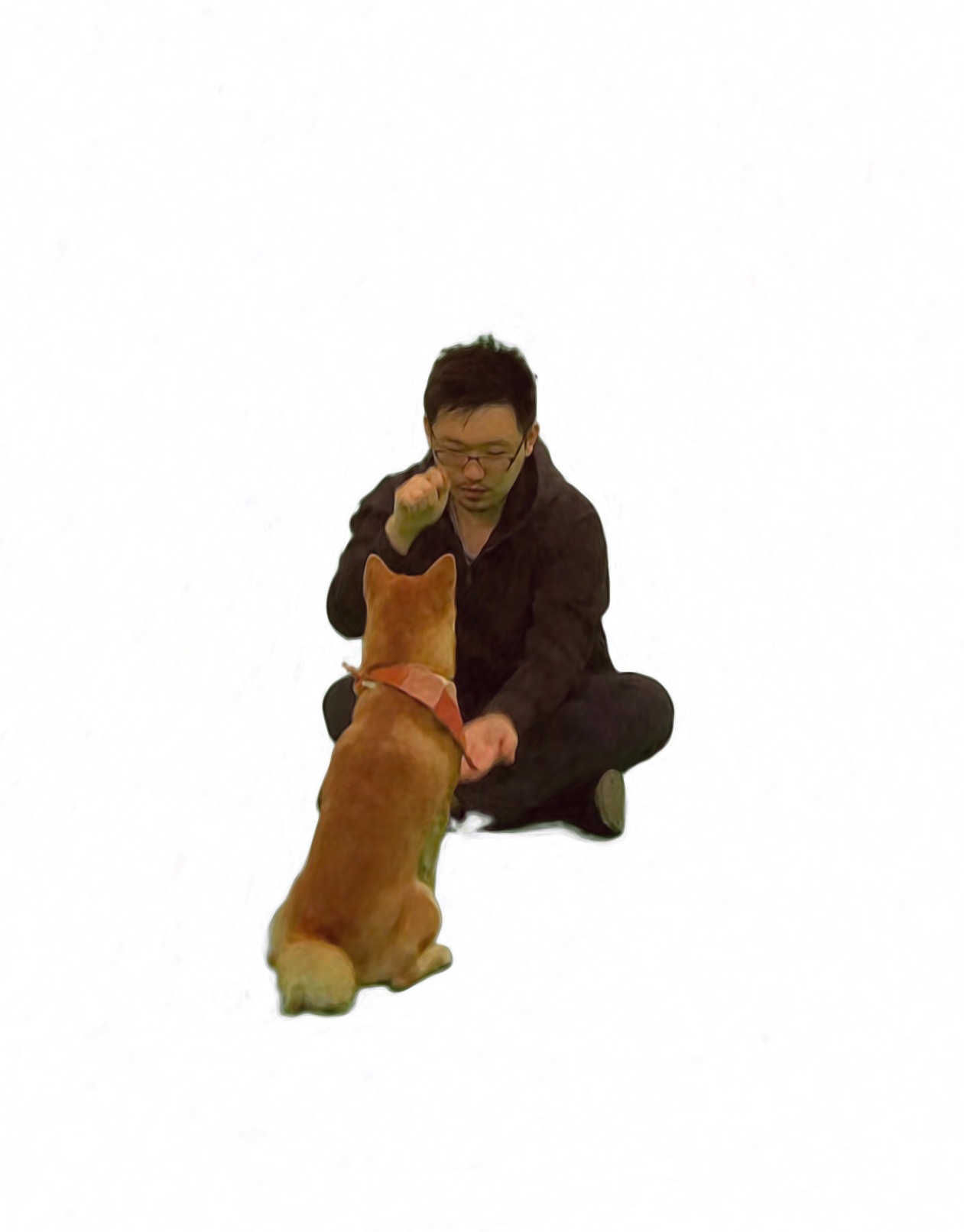}%
    ~
    \FigSevenSubfigA{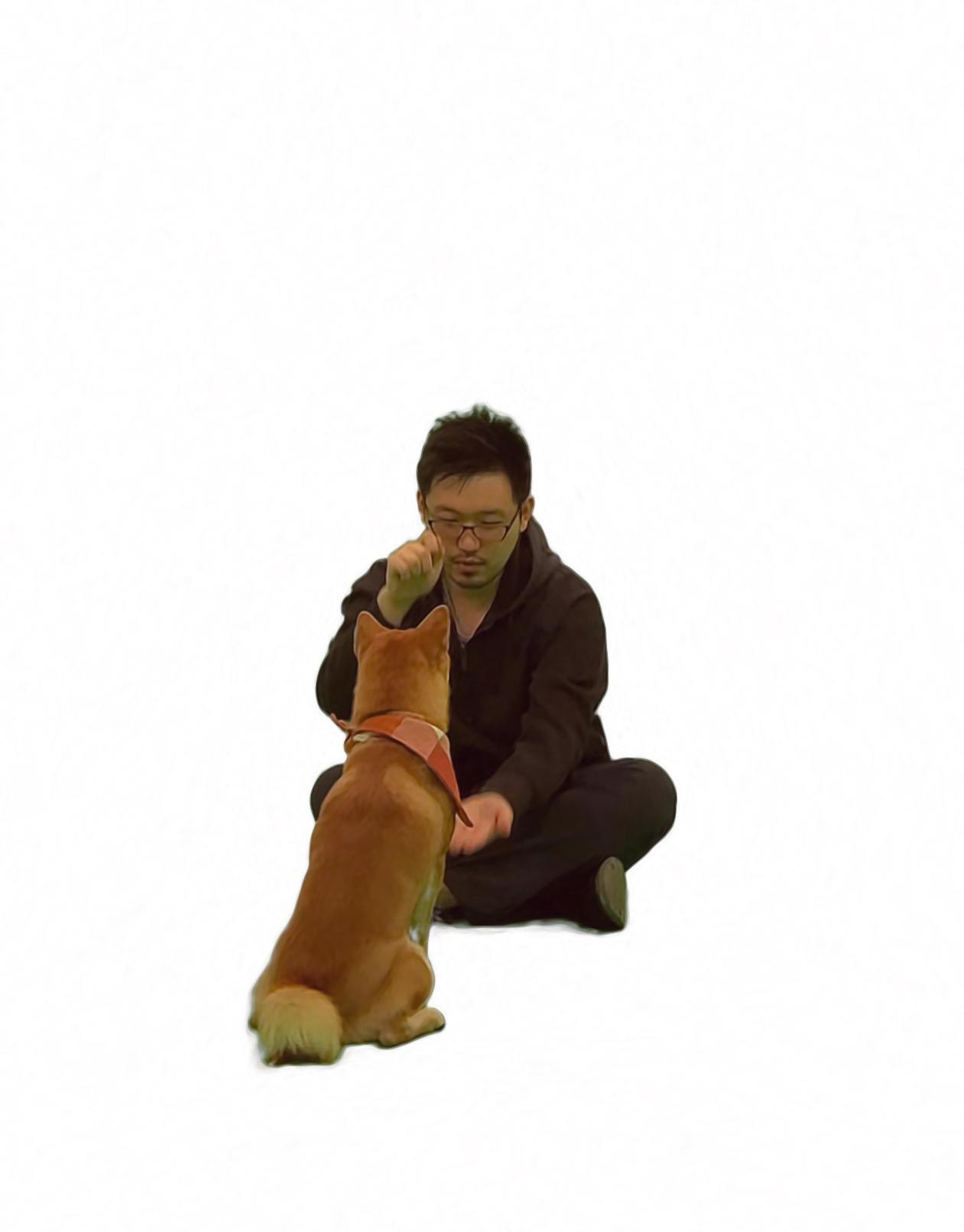}%

    \FigSixSubfigCaption{$t=0$}%
    ~
    \FigSixSubfigCaption{$t=0.25$}%
    ~
    \FigSixSubfigCaption{$t=0.5$}%
    ~
    \FigSixSubfigCaption{$t=0.75$}%
    ~
    \FigSixSubfigCaption{$t=1$}%
	\vspace{-5pt}

	\caption{{\bf Unstructured Light Field Results.} We interpolate learned codes of views $i$ and $j$ as $(1 - t)\cdot z_{i} + t\cdot z_{j}$. The camera movement between the two known views includes rotation and translation. The interpolation through INR smoothly transforms the perspective, despite having no knowledge of 3D scene structure or camera pose. Images are zoomed in for easier evaluations.}
	\label{fig:LLFFMorph}
\end{figure*}

However, there is still room for improvement if we more proactively alter the training process.
Since the fundamental issue is that $\mathcal{F}$ produces poor results when decoding novel interpolated $z_{Inter}$ unseen during training, we should explicitly encourage good results with interpolated codes during training.
As we only have access to those $N$ images and do not have ground truth interpolated frames, we propose computing the loss of interpolated results through an off-the-shelf pre-trained network $\mathcal{E}$.
Specifically, we hope the features extracted from the interpolated output are similar to the features extracted from the two source images $I_i$ and $I_j$.

Formally, with $z_{Inter} = (1 - t) \cdot z_{i} + t \cdot z_{j}$, $\triangle$ denoting all pixels in a full image frame, the interpolated output image is $I_{Inter} = \mathcal{F}(\triangle | z_{Inter})$.
We then use feature extractor $\mathcal{E}$ to compute
\begin{equation}
L_{Inter} = \| \mathcal{E}(I_{Inter}) - [(1 - t) \cdot \mathcal{E}(I_i) + t \cdot \mathcal{E}(I_j)]\|^{2},
\end{equation}
and train the INR towards minimizing this loss.

While this setup is similar to the VGGNet-based perceptual loss widely used in tasks like style transfer and image reconstruction, we find that using VGGNet as the feature extractor $\mathcal{E}$ is not adequate for our problem, as shown in Fig.~\ref{fig:CLIPCompare}.
Instead, we employ the recently released CLIP network~\cite{radford2021learning} as $\mathcal{E}$.
As shown by Fig.~\ref{fig:CLIPCompare}, we discover that the CLIP-based extractor significantly outperforms VGGNet, likely because CLIP benefits from being trained to extract semantically-consistent features from images (see~\cite{radford2021learning} for more details), whereas VGGNet is trained to extract features mainly for image classification.
Our results are analogous to previous findings~\cite{jain2021putting} that show CLIP improves NeRF training on sparse views.

In short, in addition to the original objective of minimizing $L_{Recon}$, we introduce rescaling with unit norm and the CLIP-guided interpolation loss to the process of training INR of images.
To compute $L_{Inter}$, the interpolation endpoints $z_{i}$ and $z_{j}$ are randomly selected from $\mathrm{Z}_{N}$.
We refrain from specifically sampling neighboring or adjacent viewpoints to avoid using the camera pose information and keep training as 3D-agnostic.
We would only knowingly select endpoints based on their viewpoint locations during evaluation or demonstration of the interpolation results between different viewpoints, after training is finished.
\section{Experiments} \label{Section:Experiments}
In this section, we provide more results on view interpolation and ablation studies on the techniques introduced in Section~\ref{Section:Method}.
We train VIINTER to encode real-world scenes captured under two different regimes: 4D light fields (viewpoints are on a 2D plane with the same orientation) and unstructured light fields (viewpoints are not aligned on a 2D grid and orientations might be rotated).

\paragraph{4D Planar Light Fields.}
We use scenes from Stanford Light Field Archive~\cite{StanfordLFDataset}, with $17 \times 17$ camera viewpoints on a 2D grid.
We use a $5 \times 5$ subset by taking every $4$-th image horizontally and vertically.
We render new views by selecting two trained codes and linearly interpolate them.
The resulting interpolation results are shown in Fig.~\ref{fig:4DLFMorph}, with more in the supplements.

\paragraph{Unstructured Light Fields.}
To test VIINTER on scenes with irregular camera layout, we test on the LLFF dataset~\cite{mildenhall2019local} and our own volumetric dataset.
The LLFF scenes are captured in natural indoor environments, while our own scenes come from a volumetric studio for human body captures.
We present the interpolation results in Fig.~\ref{fig:4DLFMorph}, with more in the supplements.
\begin{table}[!ht]
\caption{Quantitative results on real-world scenes with different viewpoint layouts. Our method can only render at the approximate viewpoints of ground truth, as discussed in Section~\ref{Section:Experiments}, leading to lower PSNR and SSIM values for novel views of ``Unstructured'' where the viewpoint mismatch is severe. See visual results for more comprehensive quality assessments.}
\begin{tabular}{lcccc|ccc}
\toprule
& &  \multicolumn{3}{c|}{4D Planar} &  \multicolumn{3}{c}{Unstructured} \\
\midrule
& Method & NeRF & LFN & Ours & NeRF & LFN & Ours\\
\midrule
\multirow{2}{*}{\small Known} & SSIM & 0.926 & 0.977 & 0.978 & 0.911 & 0.920 & 0.885 \\
& PSNR & 33.62 & 37.67 & 37.28 & 29.04 & 30.11 & 28.32 \\
\midrule
\multirow{2}{*}{\small $\text{Novel}^*$} & SSIM & 0.917  & 0.944  &  0.975  &  0.905 & 0.788 & 0.664 \\
& PSNR & 33.28 & 30.67 & 35.77  & 27.15  & 21.35  & 16.80  \\
\bottomrule
\end{tabular}
\label{table:table_2}
\vspace{-20pt}
\end{table}

\paragraph{Quantitative Evaluation.}
The unique challenge in evaluating our method is we cannot explicitly specify a camera pose to render at.
Nonetheless, to provide a quantitative evaluation, we approximately render at testing viewpoints by interpolating the codes from nearby known viewpoints.
For example, for the Stanford Light Field scenes, we select two viewpoints in the $5 \times 5$ training set, viewpoints $(4, 4)$ and $(4, 8)$, and interpolate their learned codes with $t = 0.5$.
Then we render the full image with the interpolated code and compare it against the actual test image (withheld from training) captured at viewpoint $(4, 6)$.
Thanks to the well-aligned structure of these 4D scenes, we can compute metrics like peak signal-to-noise ratio (PSNR) and structural similarity index measure (SSIM) against a reasonable ground truth image.

Table~\ref{table:table_1} provides the quantitative impact of our proposed techniques on known and novel viewpoints in the \textit{Lego} scene.
In Table~\ref{table:table_2} we present the evaluation results aggregated from all scenes.
Although our method is not meant to outperform methods which use 3D information, we still provide quantitative comparisons with two recent methods: NeRF~\cite{Mildenhall2020NeRF}, the most prevalent INR method for view synthesis, and LFN~\cite{sitzmann2021light}, which uses camera pose information to train an INR of 5D rays.
For results at novel viewpoints, although we can reasonably approximate the viewpoint in the \textit{4D Light Field} scenes, the approximation is very inaccurate in the \textit{Unstructured} scenes due to sparse and irregular camera layouts.
For more comprehensive assessment of the quality, please refer to the supplements.
\section{Discussion}
 \label{Section:Discussion}
In this section, we provide further discussion on the significance of the proposed method and presented results.

\paragraph{Why not NeRF (or 3D approach)?}
In general, image-based methods avoid certain issues unique to 3D approaches (\eg properly setting 3D bounding box, \# of samples per ray) that often complicate the training.
However, this paper is not intended to present a better method than the state of the art in view interpolation and synthesis, but rather to explore a new direction where a classic image manipulation problem meets the modern implicit neural representation.
We believe that 3D approaches like NeRF are currently still more appropriate in production, due to the abundance of techniques and optimizations developed to improve their performance.
For sake of transparency and thoroughness, we provide comparisons in Section~\ref{Section:Experiments} with representative methods and datasets, and we hope they help readers better contextualize this new and untested approach.

\paragraph{How is this different than image morphing?}
Although many image morphing techniques do not invoke explicit 3D knowledge about the structure or viewpoints, they rely on finding correspondence points between the images being interpolated.
Our proposed method is correspondence-free, and the interpolation happens in the space of the $z$ codes, rather than the space of image pixels.
Moreover, image morphing often applies to images of different scenes or identities (\eg face morphing between two people), but this paper is concerned with multi-view images from the same scene.
In our setting, if we do find correspondences like most morphing methods, we would essentially do keypoint matching.

\paragraph{How is this different than GAN interpolation?}
The code interpolation is seemingly similar to the latent space interpolation of GANs, but our method fundamentally differs from GANs in three ways.
First, VIINTER does not train a discriminator that provides adversarial guidance to a generator.
Second, GANs usually model a continuous Gaussian latent space while we only consider the pairwise interpolation between codes in $\mathrm{Z}_{N}$.
Third, GANs are trained for domain-specific data and are unlikely to work for out-of-distribution data.
For example, although we could project face images into the latent space of a powerful GAN trained on aligned human faces, but we cannot use it to interpolate the various categories of images used in Section~\ref{Section:Experiments}.
Our method is applicable for each separate scene, and the CLIP-based feature extraction is shown to generalize well both in prior work~\cite{jain2021putting} and our experiments with scenes containing vastly different visual attributes.

\begin{figure}[!t]
    \centering
    \FigLimitationSubfig{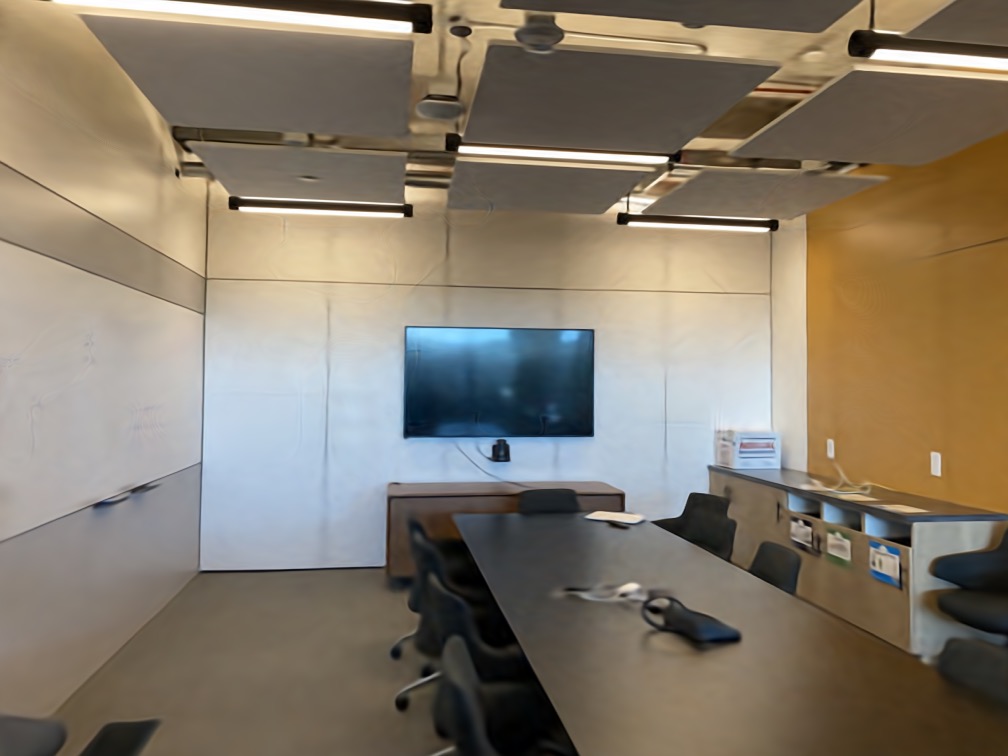}{$t=0$}%
    ~
    \FigLimitationSubfig{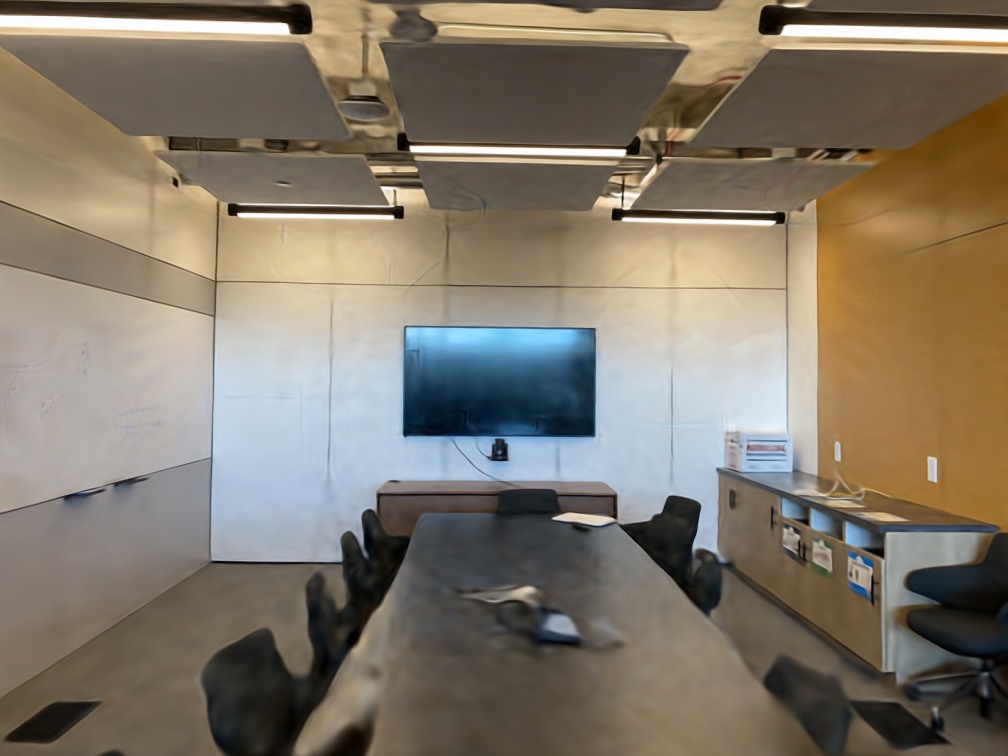}{$t=0.5$}%
    ~
    \FigLimitationSubfig{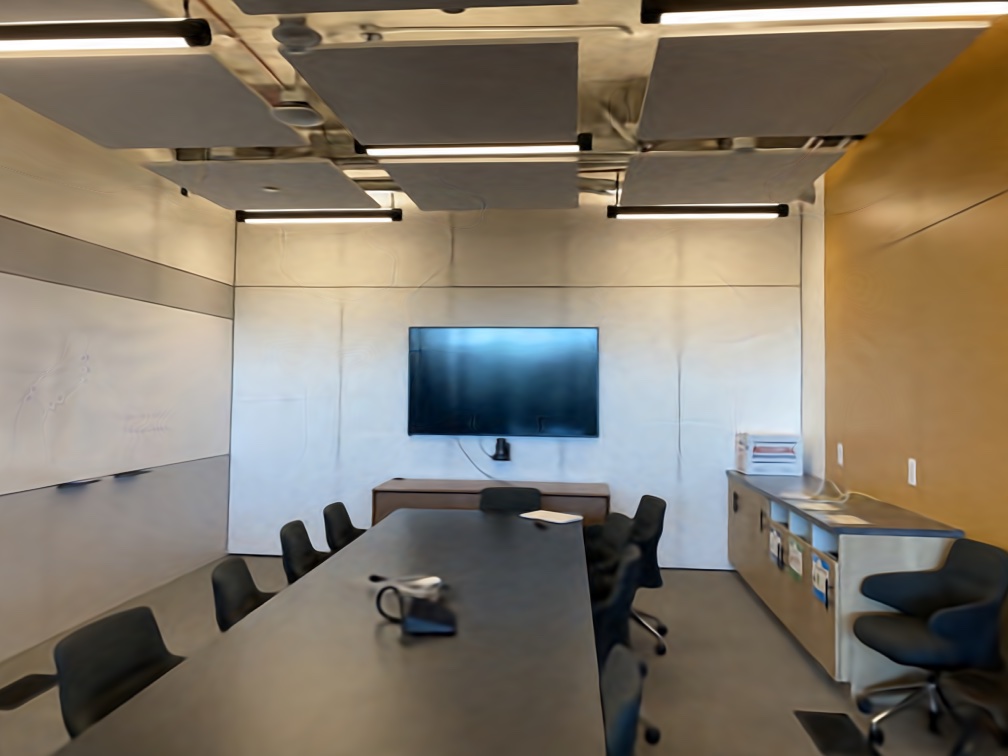}{$t=1$}%
    $\underbracket[0pt][1mm]{\hspace{65pt}}_%
    {\hspace{-15pt}\substack{\vspace{-70pt}\\ {\tiny \mybox{\quad \quad}}}}$
    \vspace{-30pt}
    \caption{{\bf Limitations.} When the disparity is too large due to insufficient viewpoint density, our method might not produce plausible interpolation, likely because the training views do not provide the INR with enough information to perform implicit 2D interpolation without 3D knowledge.}
    \vspace{-10pt}
    \label{fig:Limitation}
\end{figure}

\paragraph{What is the implication of the norm of $z$?}
We point out that our strategy is not adding a penalty on the norm of $z$, but rather strictly enforcing the norm of $z = 1$.
Controlling the norm of $z$ ensures that the learned codes exists on a well-defined region in $\mathbb{R}^{M}$.
In the case of $M = 2$, $2$-norm ensures all 2D points lie on a circle, whereas $1$-norm ensures all 2D points lie on a square inside that circle.
As we increase $M$ to higher dimensions, the difference between $1$-norm and $2$-norm intensifies as the gap between that ``circle'' and ``square'' enlarges.
As a result, the codes learned with $1$-norm is more compact and likely more conducive for interpolation.

\paragraph{What are the limitations?}
The method may produce obvious artifacts when interpolating scenes with large disparity (Fig.~\ref{fig:Limitation}).
{Additional experiments on more varied data would be necessary for a comprehensive assessment of its performance on challenging data }
Another limitation is it has no sense of 3D locations or camera pose.
As a result, it can only interpolate known viewpoints and cannot render at arbitrary locations.
Nonetheless, this limitation would not be a deal breaker for many practical use cases that only interpolate between known viewpoints, like 4D light field rendering where viewpoints are fixed on a 2D plane.
Another notable use case is event replay for TV viewers, which produces a fly-through effect by interpolating between known cameras.
Finally, the proposed method is limited by the training speed of INR, and training the basic INR implemented in this work (with loss from feature extractor) takes a few hours.
We believe future work can significantly alleviate this limitation by incorporating recent techniques~\cite{martel2021acorn, muller2022instant} to speed up INR training.
\section{Conclusion}
 \label{Section:Conclusion}
Images have been an indispensable data primitive in graphics and vision.
Exciting recent developments are advancing INR of images towards two goals: image fitting and view synthesis.
{Instead of pushing further ahead along either direction with increasingly specialized techniques, we look sideways} to explore a new possibility of fusing those two directions together.
{Results from our study show that with careful modifications, INRs can perform view interpolation through code interpolation in appropriate scenarios}.
Although the method is limited by its inherent lack of 3D knowledge, our study presents a proof of concept revealing an unrealized potential of INR of images.
Our success in adapting CLIP, a pre-trained deep network, to guide the INR training also suggests that future developments of INR could further benefit from absorbing concurrent progresses in other areas of deep learning.
As INRs of images evolve to be more accurate and efficient, with this paper, we offer a promising outlook on employing INRs for image manipulation tasks beyond fitting and super-resolving known images.

\begin{acks}
We sincerely thank the anonymous reviewers for their valuable suggestions to improve the paper.
We thank Jonathan Heagerty, Sida Li, Eric Lee, Barbara Brawn, and Maria Herd for developing our light field datasets.
This work has been supported in part by the NSF Grants 18-23321 and 21-37229, and the State of Maryland’s MPower initiative. 
Any opinions, findings, conclusions, or recommendations expressed in this article are those of the authors and do not necessarily reflect the views of the research sponsors.
\end{acks}

\bibliographystyle{ACM-Reference-Format}
\bibliography{main}

\clearpage
\newpage
\section{Supplementary Information}
\subsection{Training Details.}
\subsubsection{Hyperparameters.}
We use SIREN~\cite{sitzmann2020implicit} as the network backbone and randomly initialize the weights and the codes $z$ by default PyTorch settings.
In the main experiments, we set length of $z$ as $M=128$.
All networks have eight intermediate layers with dimension of $512$, except the three volumetric scenes where the hidden dimension of $256$, {since we found that increasing the dimension to $512$ did not meaningfully improve the quality}.
We optimize the all parameters (network weights and image code $z$) with the Adam optimizer, with a learning rate of $1^{-5}$ which decays to $1^{-6}$ with the cosine annealing schedule.
At each training iteration, the pixel batch size is $8192$.
For the 4D Planar scenes, the network is trained for $300,000$ iterations, while it is trained for $200,000$ iterations for the other scenes, {since we did not find quality improvements even if we trained the network for more iterations}.
For the LLFF scenes, we $\alpha = 0.05$ except for \textit{Fortress} where $\alpha = 0.01$.
For the Stanford Light Field scenes with narrow camera baseline, we find it not necessary to apply $L_{inter}$ and thus set $\alpha = 0$.
For the volumetric scenes, we set $\alpha = 0.1$.
{Scenes with more between-view content disparities would benefit from stronger semantic regulations through CLIP. The Stanford Light Field scenes do not necessarily need such regulation because the difference between adjacent views is small.}

\subsubsection{Training with CLIP-based Features.}
Without invoking CLIP to compute $L_{inter}$, training on the Stanford Light Field scenes takes around 5 hours.
For the other scenes which involve $L_{inter}$, the training time varies from 8 to 13 hours depending on the number of pixels for each scene. 
We only compute $L_{inter}$ once every two iterations to reduce the training time.

To extract the CLIP-based features, we use the public implementation provided by the original authors~\cite{radford2021learning}.
To extract the VGGNet-based features, we use the PyTorch implementation at~\url{ gist.github.com/alper111/8233cdb0414b4cb5853f2f730ab95a49}.
To address the issue that pre-trained networks require the input image to have a specific size which is different than our images, we reshape the full image into patches of $224 \times 224$ in a sliding window fashion, namely ``\textit{torch.functional.unfold(im, kernel\_size = (224, 224), stride=224}''.
Then, the features of the entire image are formed by concatenating the feature embedding of each $224 \times 224$ patch.

\subsubsection{Dataset Details.}
For the Stanford Light Field scenes, we train and test at the original resolution with a total of 25 images.
For the LLFF scenes, we use their $4\times$ downsampled version provided by the original authors.
For {our own volumetric dataset}, we train $1261 \times 1612$ for \textit{M1}, $658 \times 2246$ for \textit{M2}, and $1059 \times 182$ for \textit{W1} cropped based on their bounding boxes.
Each scene contains 30 images, {and those images are captured and included with consent from the three human participants.}

To train NeRF on the Stanford Light Field scenes, we follow~\cite{attal2021learning} and use their setup to train NeRF on these scenes.
We use NeRF with 8 hidden layers and dimension as 256 (for both coarse and fine networks), and we train for $200,000$ iterations with the batch size of $1,024$ pixels.
To train LFN, we adapt the implementation from~\cite{sitzmann2021light} to train a single network for each scene.
We use LFN with 8 hidden layers and dimension as 512, and we train for 500 epochs with the batch size of $65,536$ pixels.
The results for all methods (including ours) would likely improve further with longer training, and we tried to obtain fair results under a limited resource budget.
We note that the purpose of these comparisons is for reference, not for competition.

\begin{algorithm}[!th]
\caption{VIINTER Training Procedure}
\label{alg:recon}
\SetKwProg{generate}{\textbf{For each training iteration:}}{}{end}
\SetKwProg{Inp}{Data}{}
\Inp{\textbf{Data}: $N$ images of different views $\{I_{n}\}_{n=1}^{N}$ each with pixels $\mathcal{P}_{n}$. Each $p \in \mathcal{P}_{n}$ has coordinate $(p_{x}, p_{y})$ and color $p_{c | n}^{GT}$. Feature extracting network $\mathcal{E}.$} \\
\SetKwProg{Param}{Parameters}{}
\Inp{\textbf{Parameters}: Weights of $\mathcal{F}$ and $\{z_{n} \in \mathbb{R}^{M}\}_{n=1}^{N}$.} \\
\SetKwProg{Prep}{Prepare}{}
\Inp{\textbf{Prepare}: Extract features $\{\mathcal{E}(I_n)\}_{n=1}^{N}$ of known images.} \\
\generate{}{
    Randomly select $i, j = \{1, ..., N\}$\\
    $1$-norm constraint $z_{i}, z_{j} = \frac{z_{i}}{\|z_{i}\|_{1}}, \frac{z_{j}}{\|z_{j}\|_{1}}$ \\
    Sample $B$ pixels as $\mathcal{P}_{i}^{batch}$ from $\mathcal{P}_{i}$.
    $\forall p \in \mathcal{P}_{i}^{batch}$, get $\mathcal{F}(p_{x}, p_{y} \mid z_{i}) = p_{c|i}$ and loss $L_{Recon}$ with $p_{c | i}^{GT}$\\
    Randomly select an interpolation weight $t = [0, 1]$ such that $z_{Inter}$ = $(1-t) \cdot z_{i} + t \cdot z{j}$ \\
    $\forall p \in \mathcal{P}_{i}$, compute $\mathcal{F}(p_{x}, p_{y} \mid z_{Inter}) = p_{c|Inter}$ \\
    Reshape the output $\{p_{c|Inter}\}_{\forall p \in \mathcal{P}_{i}}$ as 2D image $I_{Inter}$ \\
    Extract features $\mathcal{E}(I_{Inter})$ and compute loss $L_{Inter}$ with $(1 - t) \cdot \mathcal{E}(I_{i}) + t \cdot \mathcal{E}(I_{j})$ \\
    Update $\mathcal{F}, z_{i}, z{j}$ based on loss terms $L_{Recon}$ and $L_{Inter}$
}
\end{algorithm}
\vspace{-10pt}

\begin{figure*}[t]
    \centering
    \includegraphics[height=30mm]{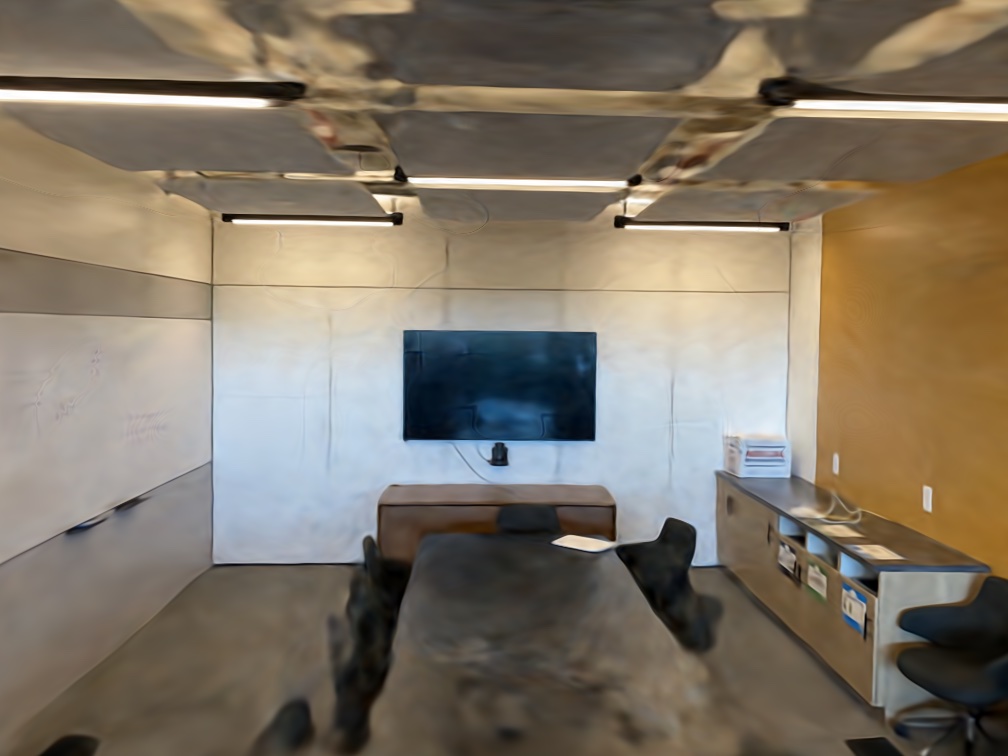}%
    ~
    \includegraphics[height=30mm]{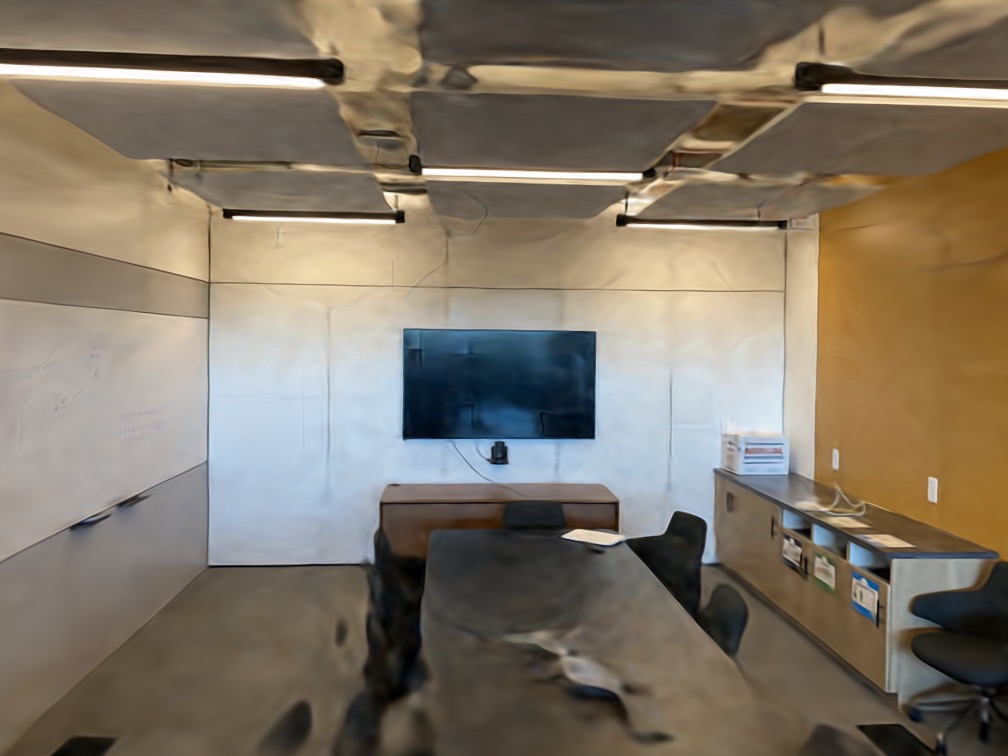}%
    ~
    \includegraphics[height=30mm]{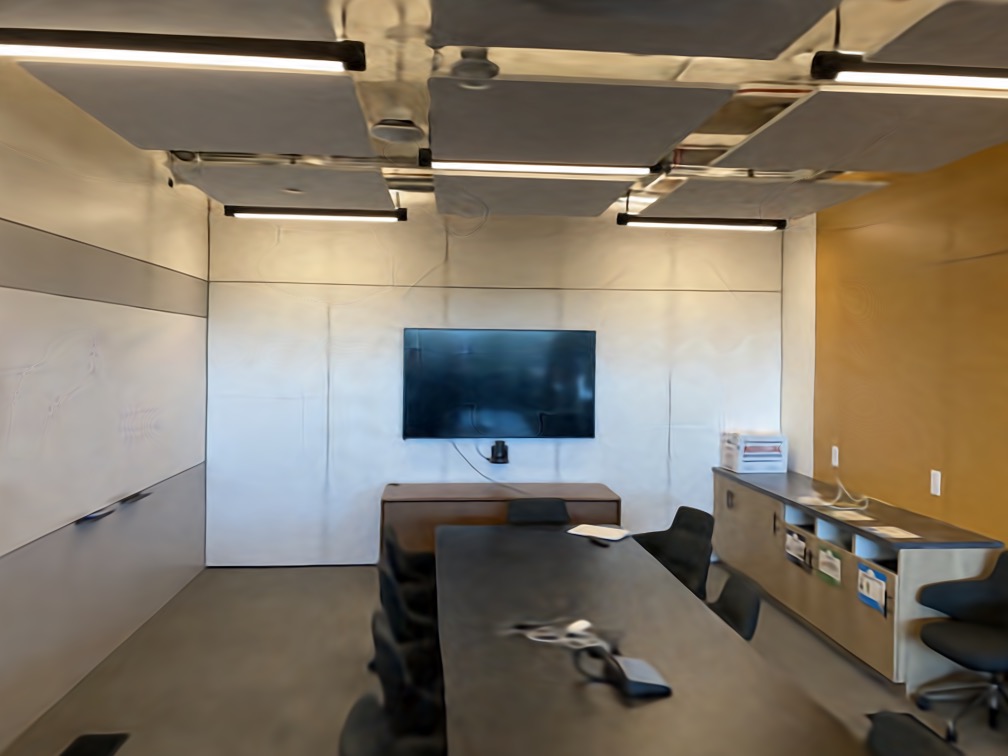}%
    ~
    \includegraphics[height=30mm]{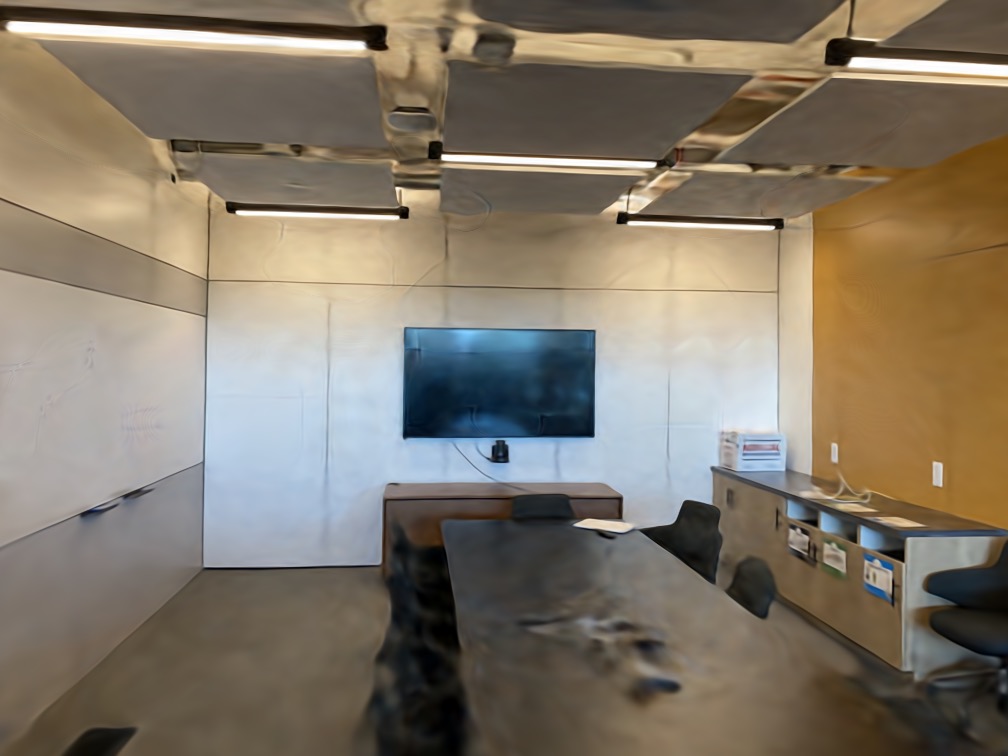}%

    \includegraphics[height=30mm]{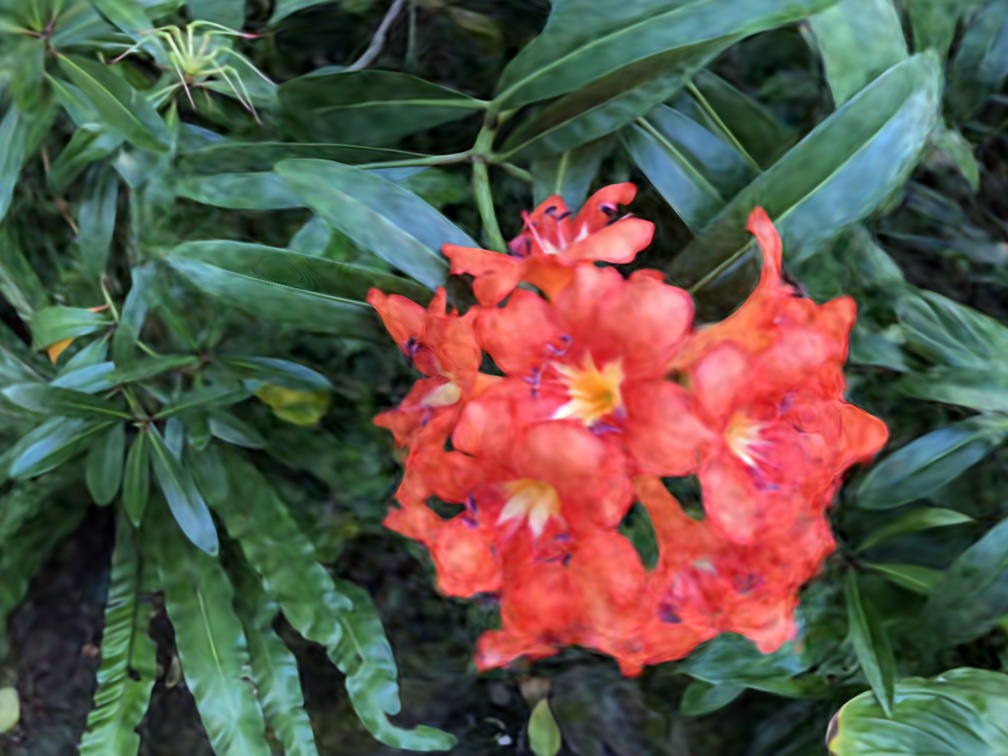}%
    ~
    \includegraphics[height=30mm]{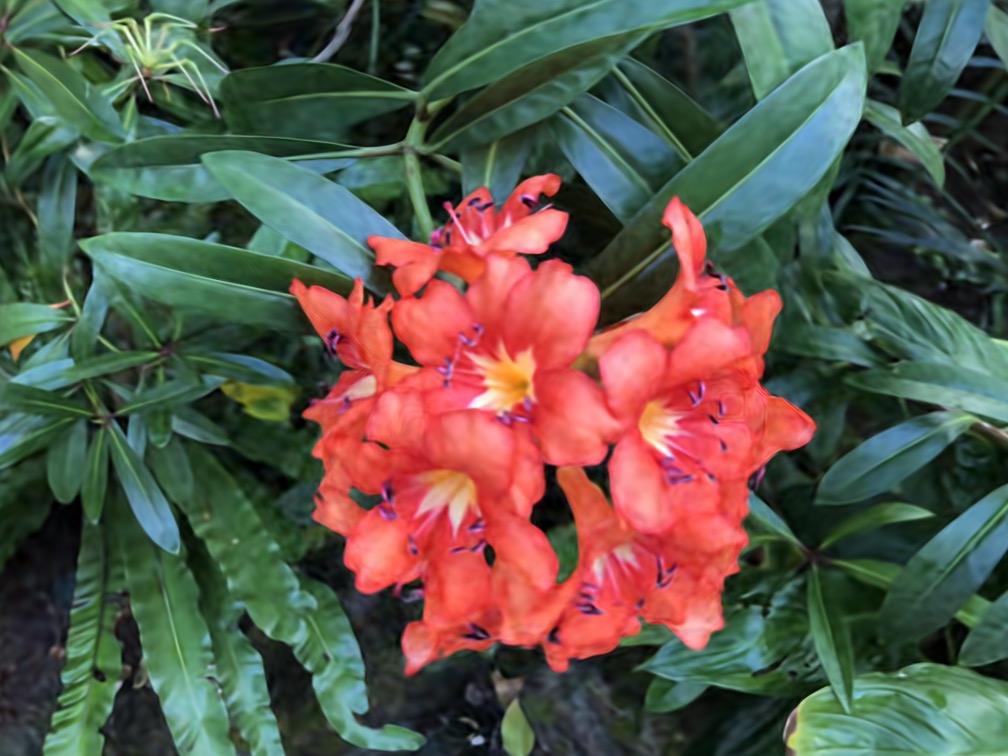}%
    ~
    \includegraphics[height=30mm]{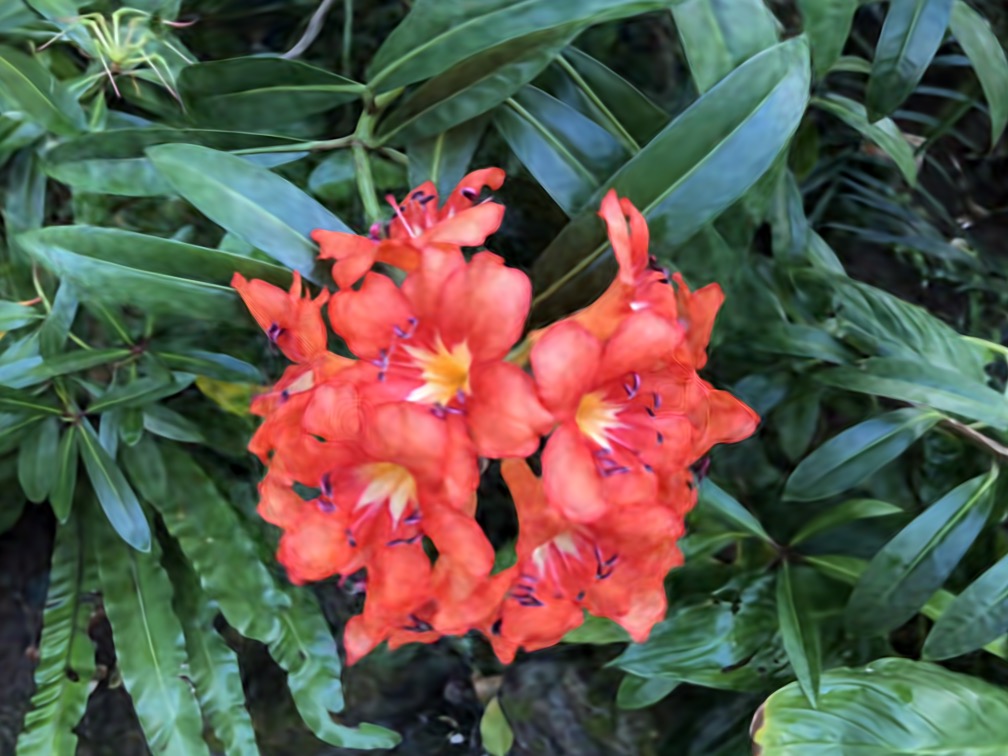}%
    ~
    \includegraphics[height=30mm]{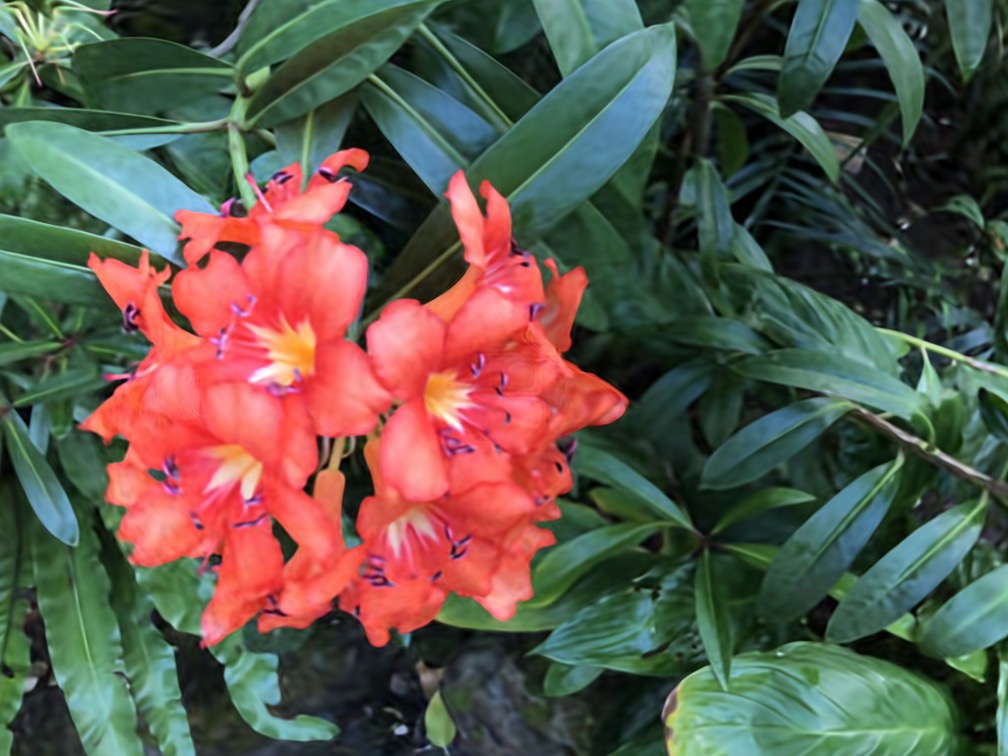}%

    \vspace{-3pt}
    \caption{{Interpolated Between Interpolated Latent Codes.} After interpolating the latent codes for two training views (Column 1 and 4), we can further interpolate between those interpolated latent codes (Column 2 and 3). This additional step effectively leads to more viewpoints that can be expressed by our INR.}
    \label{fig:ABCDMorph}
\end{figure*}

\begin{figure*}[t]
    \centering
    \includegraphics[width=32mm]{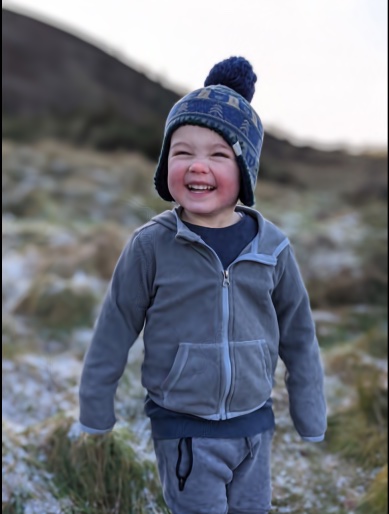}%
    ~
    \includegraphics[width=32mm]{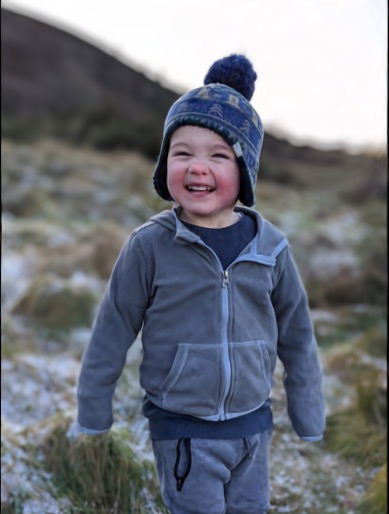}%
    ~    ~
    \includegraphics[width=32mm]{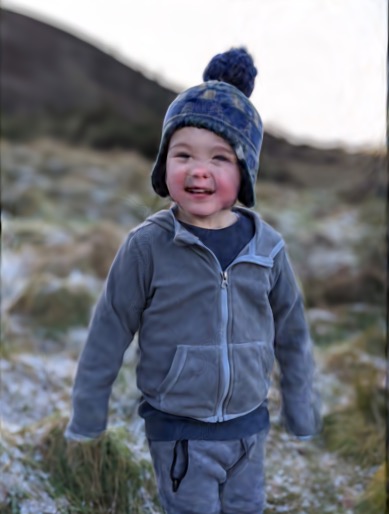}%
    ~
    \includegraphics[width=32mm]{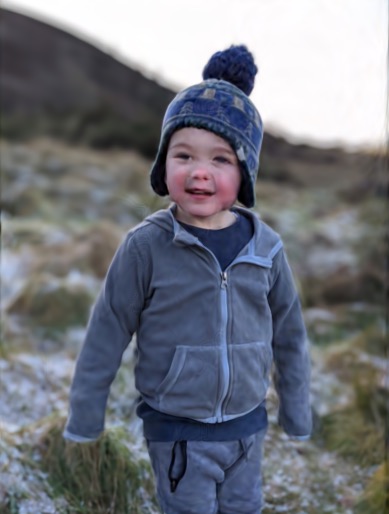}%
    ~
    \includegraphics[width=32mm]{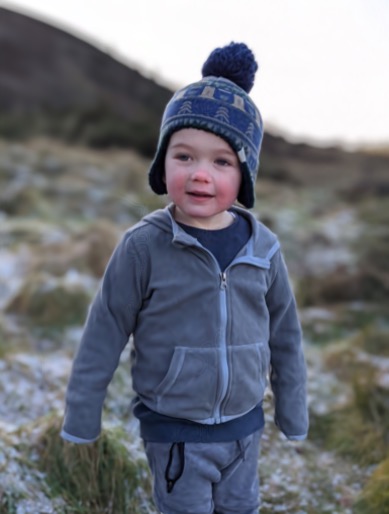}%

    \includegraphics[width=32mm]{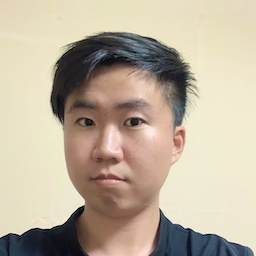}%
    ~
    \includegraphics[width=32mm]{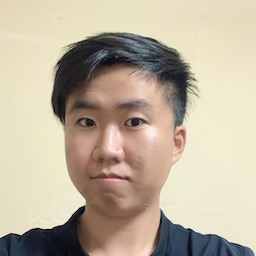}%
    ~    ~
    \includegraphics[width=32mm]{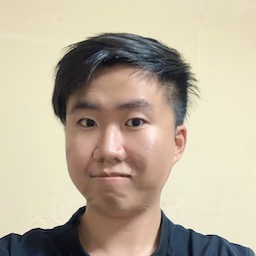}%
    ~
    \includegraphics[width=32mm]{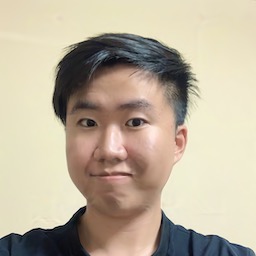}%
    ~
    \includegraphics[width=32mm]{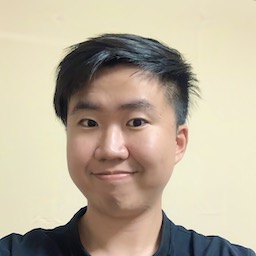}%

    \vspace{-3pt}
    \caption{{Frame Interpolation With Human Faces.} Inspired by~\cite{reda2022film}, we deploy VIINTER on pairs of human portrait images with different expressions (Column 1 and 5). We then render the INR with interpolated latent codes between those two training views, and the resulting images (Column 2, 3 4) exhibit smooth transitions between the two expressions.}
    \label{fig:FaceMorph}
\end{figure*}

\begin{figure*}[t]
    \centering
    \includegraphics[width=32mm]{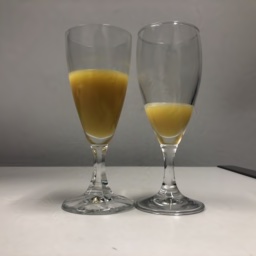}%
    ~
    \includegraphics[width=32mm]{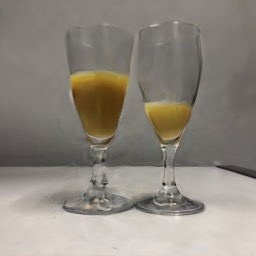}%
    ~    ~
    \includegraphics[width=32mm]{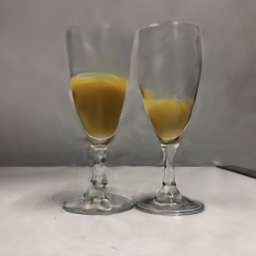}%
    ~
    \includegraphics[width=32mm]{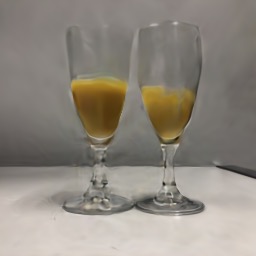}%
    ~
    \includegraphics[width=32mm]{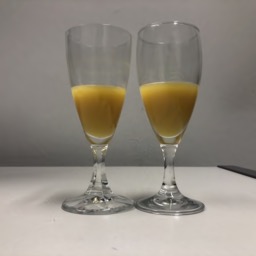}%

    \includegraphics[width=32mm]{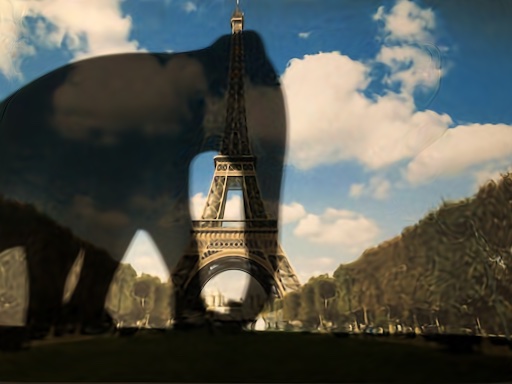}%
    ~
    \includegraphics[width=32mm]{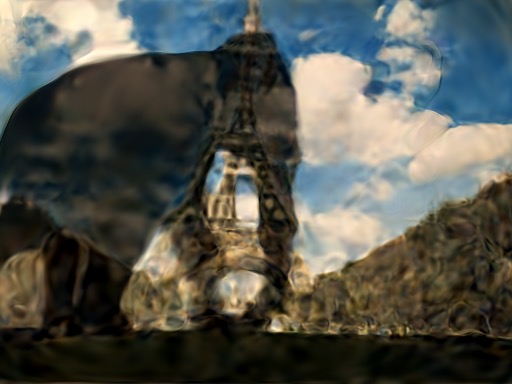}%
    ~    ~
    \includegraphics[width=32mm]{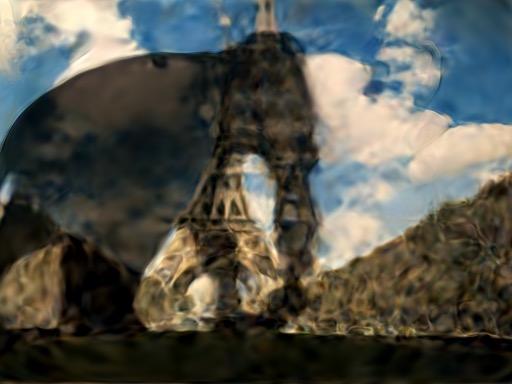}%
    ~
    \includegraphics[width=32mm]{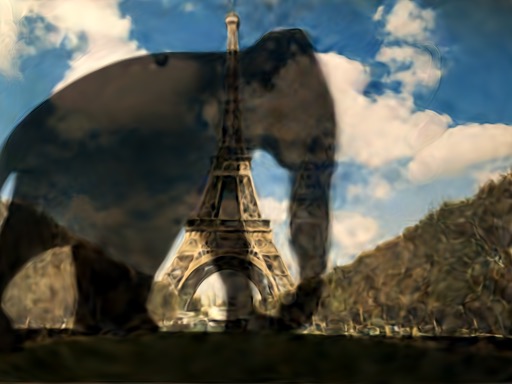}%
    ~
    \includegraphics[width=32mm]{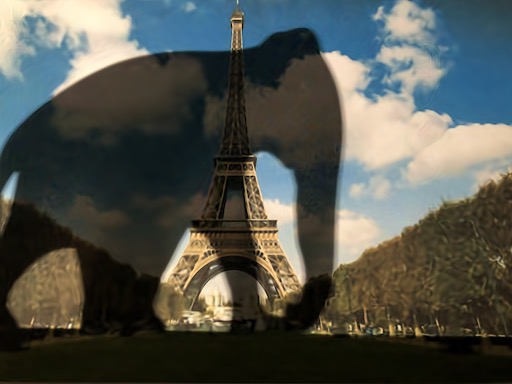}%

    \vspace{-3pt}
    \caption{{Frame Interpolation on General Images.} We deploy VIINTER on two images (Column 1 and 5) captured at different timesteps provided by X-Fields~\cite{Bemana2020xfields}. We then render the INR with interpolated latent codes between those two training views, and the resulting images (Column 2, 3, 4) exhibit smooth transitions between the two expressions.}
    \label{fig:XFields}
\end{figure*}

\begin{figure*}[!ht]

    \FigThreeSubfig{./figures/MCompare/64/0.jpeg}{$t=0$}%
    ~
    \FigThreeSubfig{./figures/MCompare/64/14.jpeg}{$t=0.5$}%
    ~
    \FigThreeSubfig{./figures/MCompare/64/29.jpeg}{$t=1$}%
    ~
    \FigThreeSubfig{./figures/MCompare/128/0.jpeg}{$t=0$}%
    ~
    \FigThreeSubfig{./figures/MCompare/128/14.jpeg}{$t=0.5$}%
    ~
    \FigThreeSubfig{./figures/MCompare/128/29.jpeg}{$t=1$}%
    
    \vspace{-12pt}\hspace{0.1cm}$\underbracket[0.5pt][3pt]{\hspace{8.1cm}}_%
    {\substack{\vspace{-10pt}\\ \colorbox{white}{$M=64$}}}$    \vspace{-7pt}
    \hspace{0.8cm}$\underbracket[0.5pt][3pt]{\hspace{8cm}}_%
    {\substack{\vspace{-10pt}\\ \colorbox{white}{$M=128$}}}$
    \vspace{7pt}

    \FigThreeSubfig{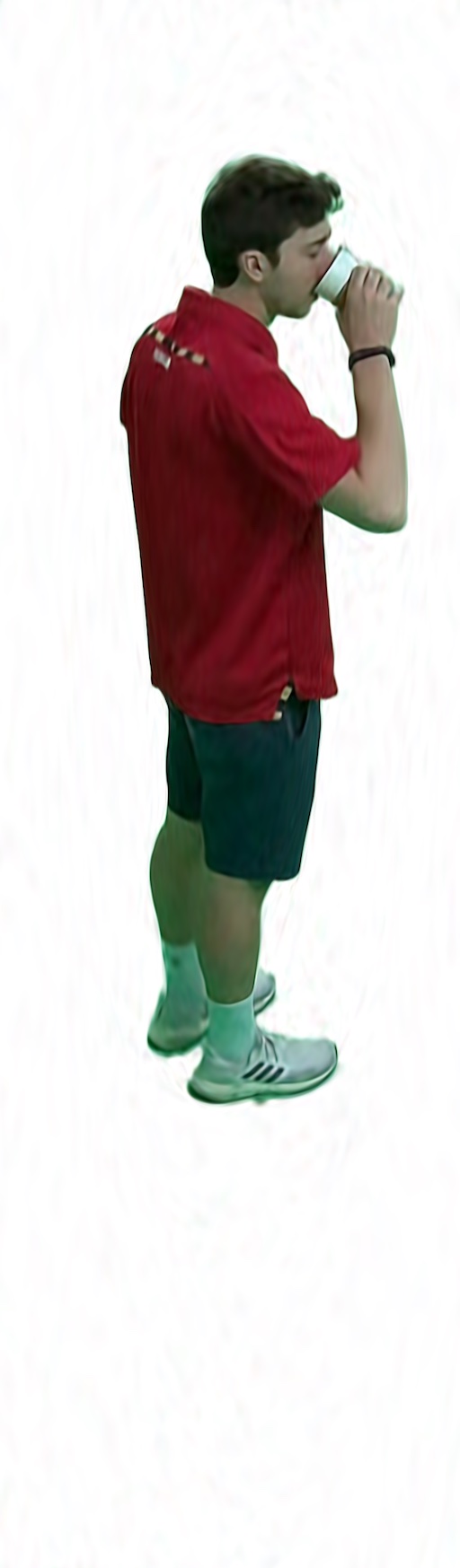}{$t=0$}%
    ~
    \FigThreeSubfig{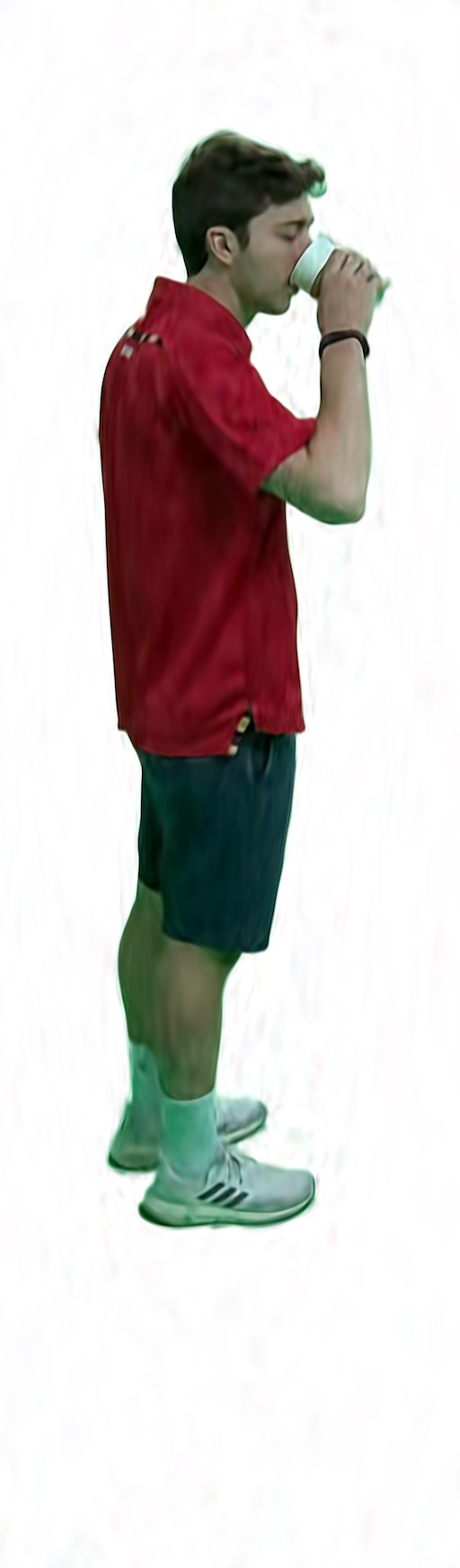}{$t=0.5$}%
    ~
    \FigThreeSubfig{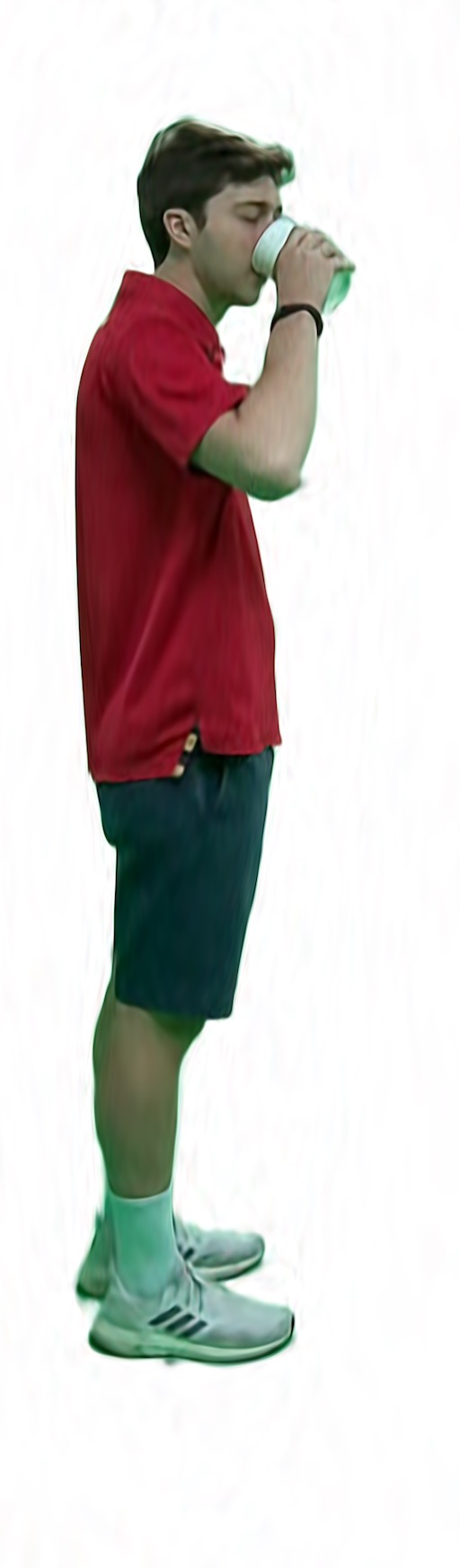}{$t=1$}%
    ~
    \FigThreeSubfig{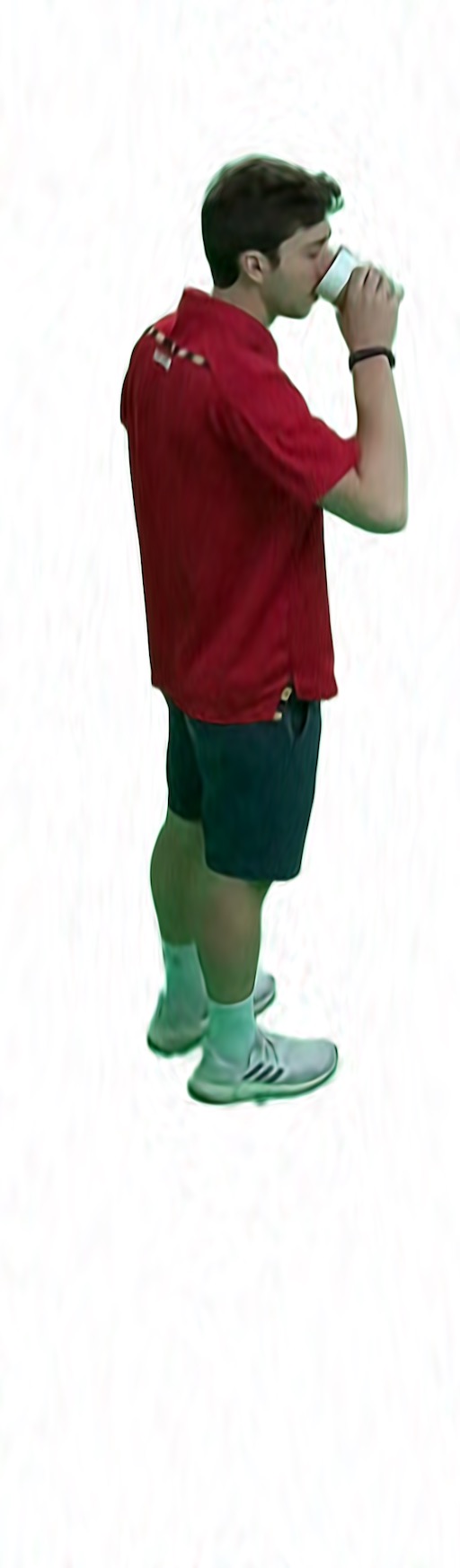}{$t=0$}%
    ~
    \FigThreeSubfig{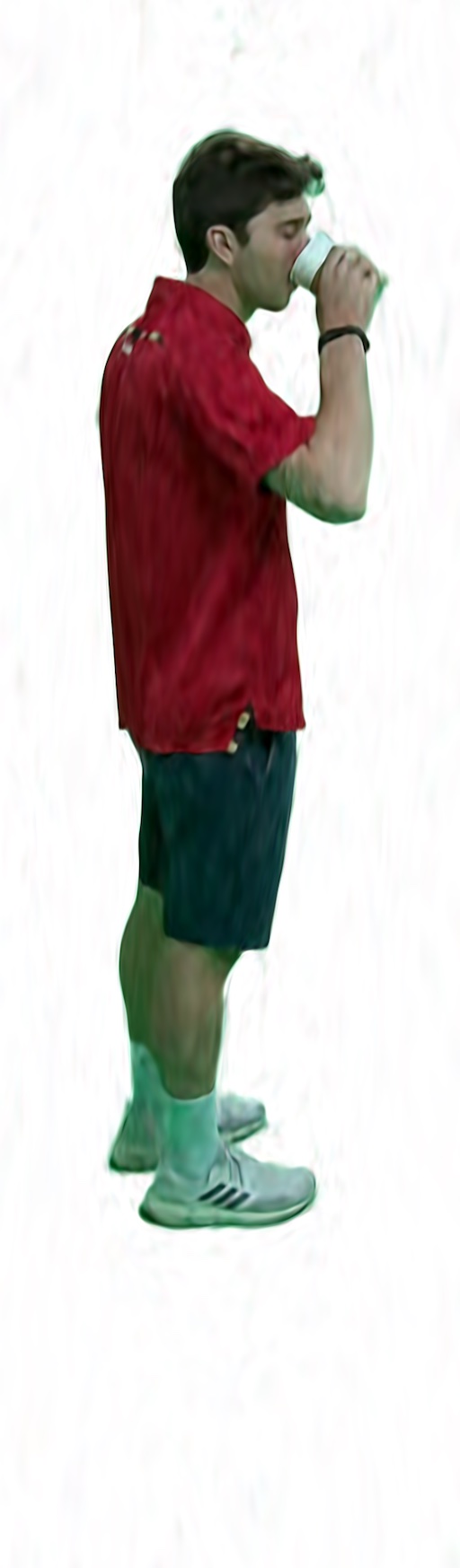}{$t=0.5$}%
    ~
    \FigThreeSubfig{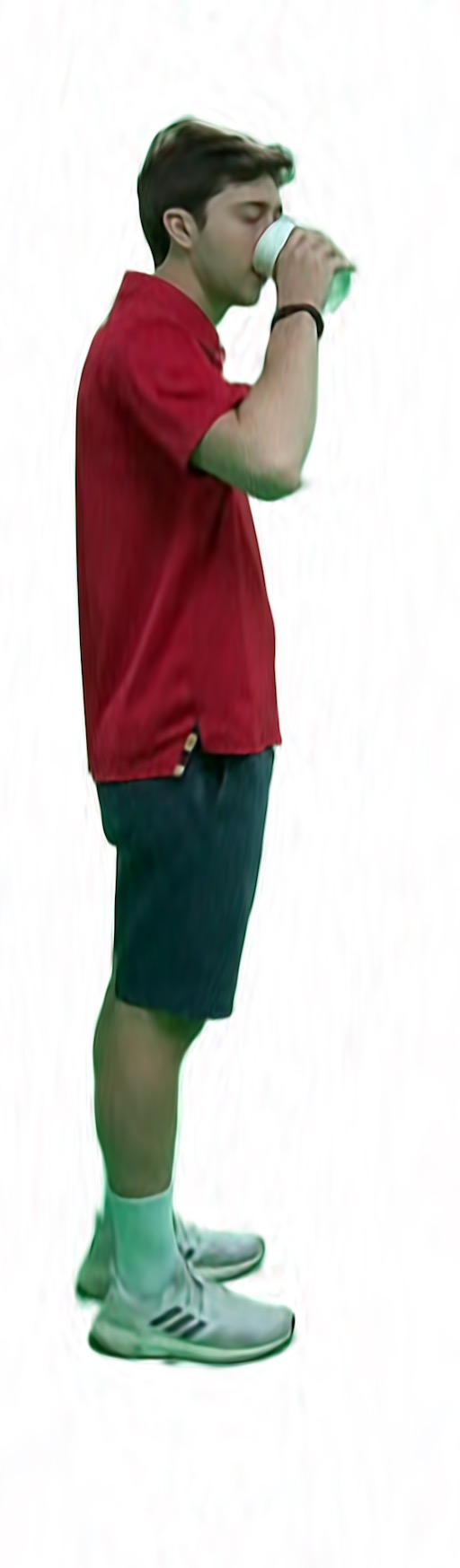}{$t=1$}%
    
    \vspace{-12pt}\hspace{0.1cm}$\underbracket[0.5pt][3pt]{\hspace{8.1cm}}_%
    {\substack{\vspace{-10pt}\\ \colorbox{white}{$M=256$}}}$    \vspace{-7pt}
    \hspace{0.8cm}$\underbracket[0.5pt][3pt]{\hspace{8cm}}_%
    {\substack{\vspace{-10pt}\\ \colorbox{white}{$M=512$}}}$
	\caption{{Additional Results Similar to Figure~\ref{fig:CodeCompare}. Increasing $M$ beyond $128$ leads to marginal impact in our experiments.}}
	\label{fig:SuppCodeCompare}
\end{figure*}

\vspace{-10pt}
{
\subsection{Additional Results}
}
\subsubsection{{More Ablation Results.}}
\begin{figure}[!t]
    \centering
    \SuppResolutionCompareSubfig{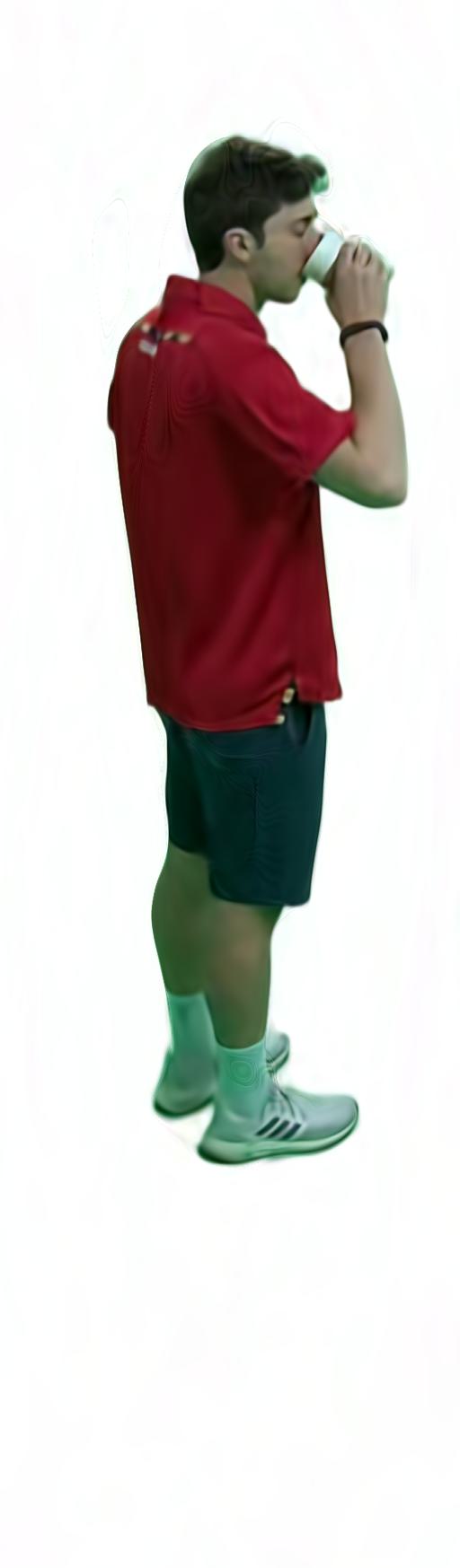}%
    ~
    \SuppResolutionCompareSubfig{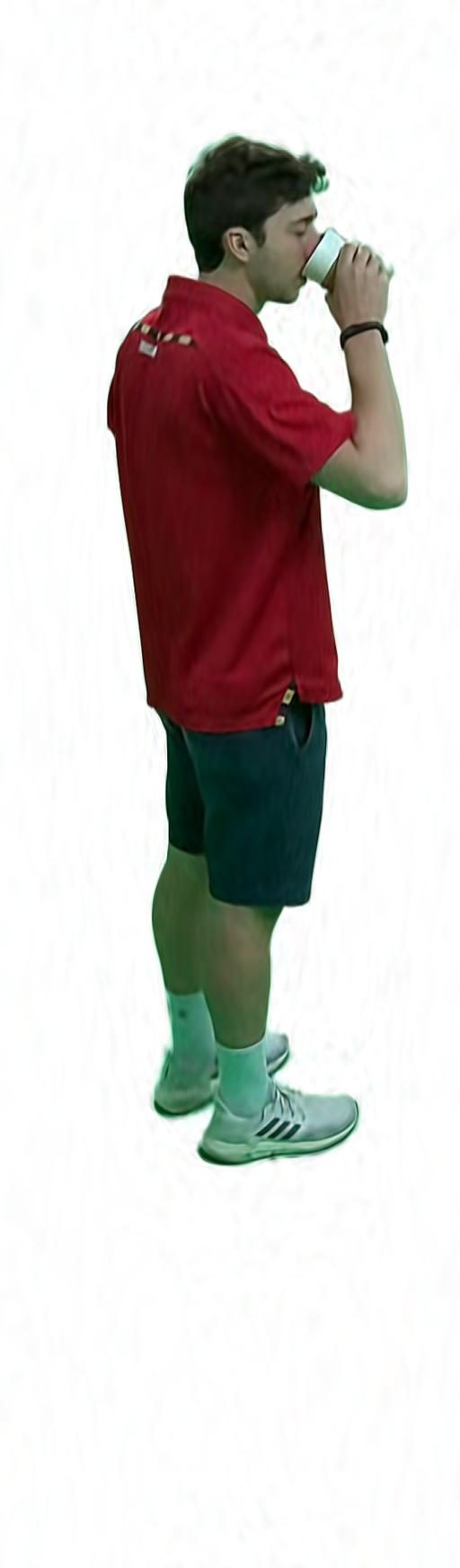}%
    \vspace{-3pt}
    \caption{{Left: INR trained with $L_{Inter}$. Right: INR trained without $L_{Inter}$. Despite smoother interpolation, training with $L_{Inter}$ could restrict the INR's to ability to preserve sharp details. The training set PSNR drops from 32.36 to 30.39.}}
    \label{fig:SuppLimitation}
\end{figure}

In Figure~\ref{fig:SuppCodeCompare}, we provide more results on increasing the latent code length $M$ beyond the default value of 128.
In our experiments, we did not find meaningful benefits of having longer latent code.
While the INR trained to almost perfectly reconstruct the training views is unlikely to allow for smooth interpolation, our method significantly improves interpolation, but sometimes at the expense of some sharp details in the reconstruction, as shown in Figure~\ref{fig:SuppLimitation}.

\subsubsection{Detailed Reconstructed and Novel Views.}
\begin{figure*}[!h]
    \SuppSixSubfig{0 0 0 0}{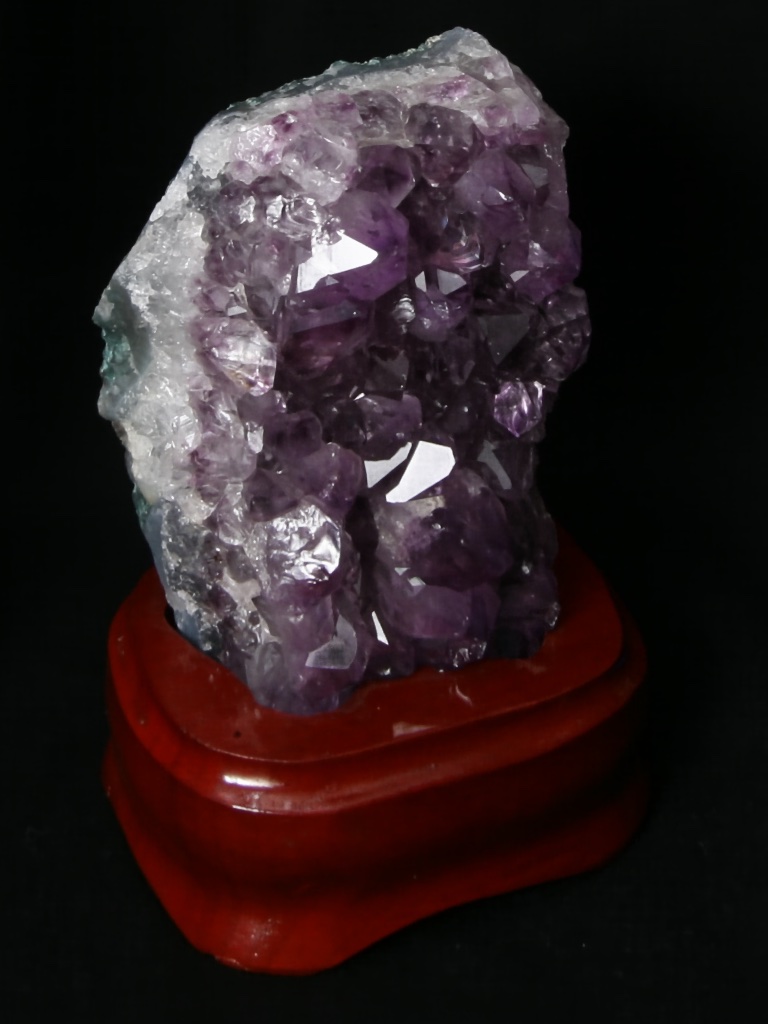}%
    ~
    \SuppSixSubfig{0 0 0 0}{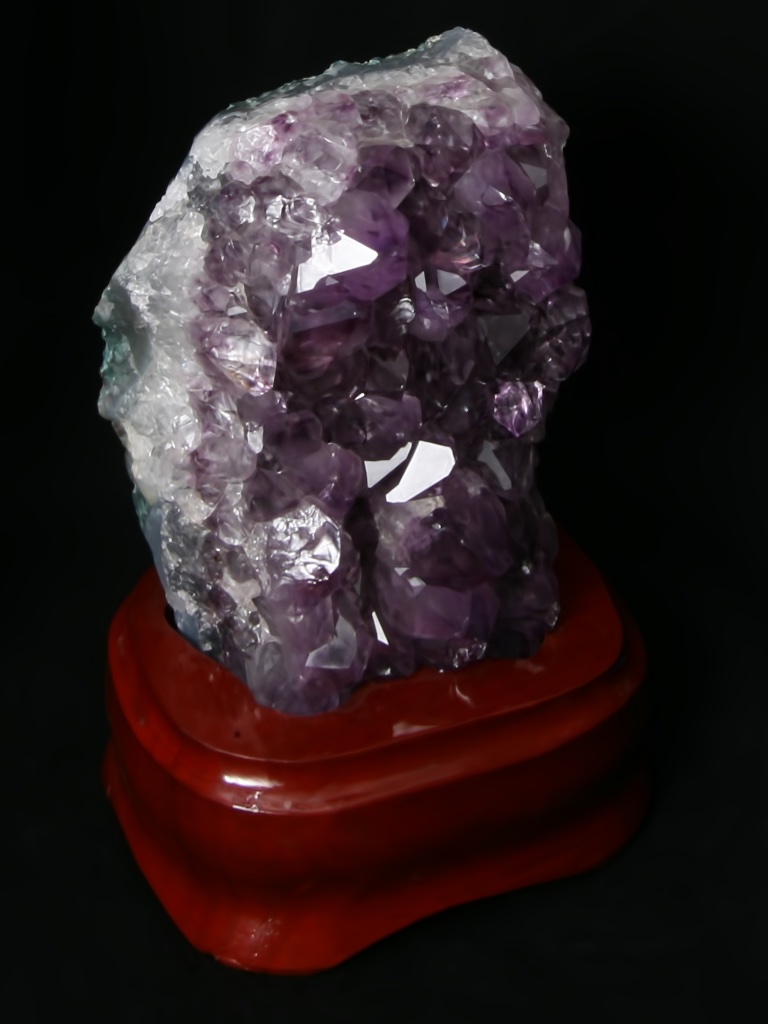}%
    ~
    \SuppSixSubfig{0 0 0 0}{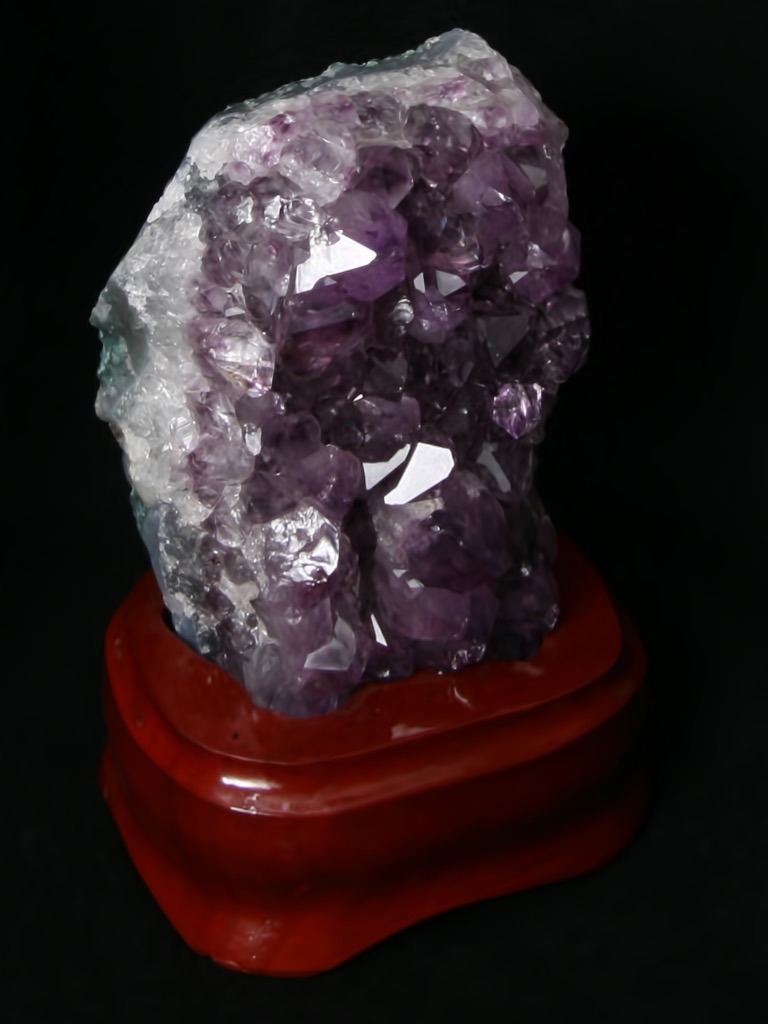}%
    ~
    \SuppSixSubfig{0 0 0 0}{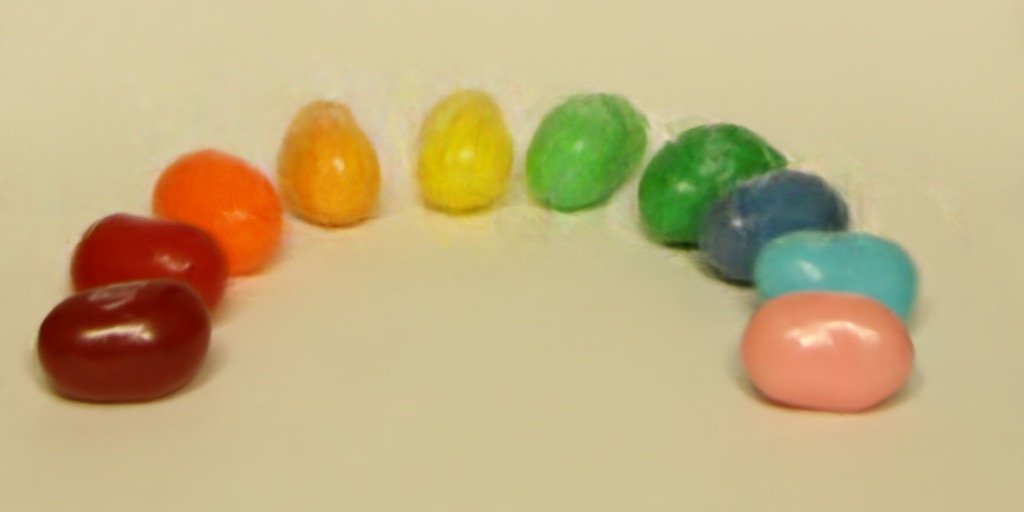}%
    ~
    \SuppSixSubfig{0 0 0 0}{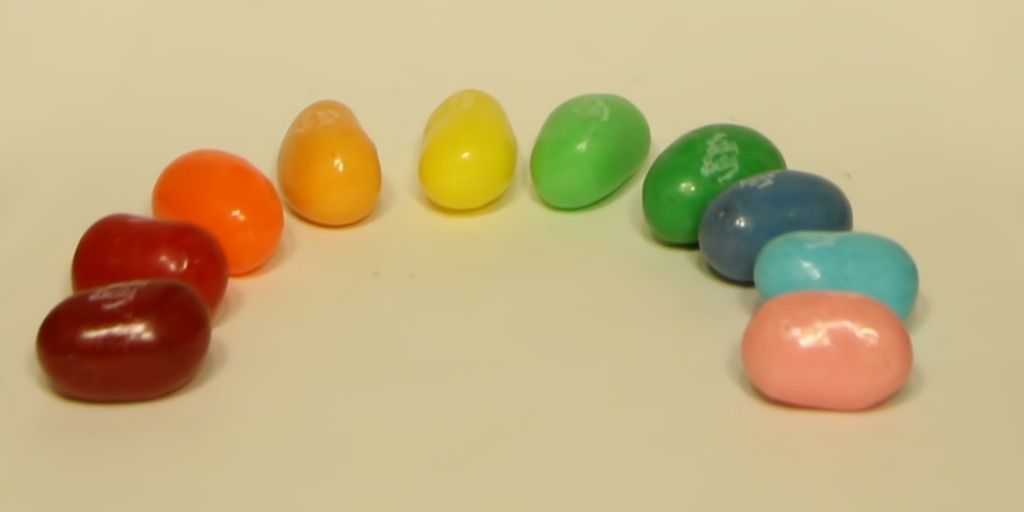}%
    ~
    \SuppSixSubfig{0 0 0 0}{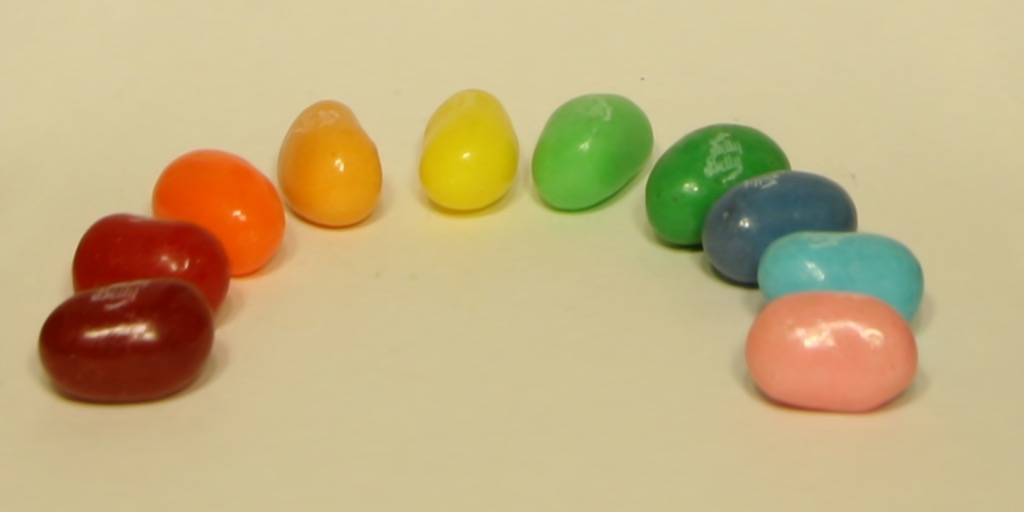}%
    
    \SuppSixSubfig{0 0 0 0}{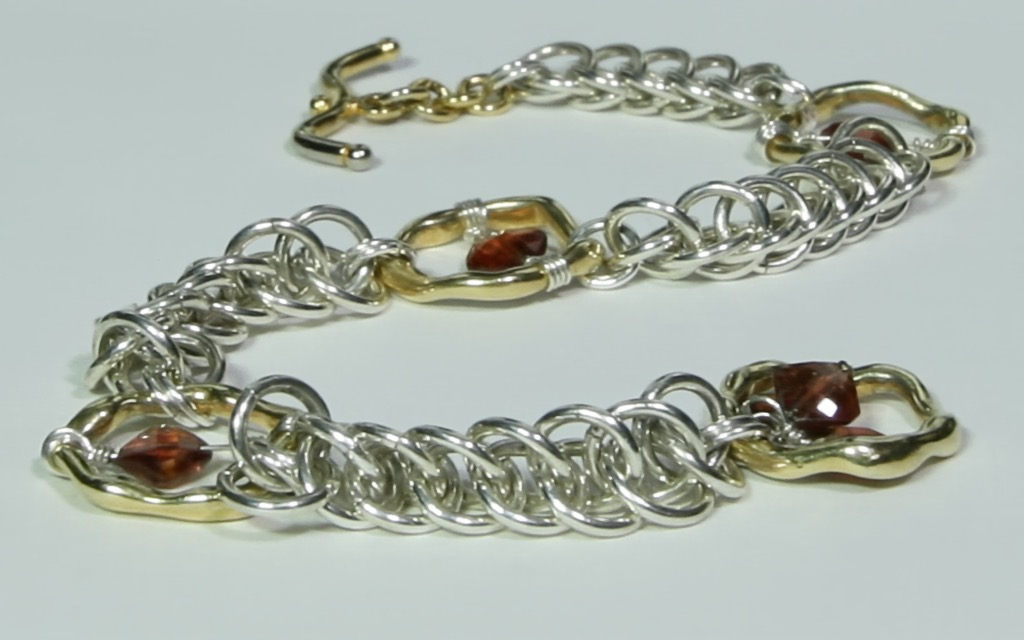}%
    ~
    \SuppSixSubfig{0 0 0 0}{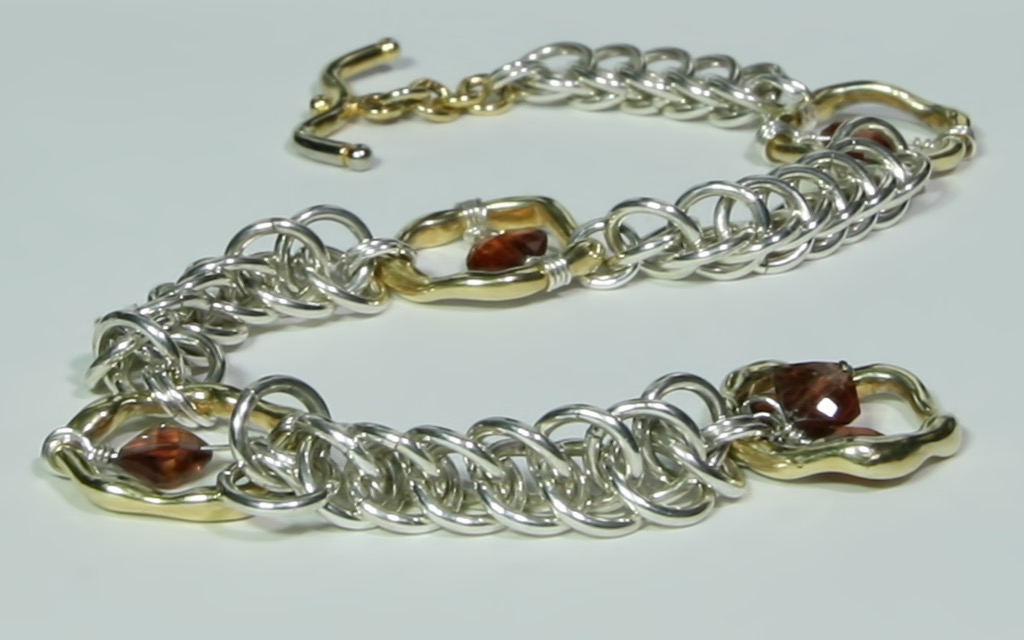}%
    ~
    \SuppSixSubfig{0 0 0 0}{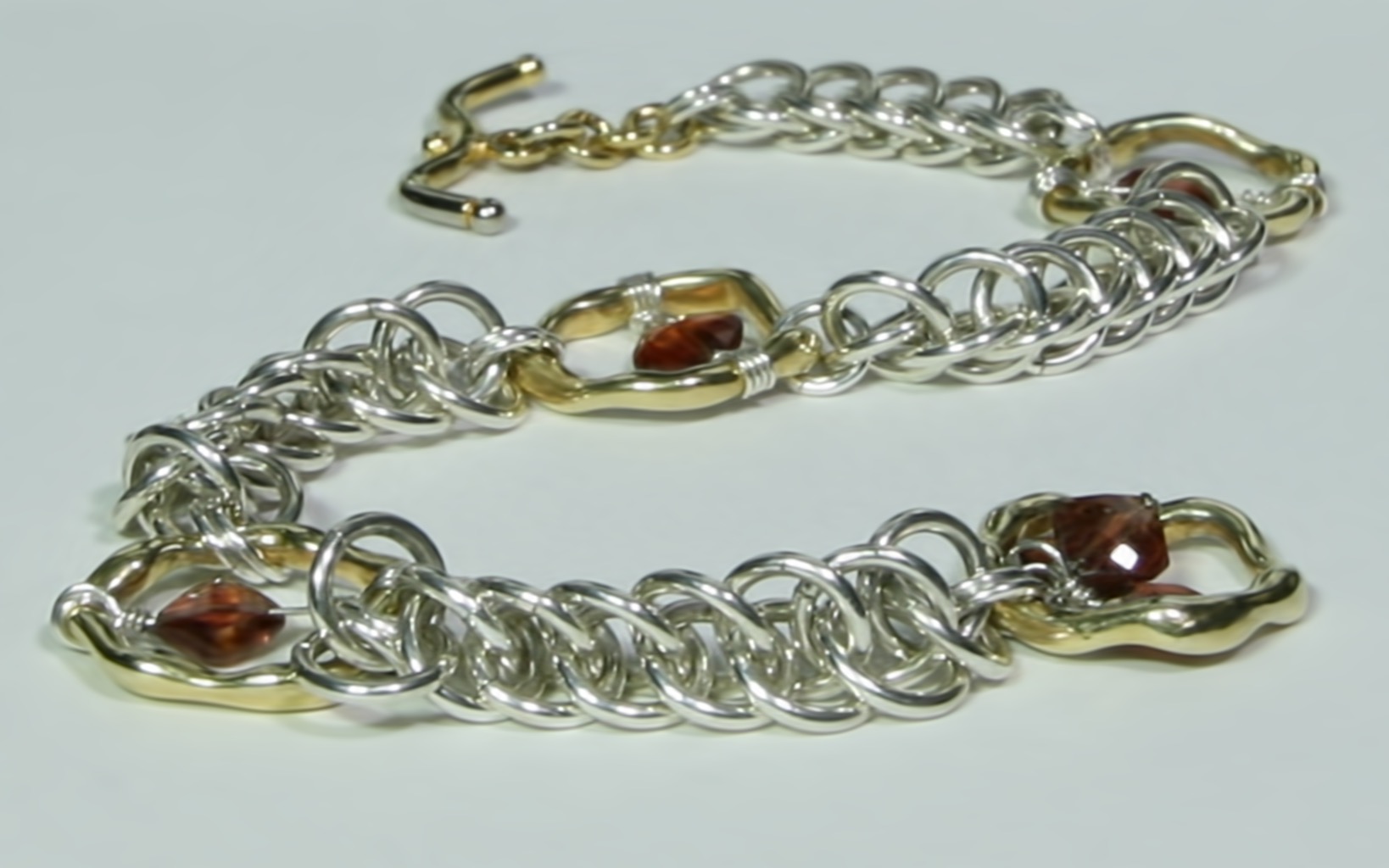}%
    ~
    \SuppSixSubfig{0 0 0 0}{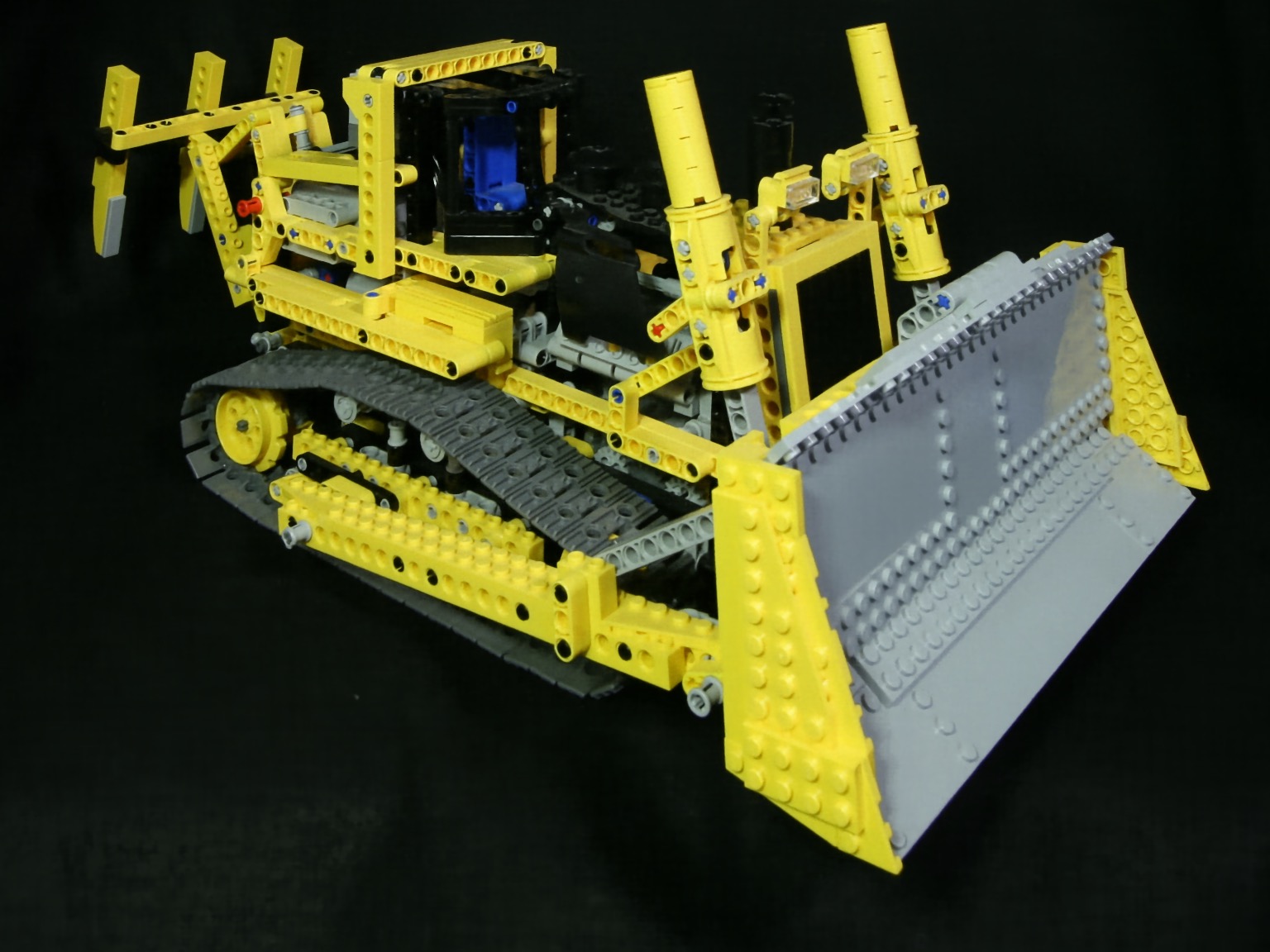}%
    ~
    \SuppSixSubfig{0 0 0 0}{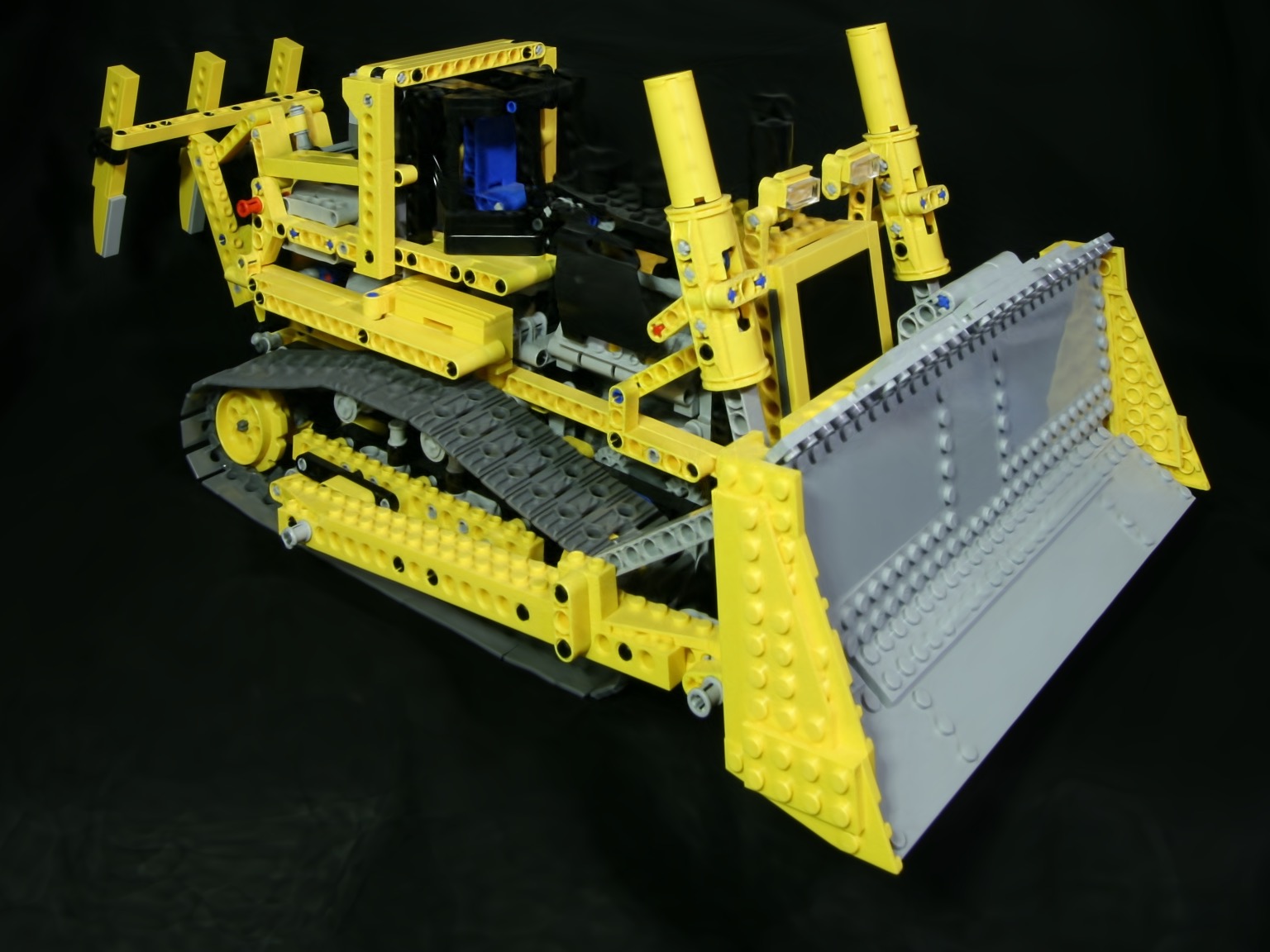}%
    ~
    \SuppSixSubfig{0 0 0 0}{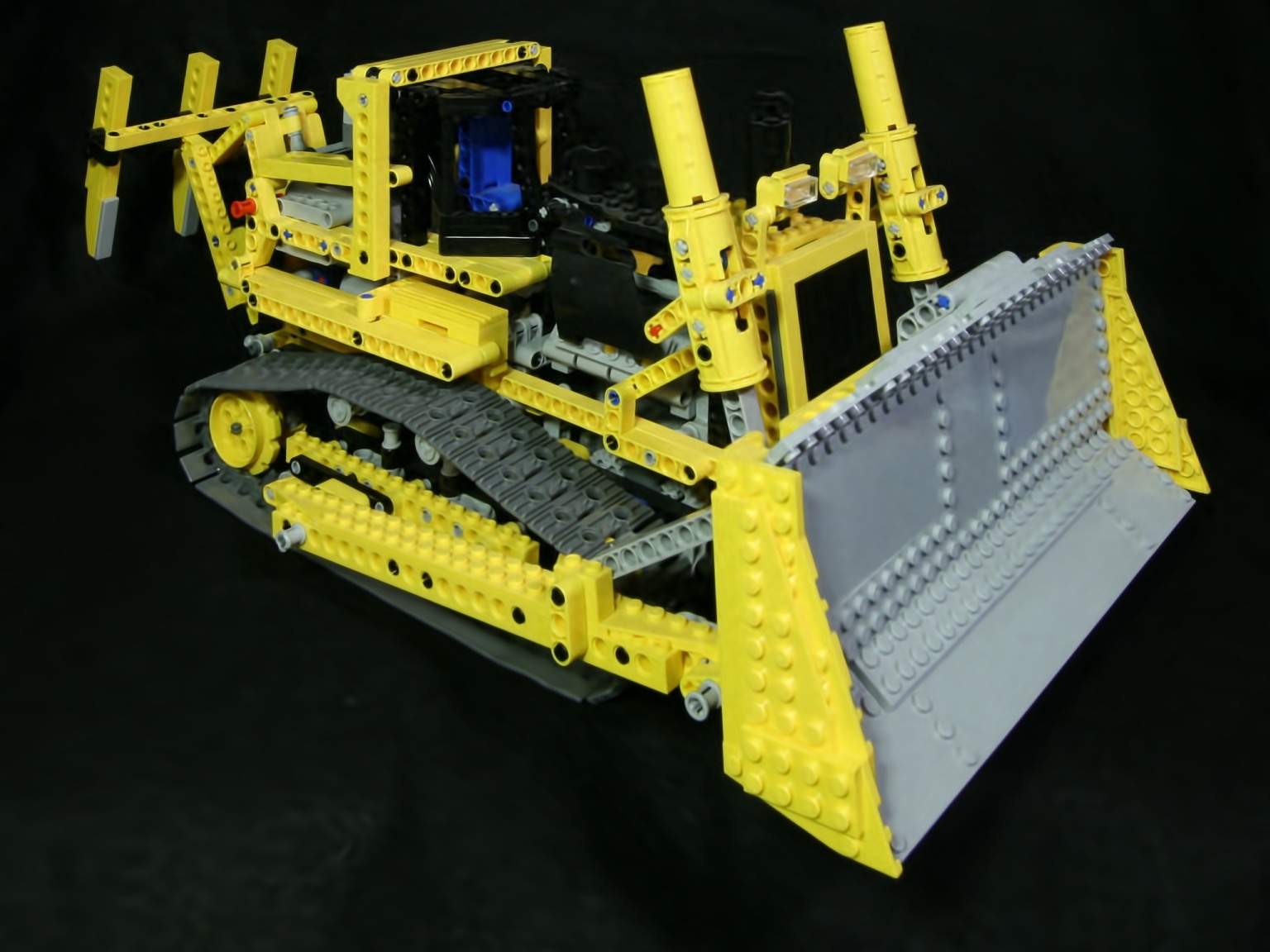}%
    
    \SuppSixSubfig{0 0 0 0}{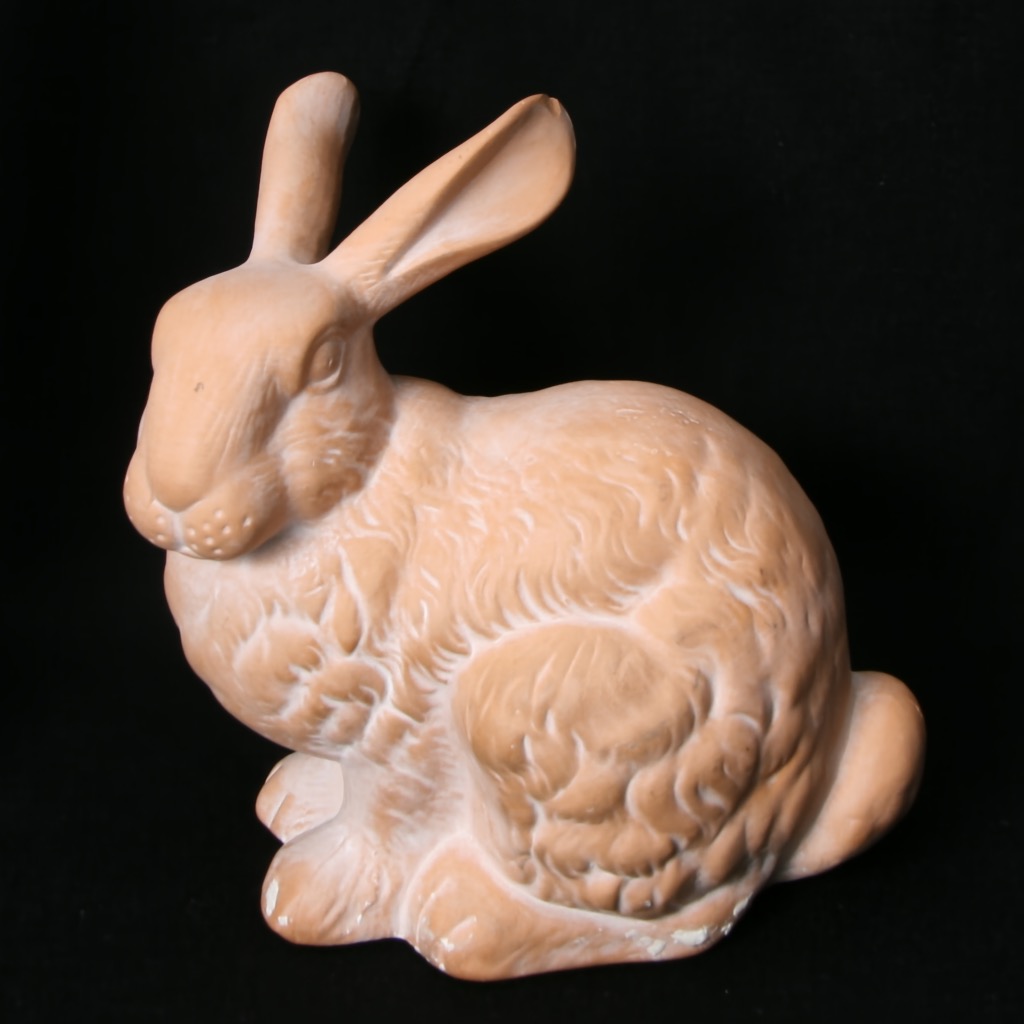}%
    ~
    \SuppSixSubfig{0 0 0 0}{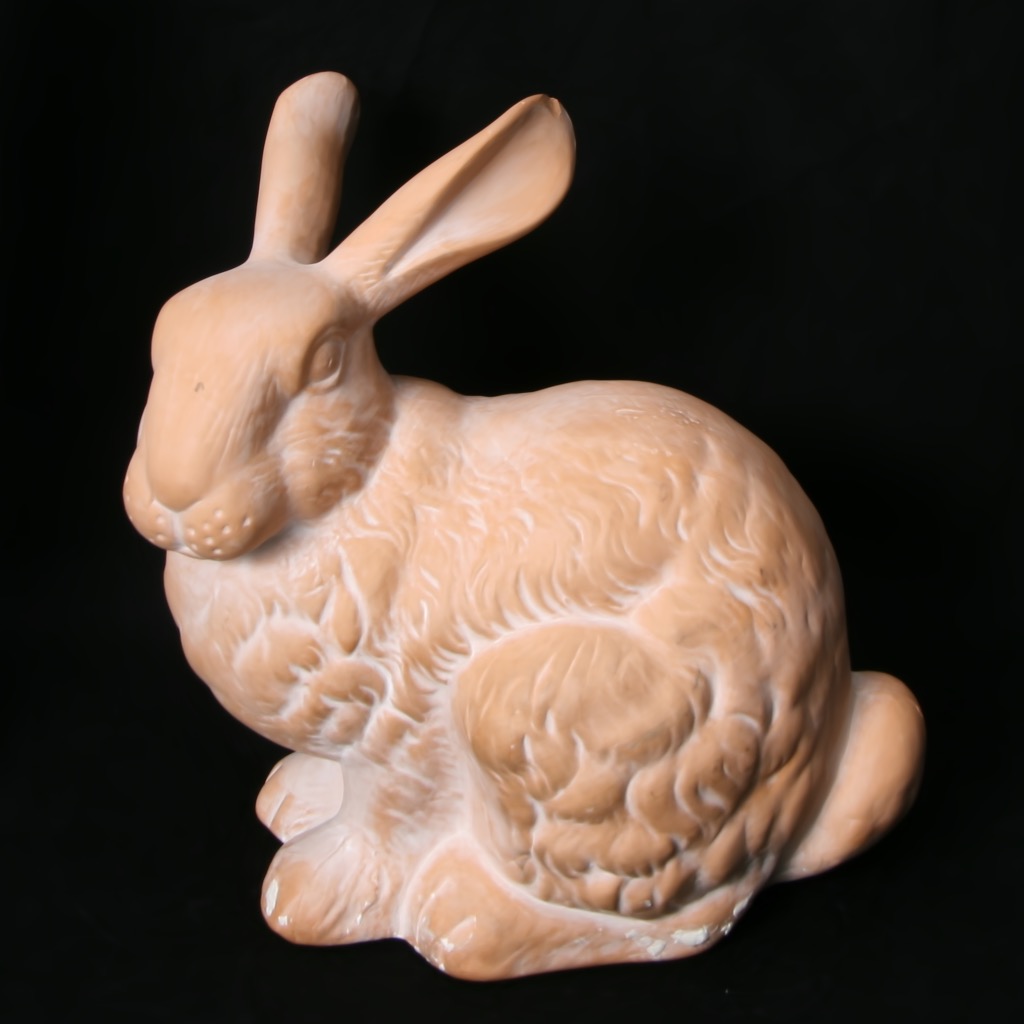}%
    ~
    \SuppSixSubfig{0 0 0 0}{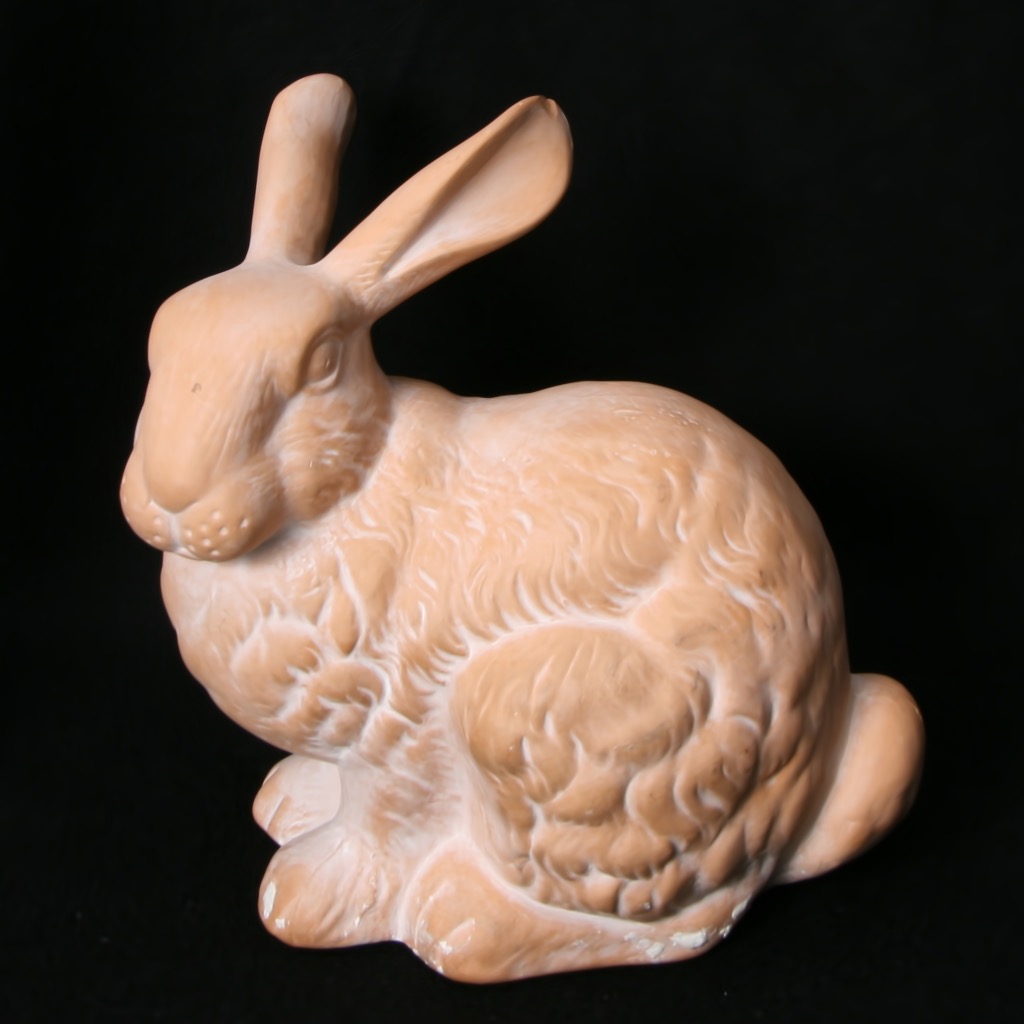}%
    ~
    \SuppSixSubfig{0 0 0 0}{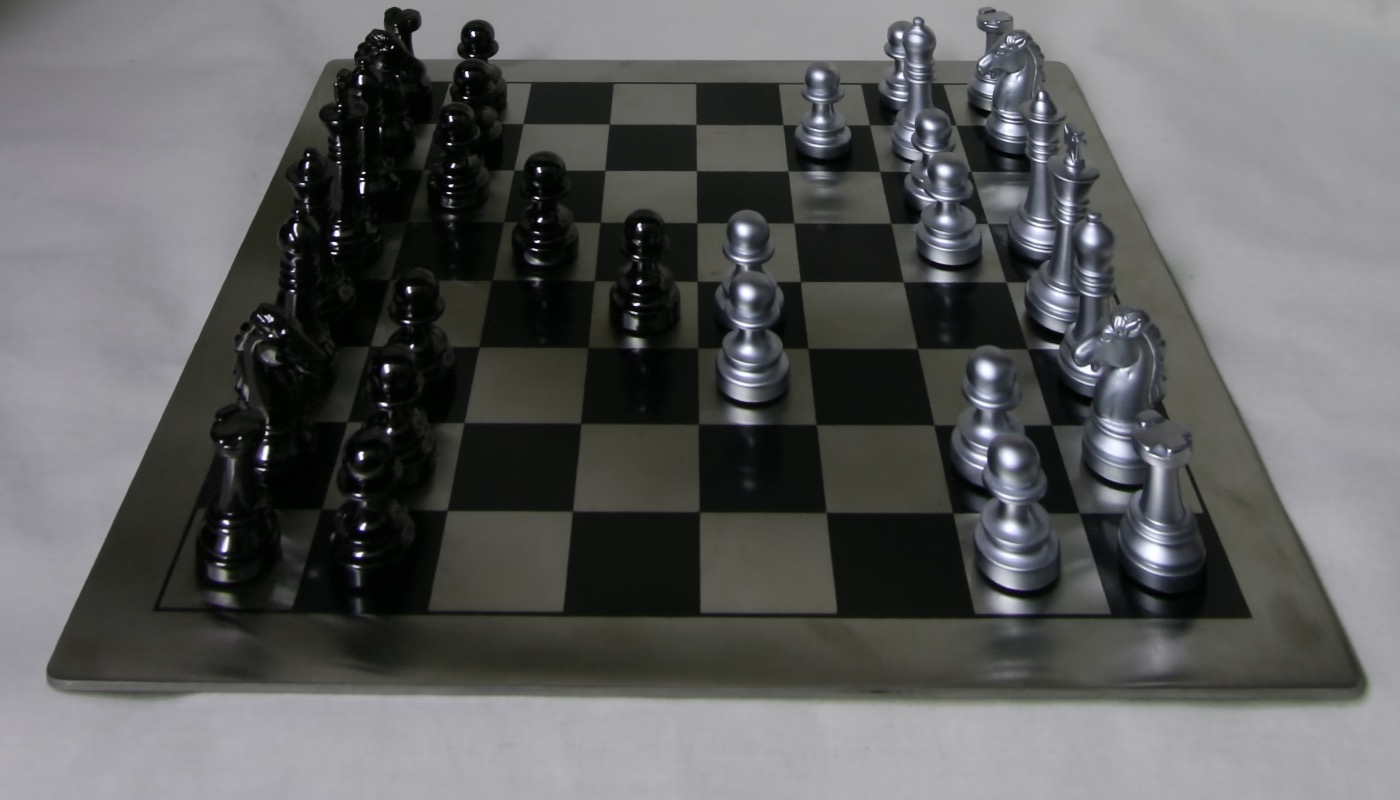}%
    ~
    \SuppSixSubfig{0 0 0 0}{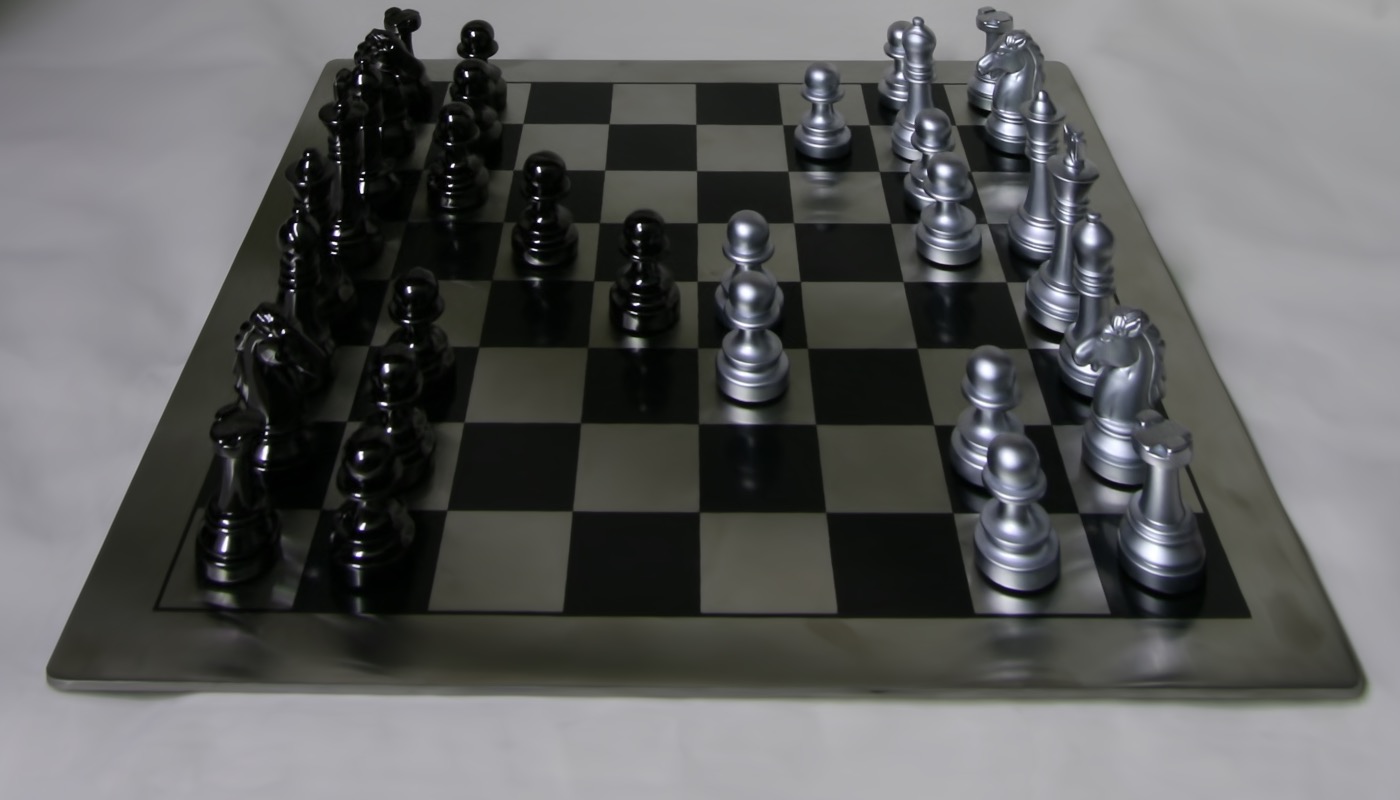}%
    ~
    \SuppSixSubfig{0 0 0 0}{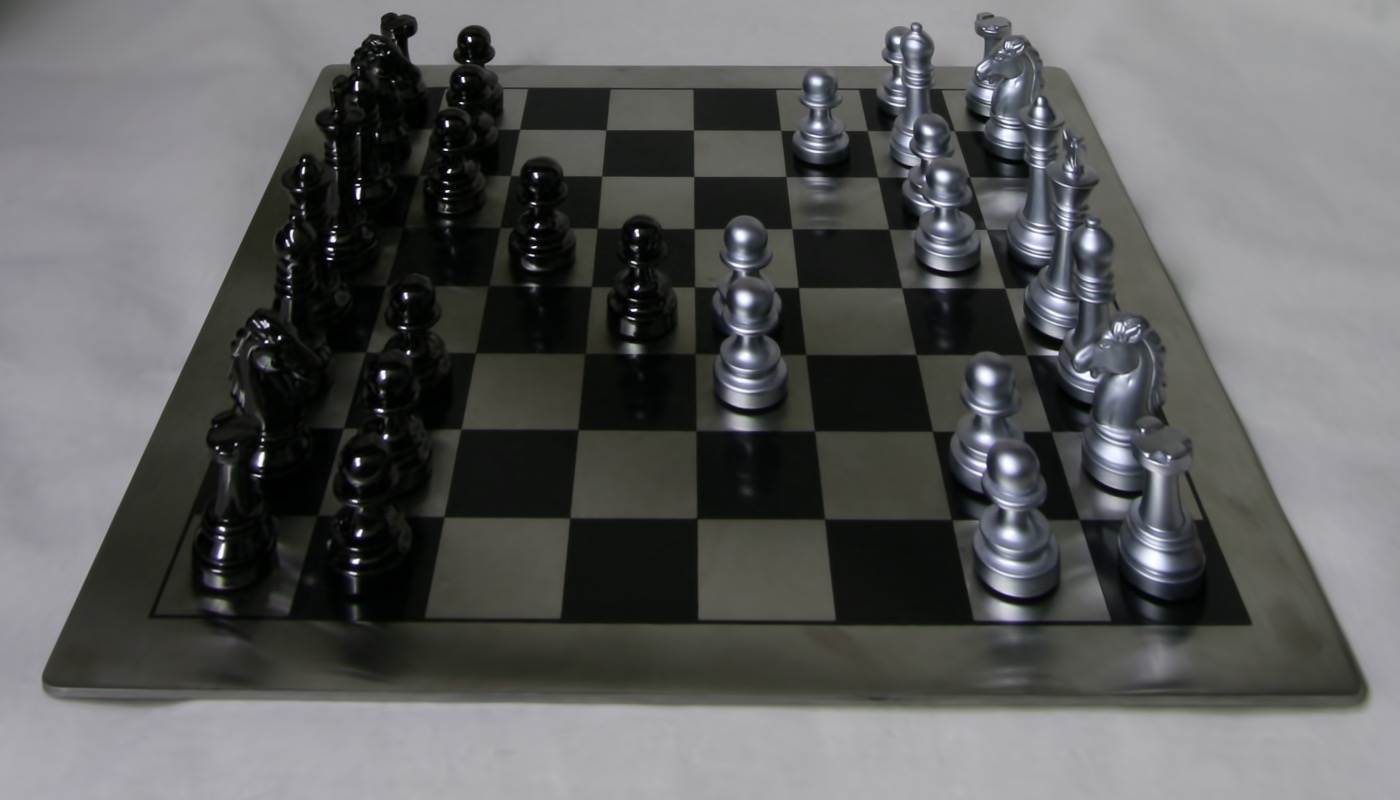}%
    
    \SuppSixSubfig{0 0 0 0}{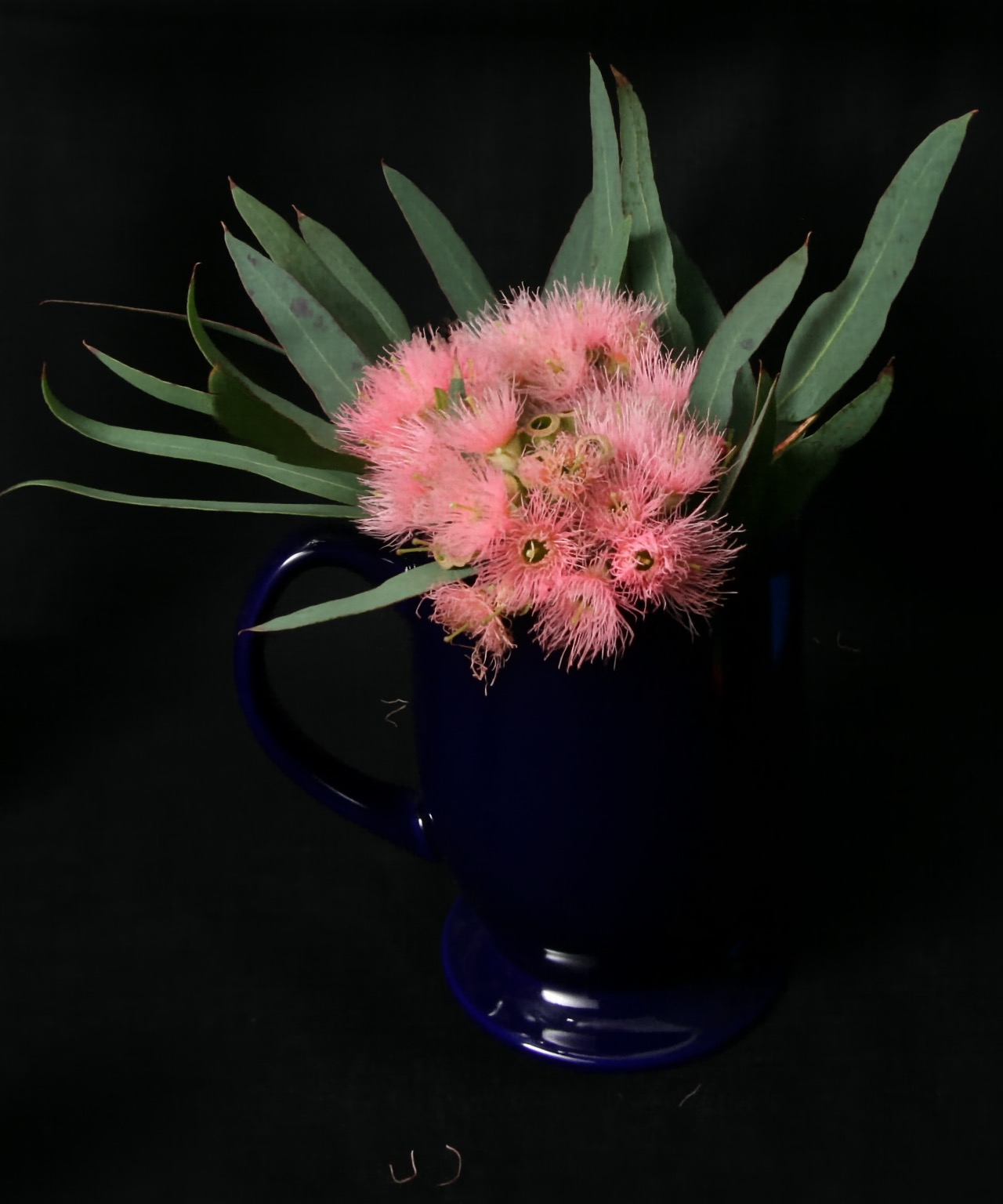}%
    ~
    \SuppSixSubfig{0 0 0 0}{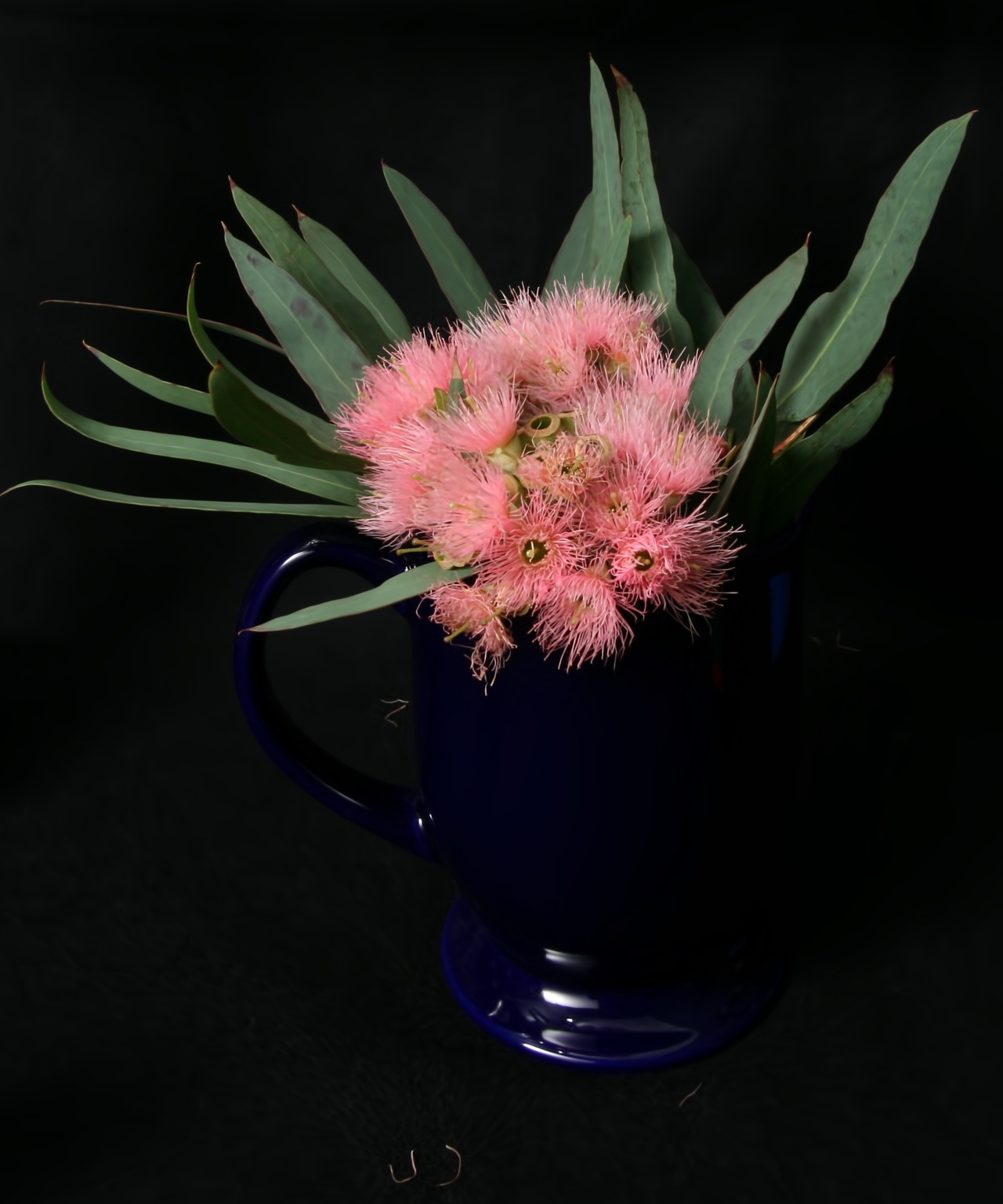}%
    ~
    \SuppSixSubfig{0 0 0 0}{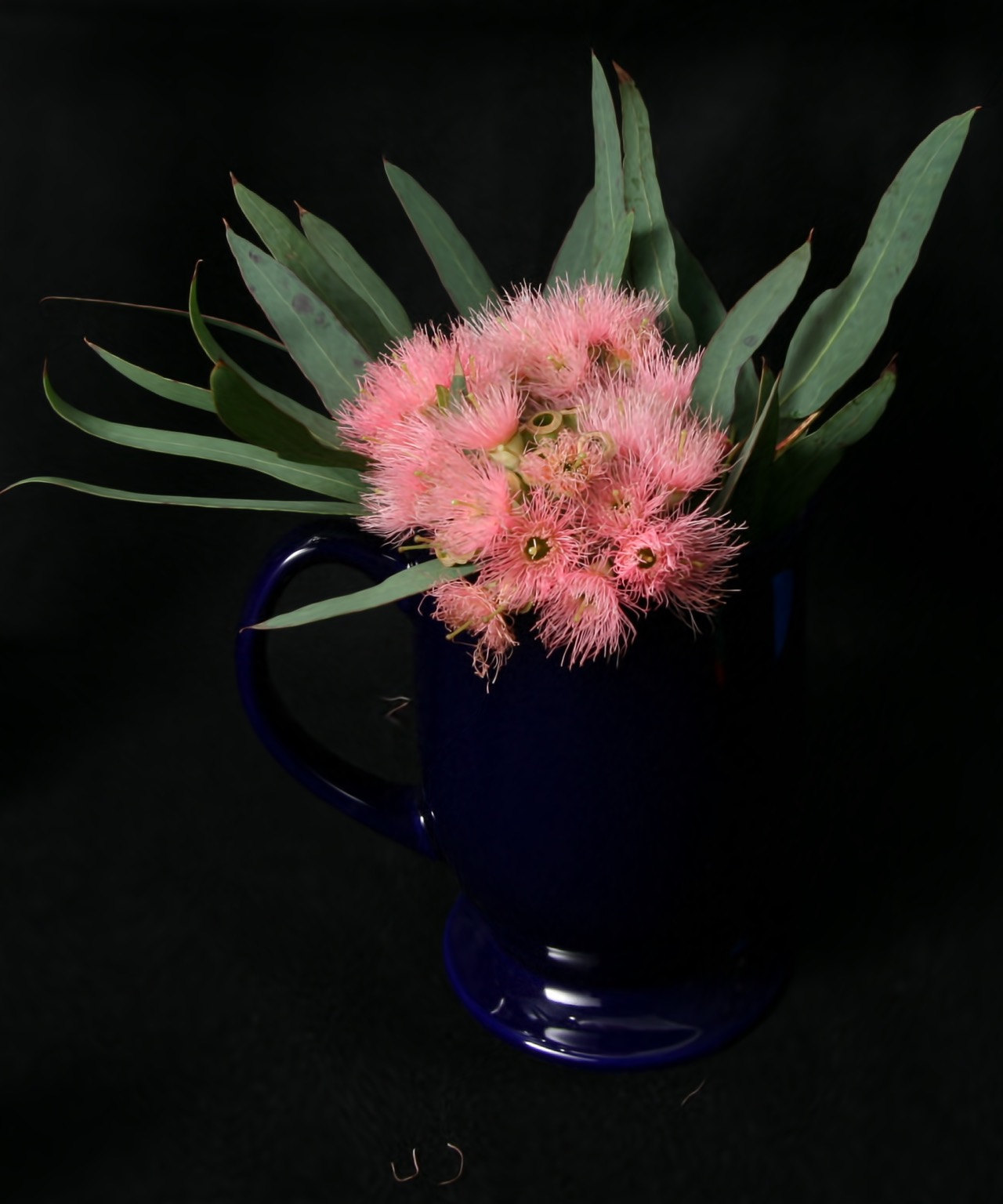}%
    ~
    \SuppSixSubfig{0 0 0 0}{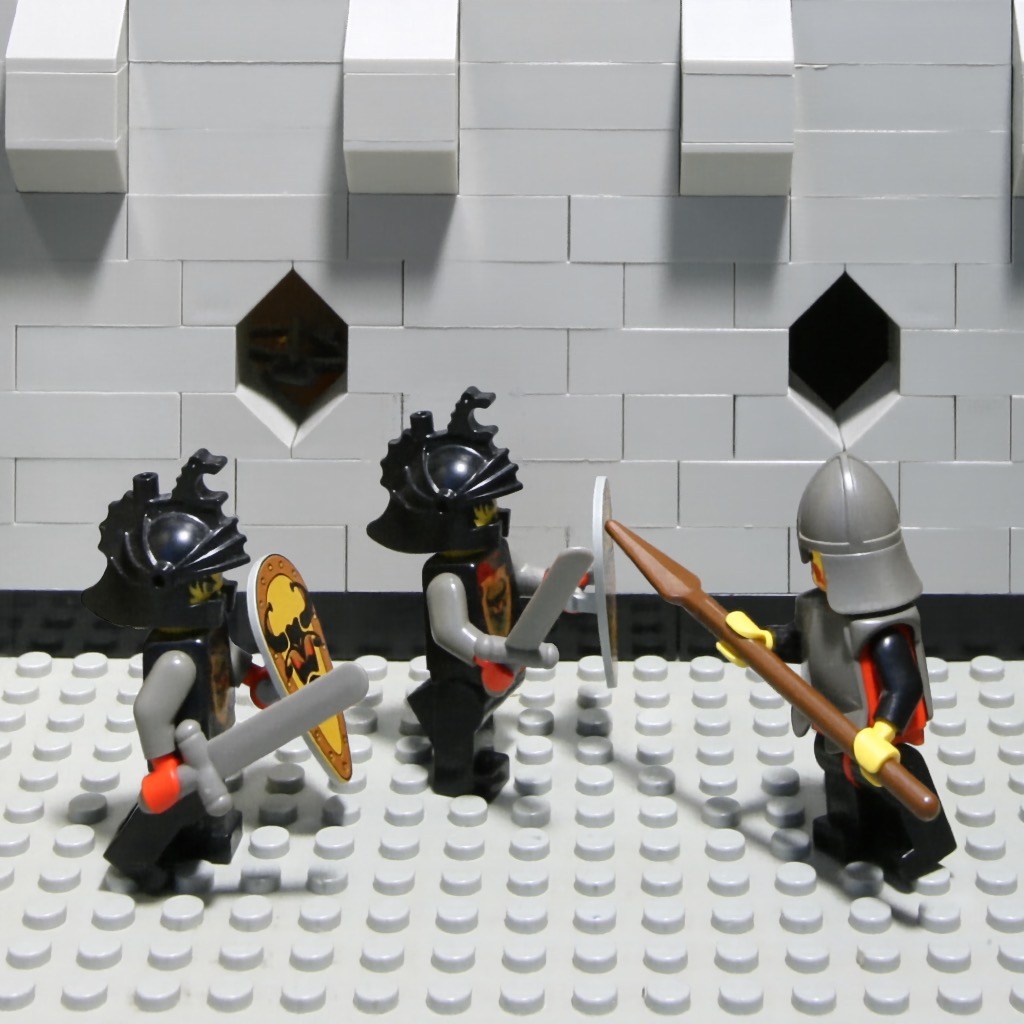}%
    ~
    \SuppSixSubfig{0 0 0 0}{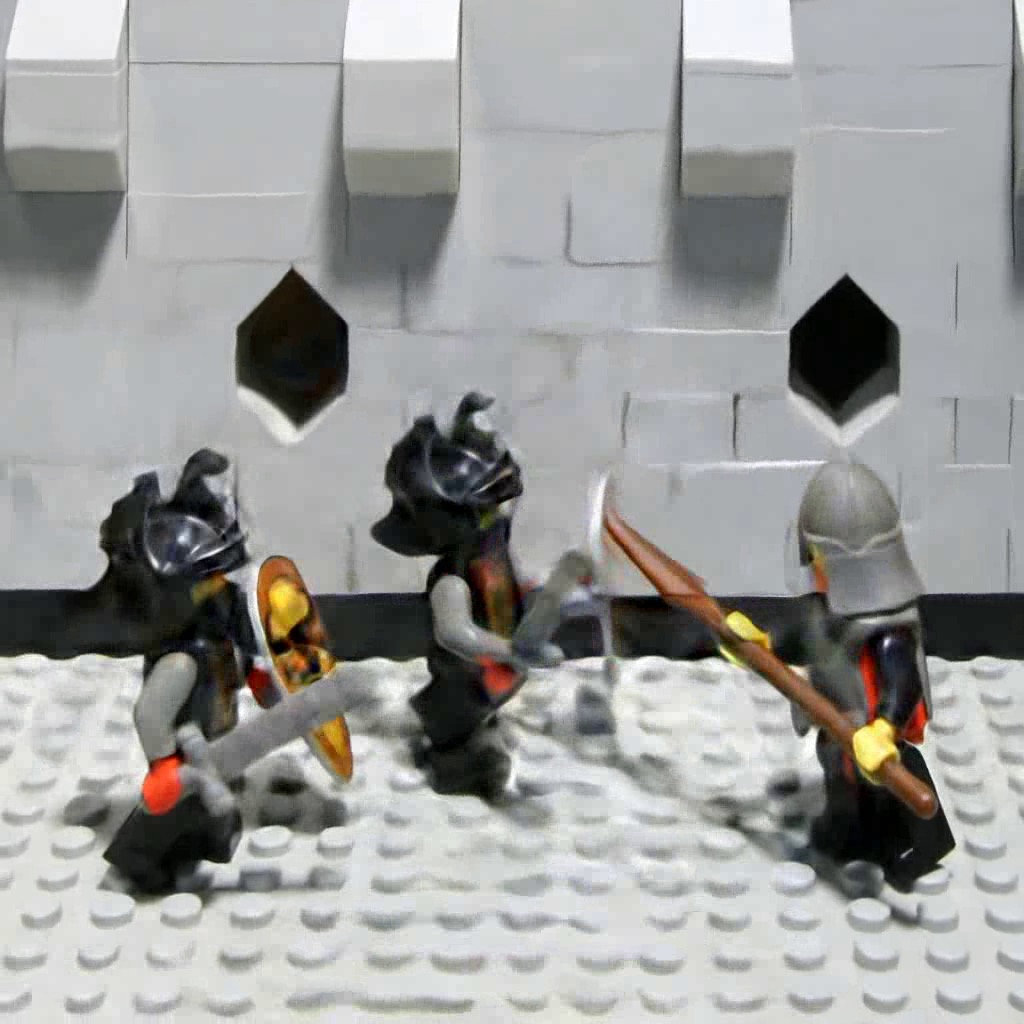}%
    ~
    \SuppSixSubfig{0 0 0 0}{./figures/4DLF/lego/14.jpeg}%
    
    \SuppSixSubfig{0 0 0 0}{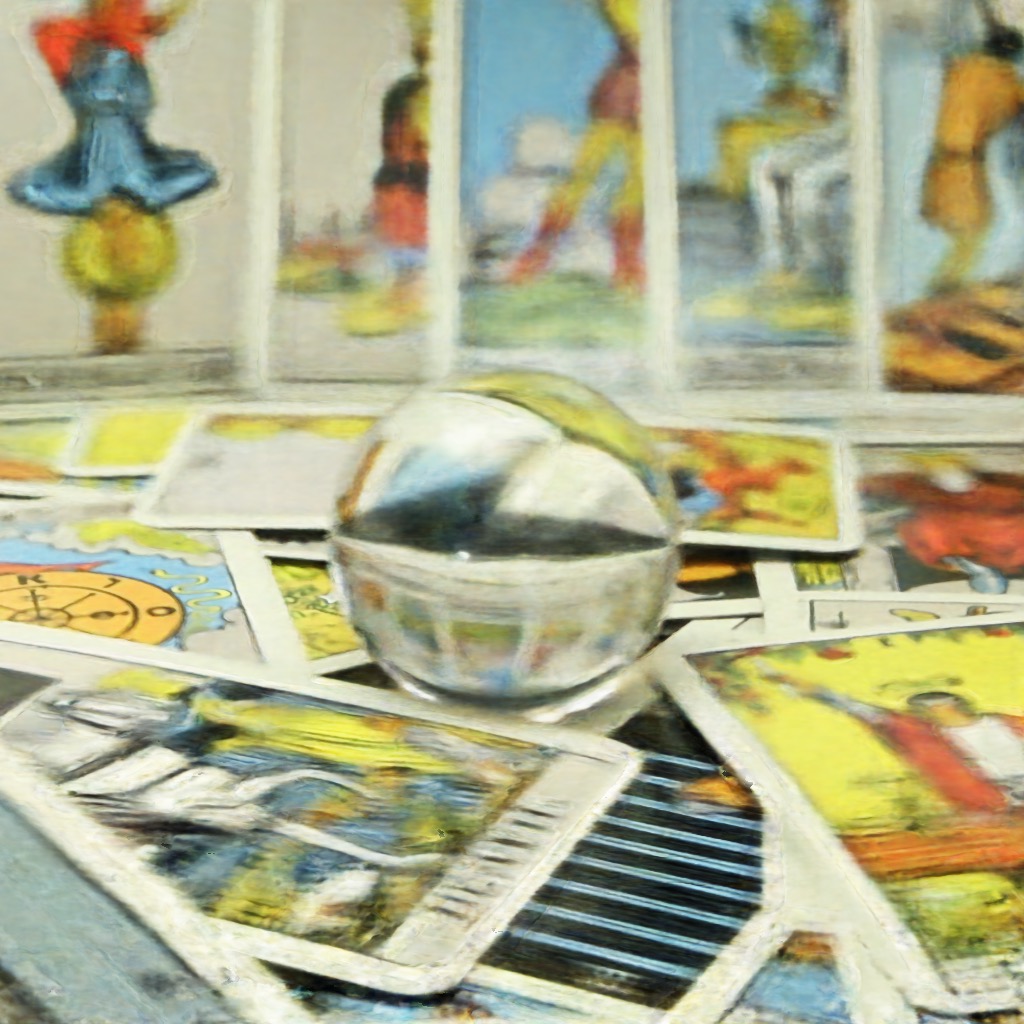}%
    ~
    \SuppSixSubfig{0 0 0 0}{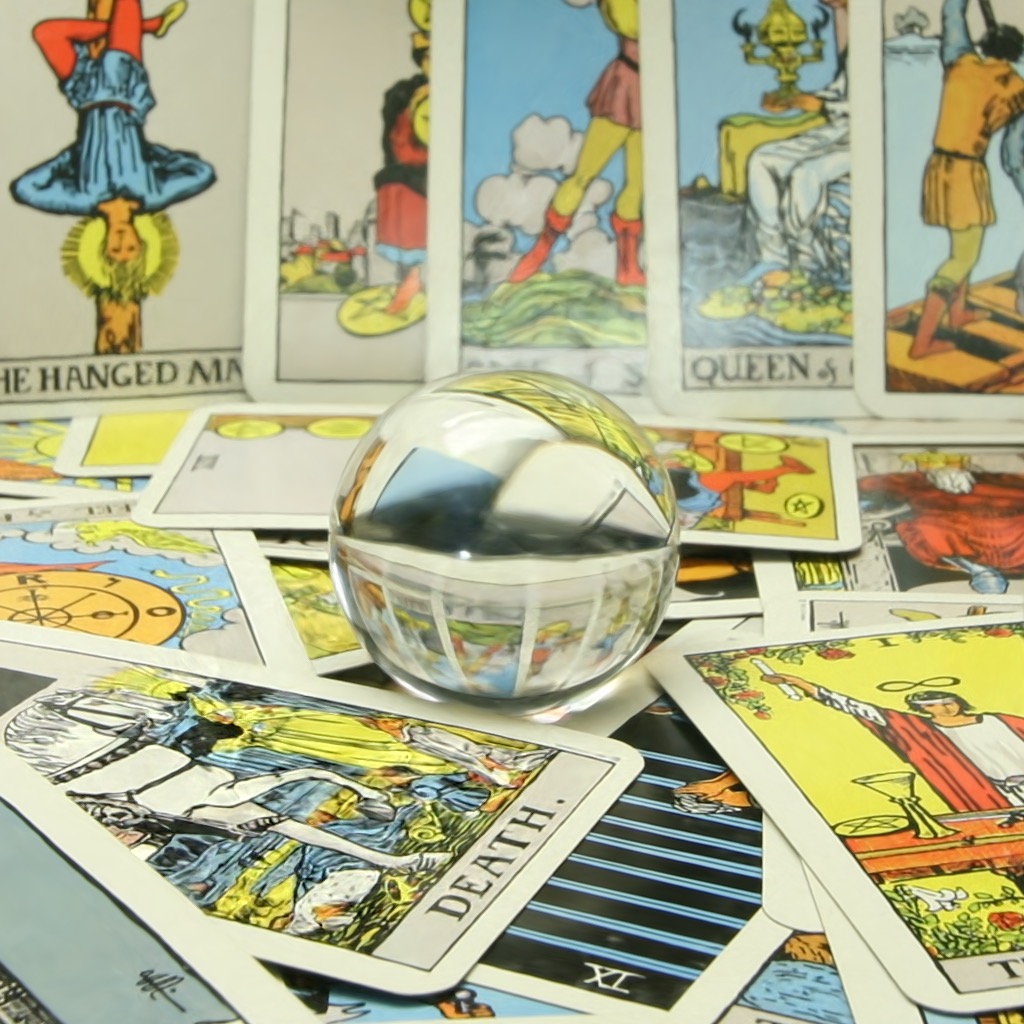}%
    ~
    \SuppSixSubfig{0 0 0 0}{./figures/4DLF/tarot/14.jpeg}%
    ~
    \SuppSixSubfig{0 0 0 0}{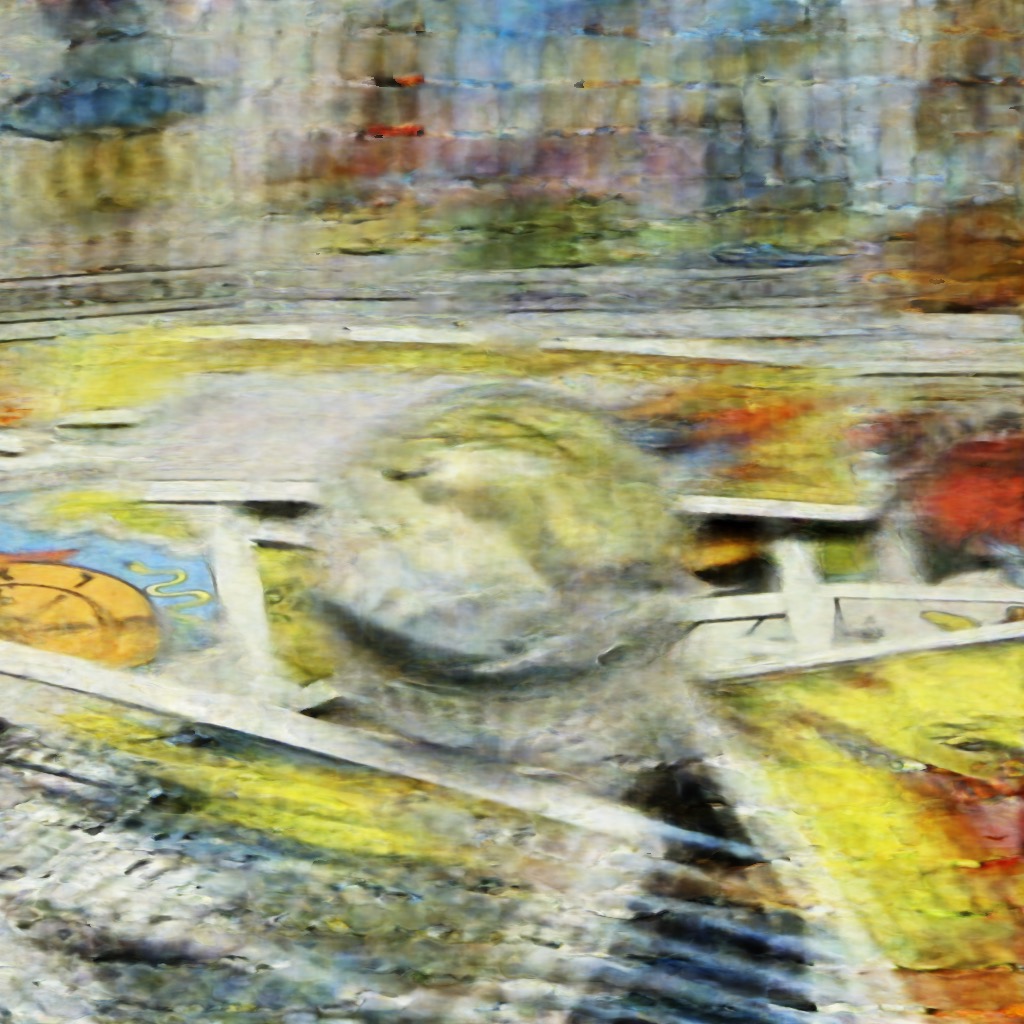}%
    ~
    \SuppSixSubfig{0 0 0 0}{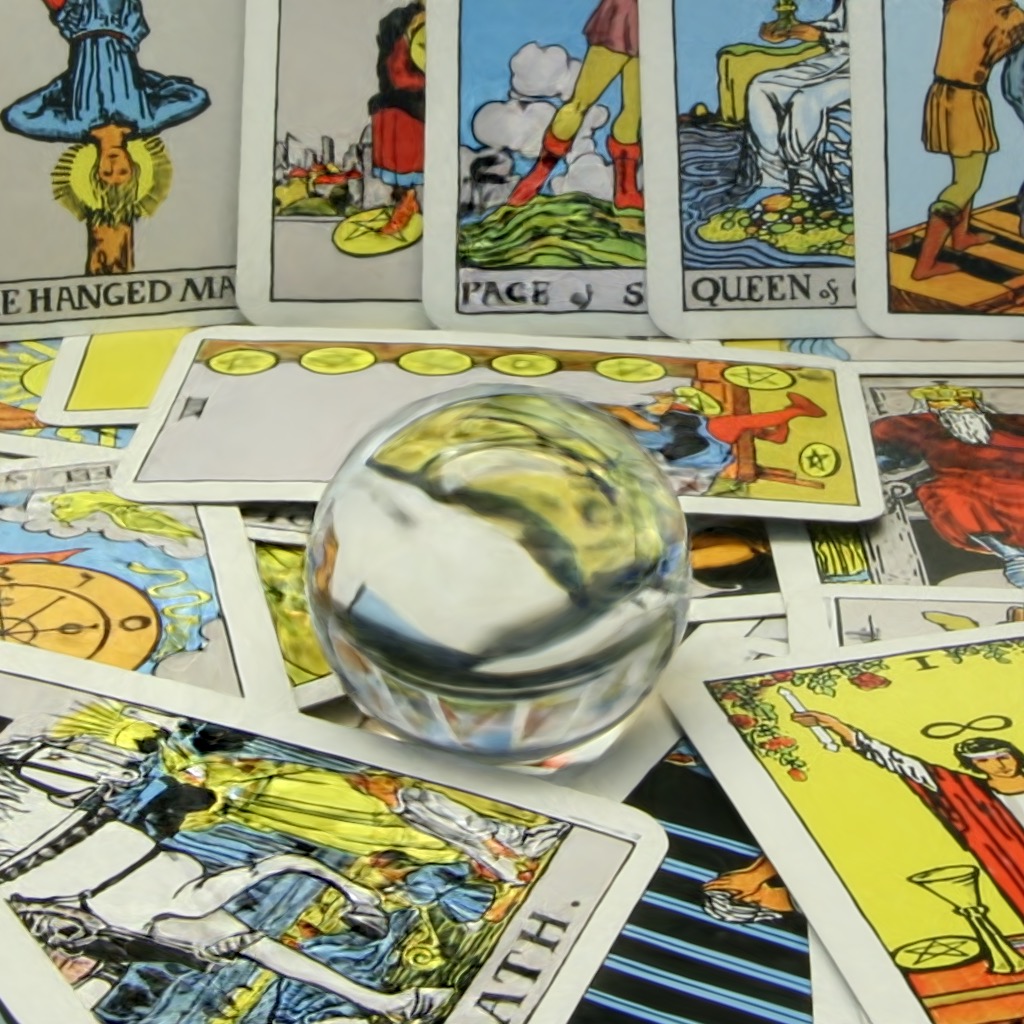}%
    ~
    \SuppSixSubfig{0 0 0 0}{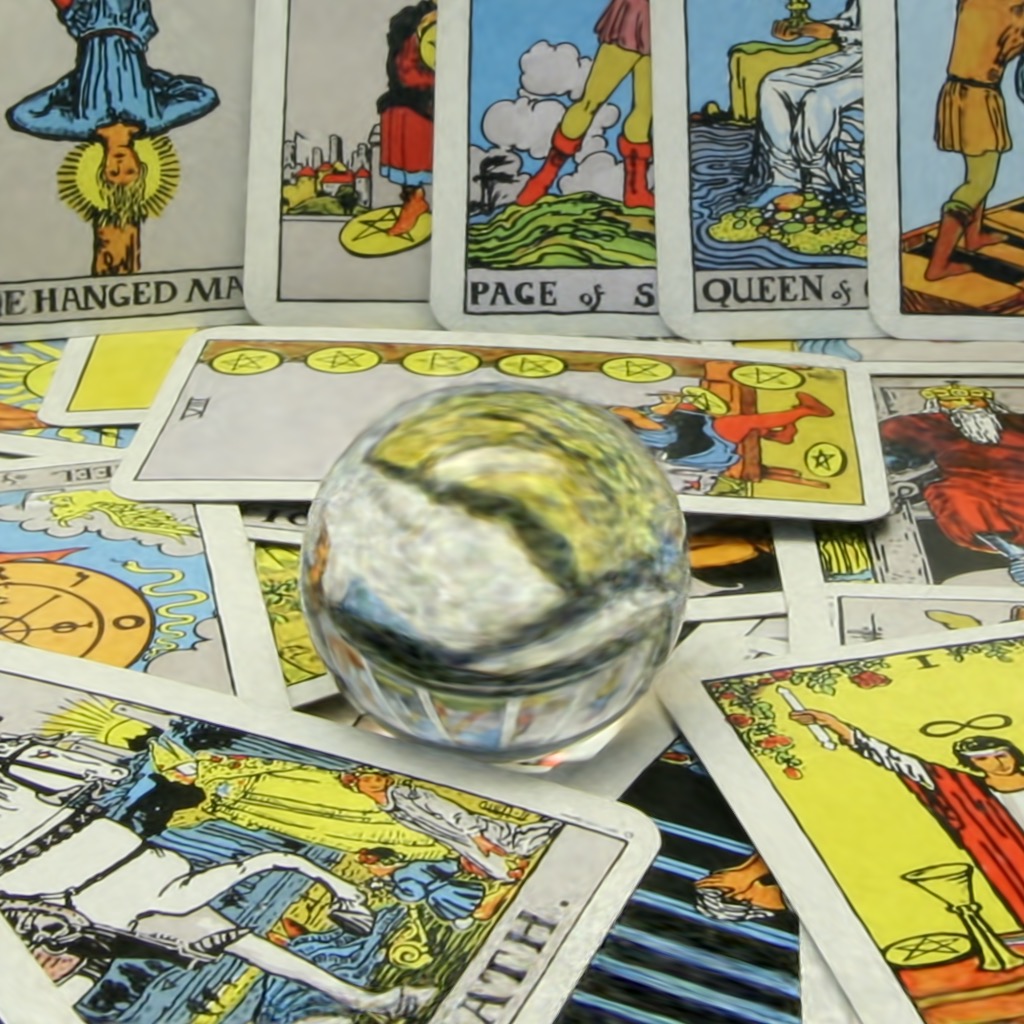}%
    
    \SuppSixSubfig{0 0 0 0}{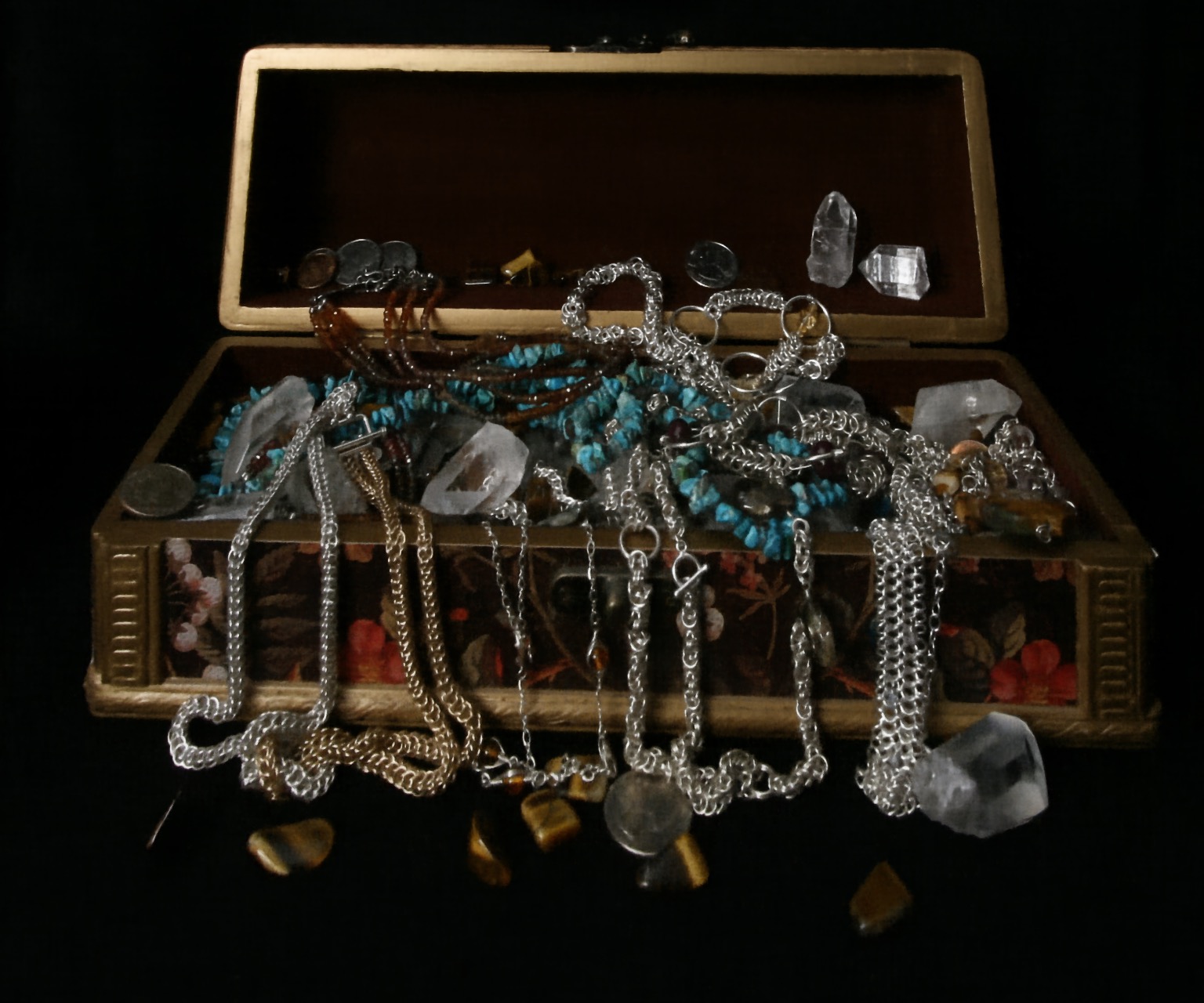}%
    ~
    \SuppSixSubfig{0 0 0 0}{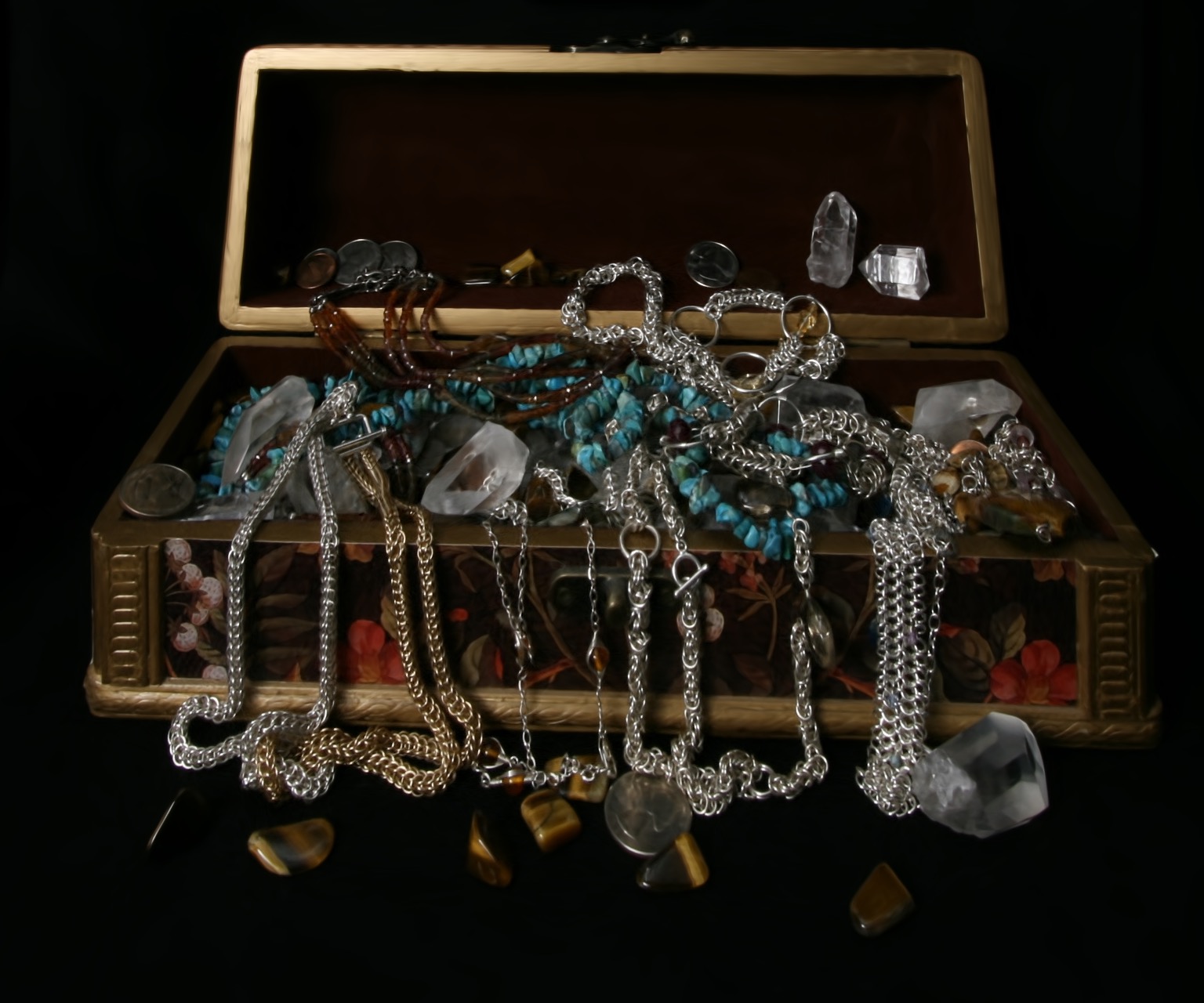}%
    ~
    \SuppSixSubfig{0 0 0 0}{./figures/4DLF/treasure/14.jpeg}%
    ~
    \SuppSixSubfig{0 0 0 0}{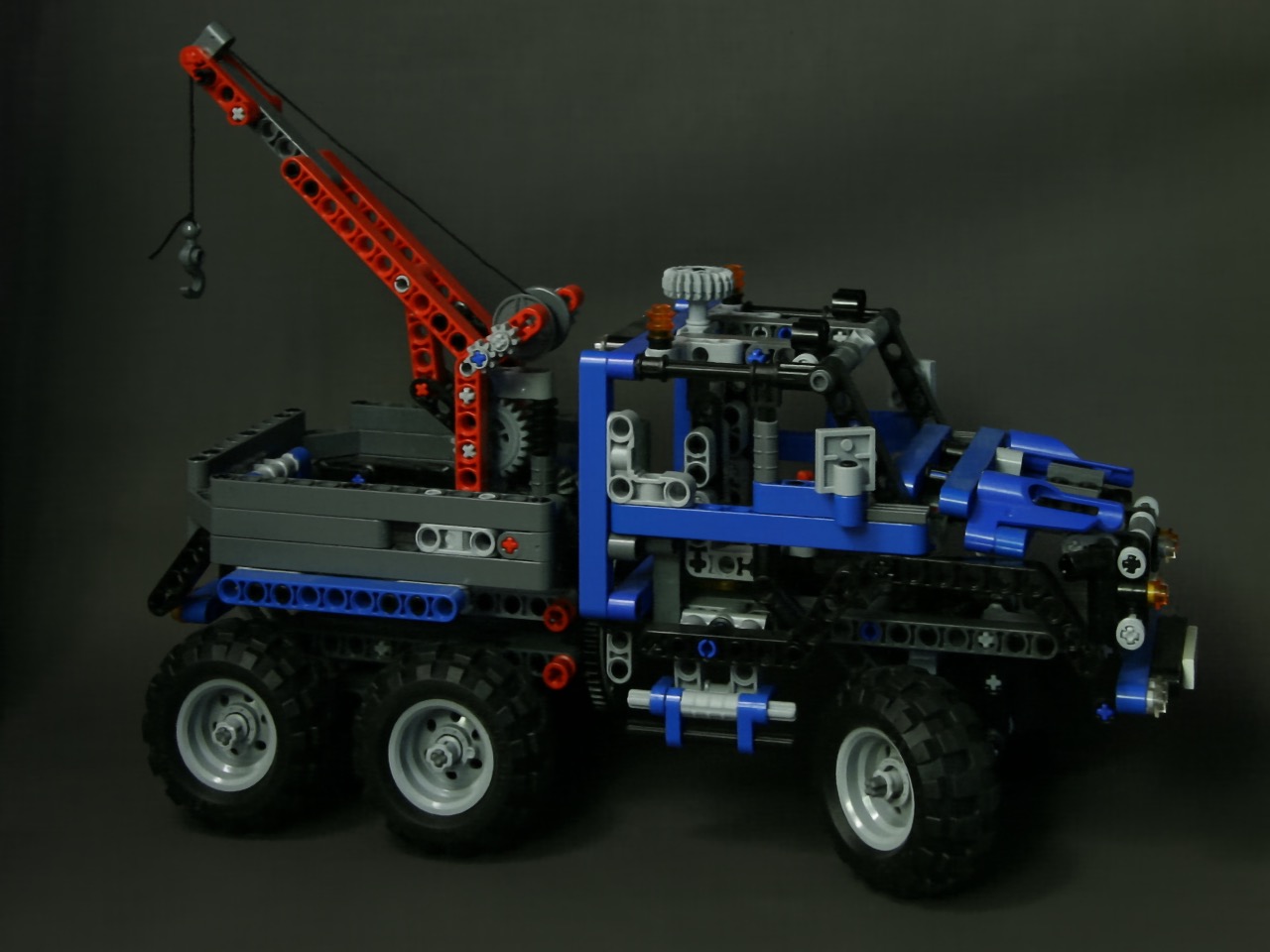}%
    ~
    \SuppSixSubfig{0 0 0 0}{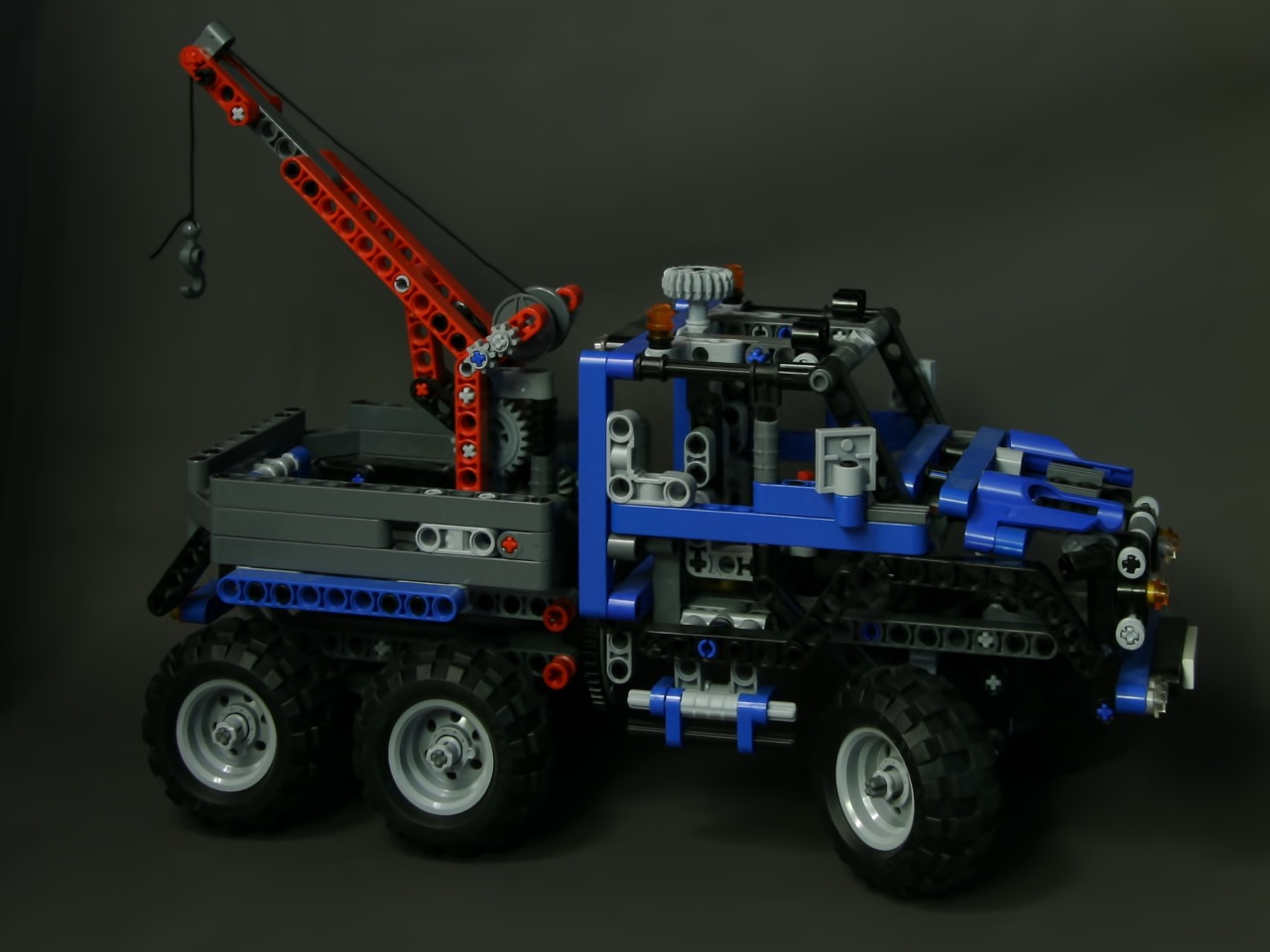}%
    ~
    \SuppSixSubfig{0 0 0 0}{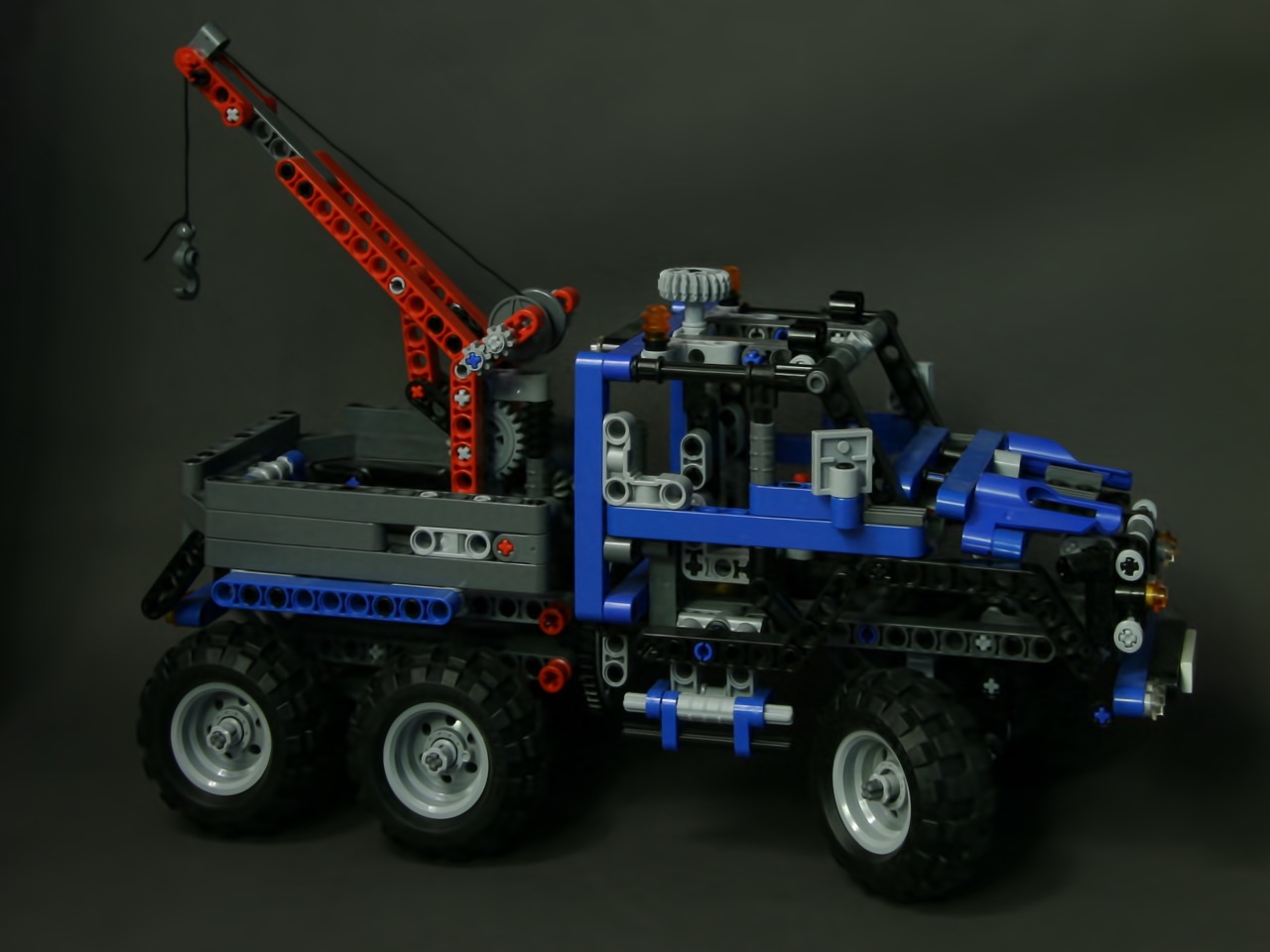}%
    \vspace{-5pt}
	\caption{Comparison With Baselines. From left to right: NeRF, LFN, and Ours.}
    \label{fig:SuppCompareBegin}
\end{figure*}

\begin{figure*}[!h]
    \SuppThreeSubfig{200 0 300 0}{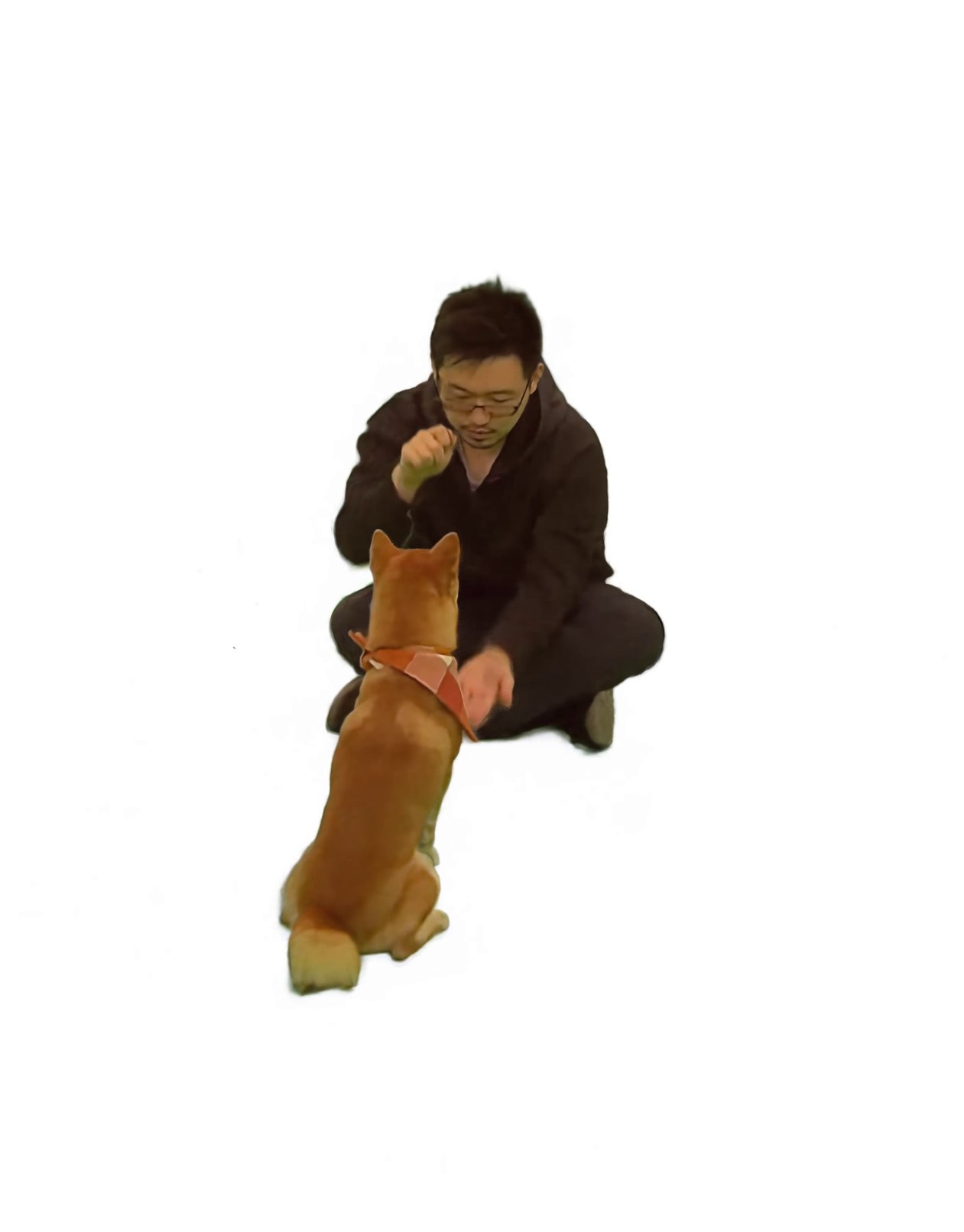}%
    ~
    \SuppThreeSubfig{200 0 300 0}{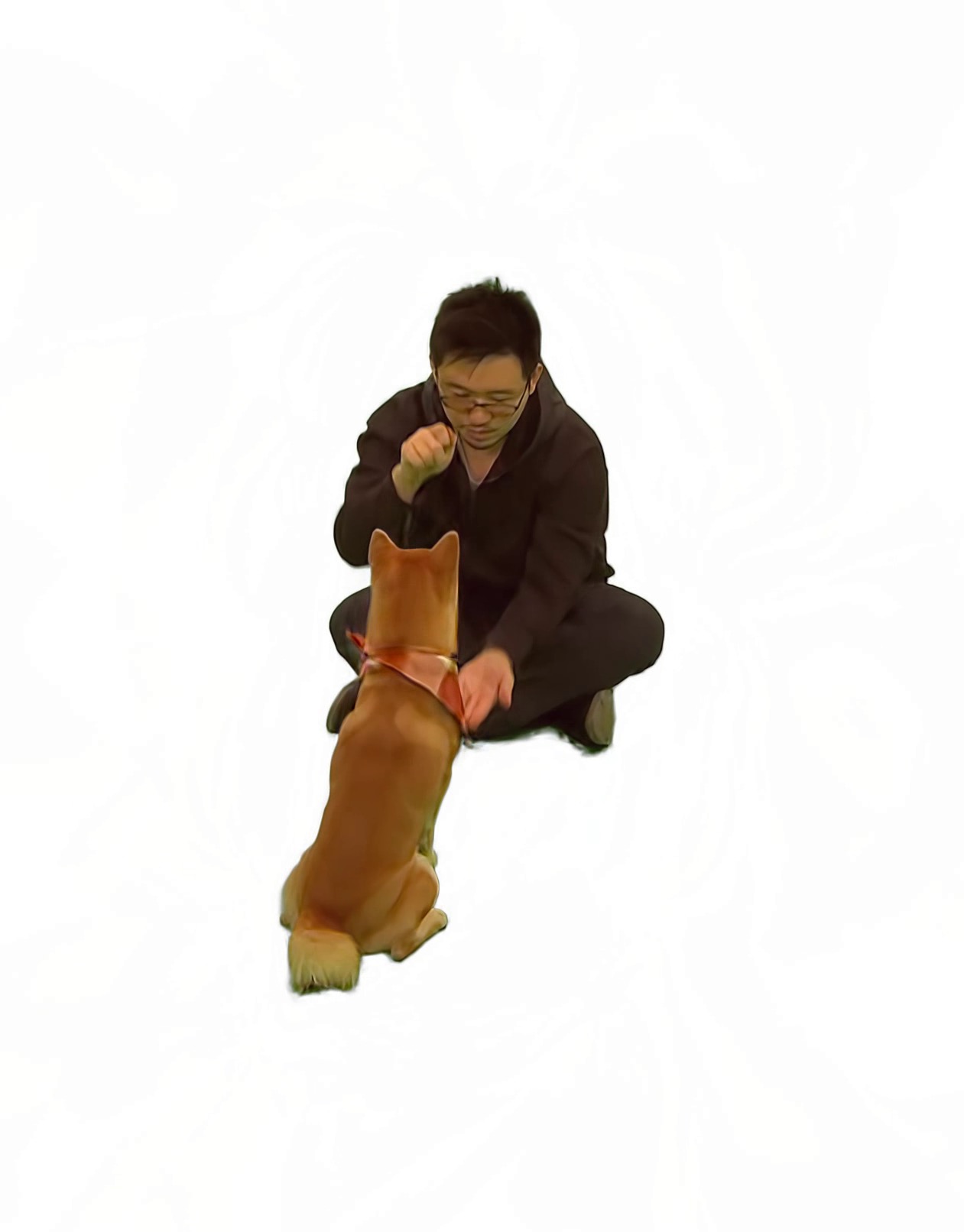}%
    ~
    \SuppThreeSubfig{200 0 300 0}{./figures/Holo/Shiba/02.jpeg}%

    \SuppThreeSubfig{0 800 0 100}{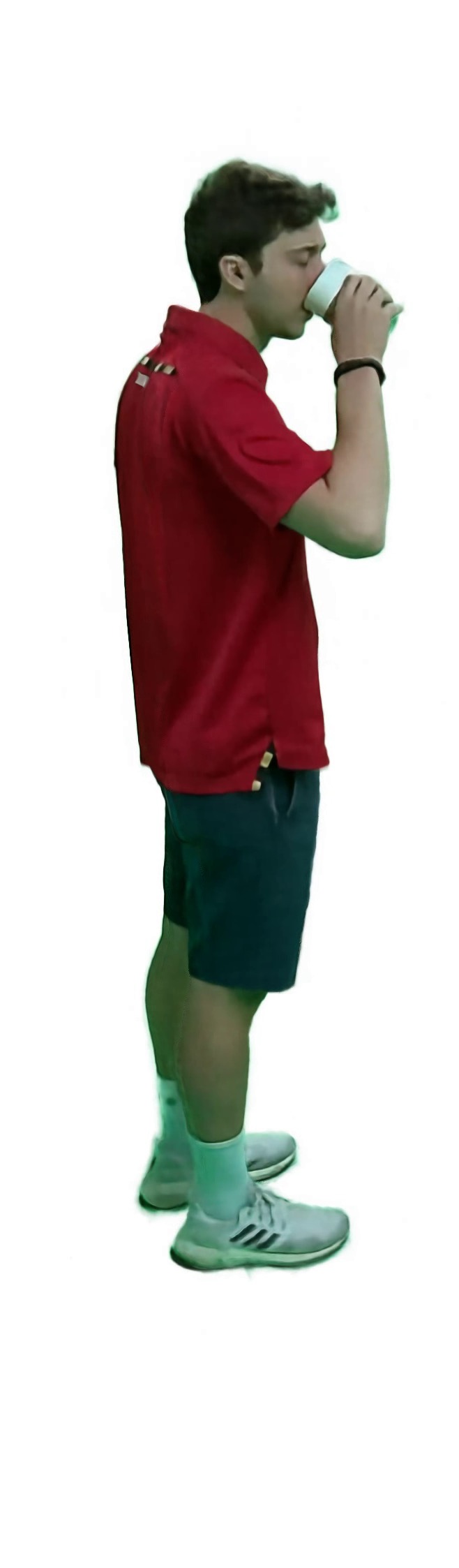}%
    ~
    \SuppThreeSubfig{0 800 0 100}{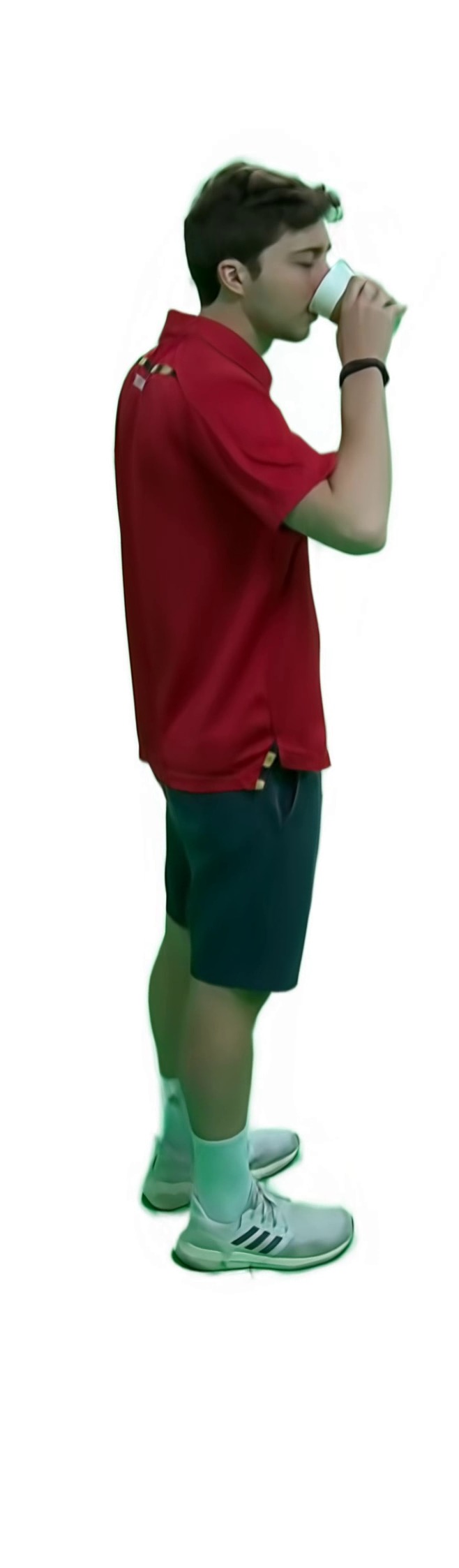}%
    ~
    \SuppThreeSubfig{0 800 0 100}{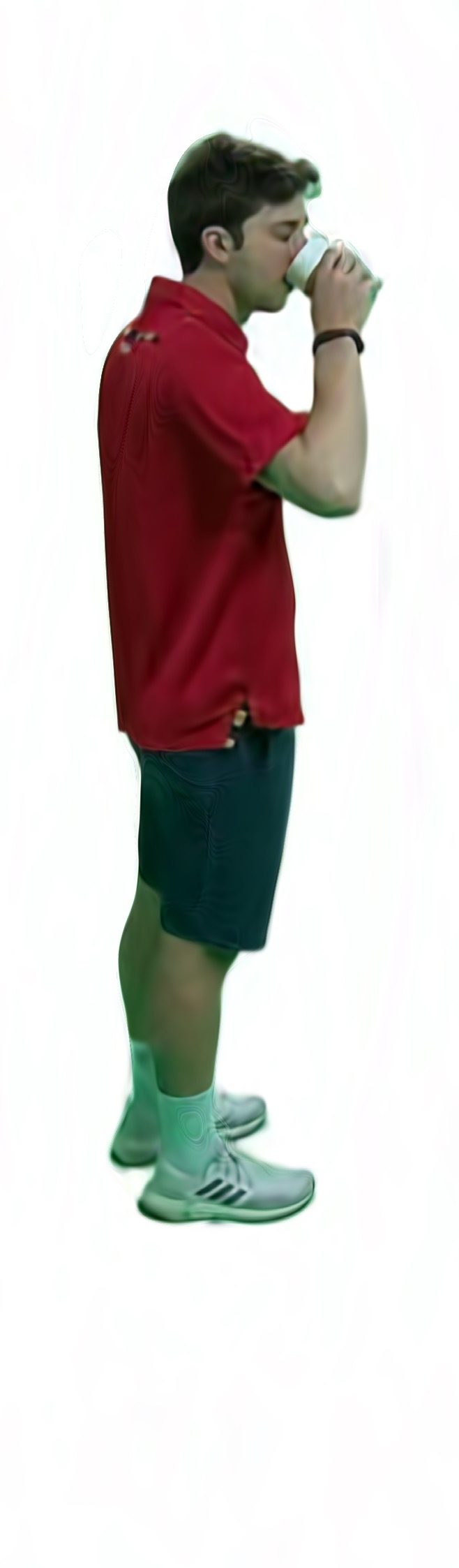}%

    \SuppThreeSubfig{0 0 0 0}{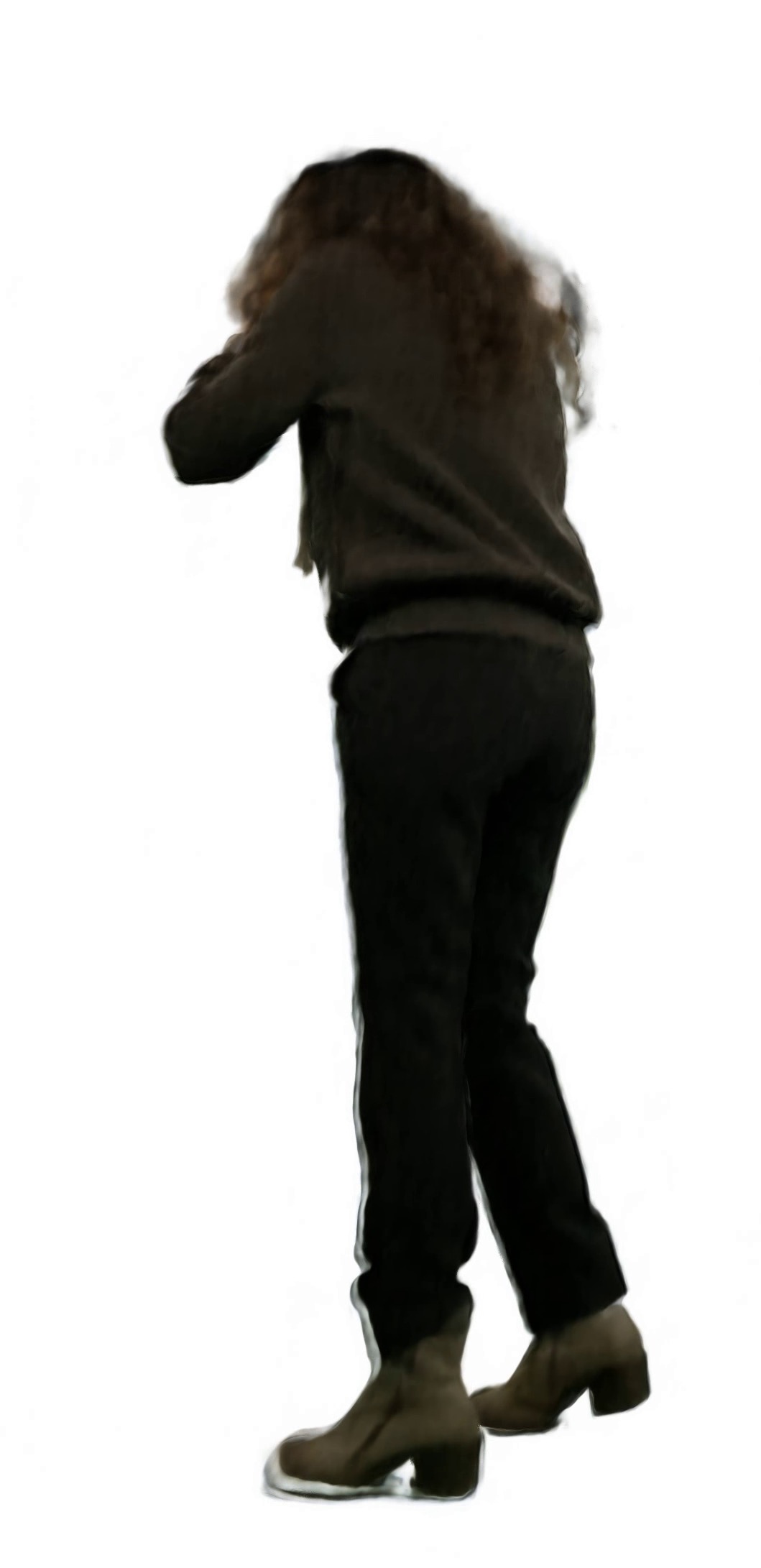}%
    ~
    \SuppThreeSubfig{0 0 0 0}{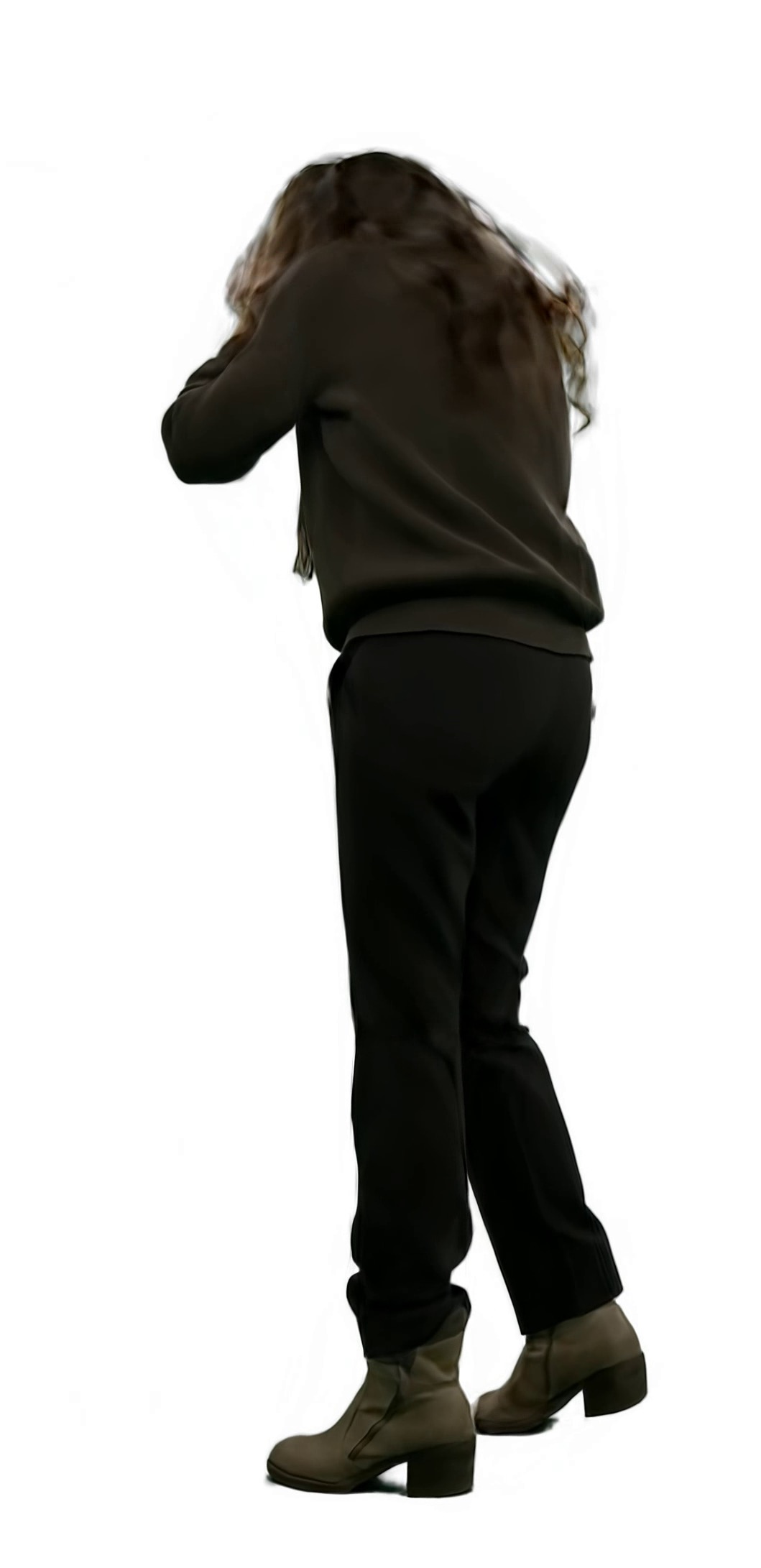}%
    ~
    \SuppThreeSubfig{0 0 0 0}{./figures/Holo/Maria/02.jpeg}%
    
    \vspace{-15pt}
	\caption{Comparison With Baselines. From left to right: NeRF, LFN, and Ours. The difference in perspective is partially due to the difference between 3D-based viewpoint movement v.s. 2D-based image morphing.}
    \label{fig:SuppCompareEnd}

\end{figure*}

\begin{figure*}[!h]
    \SuppLFSubfig{0 0 0 0}{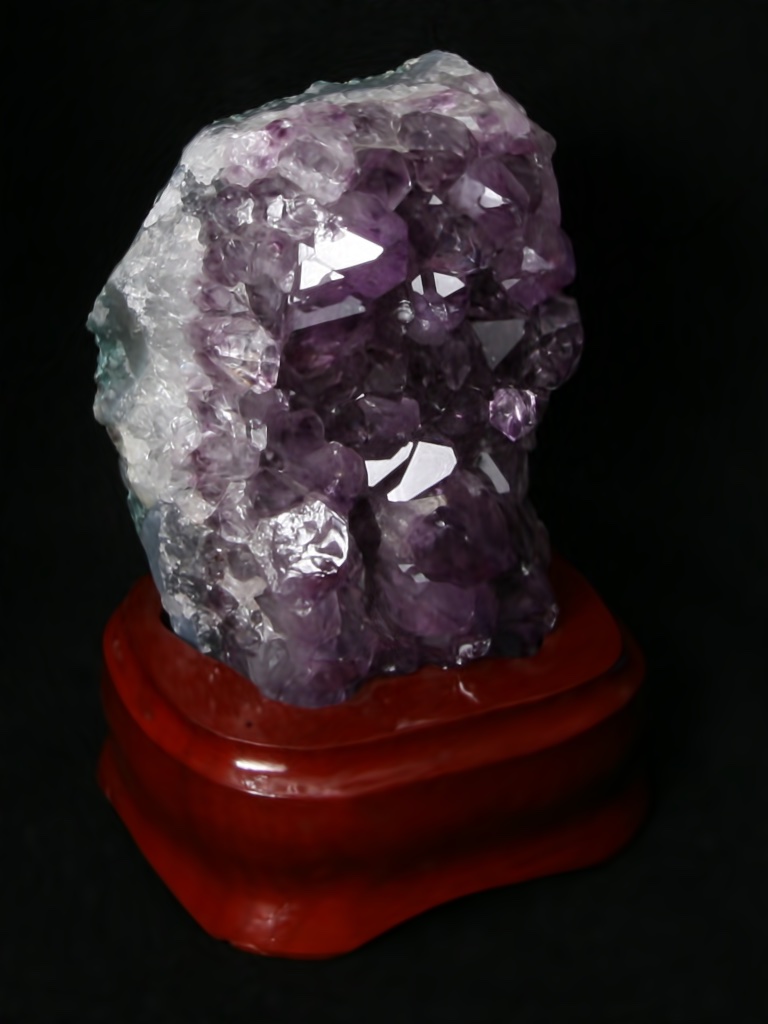}%
    ~
    \SuppLFSubfig{0 0 0 0}{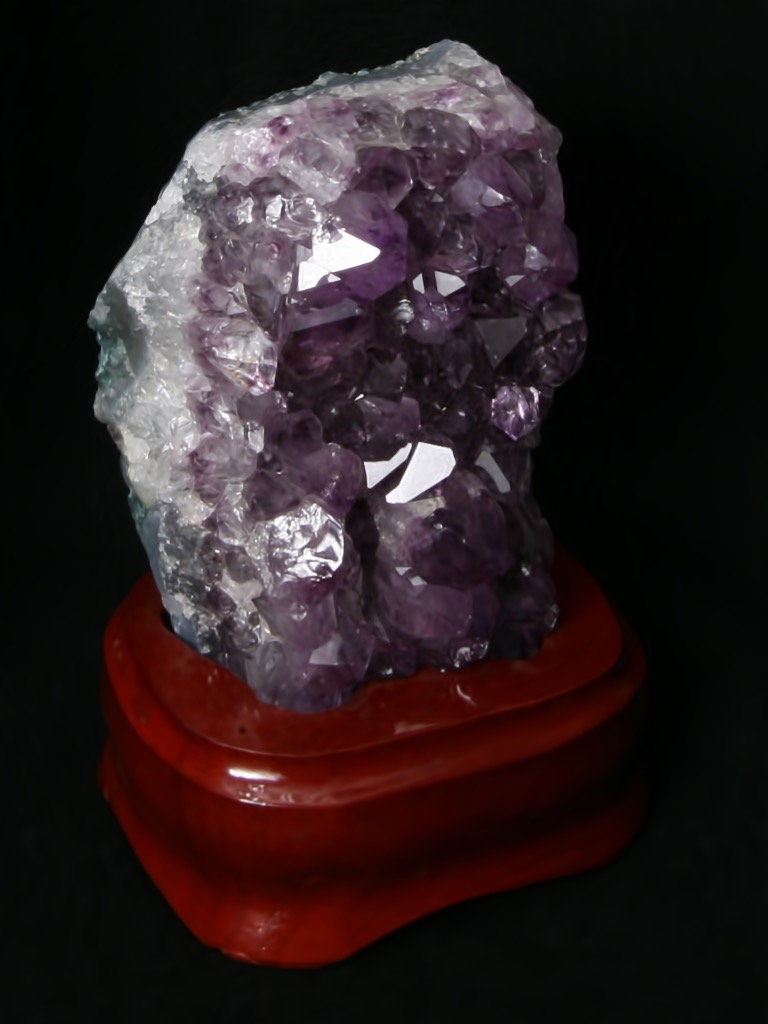}%
    ~
    \SuppLFSubfig{0 0 0 0}{./figures/4DLF/amethyst/14.jpeg}%
    ~
    \SuppLFSubfig{0 0 0 0}{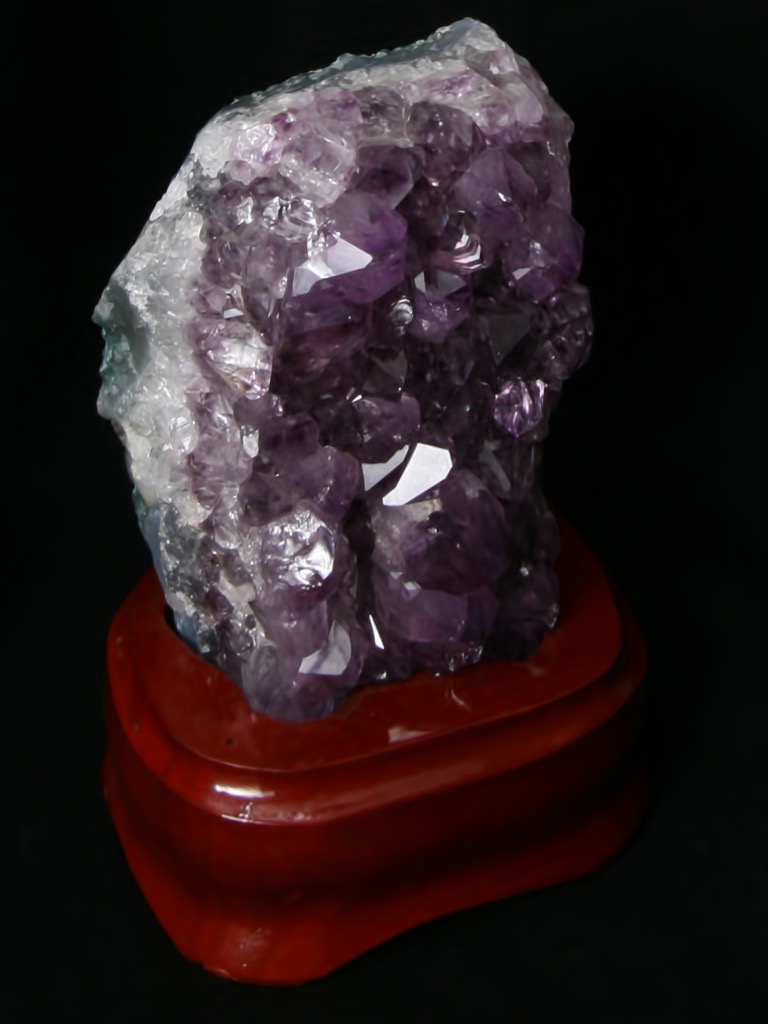}%
    ~
    \SuppLFSubfig{0 0 0 0}{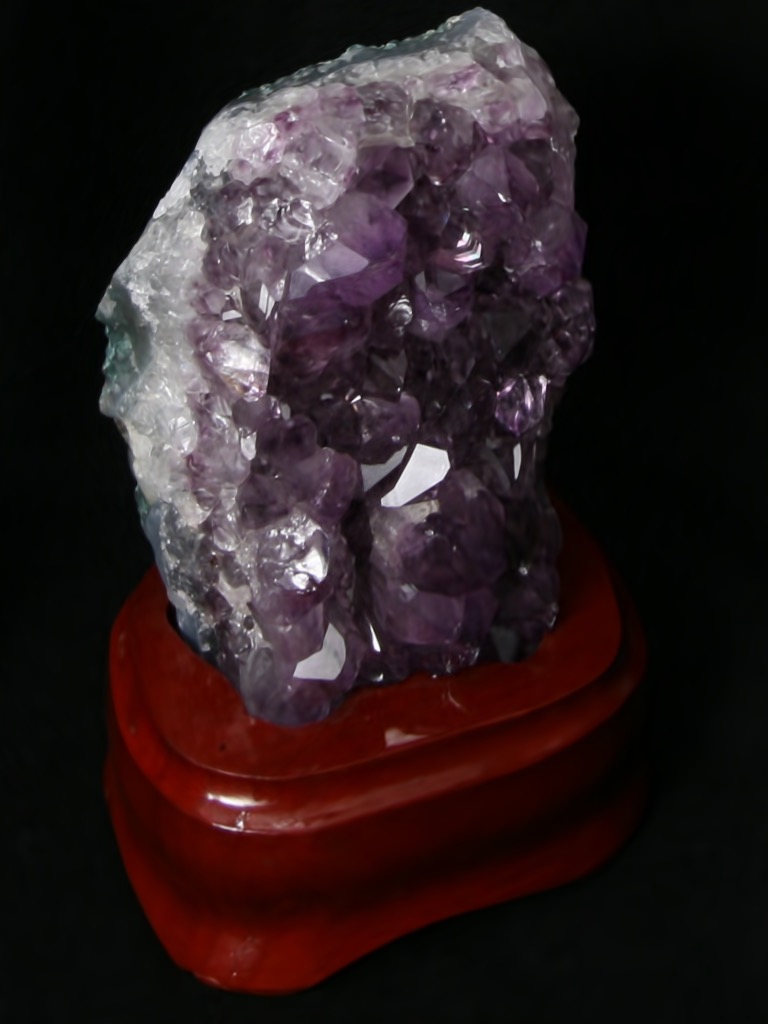}%

    \SuppLFSubfig{0 0 0 0}{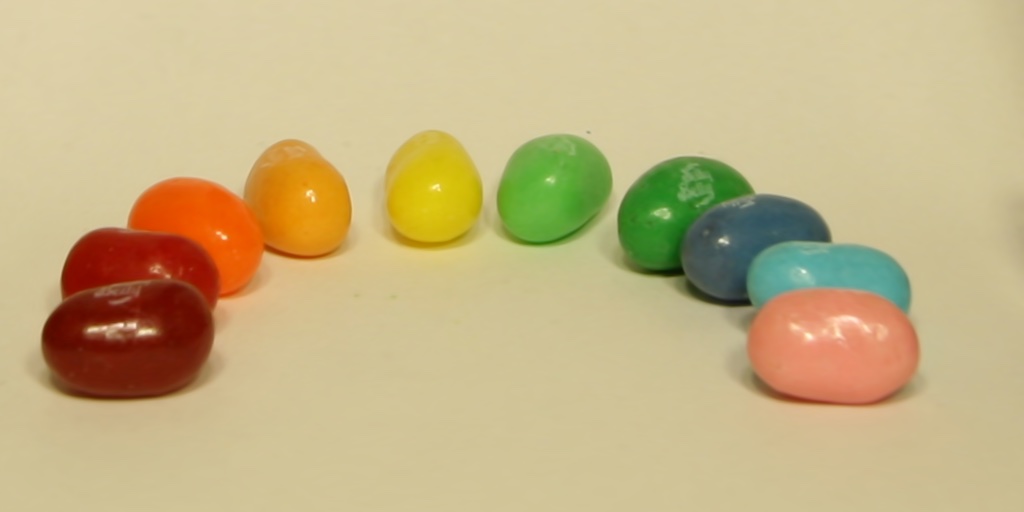}%
    ~
    \SuppLFSubfig{0 0 0 0}{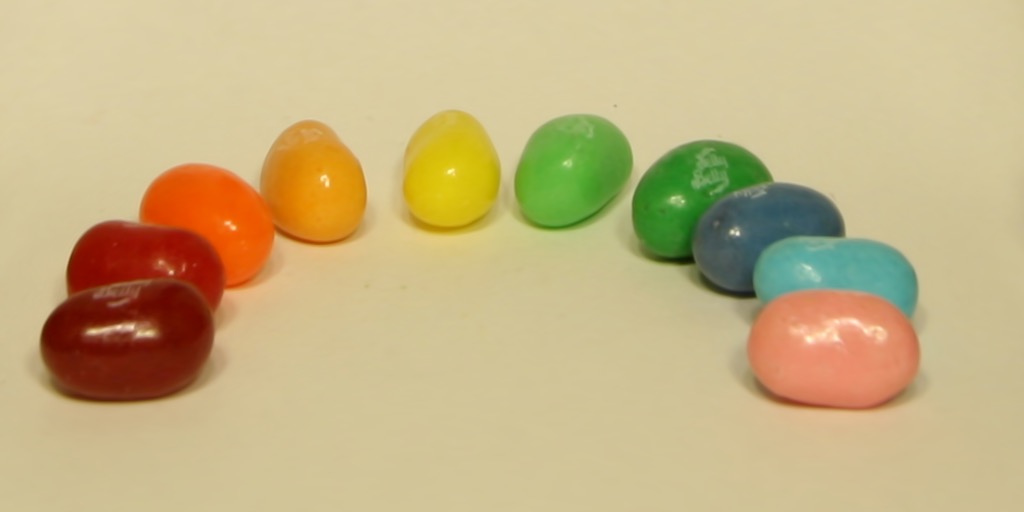}%
    ~
    \SuppLFSubfig{0 0 0 0}{./figures/4DLF/beans/14.jpeg}%
    ~
    \SuppLFSubfig{0 0 0 0}{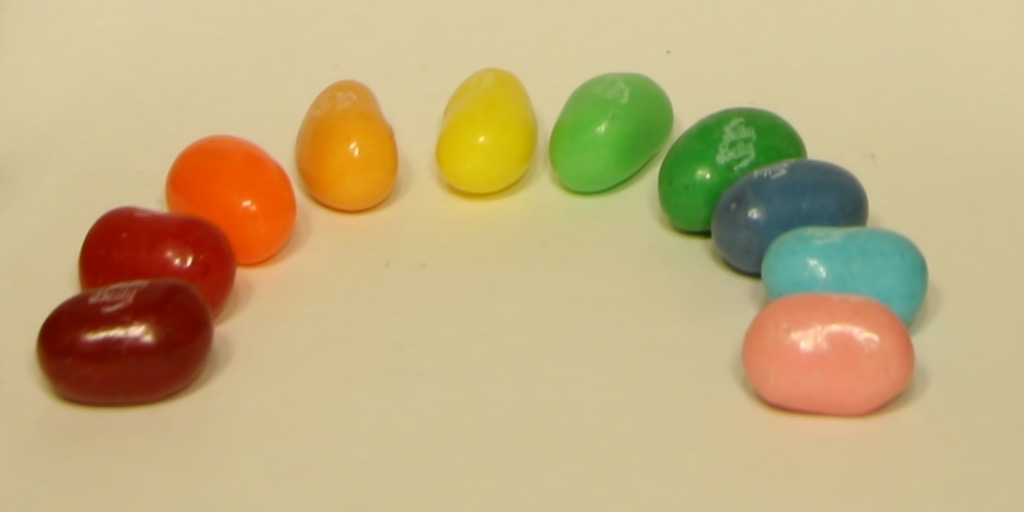}%
    ~
    \SuppLFSubfig{0 0 0 0}{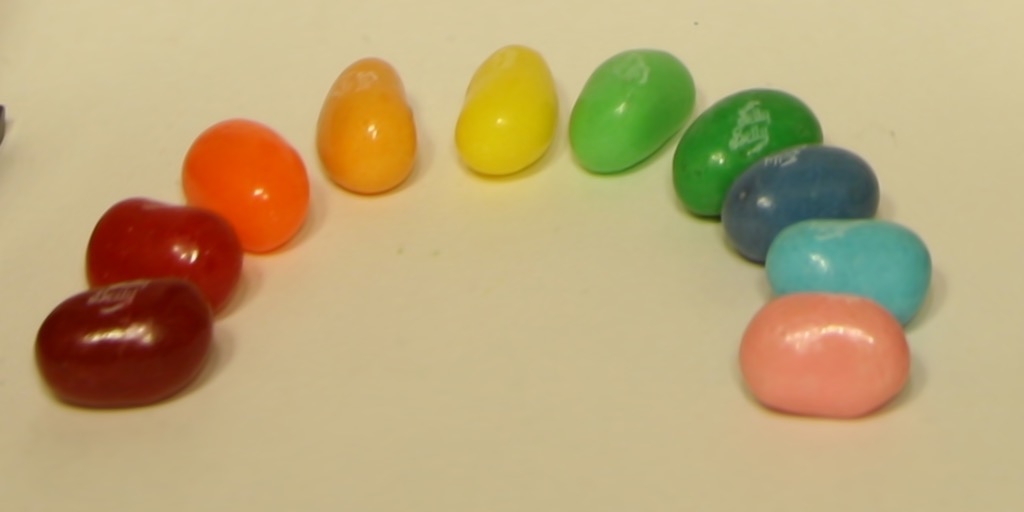}%

    \SuppLFSubfig{0 0 0 0}{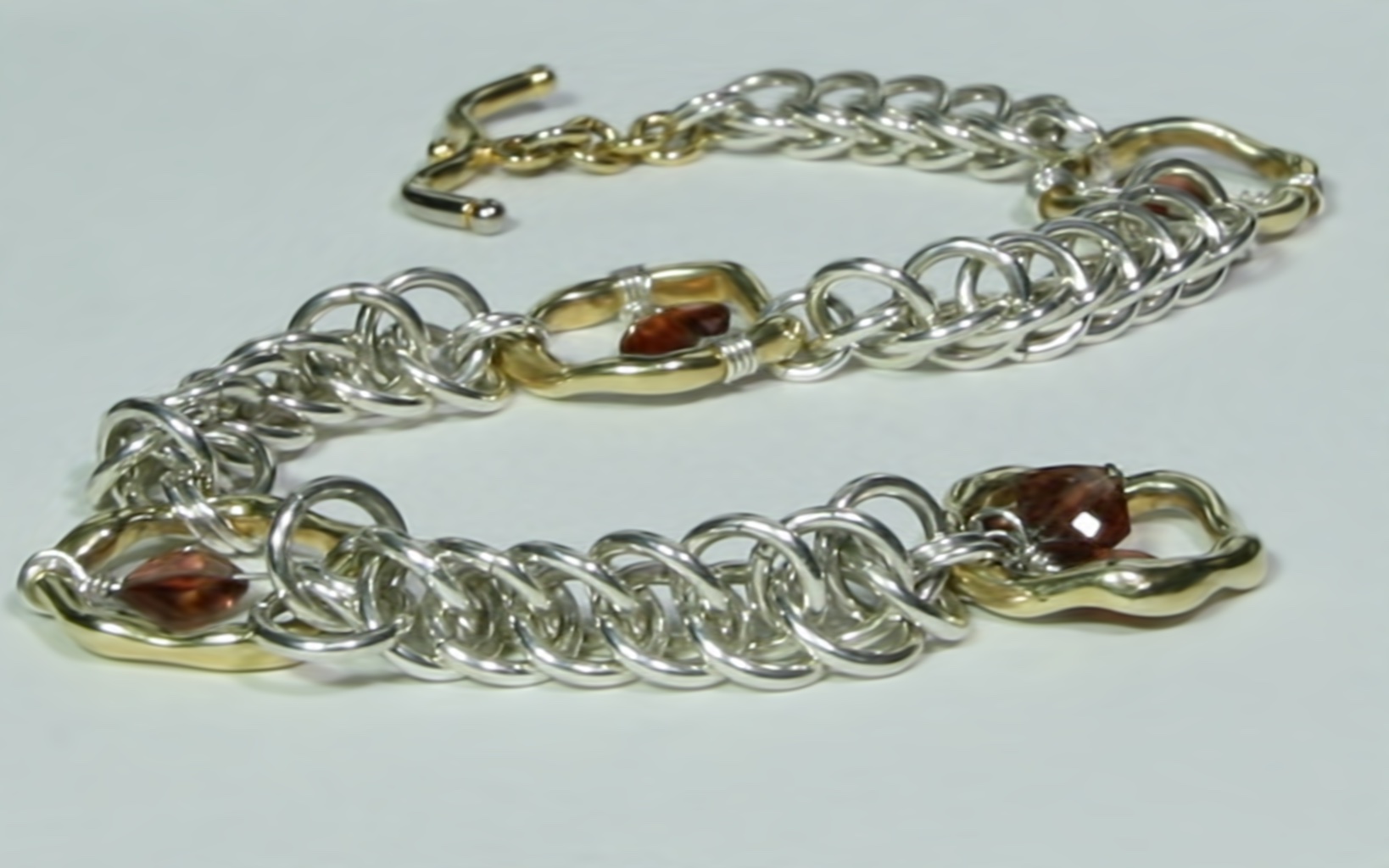}%
    ~
    \SuppLFSubfig{0 0 0 0}{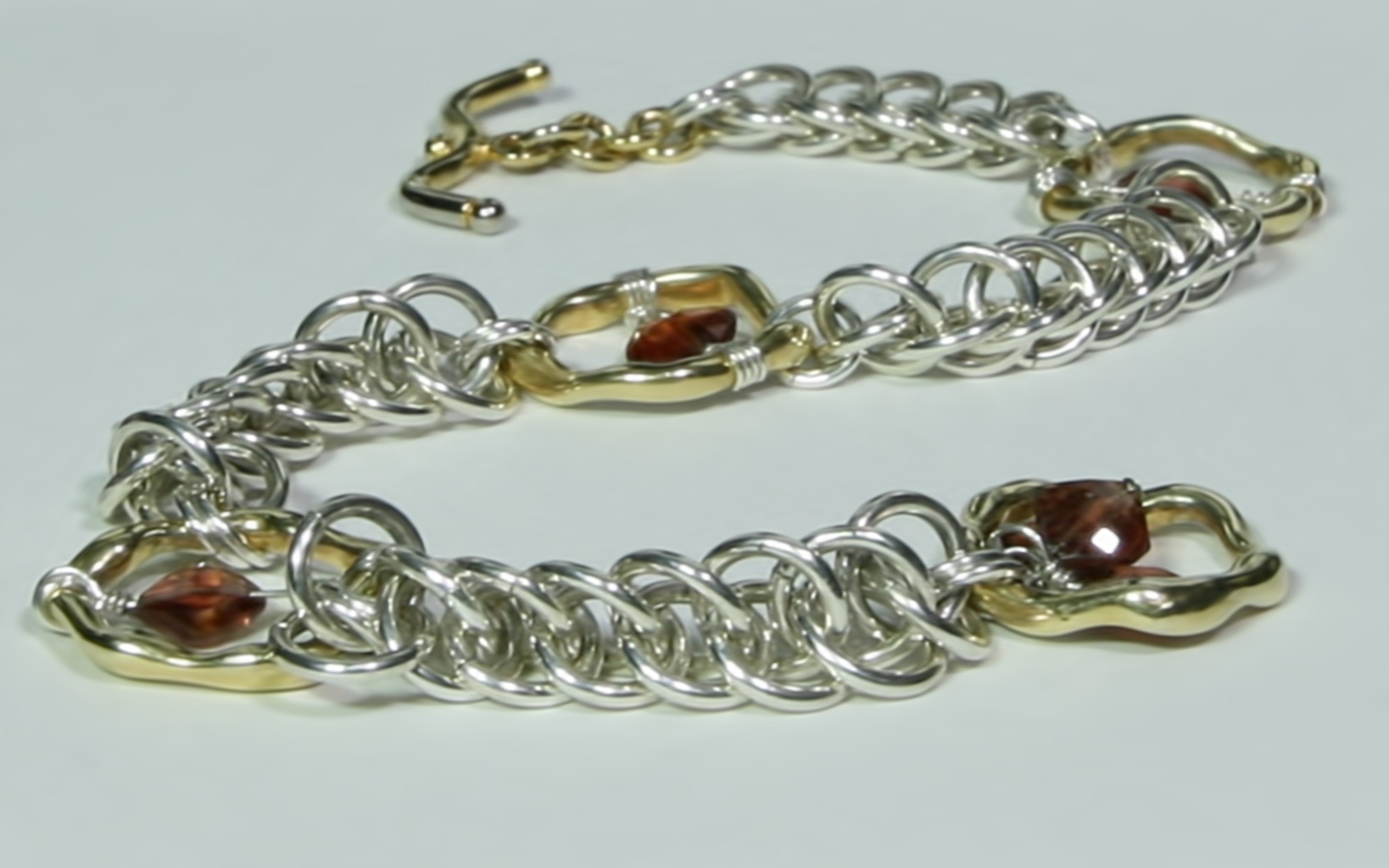}%
    ~
    \SuppLFSubfig{0 0 0 0}{./figures/4DLF/bracelet/14.jpeg}%
    ~
    \SuppLFSubfig{0 0 0 0}{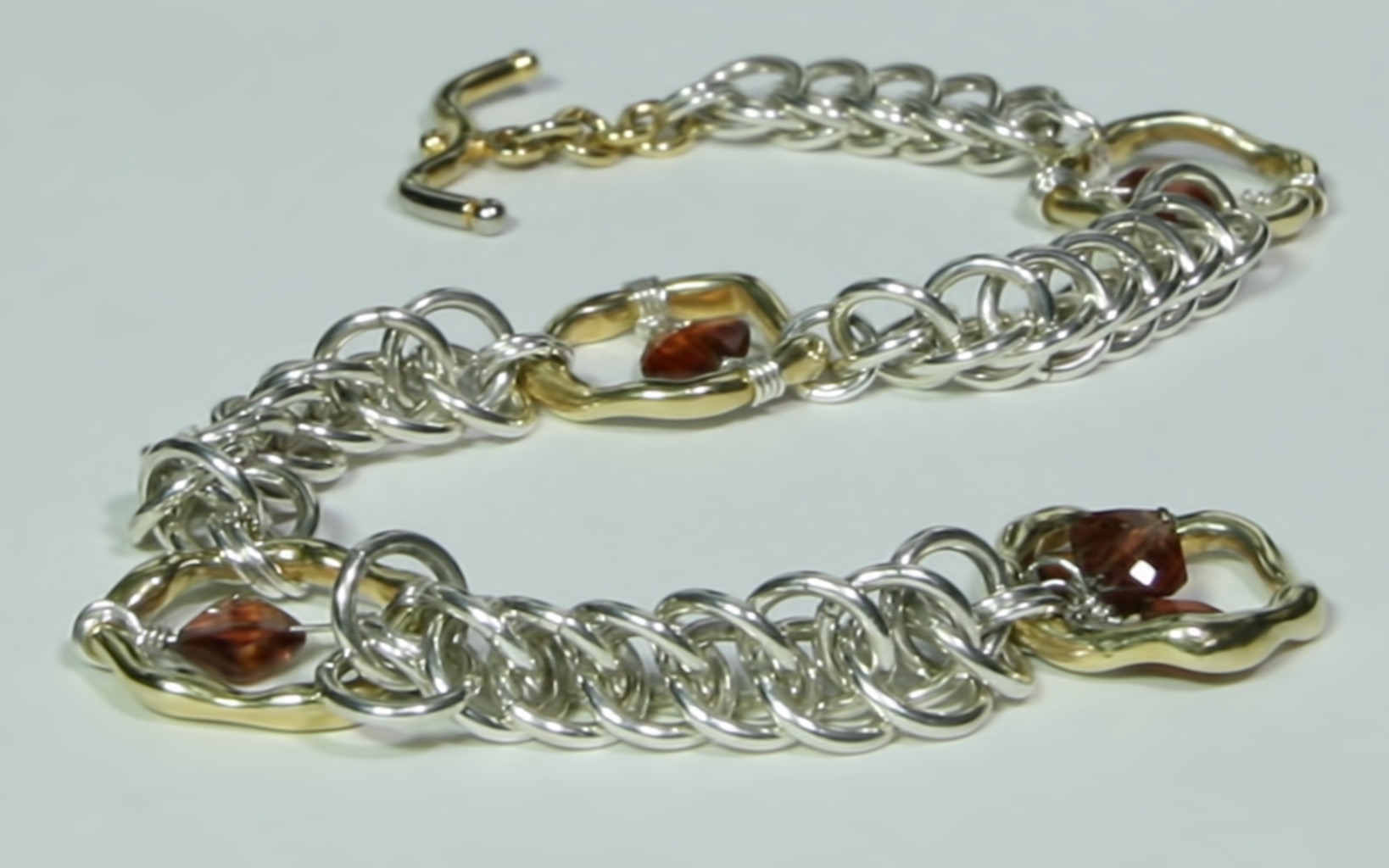}%
    ~
    \SuppLFSubfig{0 0 0 0}{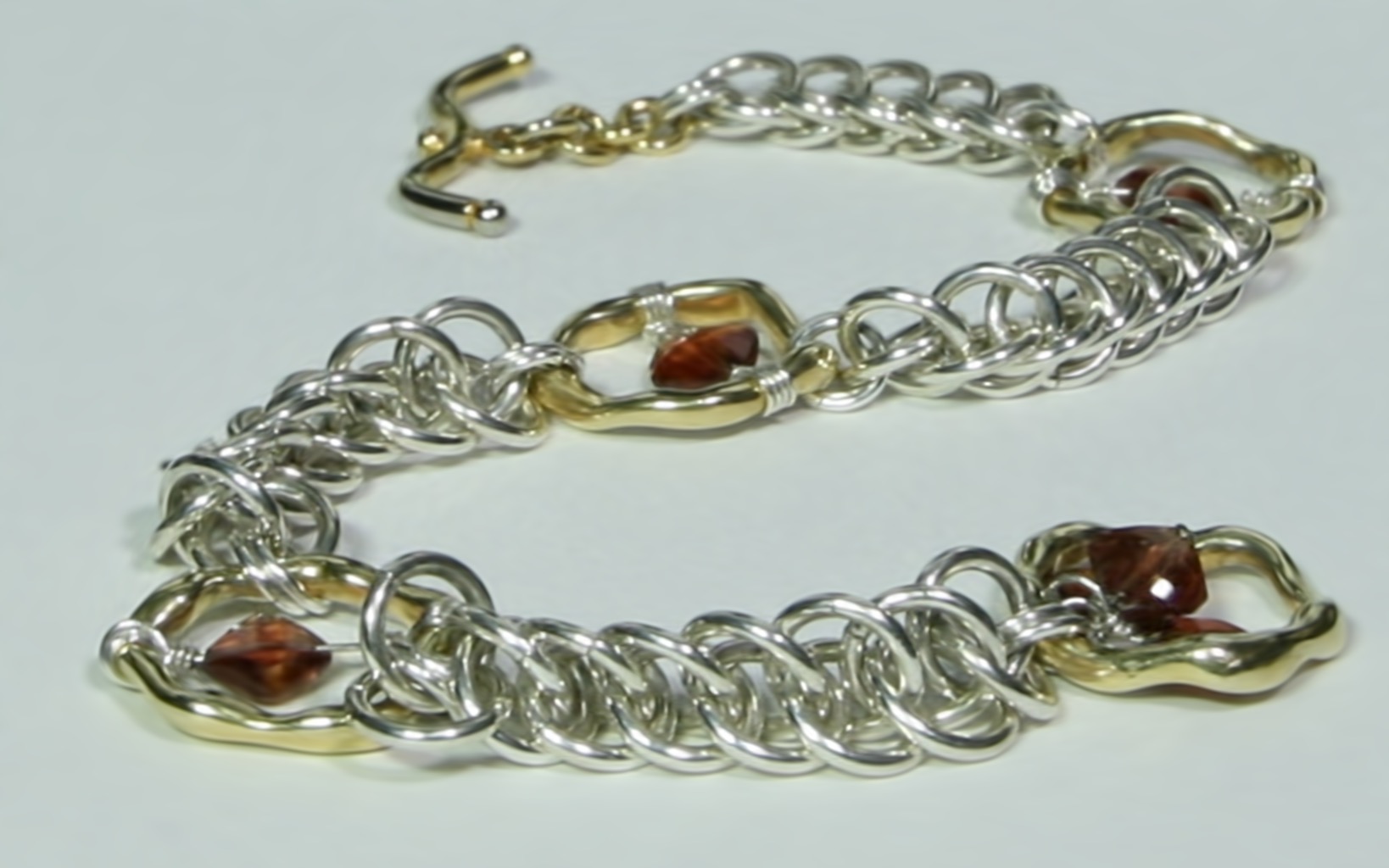}%

    \SuppLFSubfig{0 0 0 0}{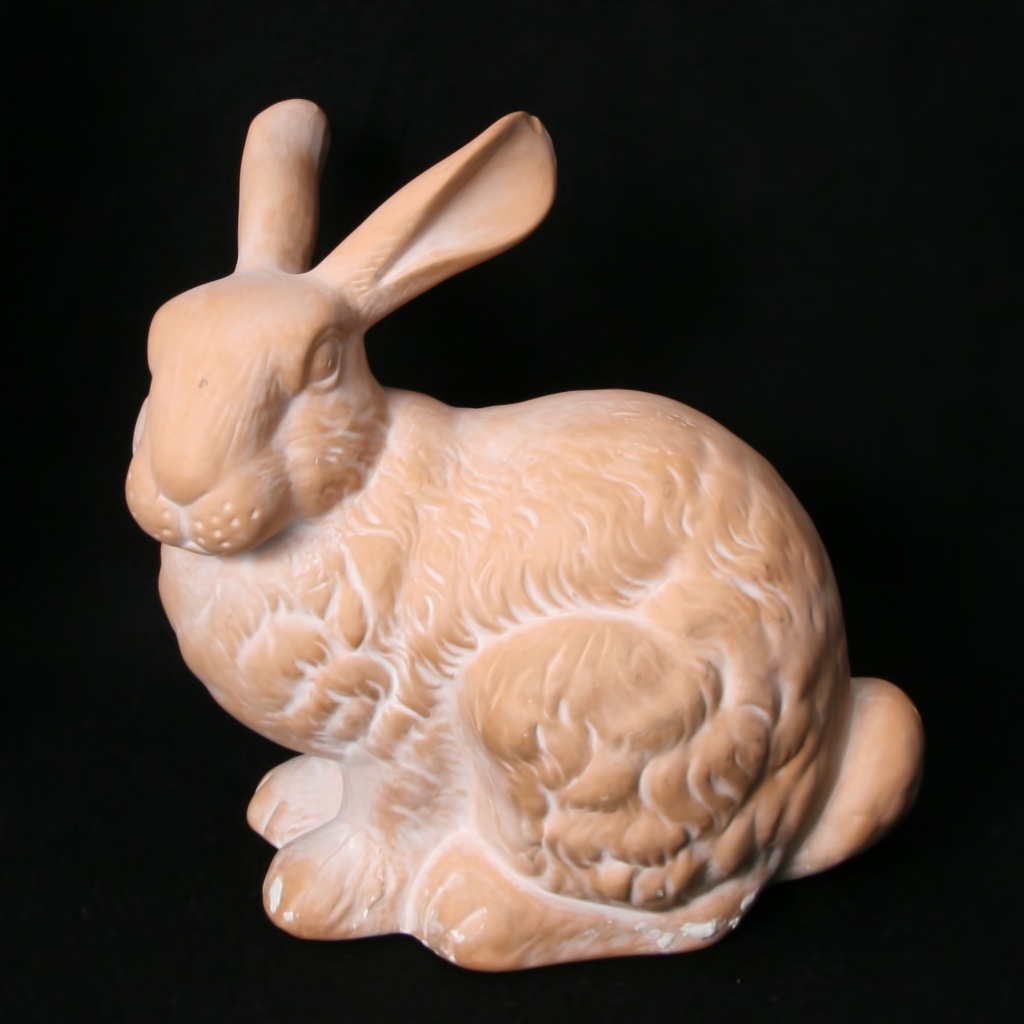}%
    ~
    \SuppLFSubfig{0 0 0 0}{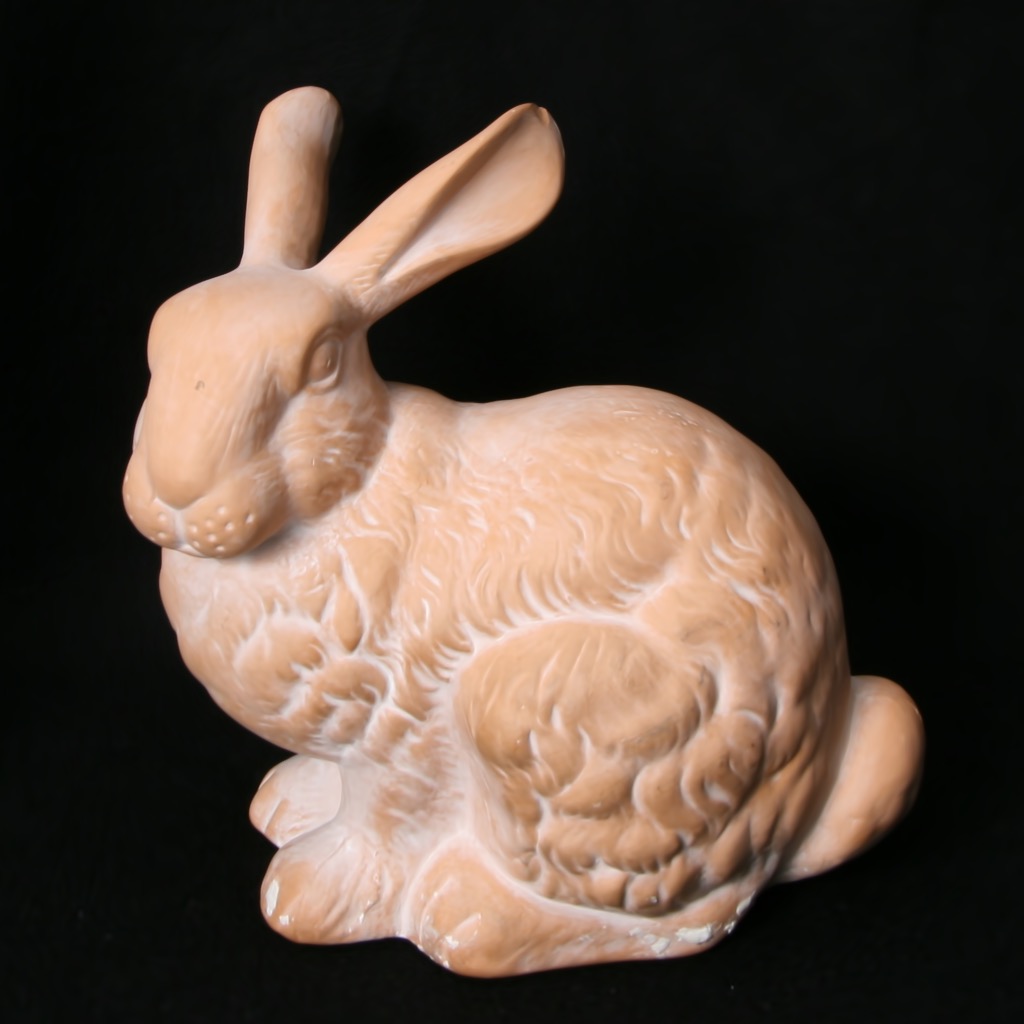}%
    ~
    \SuppLFSubfig{0 0 0 0}{./figures/4DLF/bunny/14.jpeg}%
    ~
    \SuppLFSubfig{0 0 0 0}{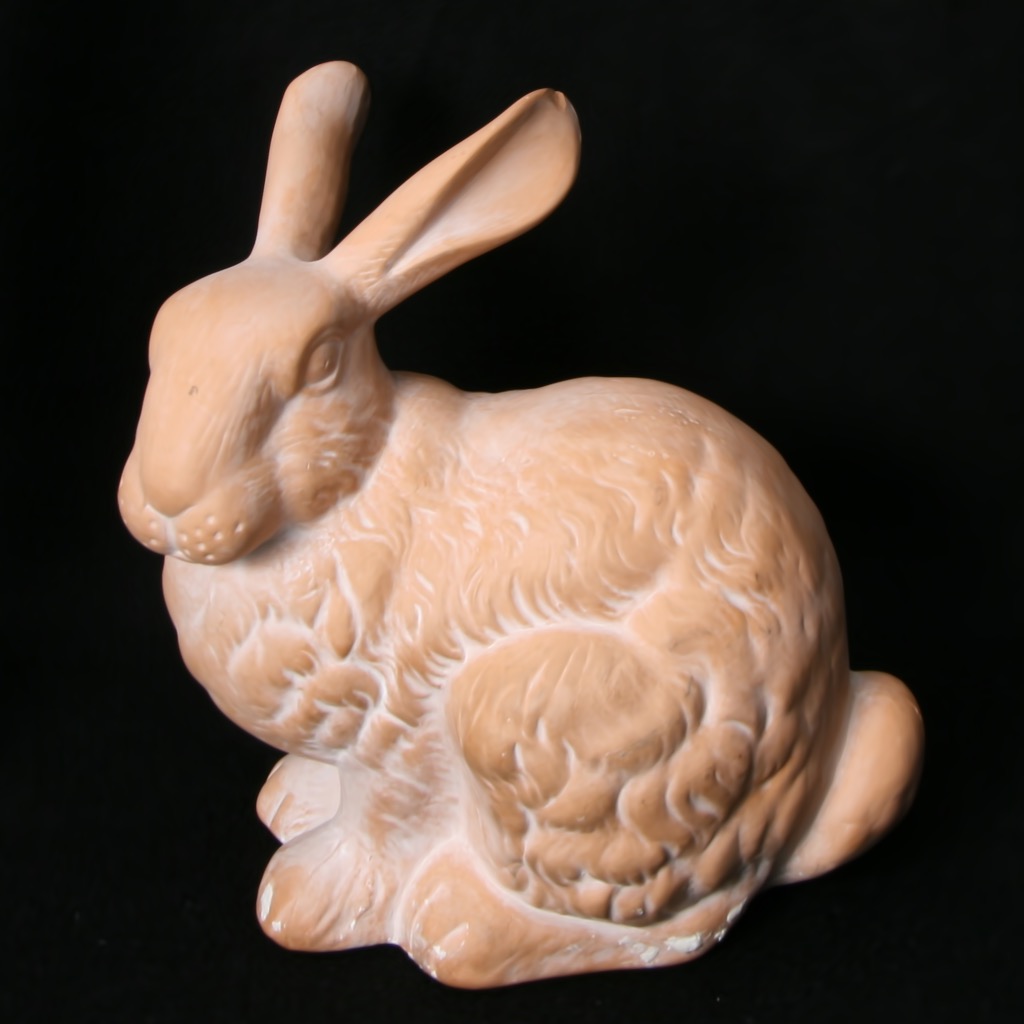}%
    ~
    \SuppLFSubfig{0 0 0 0}{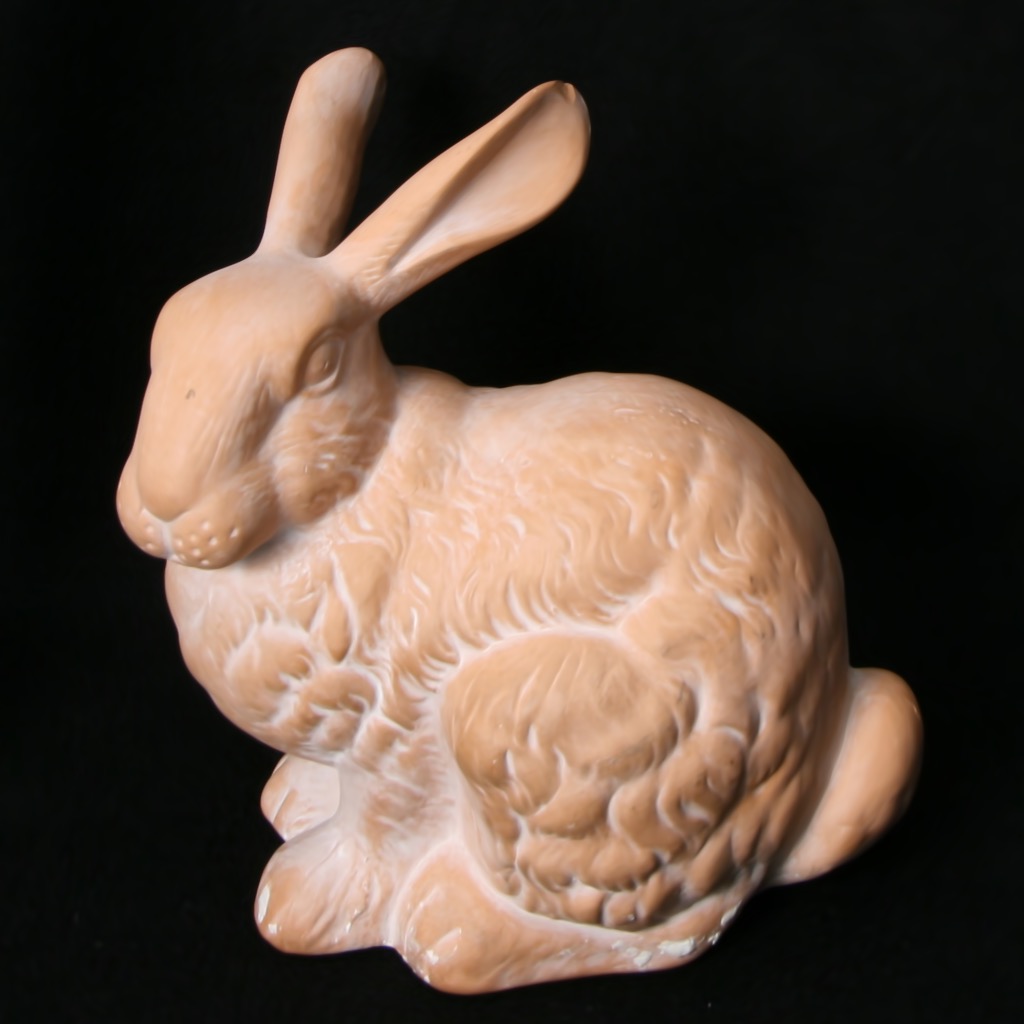}%

    \SuppLFSubfig{0 0 0 0}{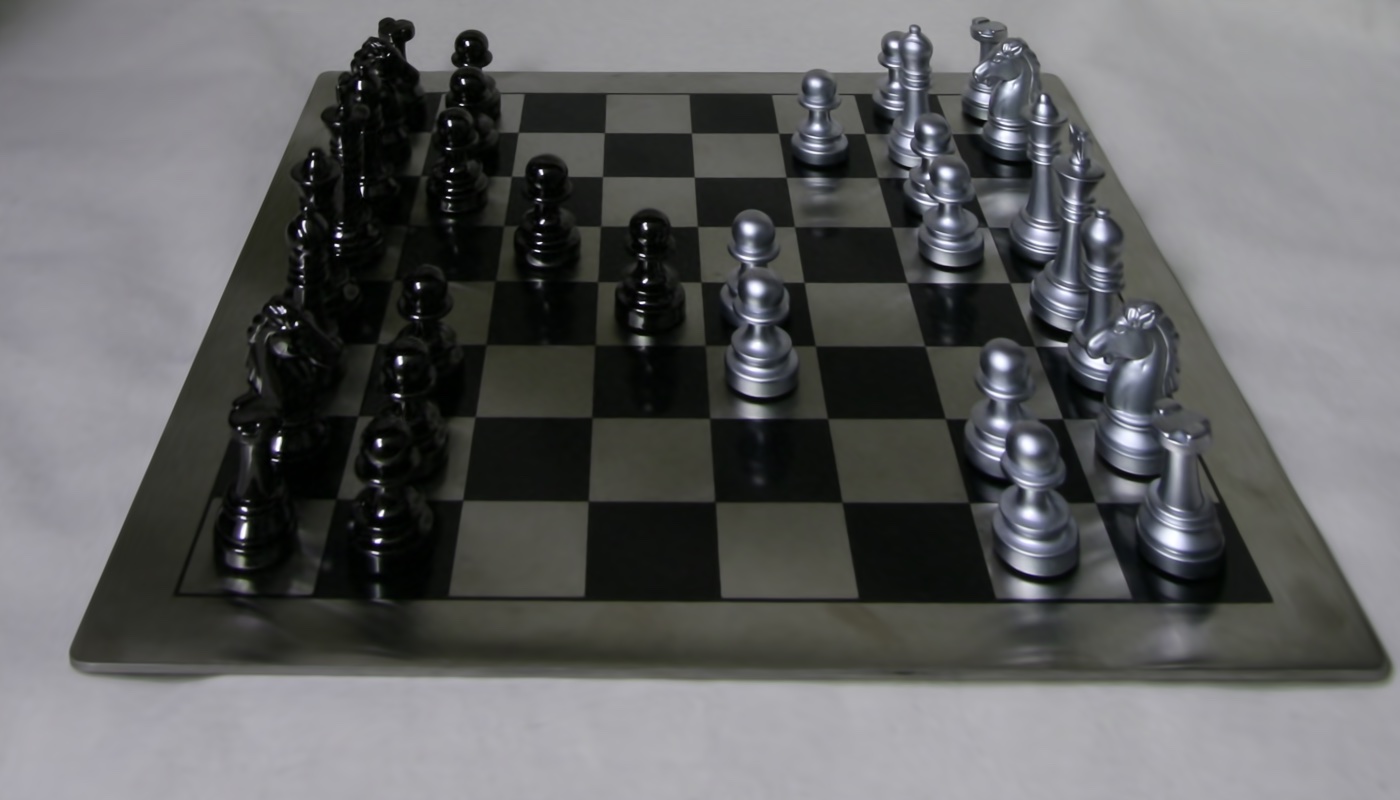}%
    ~
    \SuppLFSubfig{0 0 0 0}{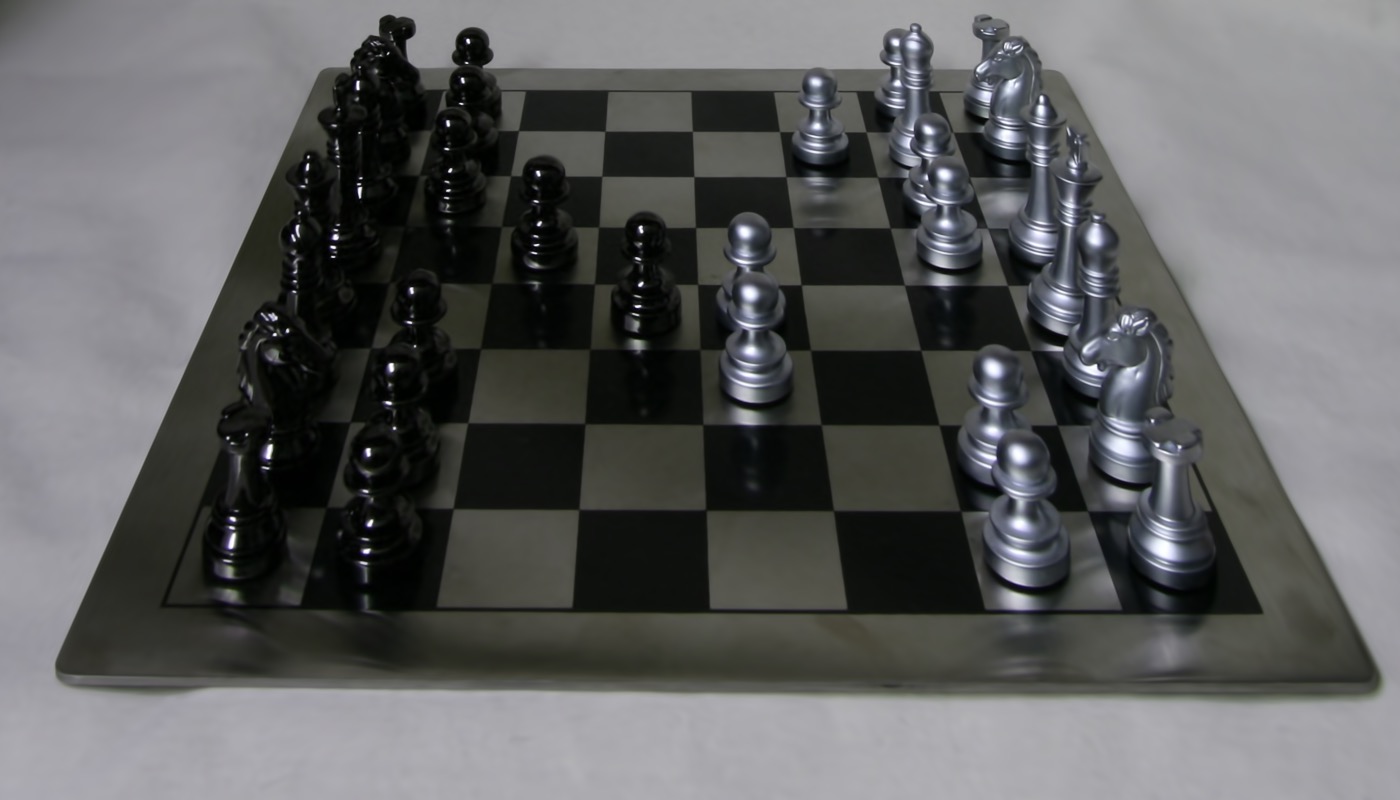}%
    ~
    \SuppLFSubfig{0 0 0 0}{./figures/4DLF/chess/14.jpeg}%
    ~
    \SuppLFSubfig{0 0 0 0}{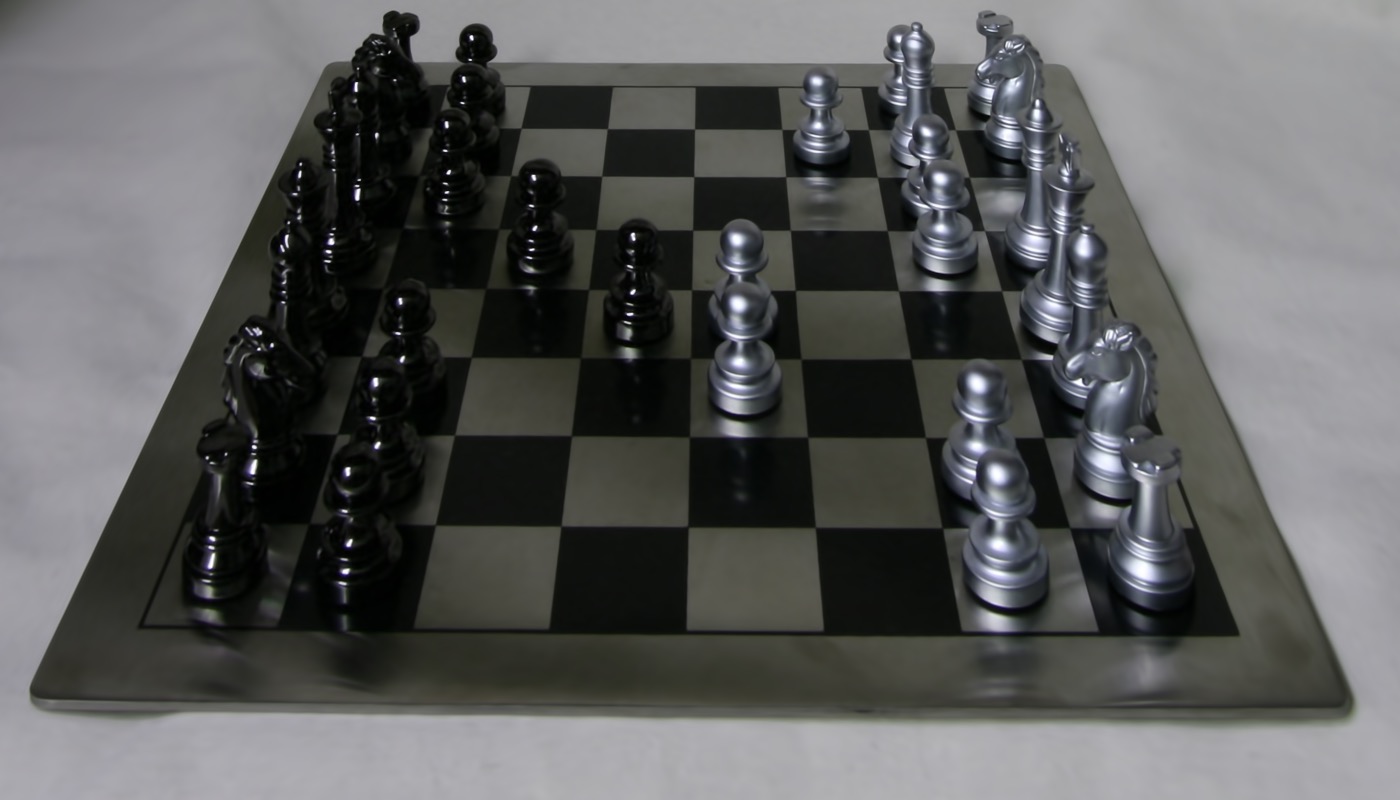}%
    ~
    \SuppLFSubfig{0 0 0 0}{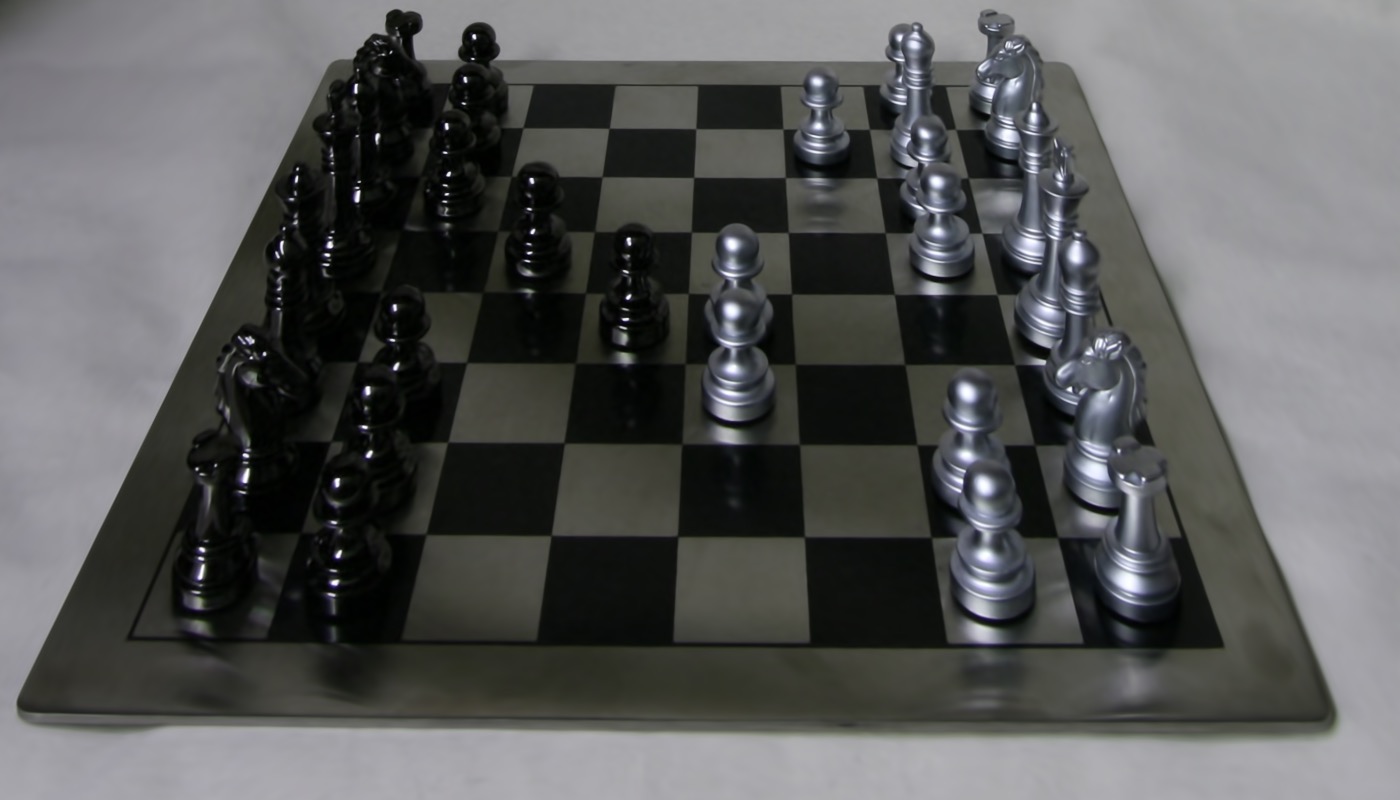}%

    \SuppLFSubfig{0 0 0 0}{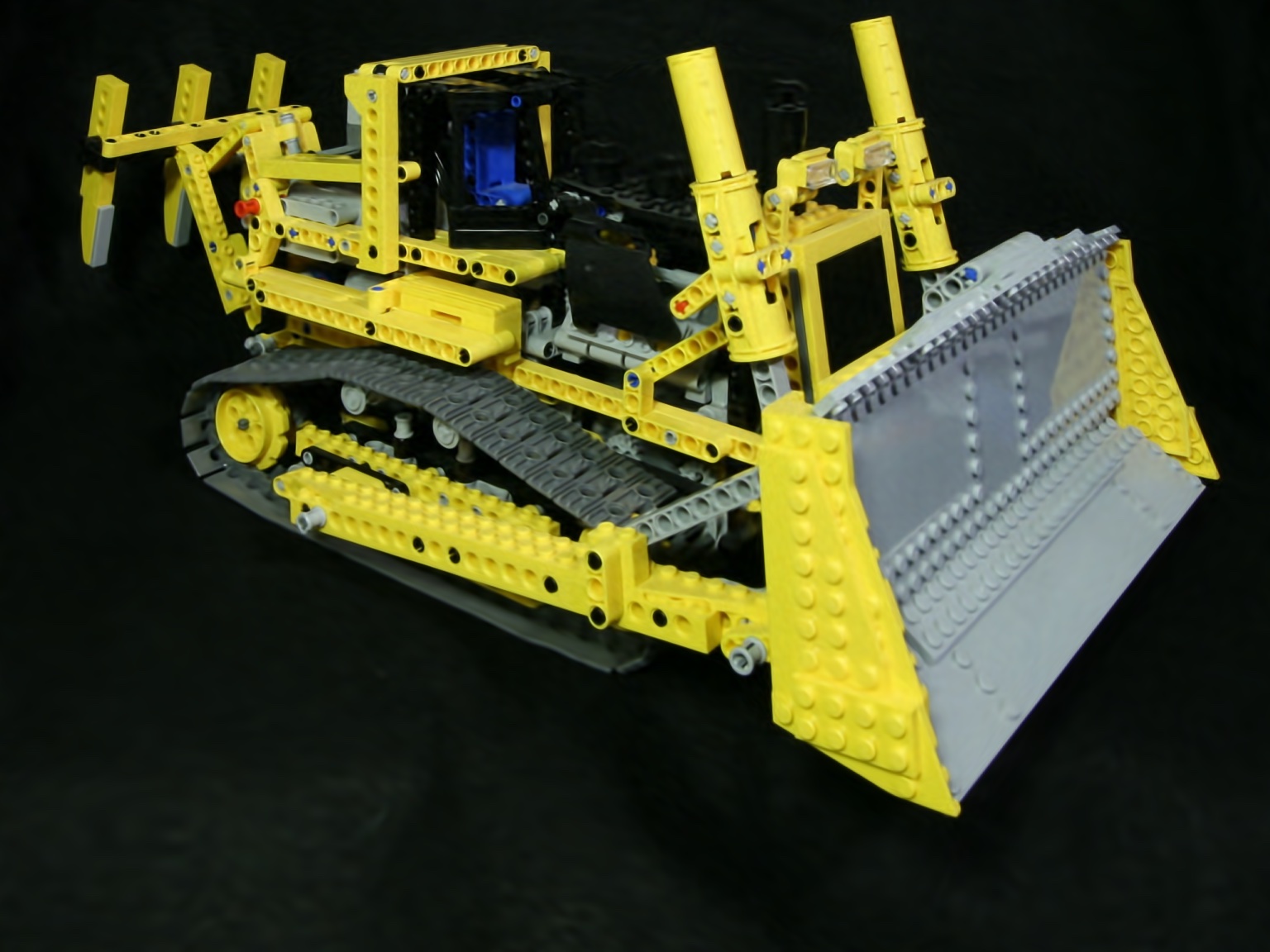}%
    ~
    \SuppLFSubfig{0 0 0 0}{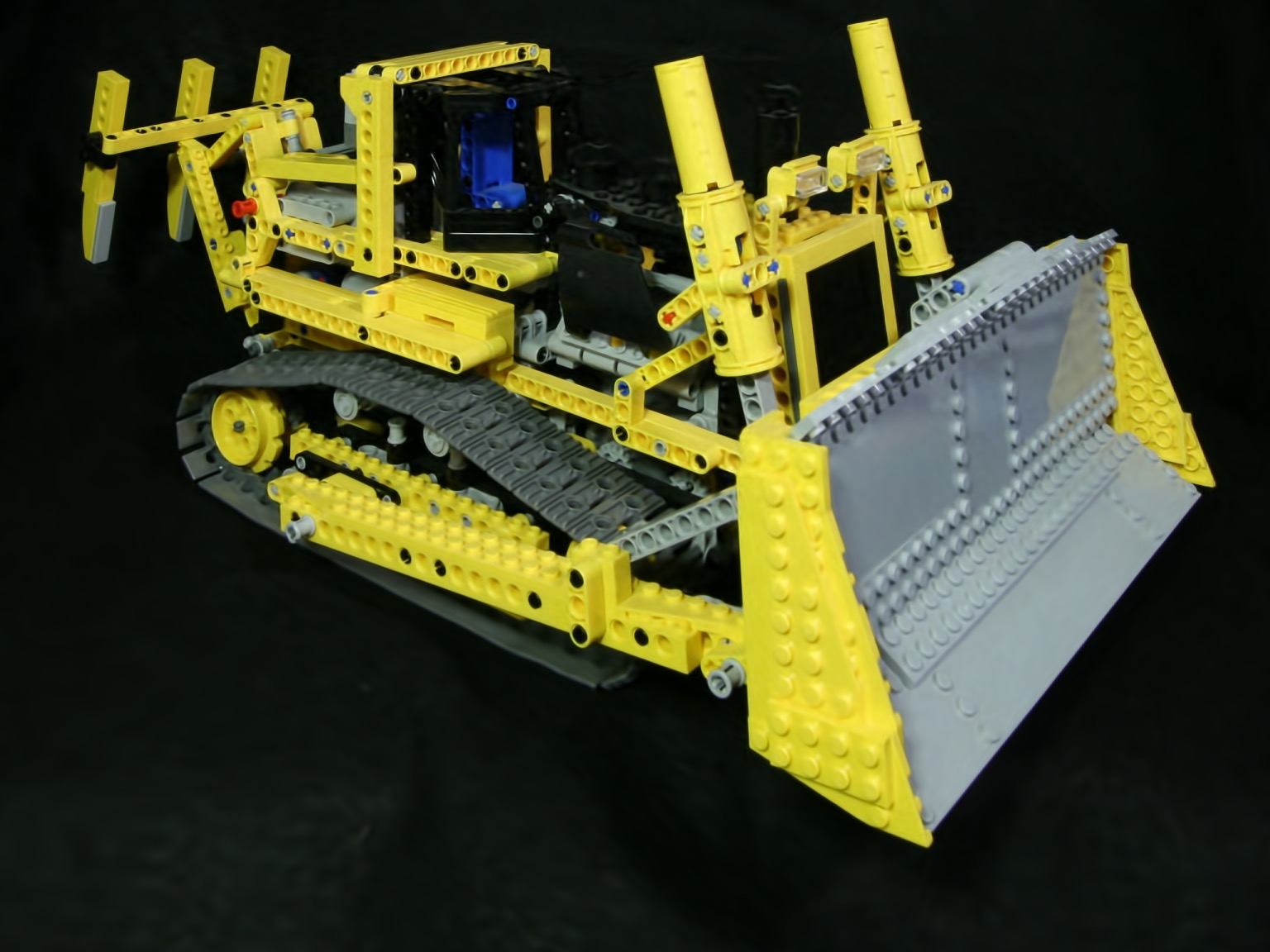}%
    ~
    \SuppLFSubfig{0 0 0 0}{./figures/4DLF/bulldozer/14.jpeg}%
    ~
    \SuppLFSubfig{0 0 0 0}{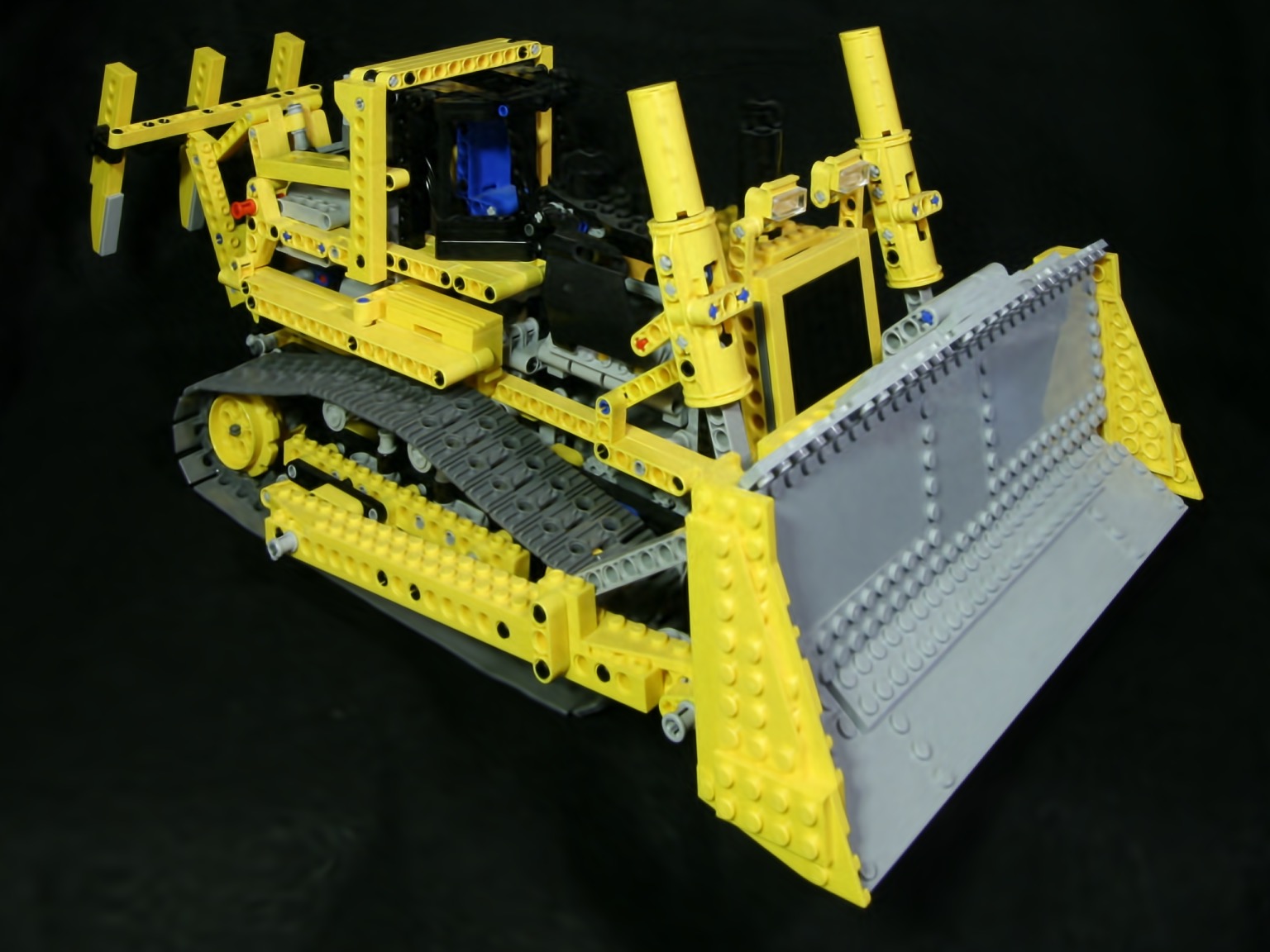}%
    ~
    \SuppLFSubfig{0 0 0 0}{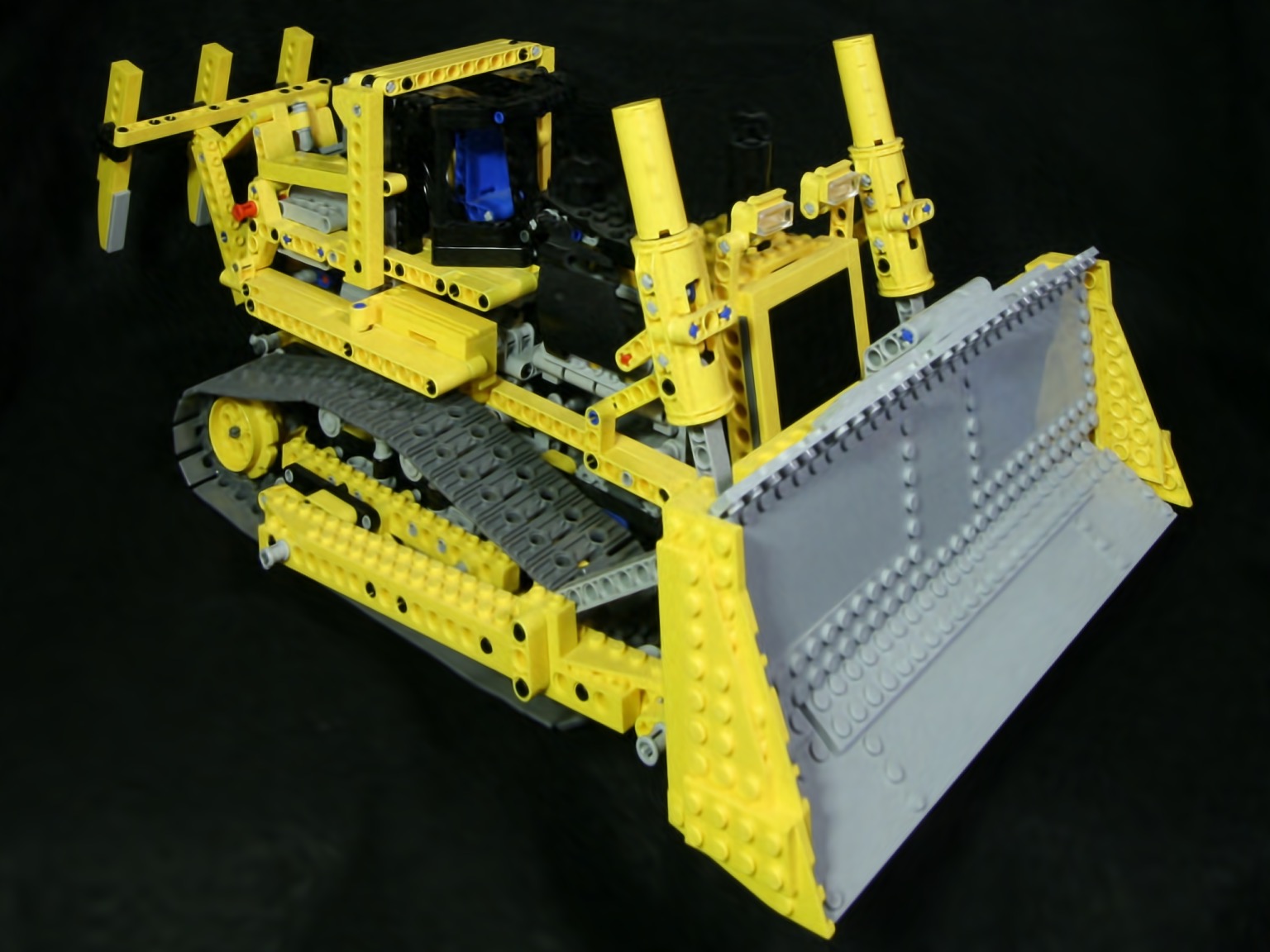}%

    \FigFiveSubfigCaption{$t=0$}%
    ~
    \FigFiveSubfigCaption{$t=0.25$}%
    ~
    \FigFiveSubfigCaption{$t=0.5$}%
    ~
    \FigFiveSubfigCaption{$t=0.75$}%
    ~
    \FigFiveSubfigCaption{$t=1$}%
    \vspace{-5pt}
	\caption{Stanford 4D Light Fields Results. We interpolate the learned codes from from two non-adjacent viewpoints (from top-left to bottom-right). See supplementary video for better contrast.}
    \label{fig:SuppInterpolateBegin}

\end{figure*}

\begin{figure*}[!ht]
    \SuppLFSubfig{250 0 300 0}{./figures/Holo/Shiba/00.jpeg}%
    ~
    \SuppLFSubfig{250 0 300 0}{./figures/Holo/Shiba/01.jpeg}%
    ~
    \SuppLFSubfig{250 0 300 0}{./figures/Holo/Shiba/02.jpeg}%
    ~
    \SuppLFSubfig{250 0 300 0}{./figures/Holo/Shiba/03.jpeg}%
    ~
    \SuppLFSubfig{250 0 300 0}{./figures/Holo/Shiba/04.jpeg}%

    \SuppHoloSubfig{0 300 0 160}{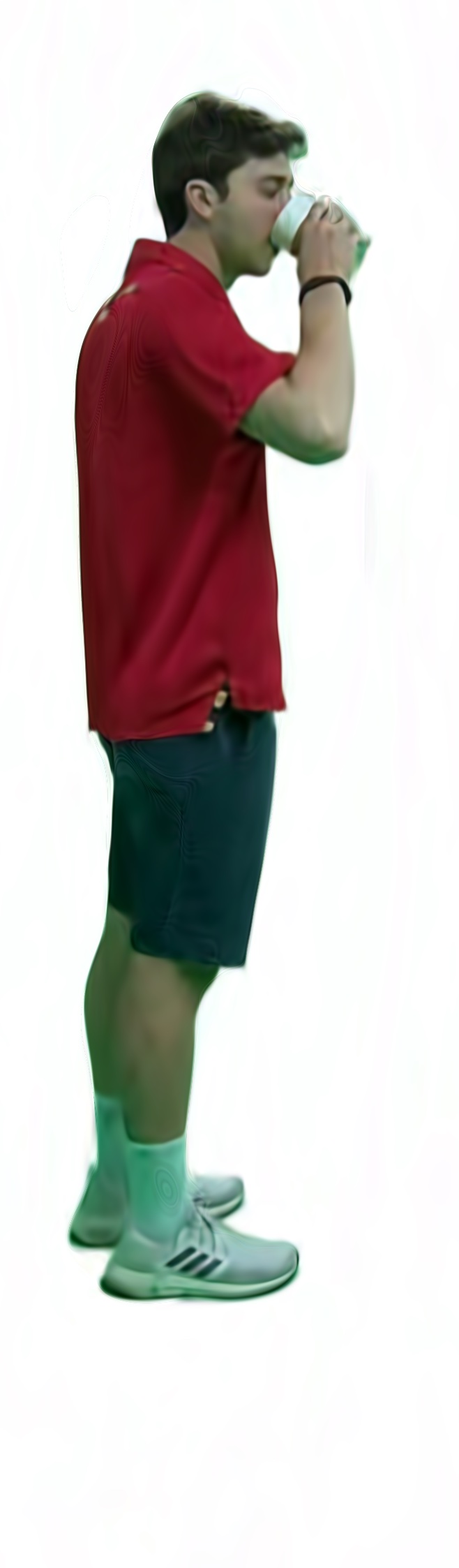}%
    ~ \hfill
    \SuppHoloSubfig{0 300 0 160}{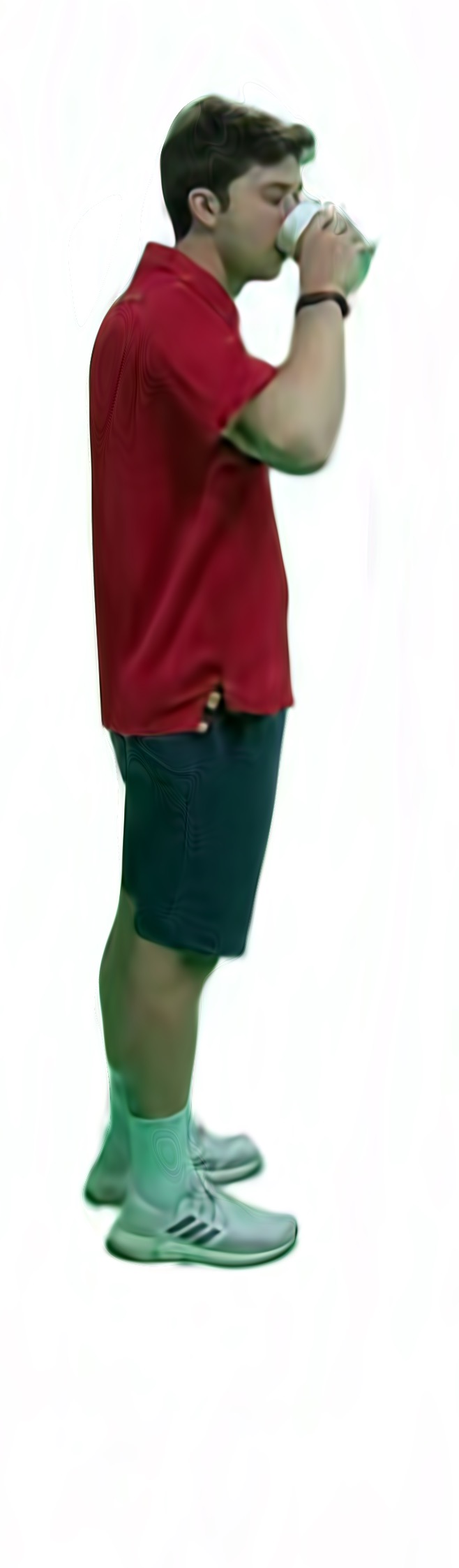}%
    ~ \hfill
    \SuppHoloSubfig{0 300 0 160}{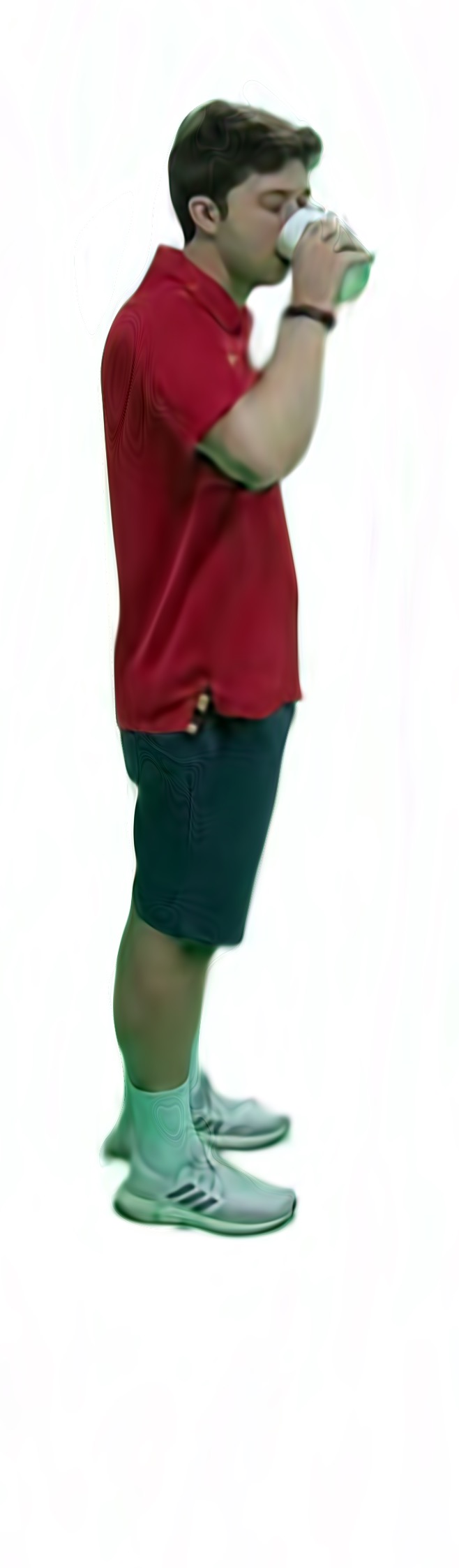}%
    ~ \hfill
    \SuppHoloSubfig{0 300 0 160}{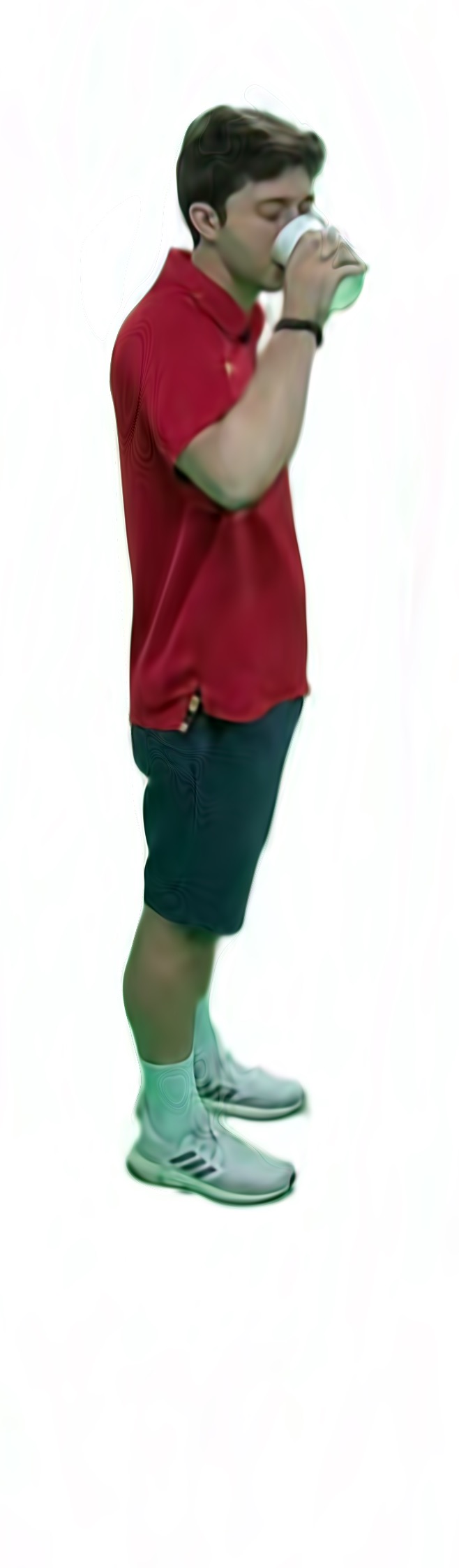}%
    ~ \hfill
    \SuppHoloSubfig{0 300 0 160}{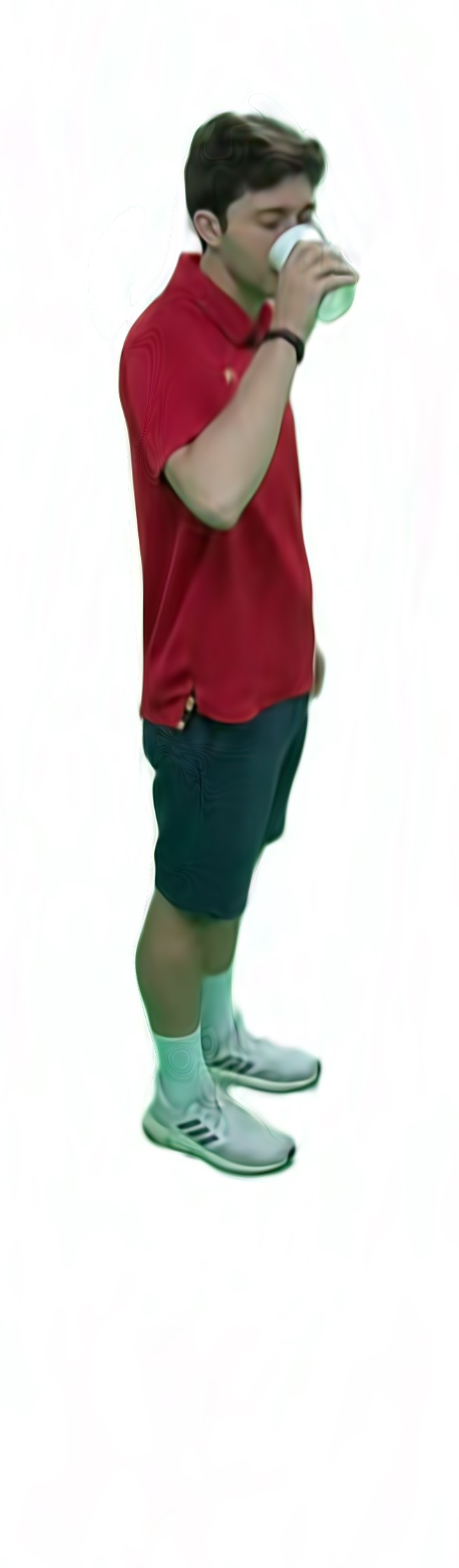}%
    \hfill

    \SuppLFSubfig{0 100 0 100}{./figures/Holo/Maria/00.jpeg}%
    ~
    \SuppLFSubfig{0 100 0 100}{./figures/Holo/Maria/01.jpeg}%
    ~
    \SuppLFSubfig{0 100 0 100}{./figures/Holo/Maria/02.jpeg}%
    ~
    \SuppLFSubfig{0 100 0 100}{./figures/Holo/Maria/03.jpeg}%
    ~
    \SuppLFSubfig{0 100 0 100}{./figures/Holo/Maria/04.jpeg}%

    \FigFiveSubfigCaption{$t=0$}%
    ~
    \FigFiveSubfigCaption{$t=0.25$}%
    ~
    \FigFiveSubfigCaption{$t=0.5$}%
    ~
    \FigFiveSubfigCaption{$t=0.75$}%
    ~
    \FigFiveSubfigCaption{$t=1$}%
    \vspace{-5pt}
    \caption{Unstructured Light Field Results. We interpolate the learned codes from two non-adjacent viewpoints. See supplementary video for better contrast. }
    \label{fig:SuppInterpolateEnd}

\end{figure*}

We present additional comparisons with LFN and NeRF from Figure~\ref{fig:SuppCompareBegin} to~\ref{fig:SuppCompareEnd}, per-scene interpolated results from Figure~\ref{fig:SuppInterpolateBegin} to~\ref{fig:SuppInterpolateEnd}, and per-scene qualitative metrics from Table~\ref{table:PerSceneBegin} to~\ref{table:PerSceneEnd}.

\subsubsection{{Beyond Interpolating Between Two Views.}}
Our proposed setting obtains the novel view $AB$ between two known views $A$ and $B$ by interpolating the latent codes associated with $A$ and $B$.
We could do the same for two other known views, $C$ and $D$, and obtain another novel view $CD$.
At this stage, we can further interpolate between $AB$ and $CD$, and in Figure~\ref{fig:ABCDMorph} we present some example results of interpolation between the two interpolated latent codes.

\begin{table*}[!ht]
\begin{tabular}{ccccc|cccc}
\toprule
& \multicolumn{4}{c|}{4D Planar} &  \multicolumn{4}{c}{Unstructured} \\
\midrule
 & NeRF & LFN & Ours & Ours-Finetuned & NeRF & LFN & Ours & Ours-Finetuned\\
\midrule
 SSIM & 0.917  & 0.944  &  0.975  & 0.977 &  0.905 & 0.788 & 0.664 & 0.802\\
PSNR & 33.28 & 30.67 & 35.77 & 36.84 & 27.15  & 21.35  & 16.80  & 24.03\\
\bottomrule
\end{tabular}
\caption{Quantitative results on novel views with an additional condidtion \textit{Ours-Finetuned}, where we render our INR with the latent code obtained after optimizing it against the ground truth test image (while freezing the network weights). Results suggest that the trained INR is capable of achieving better quantitative novel view results, but is handicapped by our inability input exact camera poses due to non-3D nature of our method.}
\label{table:table_3}
\end{table*}

\subsubsection{{Extending to Frame Interpolation.}}
Our method can be potentially modified and improved for frame interpolation using only two input images~\cite{reda2022film}.
As a proof of concept, we take two human portrait images of the same person and deploy our method on those images.
In Figure~\ref{fig:FaceMorph}, we present the intermediate frames produced by our method.
We also test our method on the data used in X-Fields~\cite{Bemana2020xfields}.
In Figure~\ref{fig:XFields}, we show the intermediate frames produced by our method.
\vspace{-5pt}
\subsubsection{{Optimizing Latent Code Given Test Images.}}
As noted in previous discussions, a major limitation of our latent interpolation method is we cannot exactly render at arbitrary camera poses.
Such a limitation also leads poor novel view performance when measured by qualitative metrics like PSNR based on pixel-wise errors against the ground truth image.
However, the poor qualitative metrics \textit{does not mean} that the INR is unable to express and decode those views.
Rather, the deficiency reflects more about our inability to find the right latent code for a novel camera pose.

Therefore, we provide additional qualitative results in Table~\ref{table:table_3} to vindicate the INR's ability to express those novel views, when it is given \textit{more appropriate latent codes}.
In this new setting, when we measure the novel view performance after training the INR as before, we assume the ground truth image is known so that we can compute the error between the INR output and the true pixel color.
We optimize the latent code to minimize such error against the ground truth without modifying the INR weights.

\begin{figure*}[!h]
    \SuppThreeSubfig{0 0 0 0}{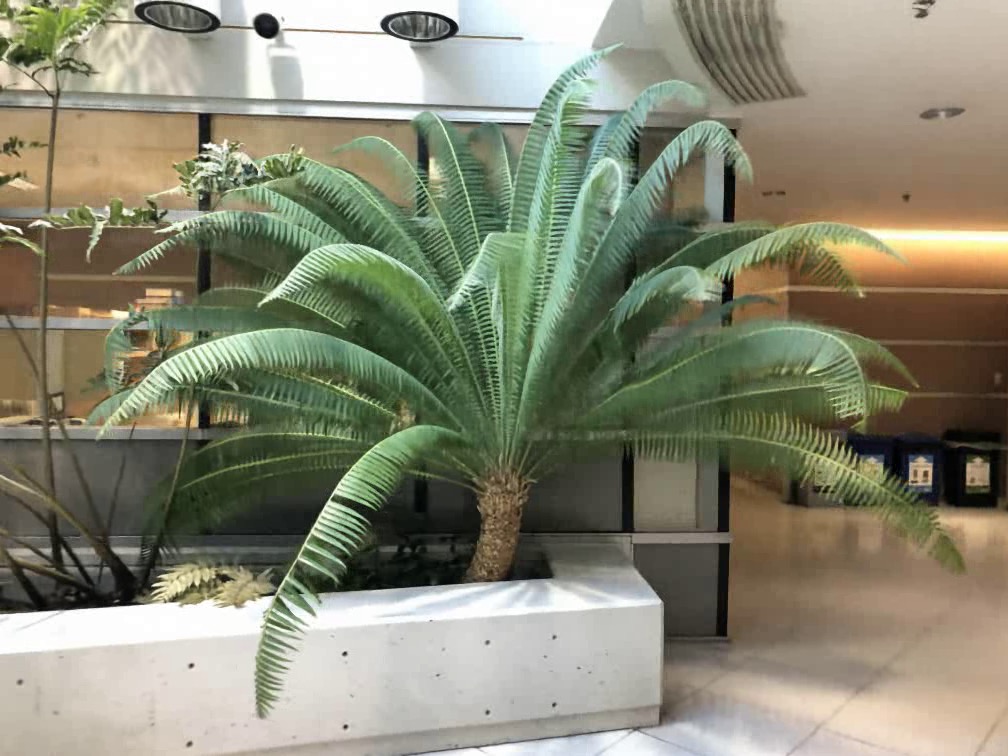}%
    ~
    \SuppThreeSubfig{0 0 0 0}{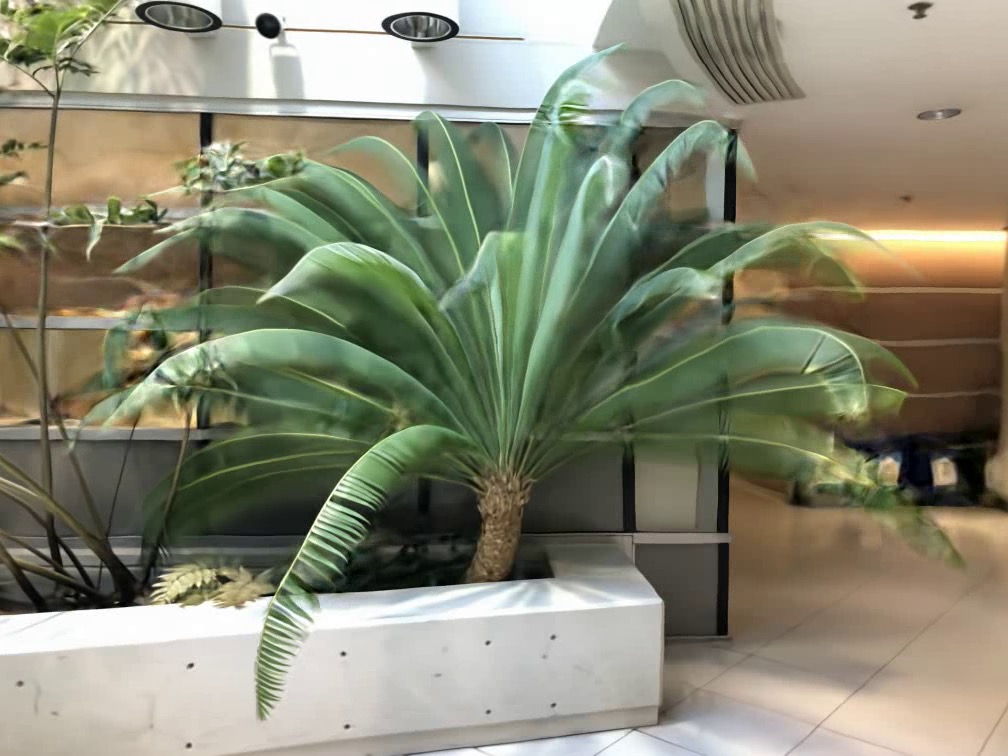}%
    ~
    \SuppThreeSubfig{0 0 0 0}{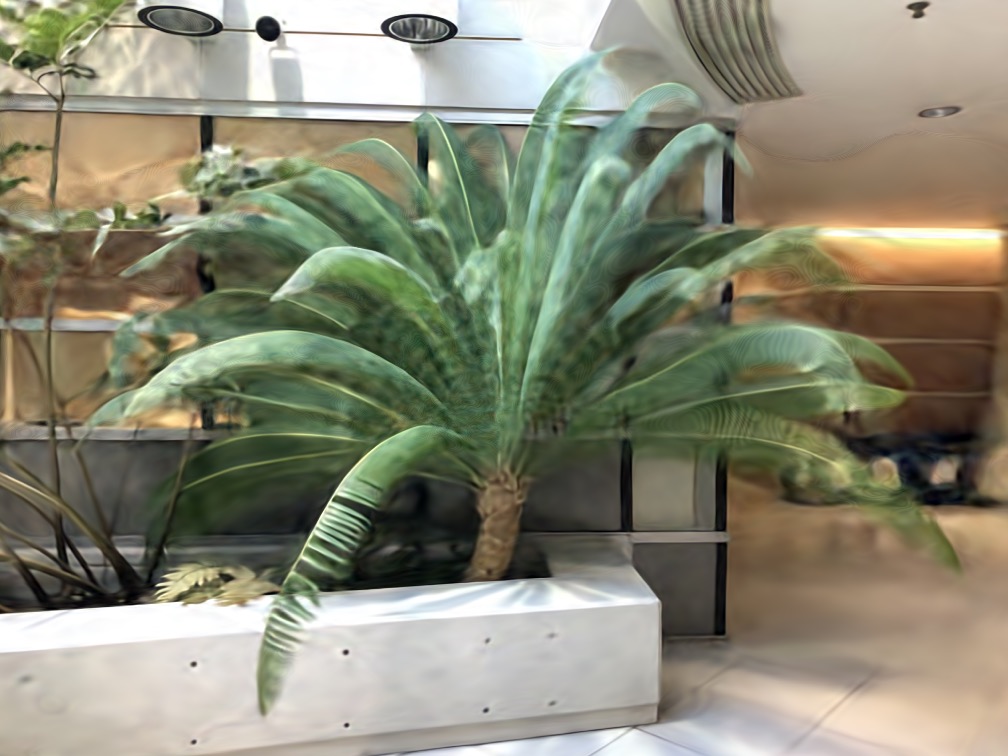}%
    
    \SuppThreeSubfig{0 0 0 0}{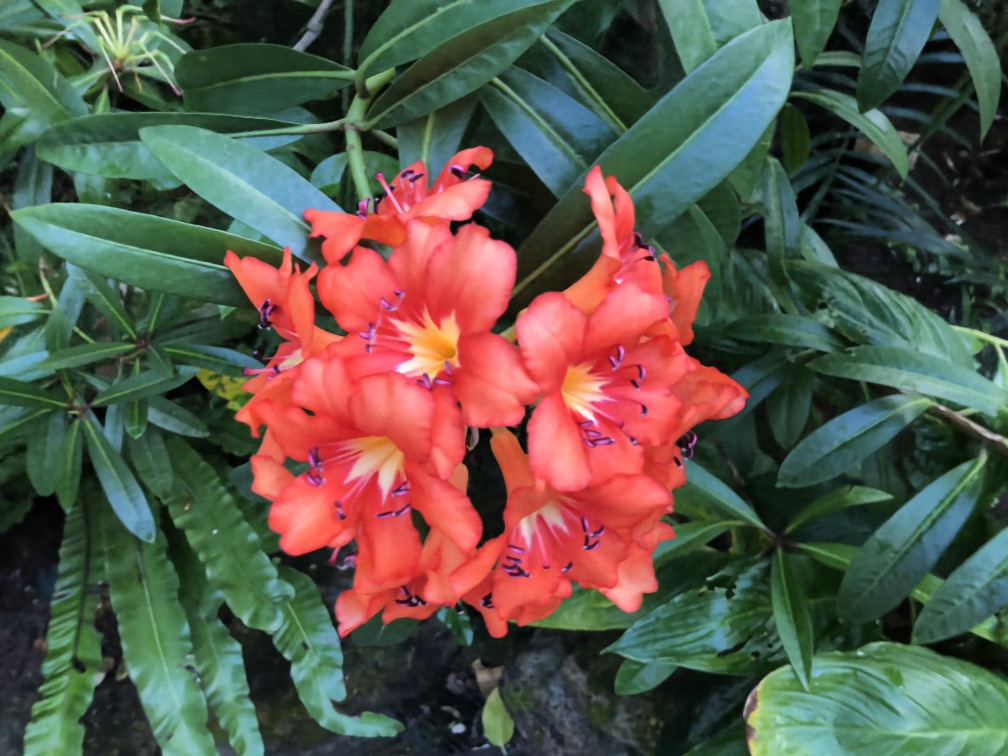}%
    ~
    \SuppThreeSubfig{0 0 0 0}{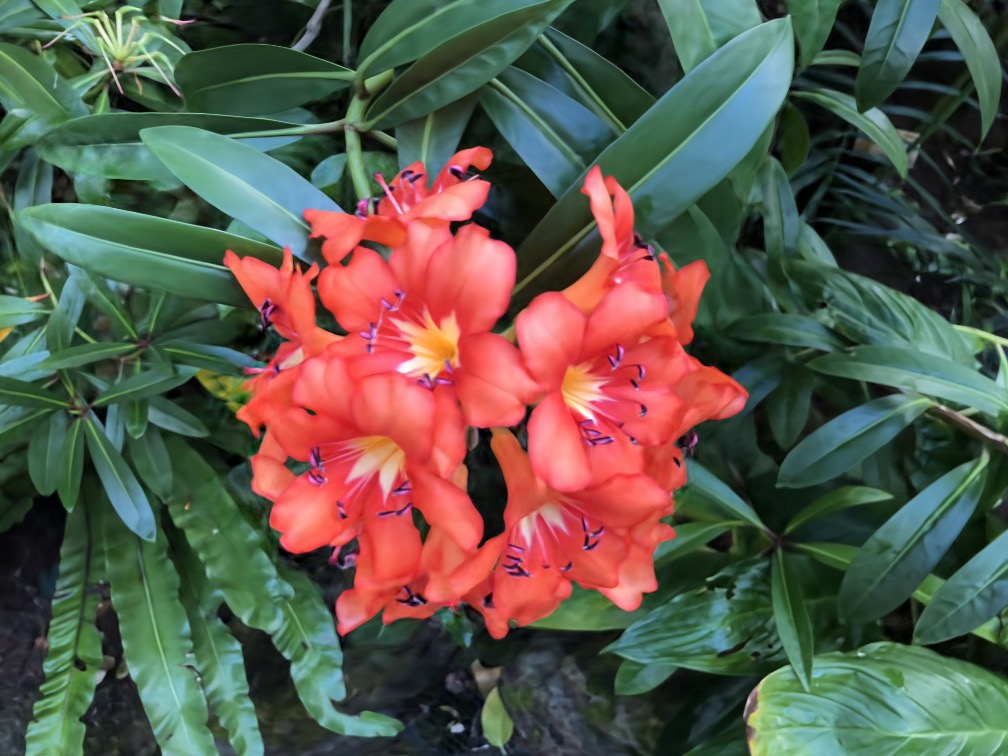}%
    ~
    \SuppThreeSubfig{0 0 0 0}{./figures/LLFF/Flower/14.jpeg}%
    
    \SuppThreeSubfig{0 0 0 0}{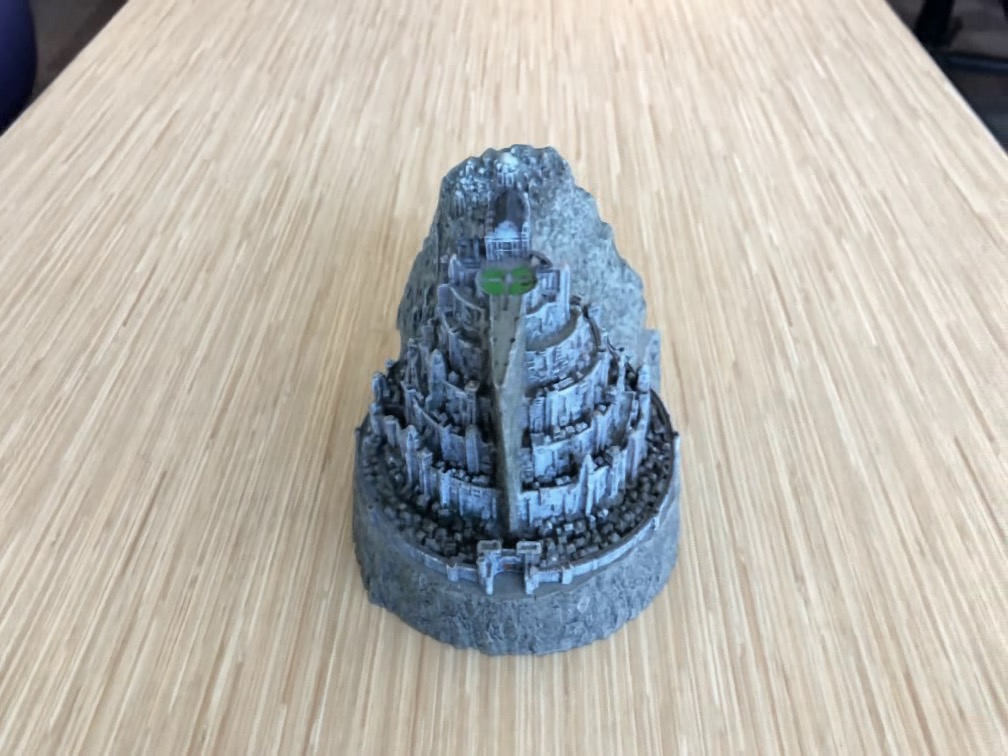}%
    ~
    \SuppThreeSubfig{0 0 0 0}{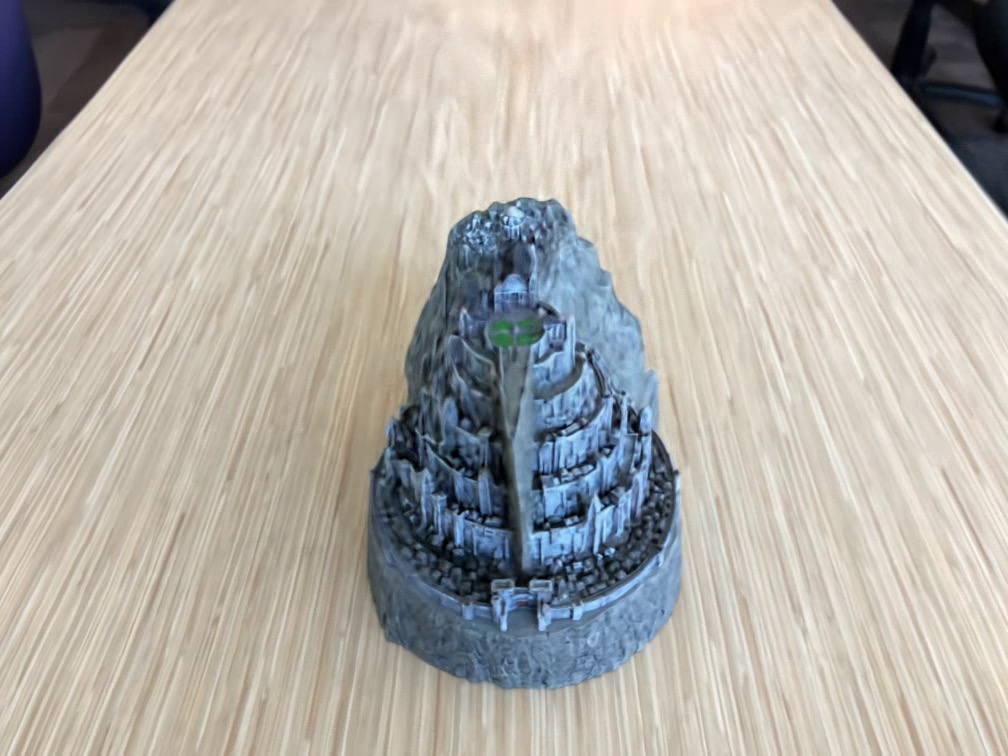}%
    ~
    \SuppThreeSubfig{0 0 0 0}{./figures/LLFF/Fortress/14.jpeg}%
    
    \SuppThreeSubfig{0 0 0 0}{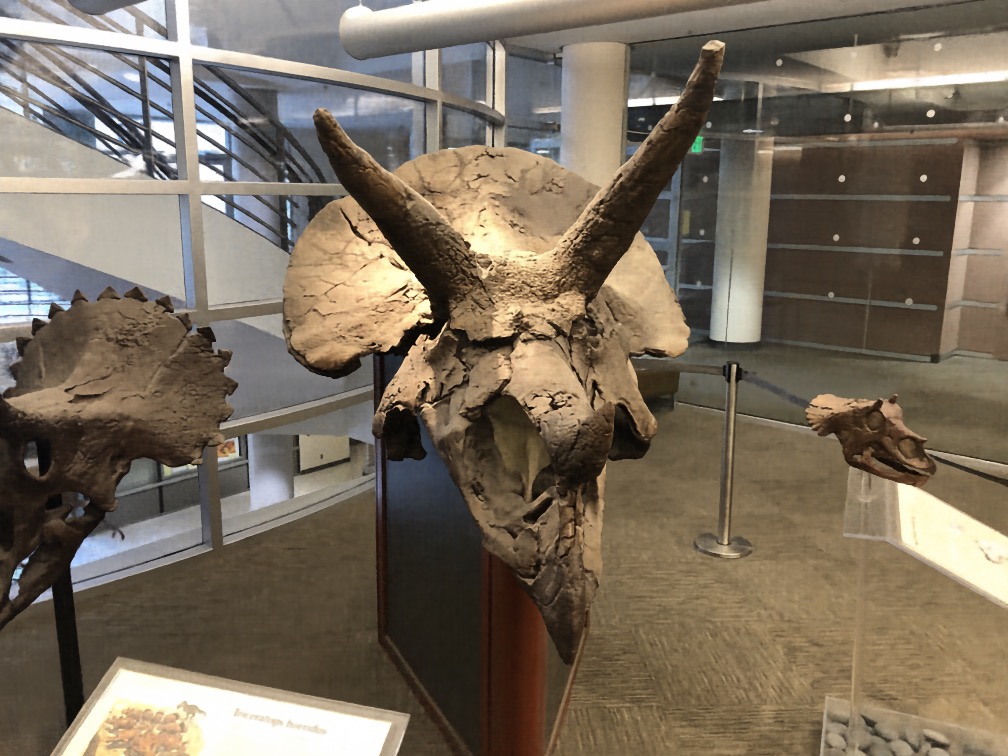}%
    ~
    \SuppThreeSubfig{0 0 0 0}{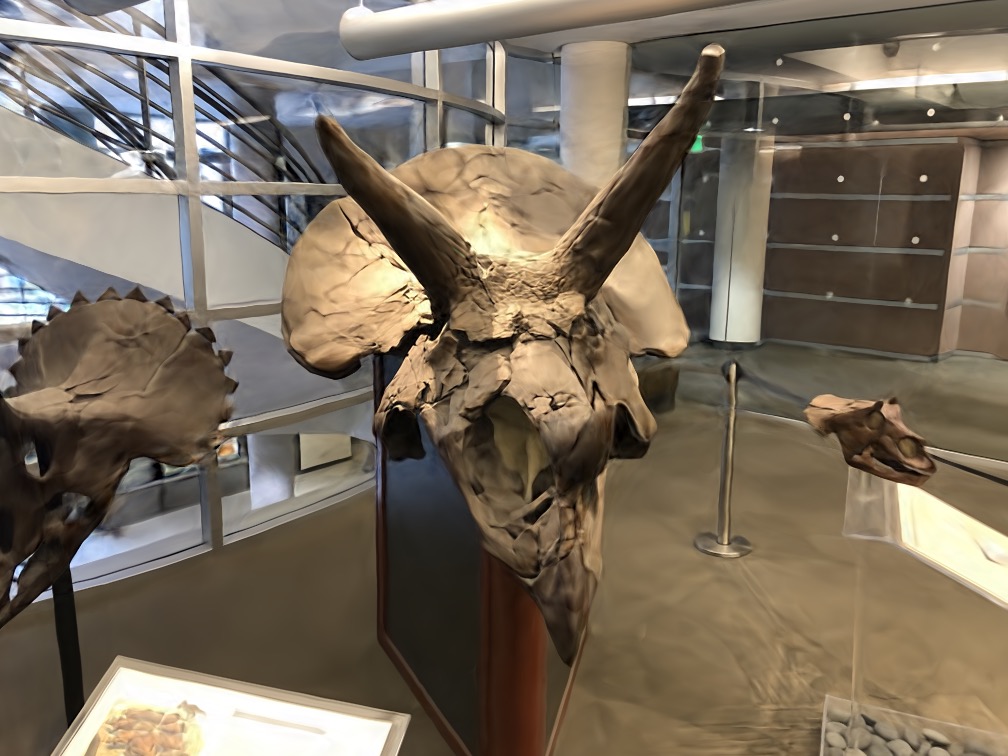}%
    ~
    \SuppThreeSubfig{0 0 0 0}{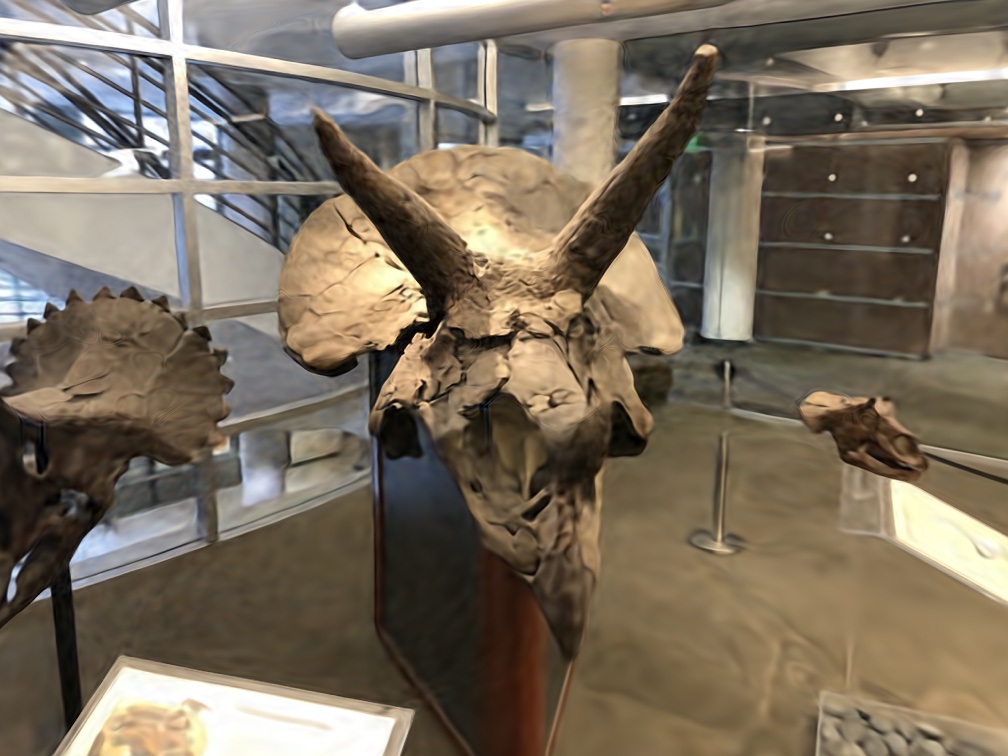}%
    
    \SuppThreeSubfig{0 0 0 0}{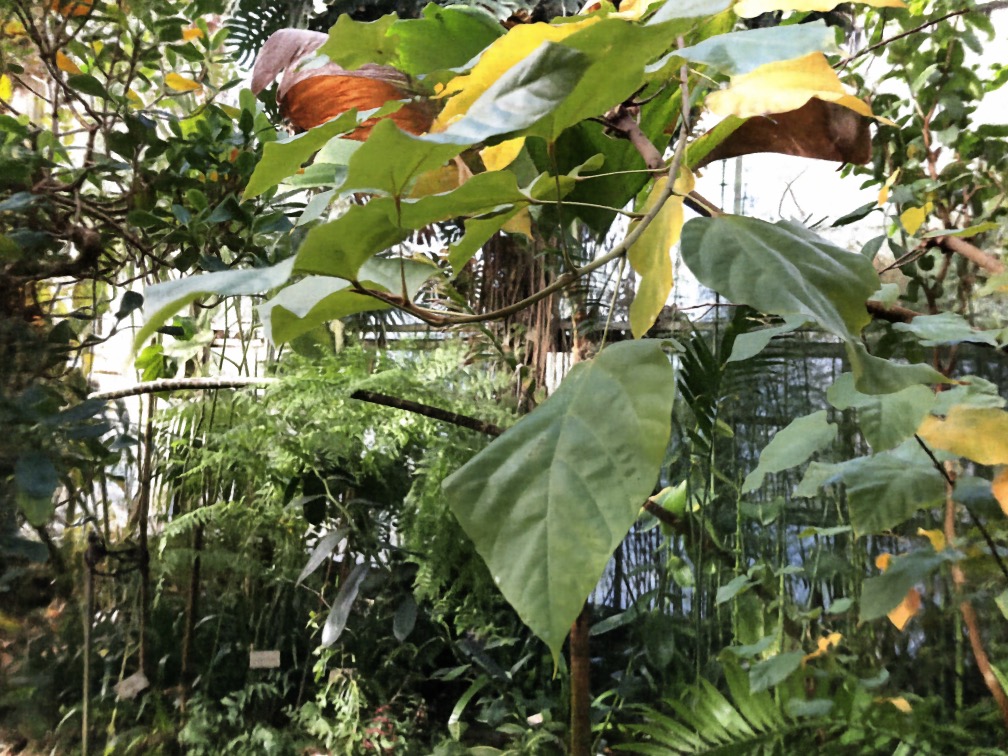}%
    ~
    \SuppThreeSubfig{0 0 0 0}{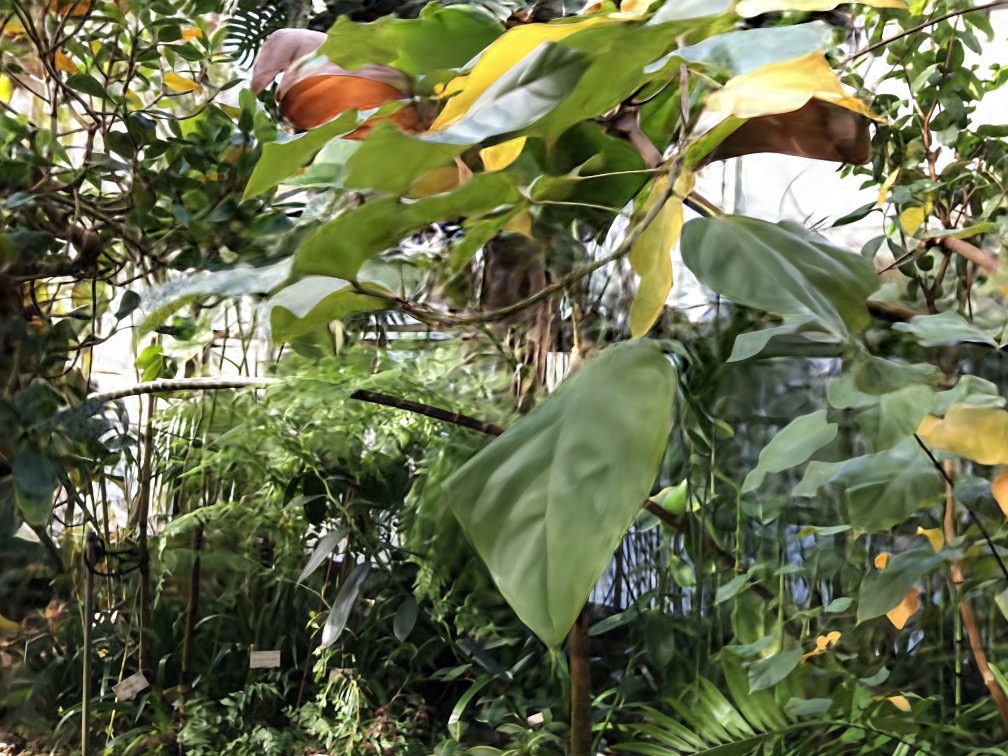}%
    ~
    \SuppThreeSubfig{0 0 0 0}{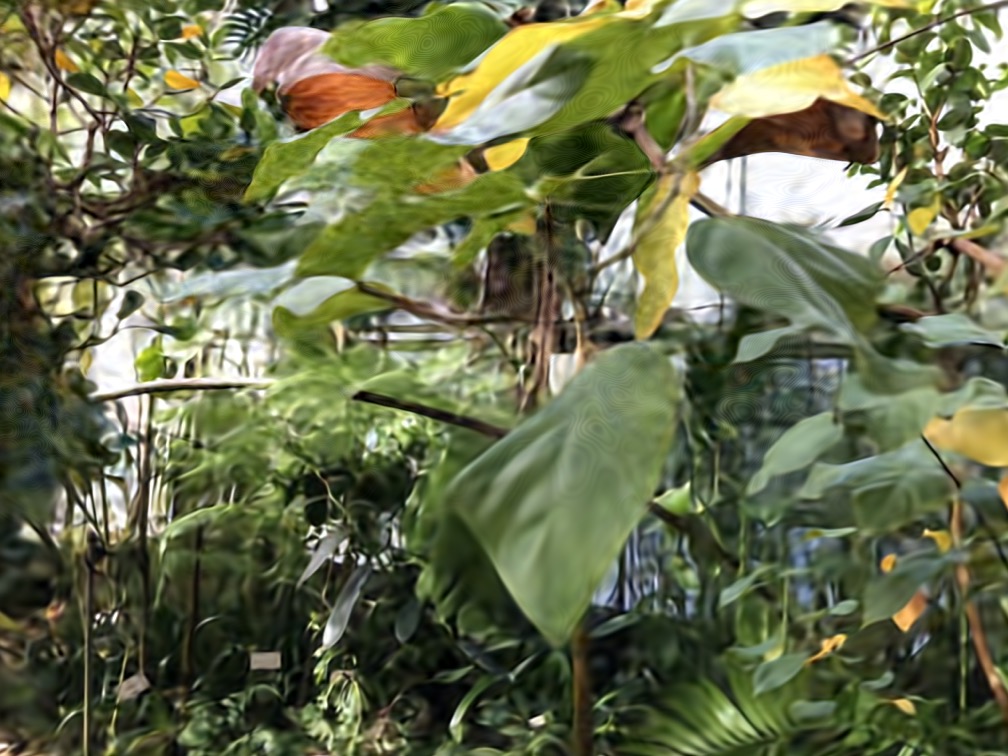}%
    
    \SuppThreeSubfig{0 0 0 0}{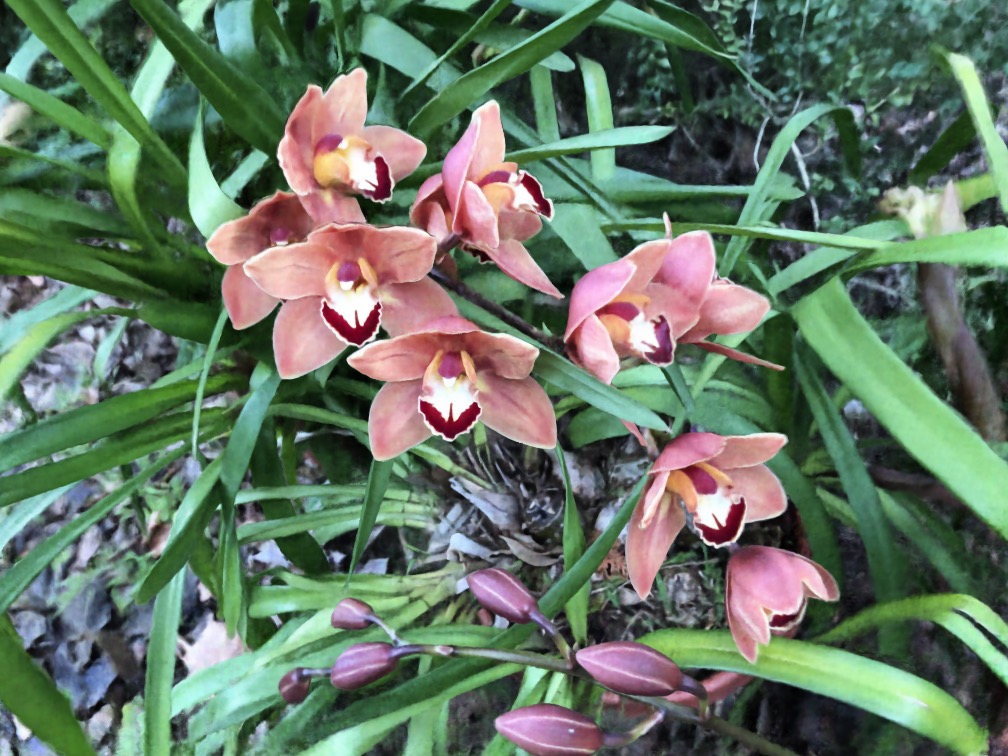}%
    ~
    \SuppThreeSubfig{0 0 0 0}{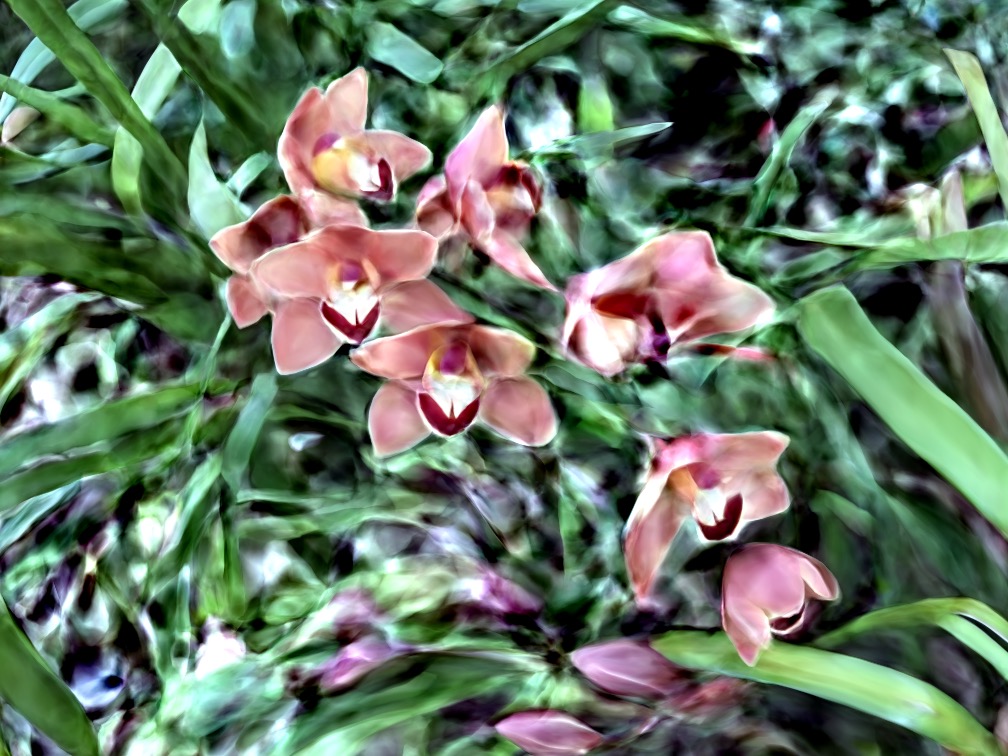}%
    ~
    \SuppThreeSubfig{0 0 0 0}{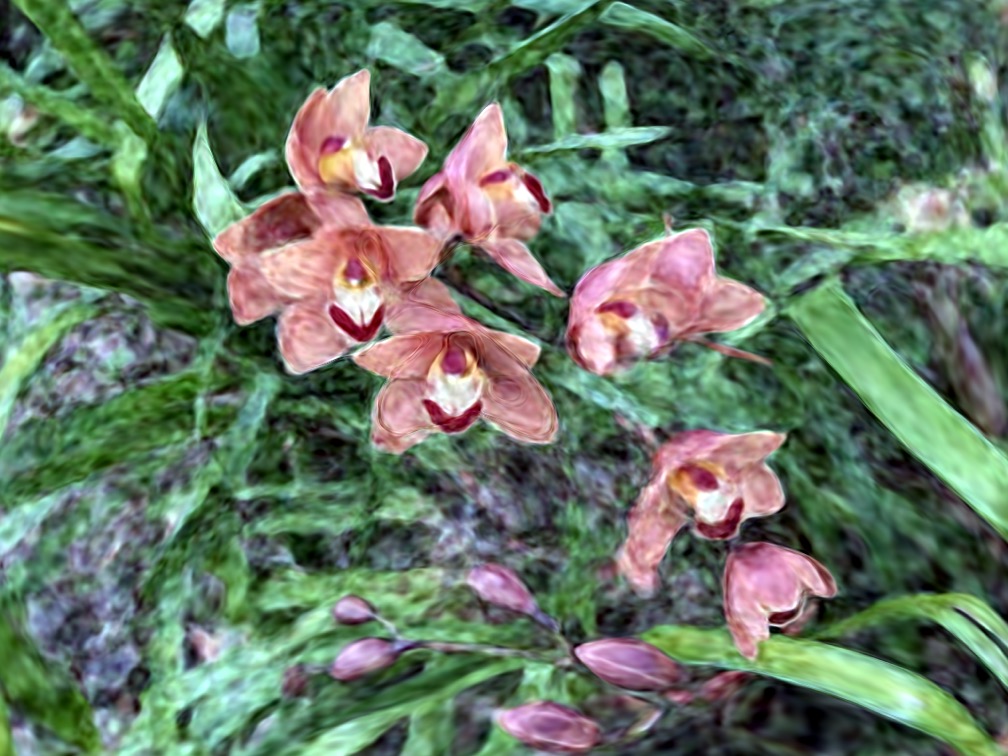}%
    
    \SuppThreeSubfig{0 0 0 0}{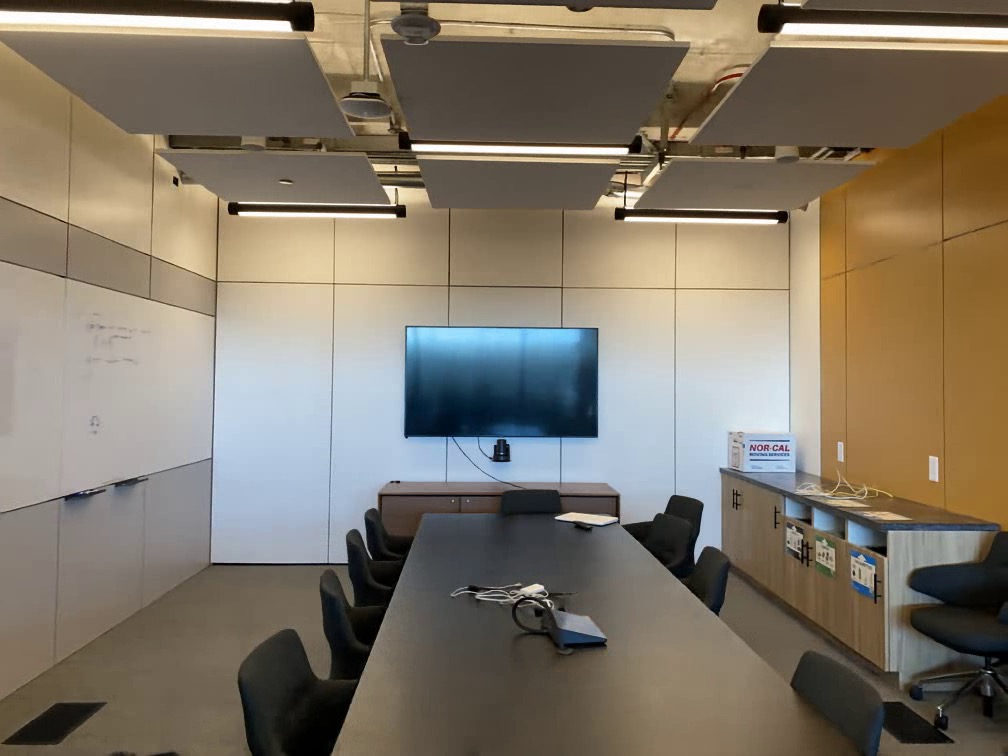}%
    ~
    \SuppThreeSubfig{0 0 0 0}{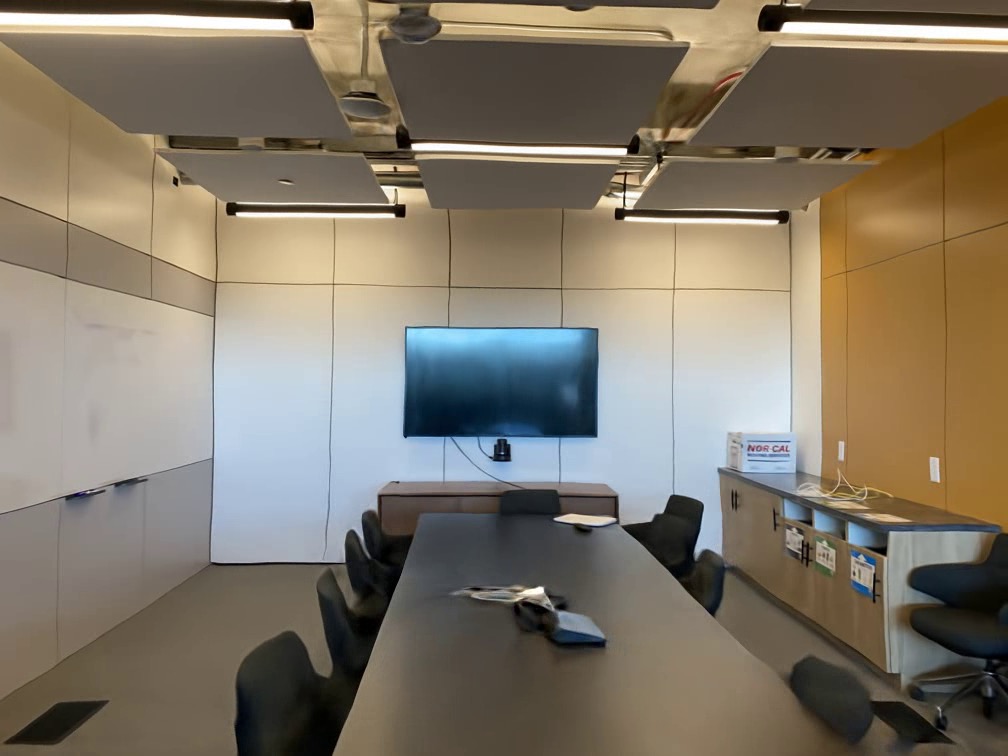}%
    ~
    \SuppThreeSubfig{0 0 0 0}{./figures/LLFF/Room/14.jpeg}%
    
    \SuppThreeSubfig{0 0 0 0}{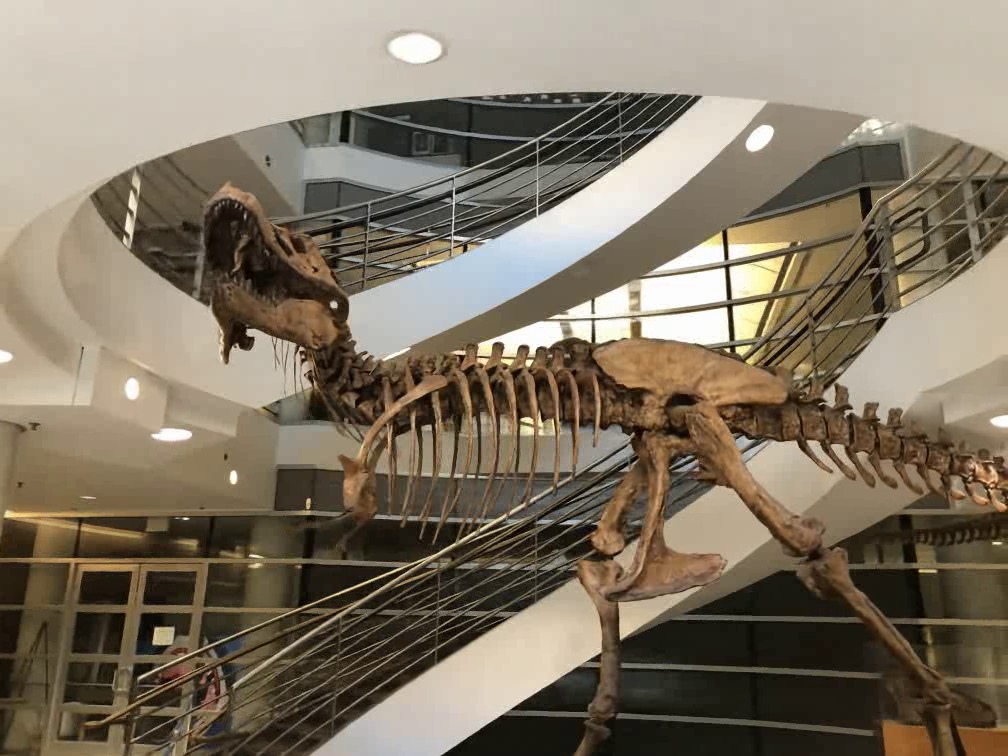}%
    ~
    \SuppThreeSubfig{0 0 0 0}{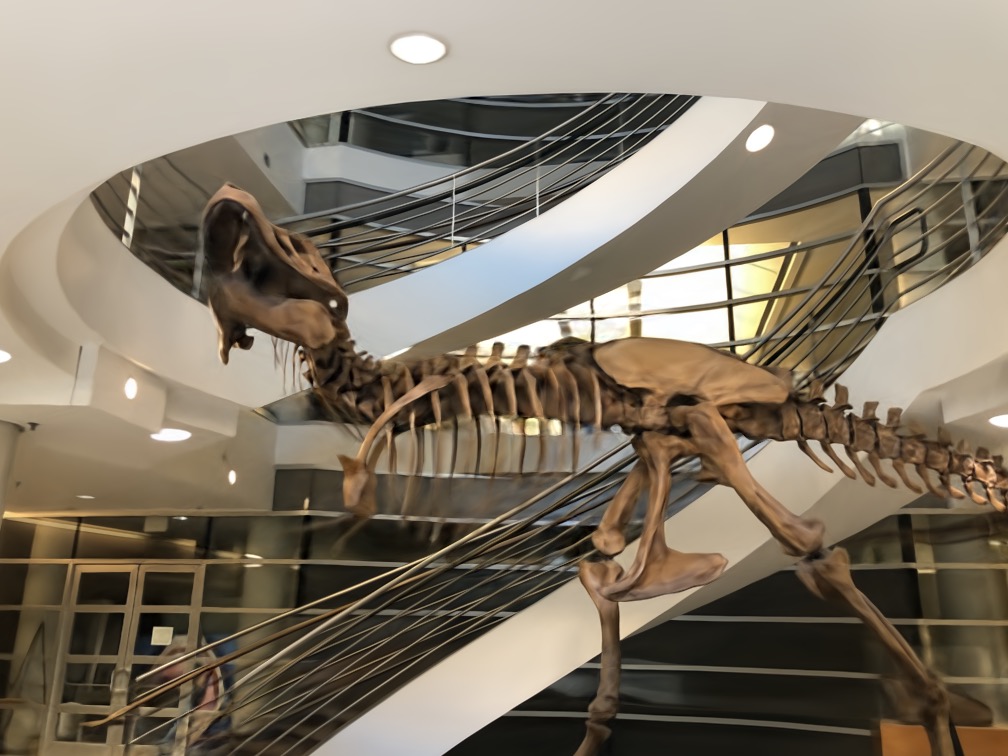}%
    ~
    \SuppThreeSubfig{0 0 0 0}{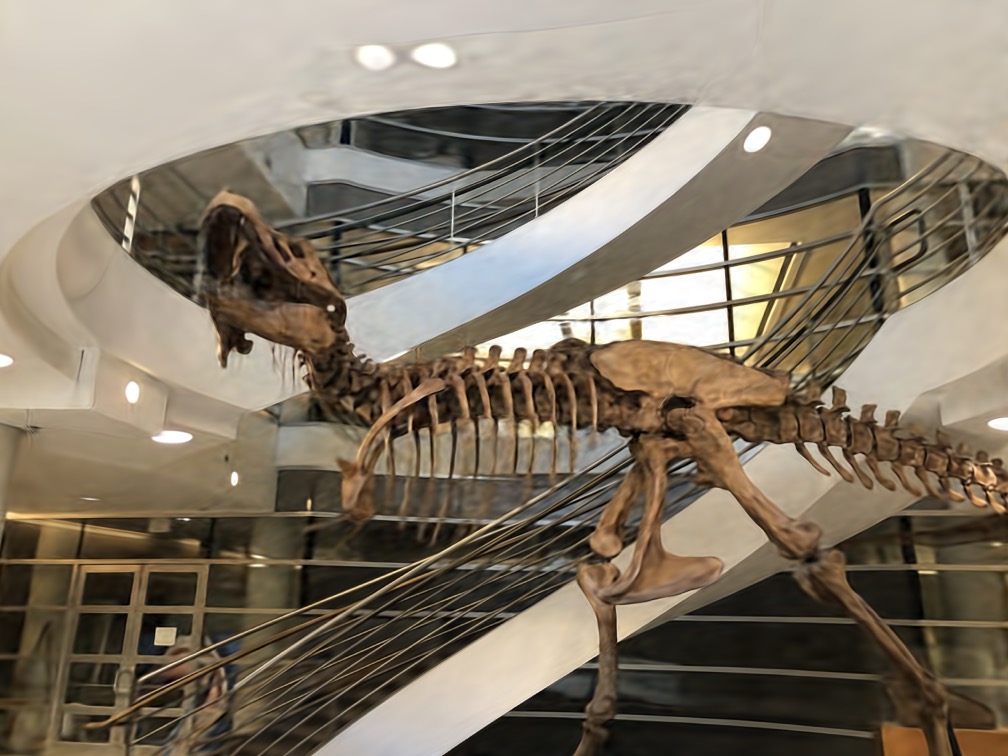}%

    \vspace{-5pt}
	\caption{Comparison With Baselines. From left to right: NeRF, LFN, and Ours.}
\end{figure*}

\clearpage

\begin{figure*}[!ht]
    \SuppLFSubfig{0 0 0 0}{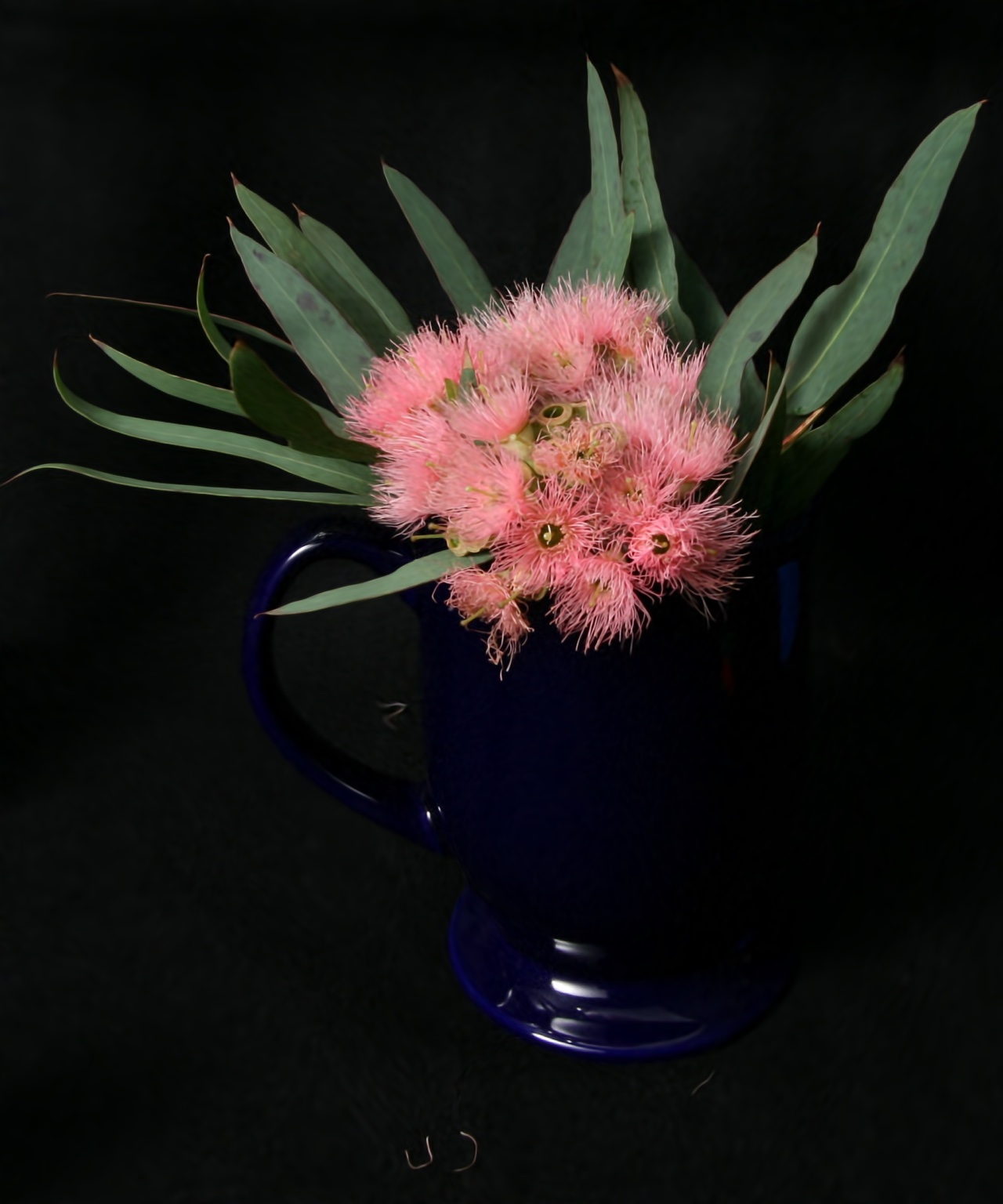}%
    ~
    \SuppLFSubfig{0 0 0 0}{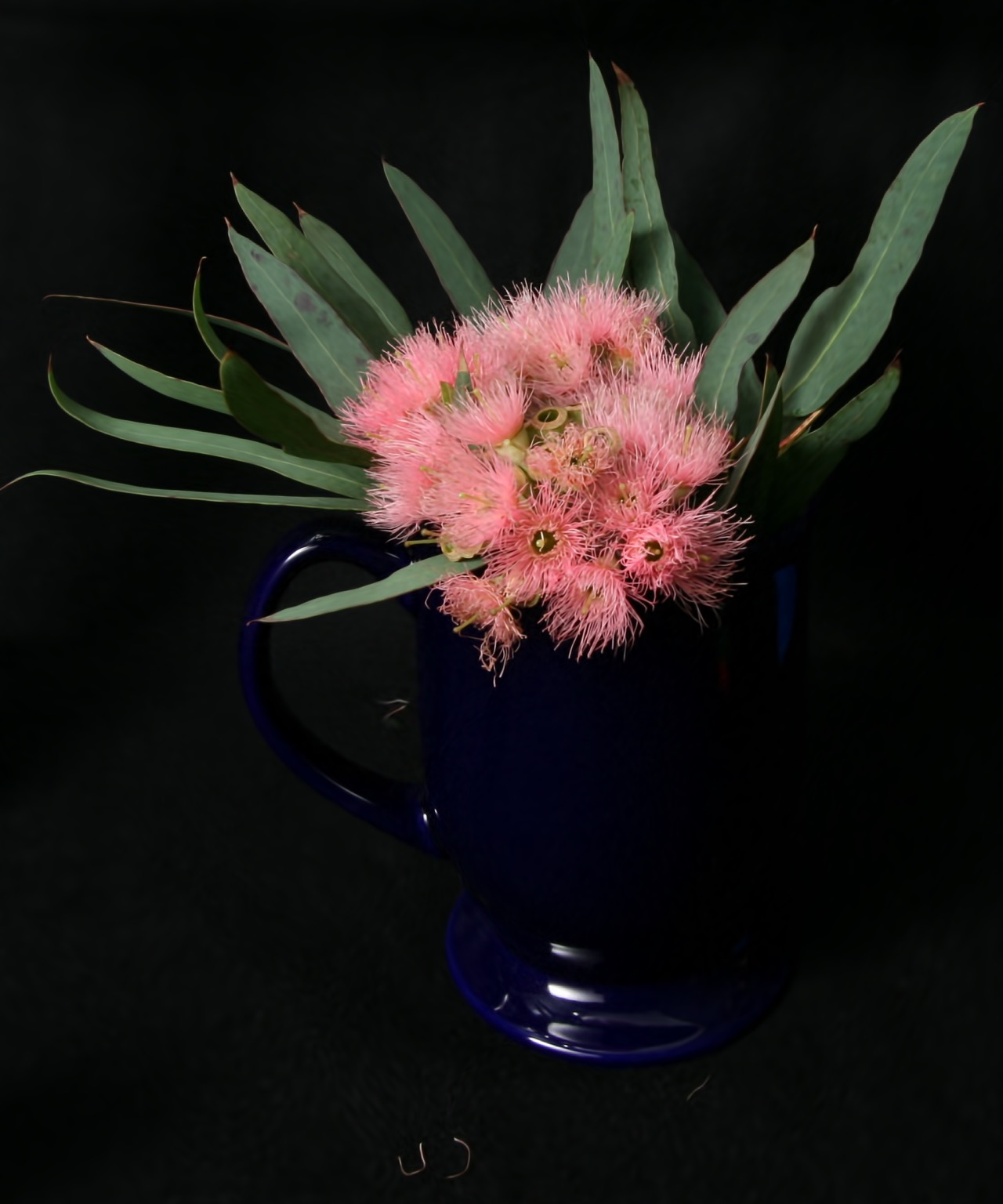}%
    ~
    \SuppLFSubfig{0 0 0 0}{./figures/4DLF/flowers/14.jpeg}%
    ~
    \SuppLFSubfig{0 0 0 0}{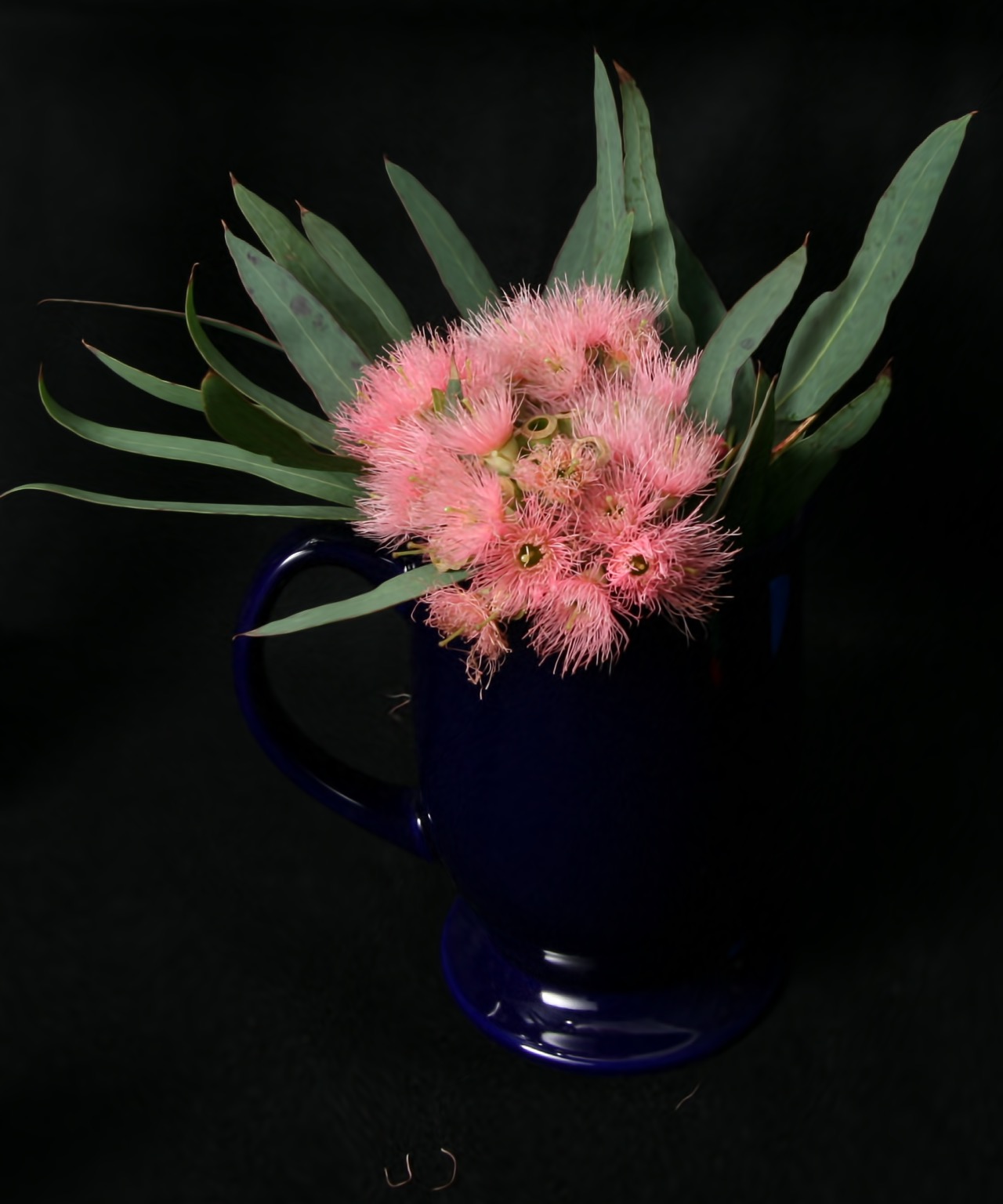}%
    ~
    \SuppLFSubfig{0 0 0 0}{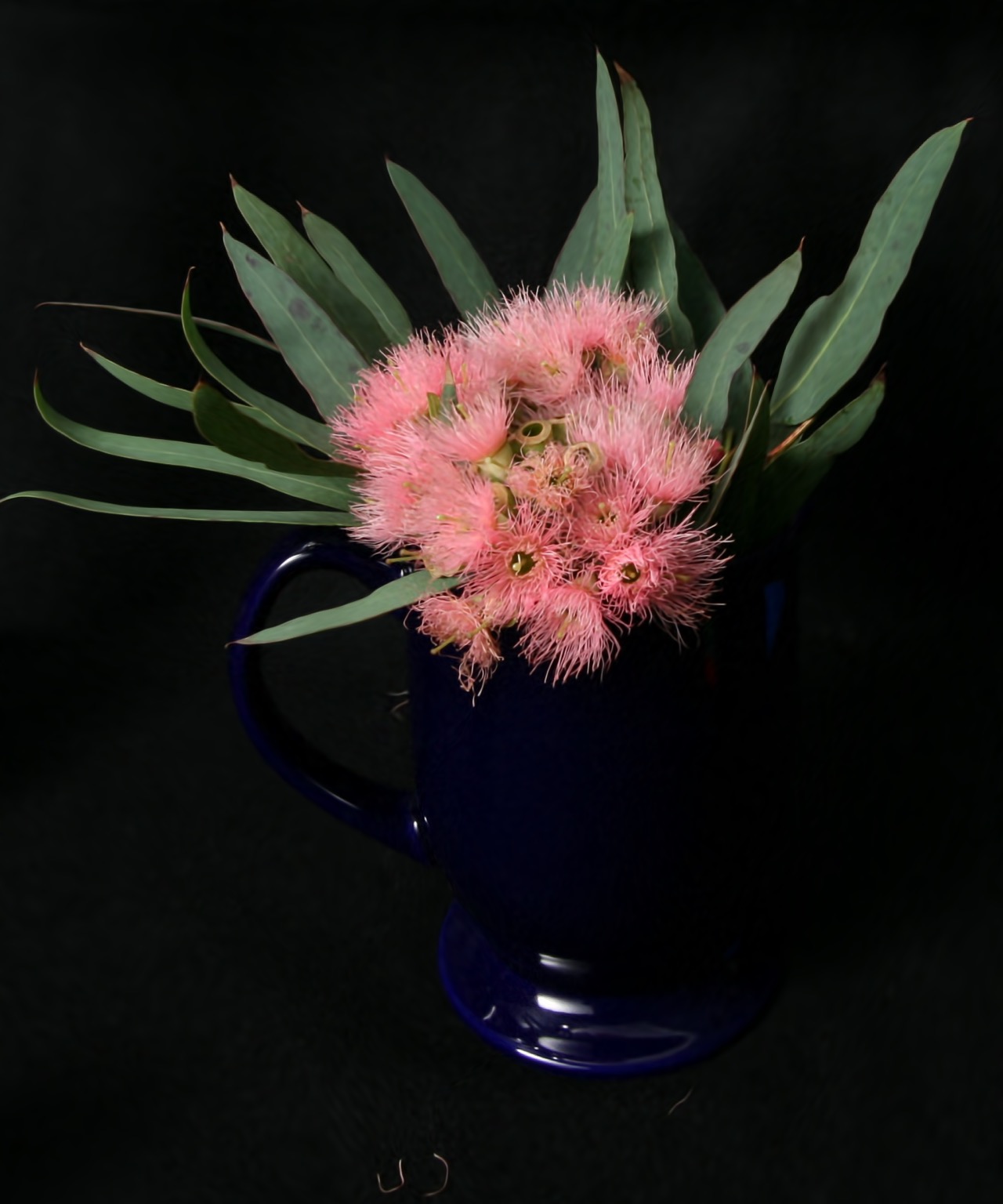}%

    \SuppLFSubfig{0 0 0 0}{./figures/4DLF/lego/0.jpeg}%
    ~
    \SuppLFSubfig{0 0 0 0}{./figures/4DLF/lego/7.jpeg}%
    ~
    \SuppLFSubfig{0 0 0 0}{./figures/4DLF/lego/14.jpeg}%
    ~
    \SuppLFSubfig{0 0 0 0}{./figures/4DLF/lego/21.jpeg}%
    ~
    \SuppLFSubfig{0 0 0 0}{./figures/4DLF/lego/29.jpeg}%

    \SuppLFSubfig{0 0 0 0}{./figures/4DLF/tarot/0.jpeg}%
    ~
    \SuppLFSubfig{0 0 0 0}{./figures/4DLF/tarot/7.jpeg}%
    ~
    \SuppLFSubfig{0 0 0 0}{./figures/4DLF/tarot/14.jpeg}%
    ~
    \SuppLFSubfig{0 0 0 0}{./figures/4DLF/tarot/21.jpeg}%
    ~
    \SuppLFSubfig{0 0 0 0}{./figures/4DLF/tarot/29.jpeg}%

    \SuppLFSubfig{0 0 0 0}{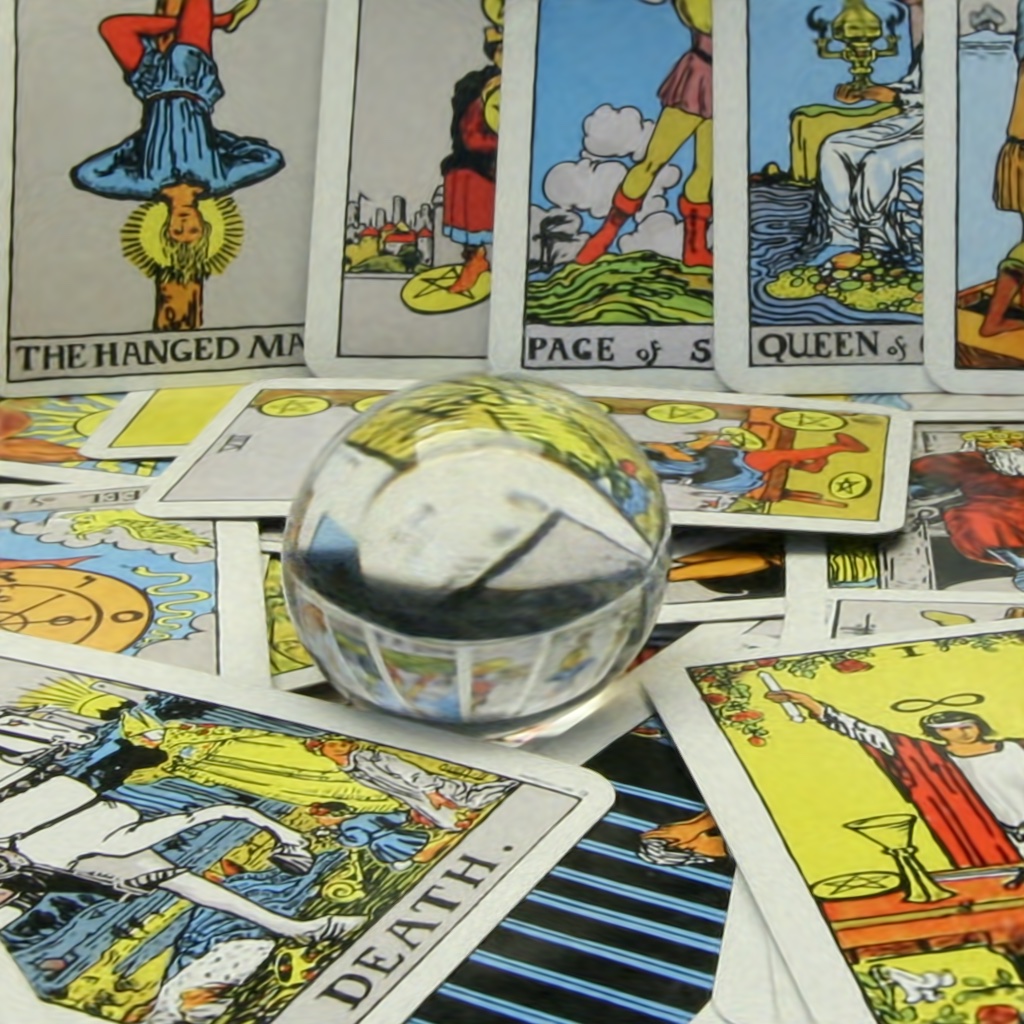}%
    ~
    \SuppLFSubfig{0 0 0 0}{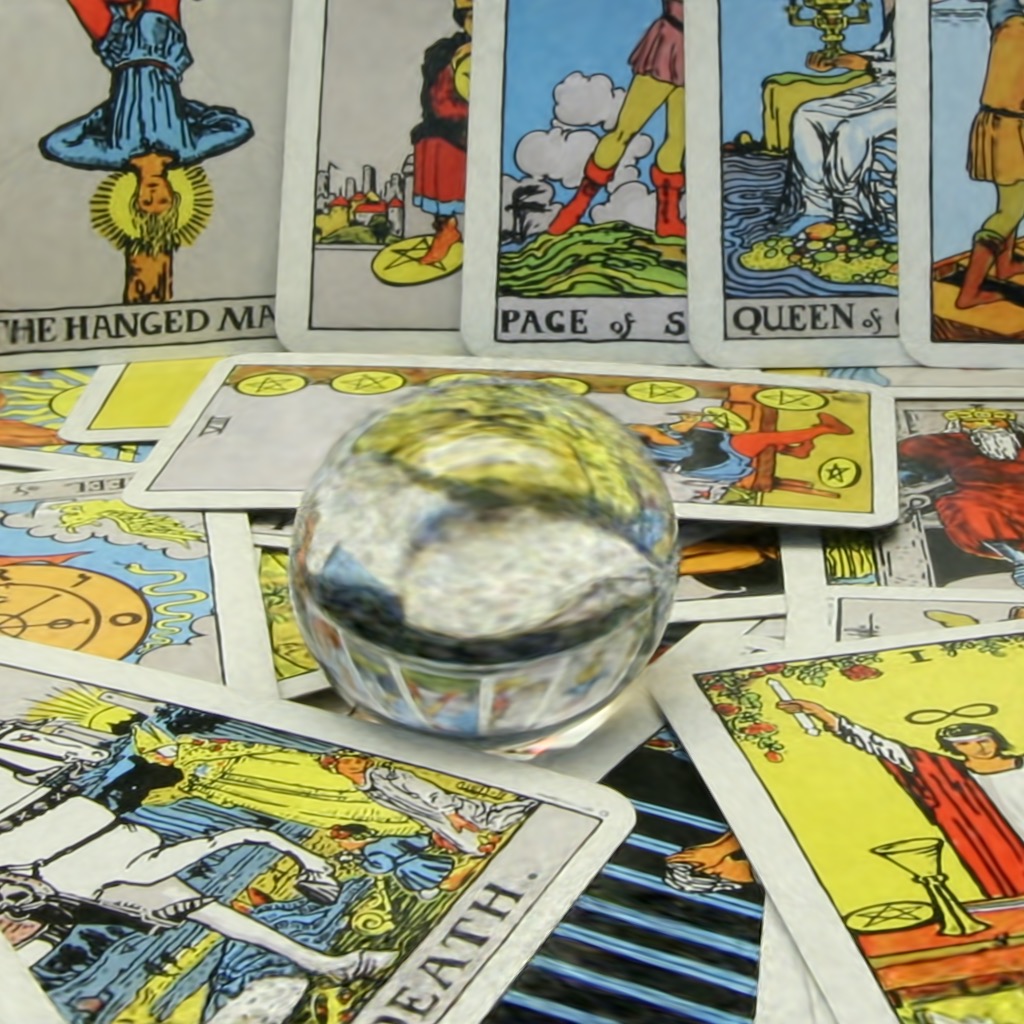}%
    ~
    \SuppLFSubfig{0 0 0 0}{./figures/4DLF/tarot_L/14.jpeg}%
    ~
    \SuppLFSubfig{0 0 0 0}{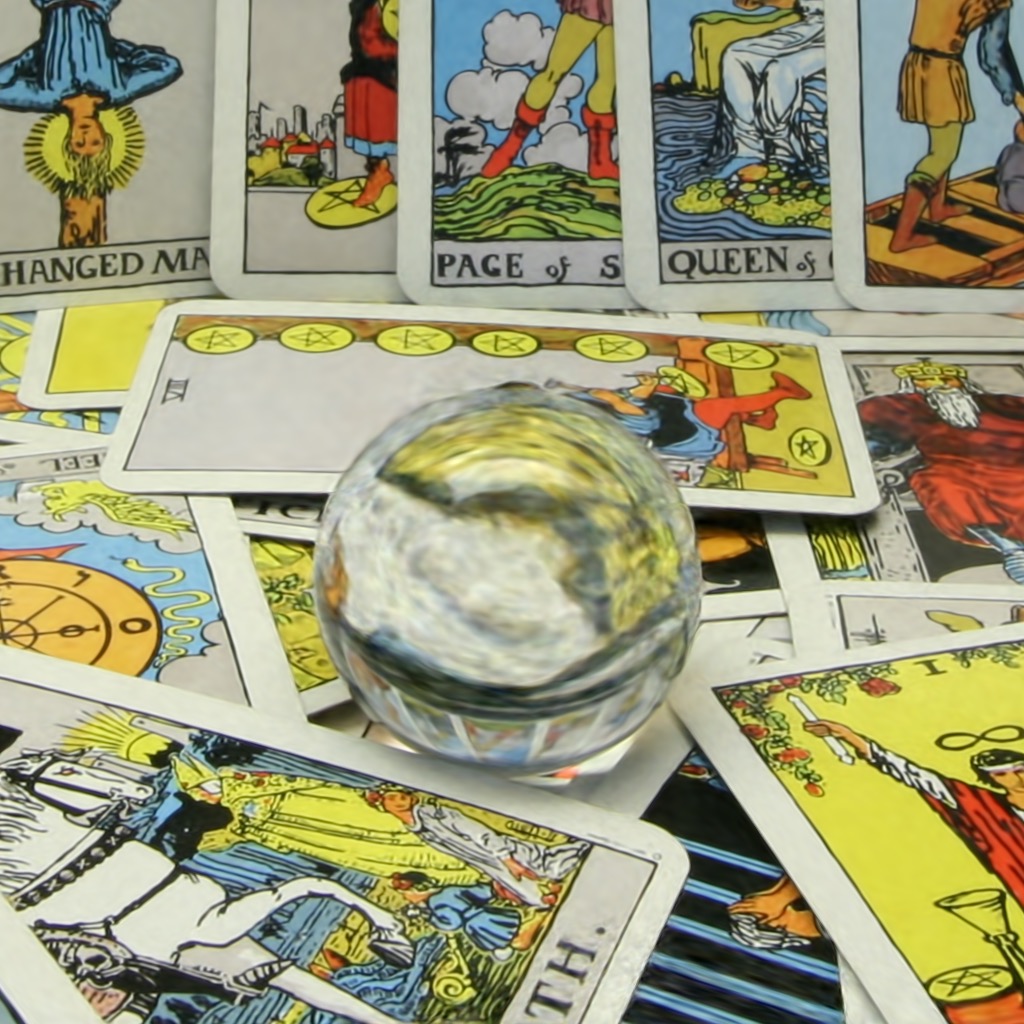}%
    ~
    \SuppLFSubfig{0 0 0 0}{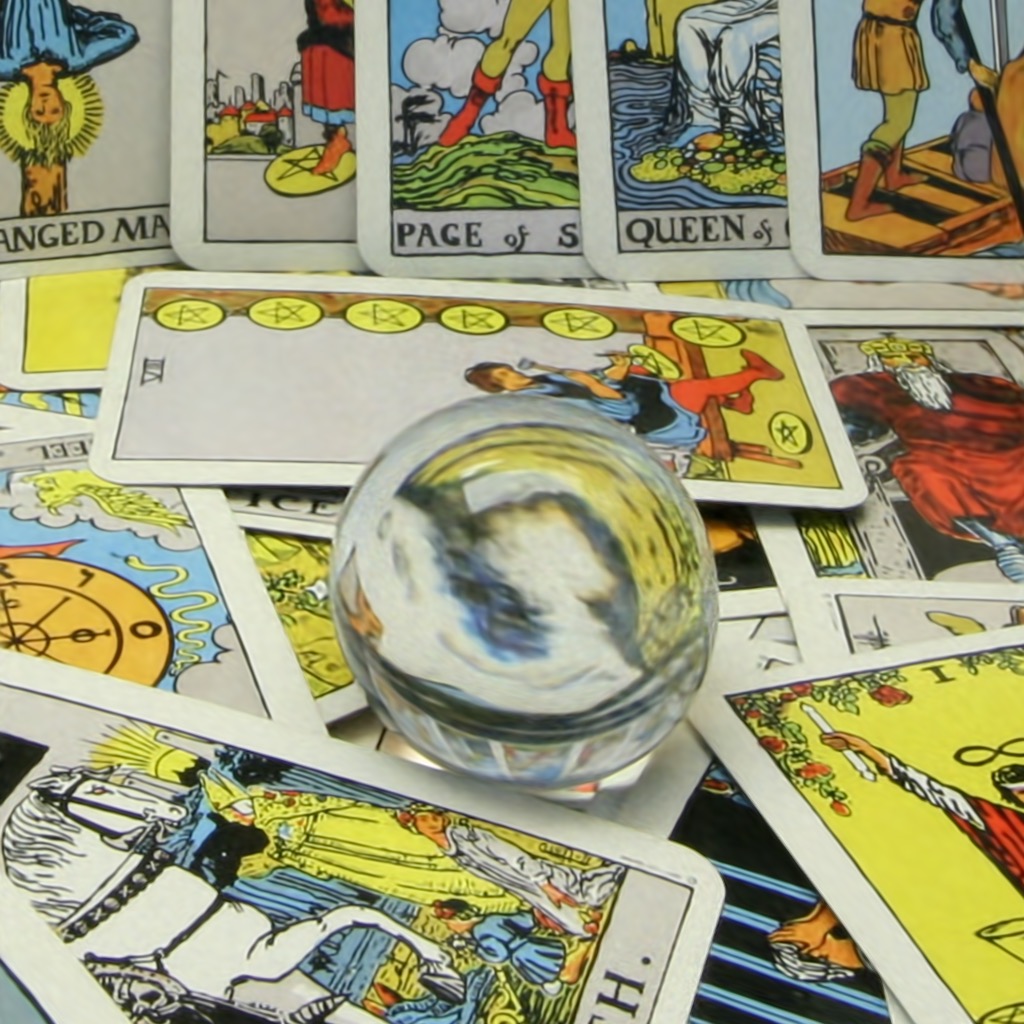}%

    \SuppLFSubfig{0 0 0 0}{./figures/4DLF/treasure/0.jpeg}%
    ~
    \SuppLFSubfig{0 0 0 0}{./figures/4DLF/treasure/7.jpeg}%
    ~
    \SuppLFSubfig{0 0 0 0}{./figures/4DLF/treasure/14.jpeg}%
    ~
    \SuppLFSubfig{0 0 0 0}{./figures/4DLF/treasure/21.jpeg}%
    ~
    \SuppLFSubfig{0 0 0 0}{./figures/4DLF/treasure/29.jpeg}%

    \SuppLFSubfig{0 0 0 0}{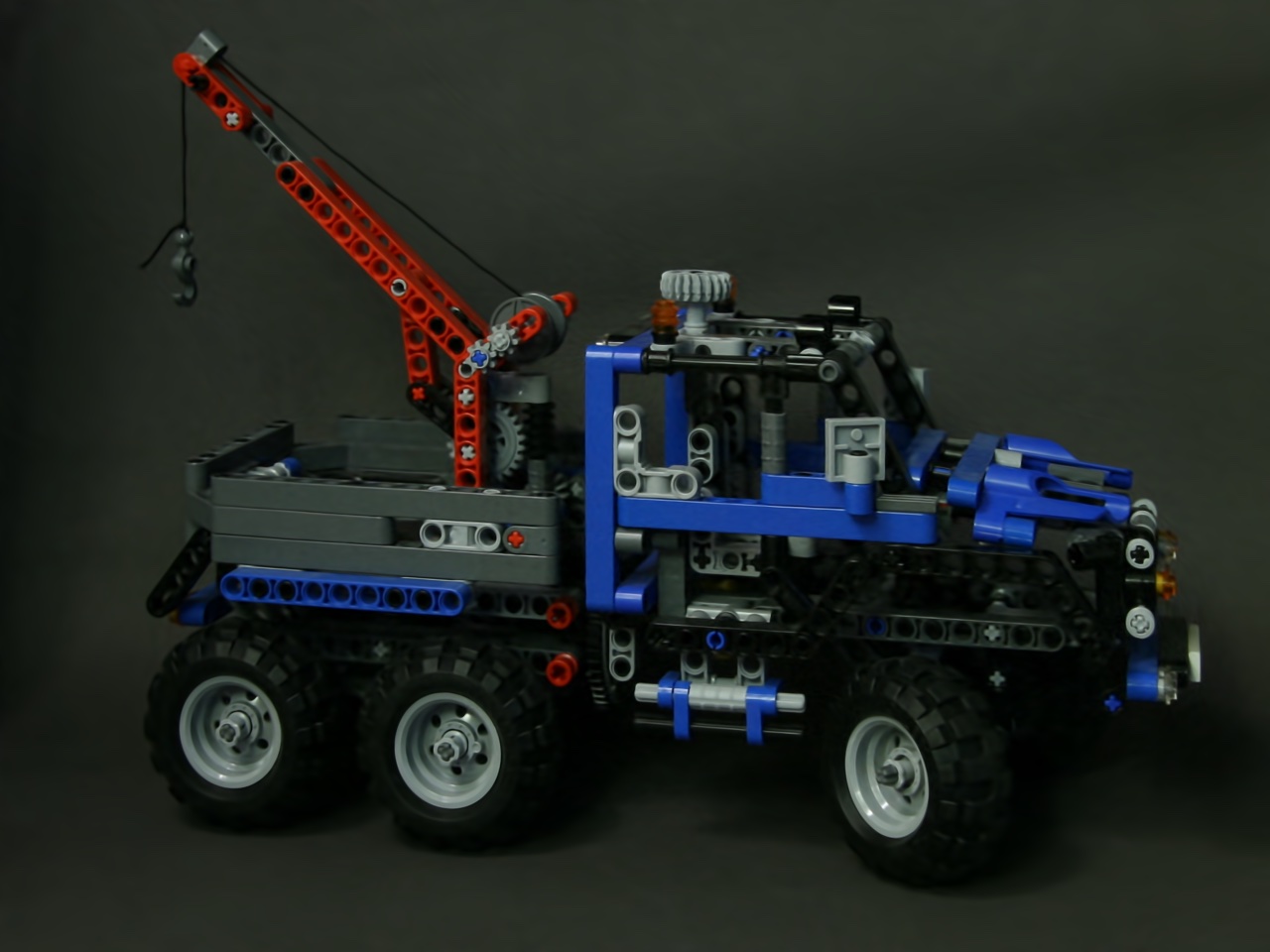}%
    ~
    \SuppLFSubfig{0 0 0 0}{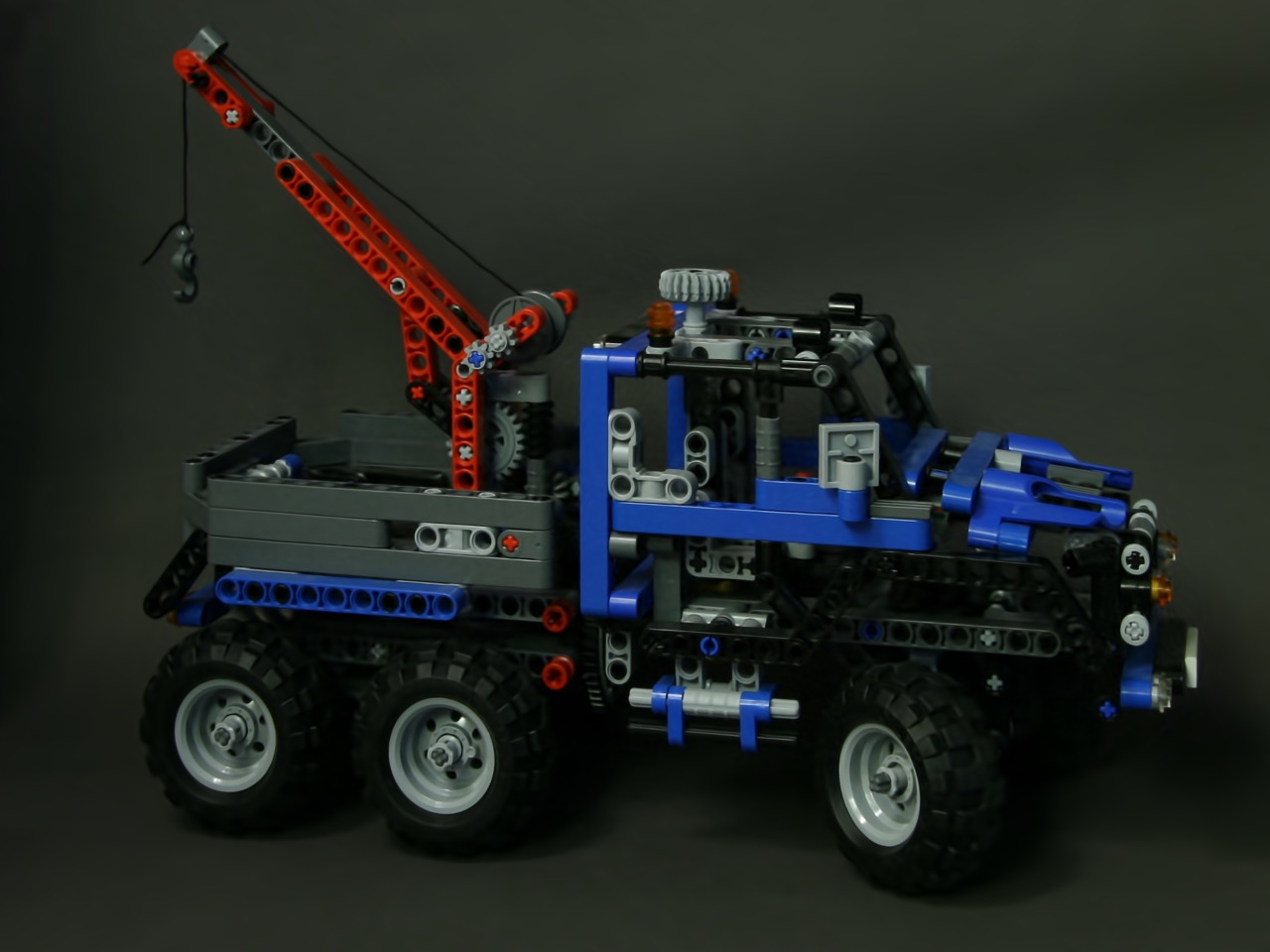}%
    ~
    \SuppLFSubfig{0 0 0 0}{./figures/4DLF/truck/14.jpeg}%
    ~
    \SuppLFSubfig{0 0 0 0}{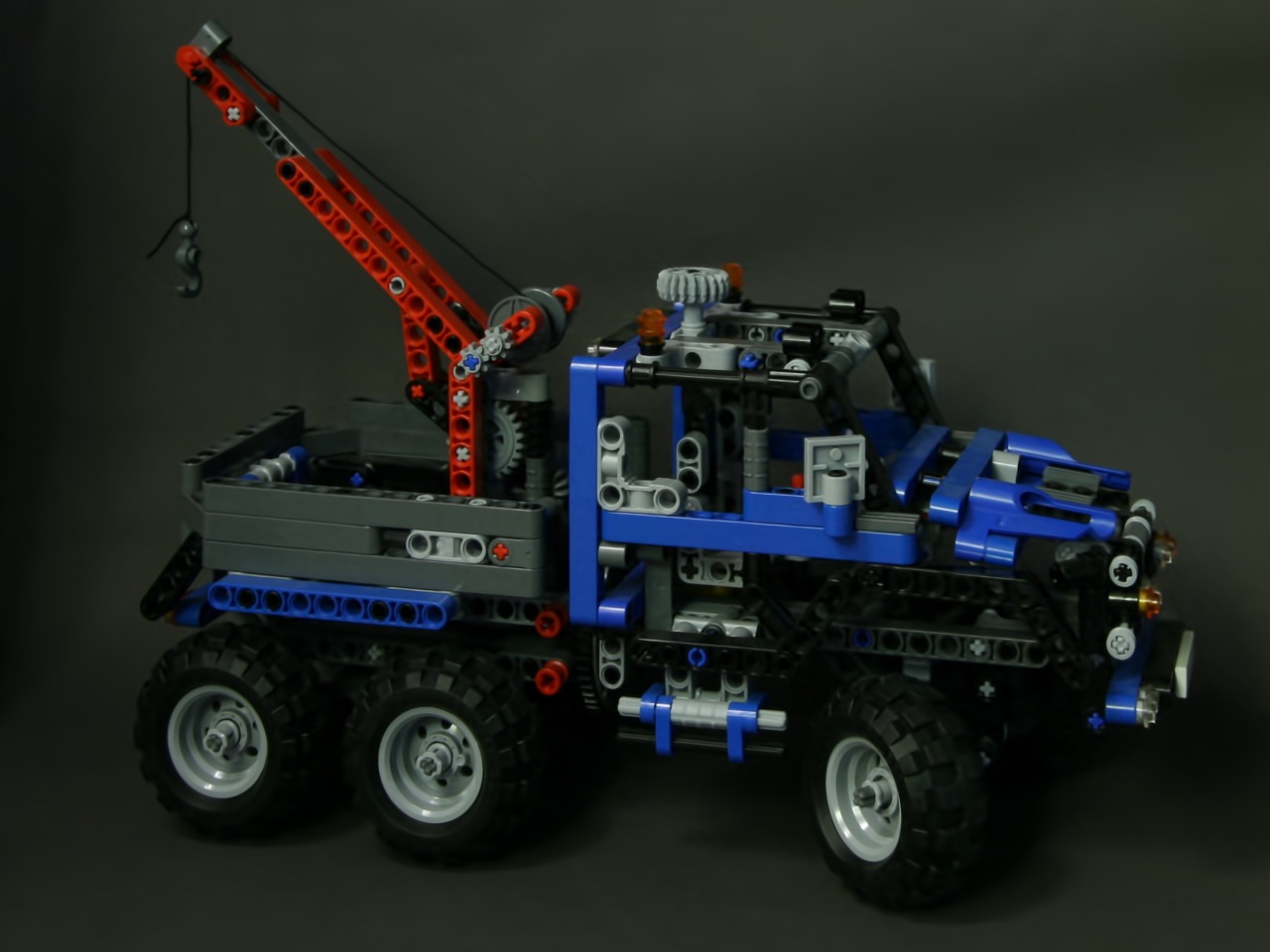}%
    ~
    \SuppLFSubfig{0 0 0 0}{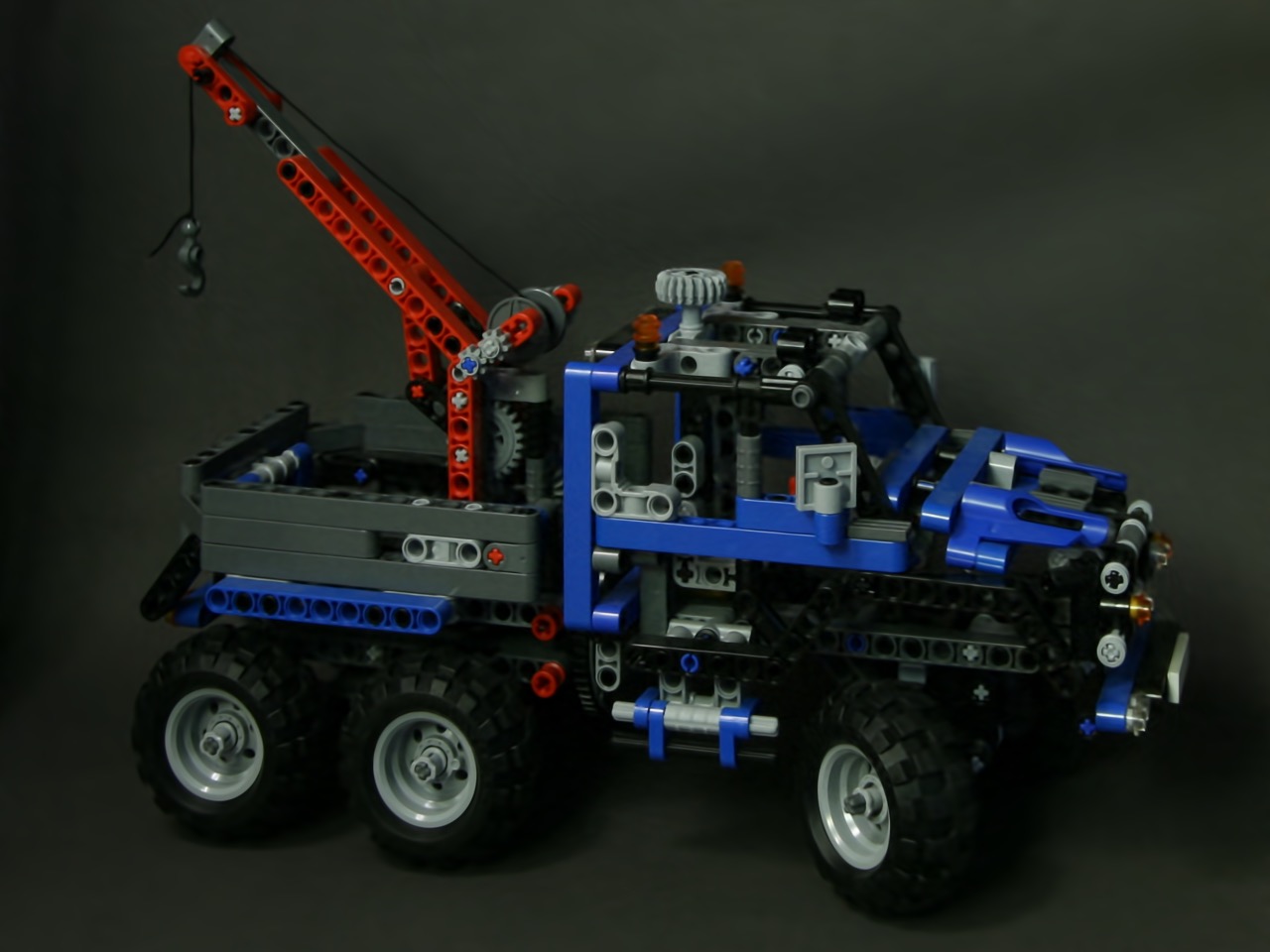}%

    \FigFiveSubfigCaption{$t=0$}%
    ~
    \FigFiveSubfigCaption{$t=0.25$}%
    ~
    \FigFiveSubfigCaption{$t=0.5$}%
    ~
    \FigFiveSubfigCaption{$t=0.75$}%
    ~
    \FigFiveSubfigCaption{$t=1$}%
    \vspace{-5pt}
	\caption{Stanford 4D Light Fields Results. We interpolate the learned codes from from two non-adjacent viewpoints (from top-left to bottom-right). See supplementary video for better contrast.}
\end{figure*}

\begin{figure*}[!ht]
    \SuppLFSubfig{0 0 0 0}{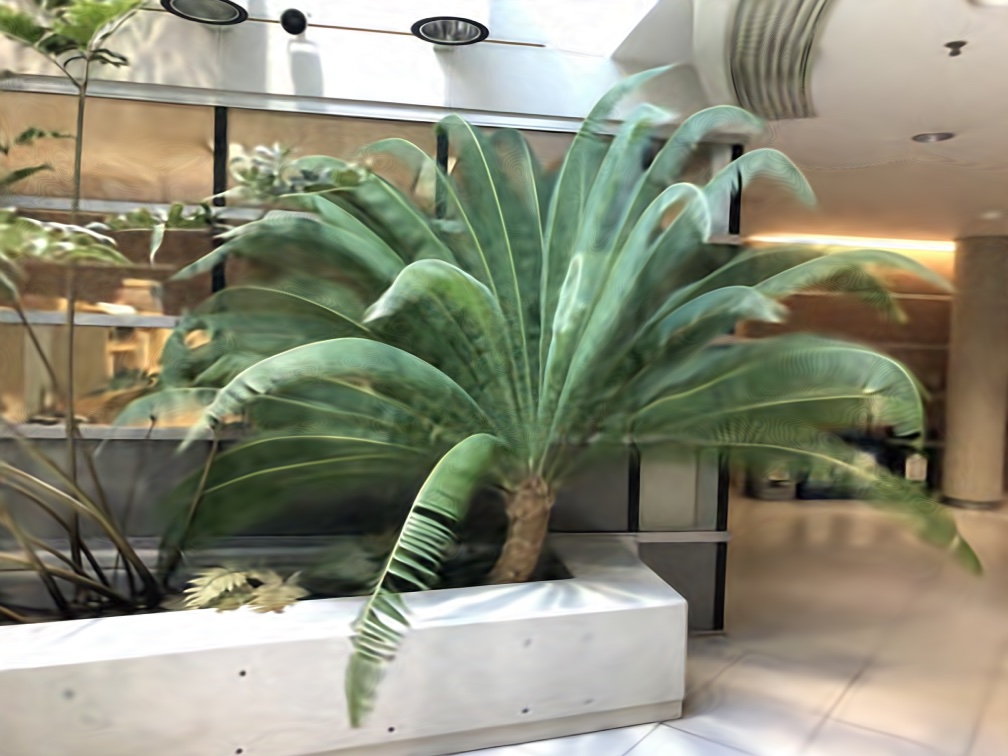}%
    ~
    \SuppLFSubfig{0 0 0 0}{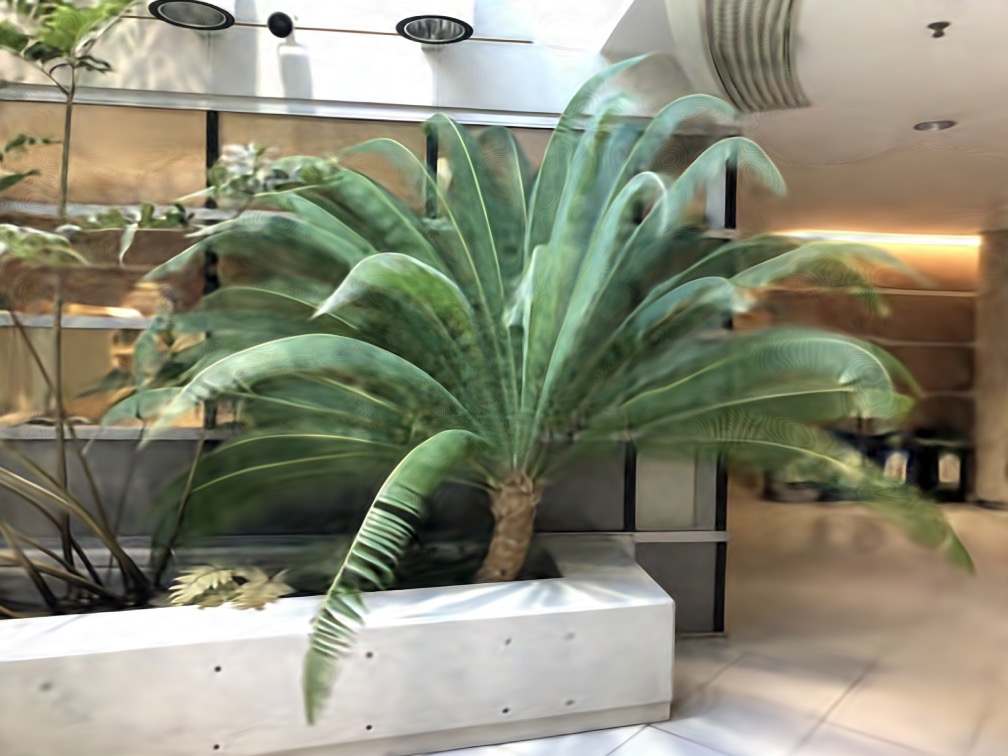}%
    ~
    \SuppLFSubfig{0 0 0 0}{./figures/LLFF/Fern/14.jpeg}%
    ~
    \SuppLFSubfig{0 0 0 0}{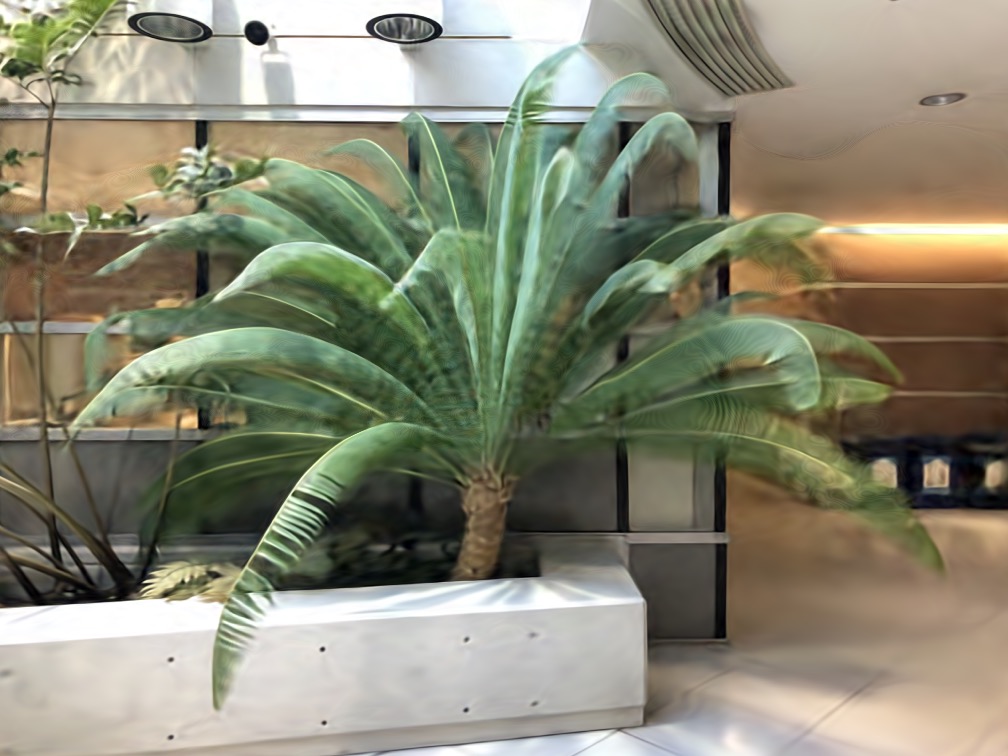}%
    ~
    \SuppLFSubfig{0 0 0 0}{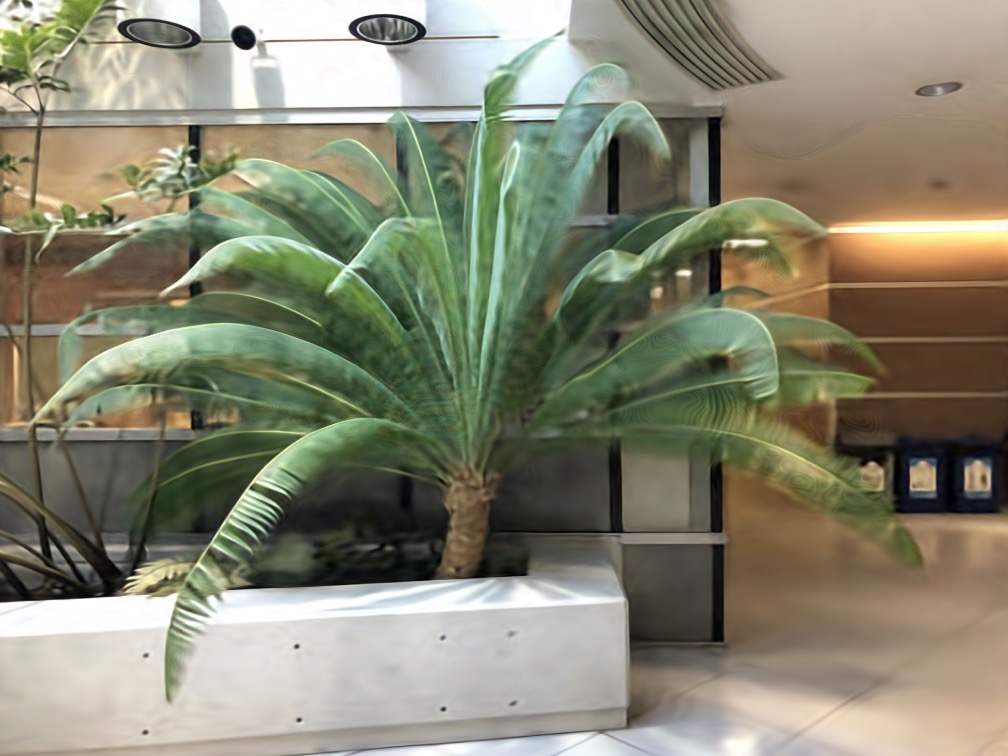}%

    \SuppLFSubfig{0 0 0 0}{./figures/LLFF/Flower/0.jpeg}%
    ~
    \SuppLFSubfig{0 0 0 0}{./figures/LLFF/Flower/7.jpeg}%
    ~
    \SuppLFSubfig{0 0 0 0}{./figures/LLFF/Flower/14.jpeg}%
    ~
    \SuppLFSubfig{0 0 0 0}{./figures/LLFF/Flower/21.jpeg}%
    ~
    \SuppLFSubfig{0 0 0 0}{./figures/LLFF/Flower/29.jpeg}%

    \SuppLFSubfig{0 0 0 0}{./figures/LLFF/Fortress/0.jpeg}%
    ~
    \SuppLFSubfig{0 0 0 0}{./figures/LLFF/Fortress/7.jpeg}%
    ~
    \SuppLFSubfig{0 0 0 0}{./figures/LLFF/Fortress/14.jpeg}%
    ~
    \SuppLFSubfig{0 0 0 0}{./figures/LLFF/Fortress/21.jpeg}%
    ~
    \SuppLFSubfig{0 0 0 0}{./figures/LLFF/Fortress/29.jpeg}%

    \SuppLFSubfig{0 0 0 0}{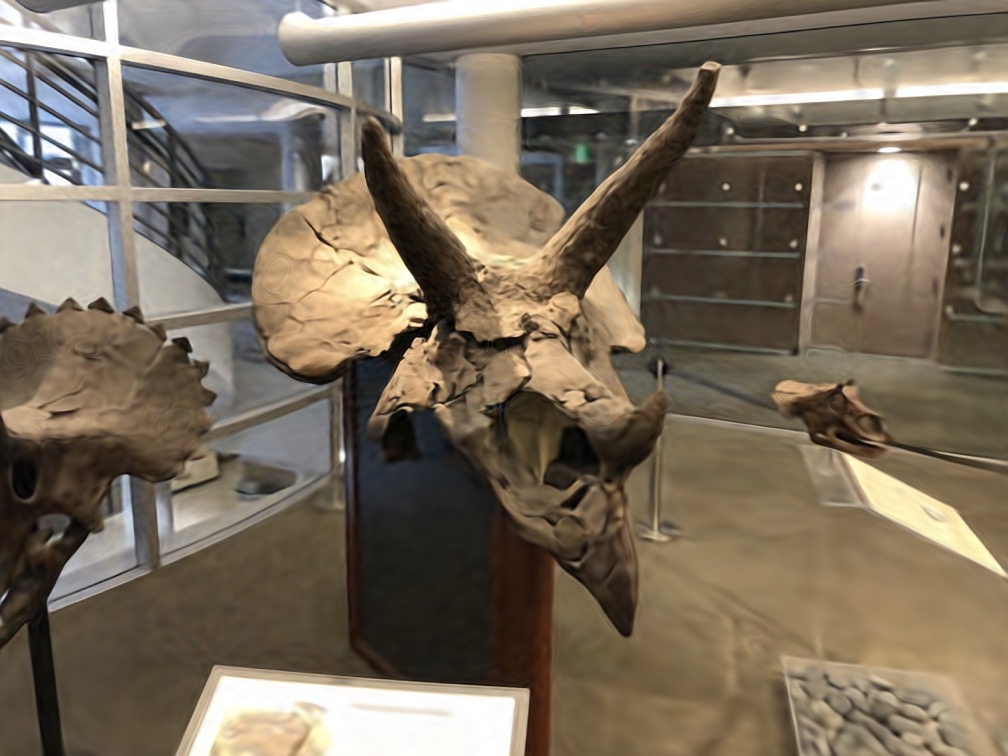}%
    ~
    \SuppLFSubfig{0 0 0 0}{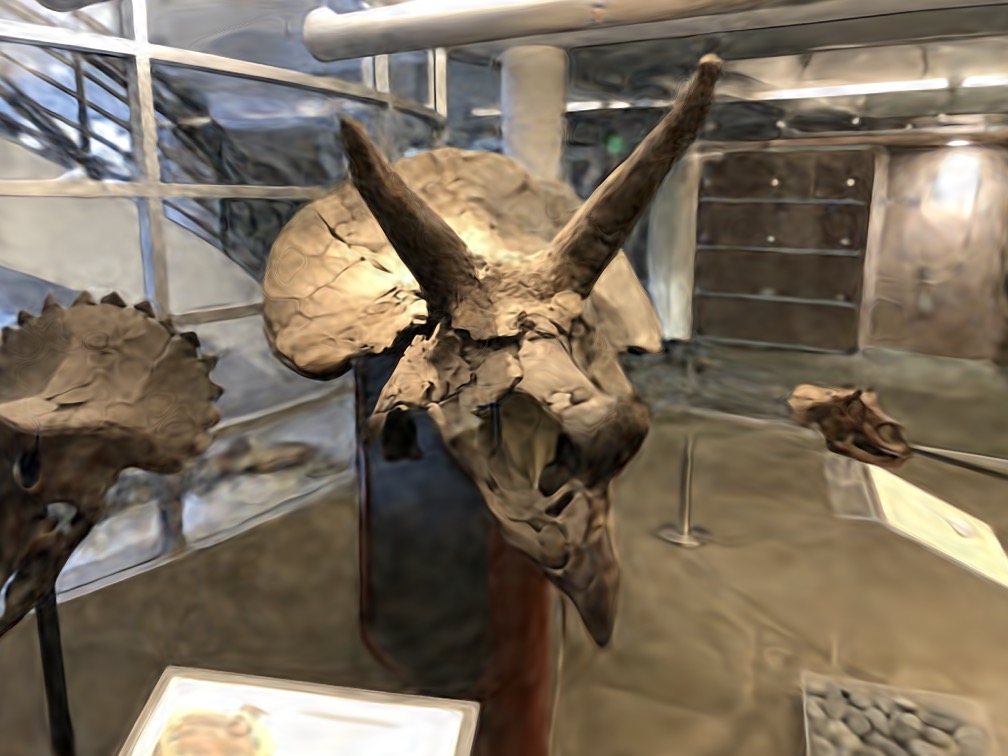}%
    ~
    \SuppLFSubfig{0 0 0 0}{./figures/LLFF/Horns/14.jpeg}%
    ~
    \SuppLFSubfig{0 0 0 0}{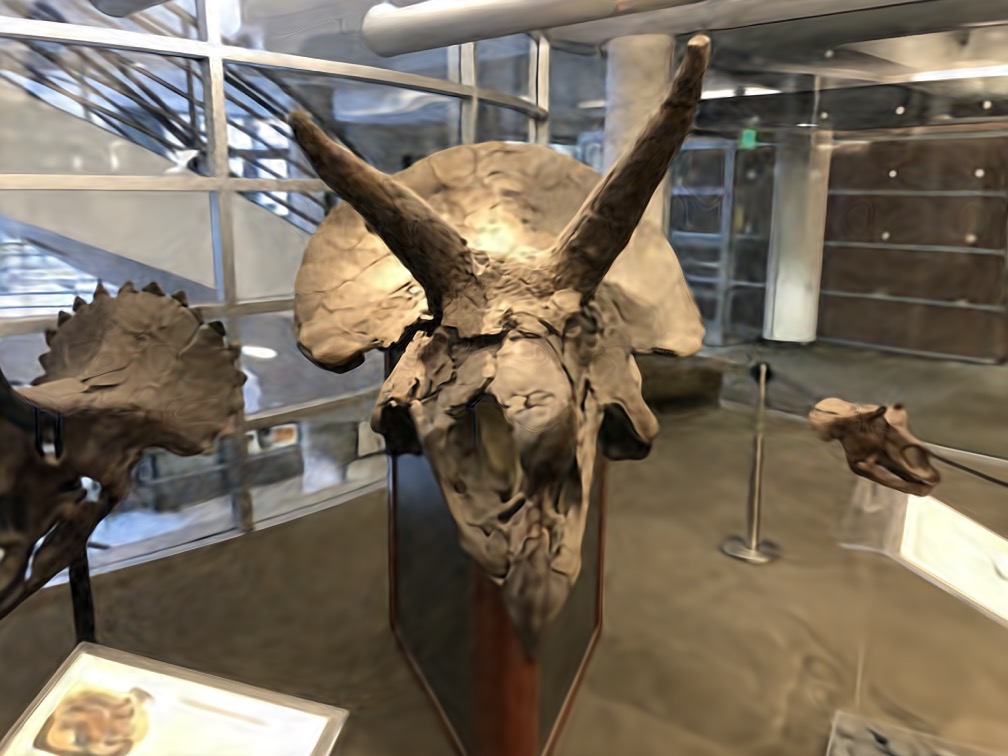}%
    ~
    \SuppLFSubfig{0 0 0 0}{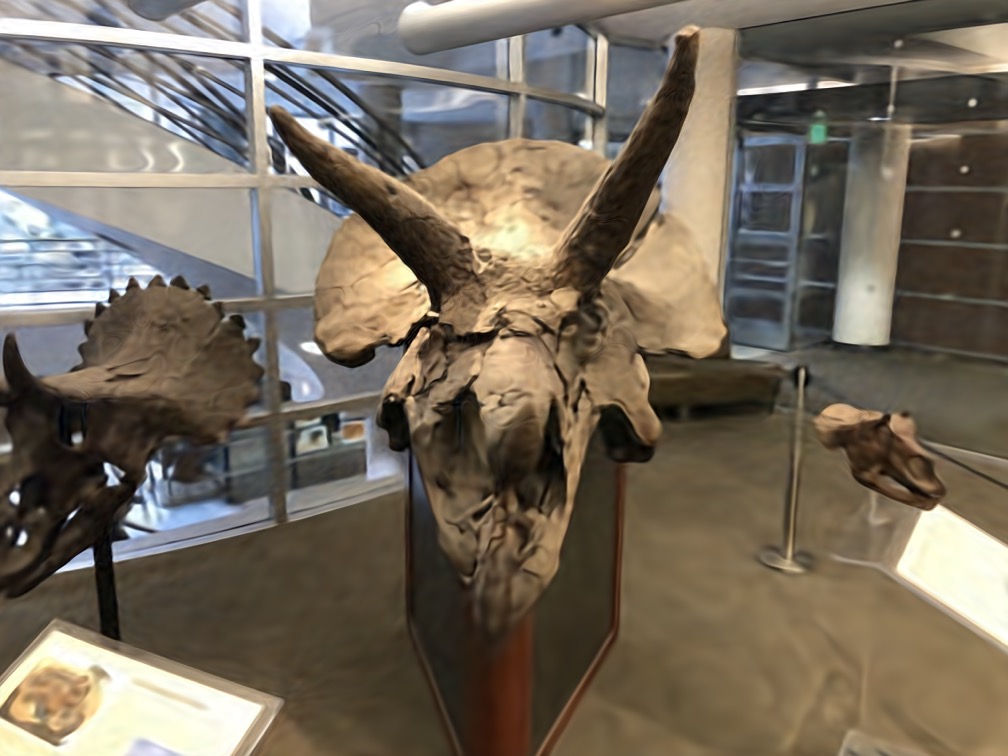}%

    \SuppLFSubfig{0 0 0 0}{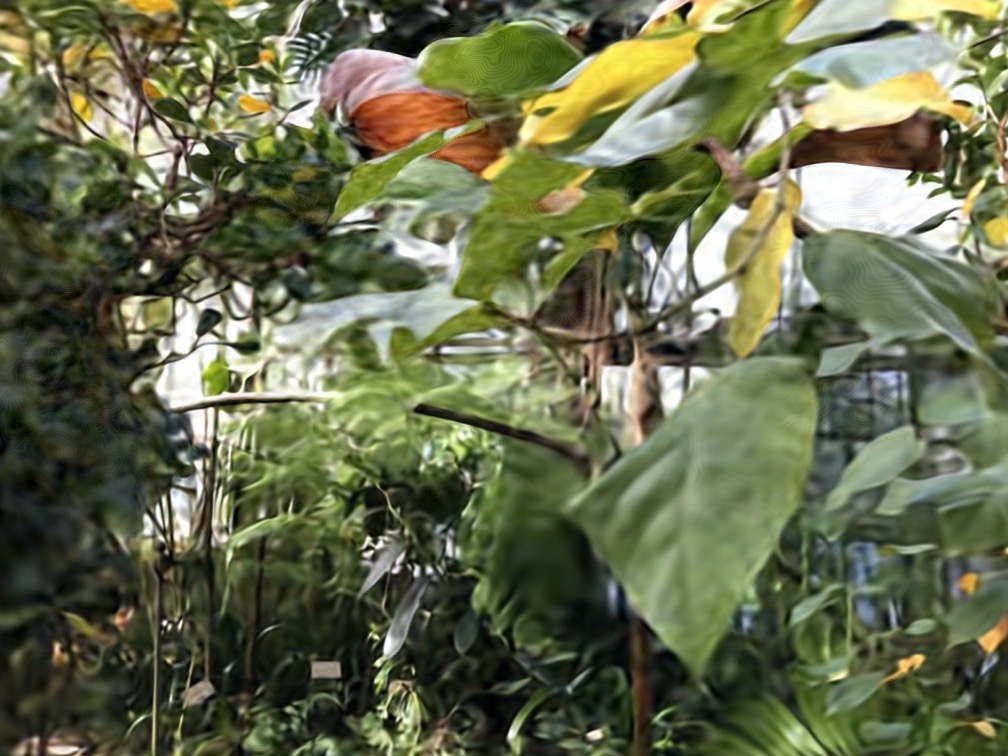}%
    ~
    \SuppLFSubfig{0 0 0 0}{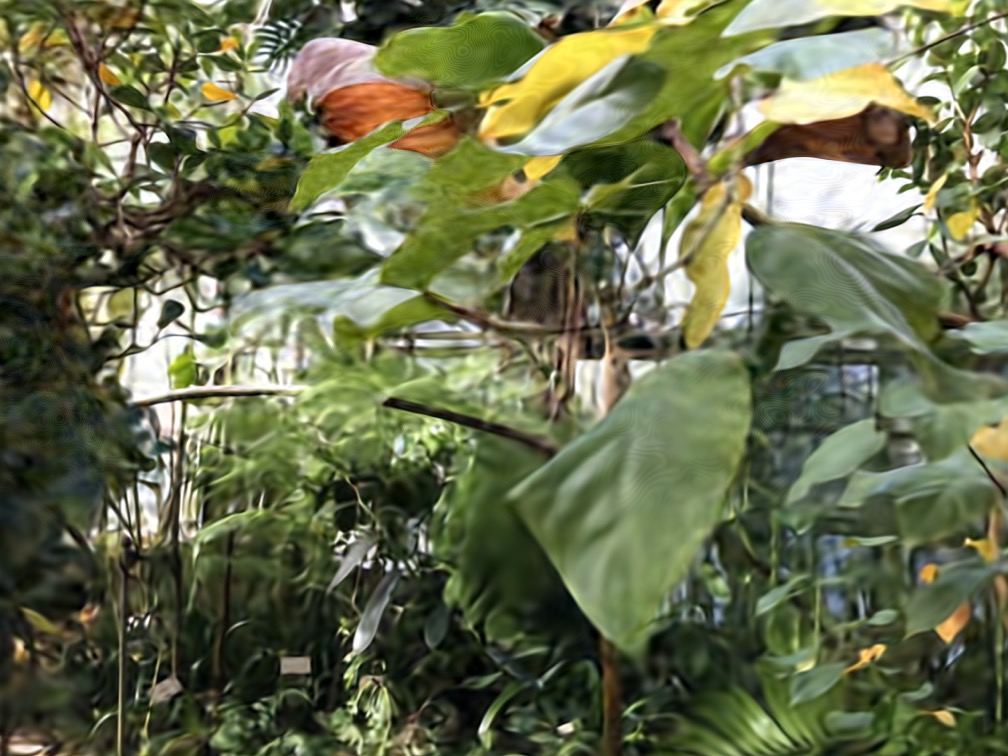}%
    ~
    \SuppLFSubfig{0 0 0 0}{./figures/LLFF/Leaves/14.jpeg}%
    ~
    \SuppLFSubfig{0 0 0 0}{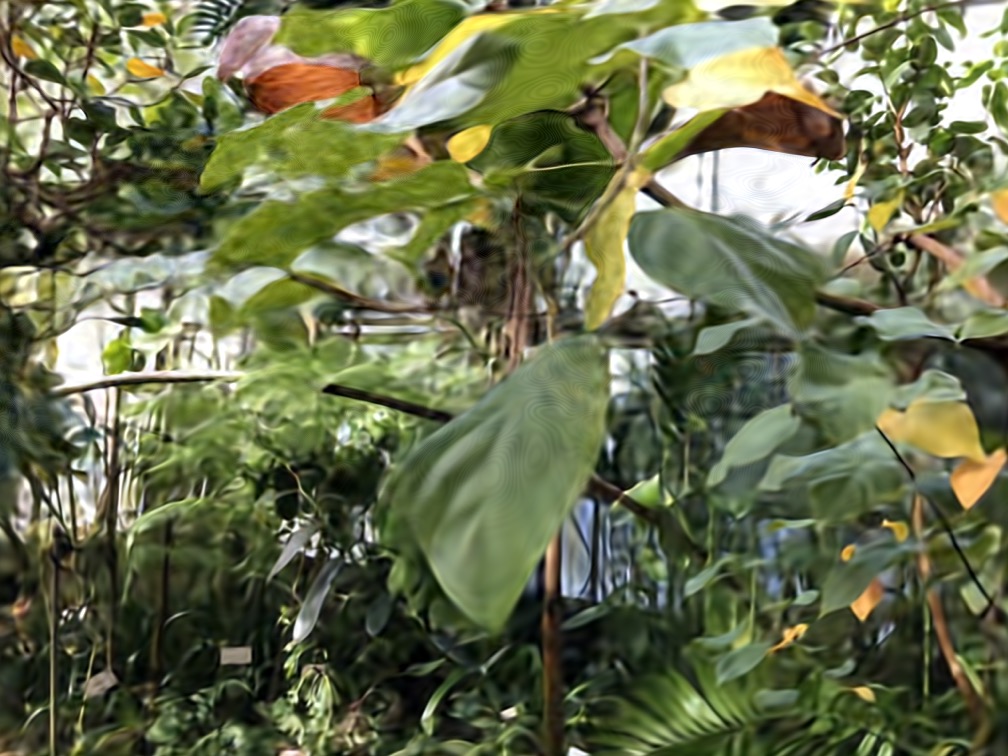}%
    ~
    \SuppLFSubfig{0 0 0 0}{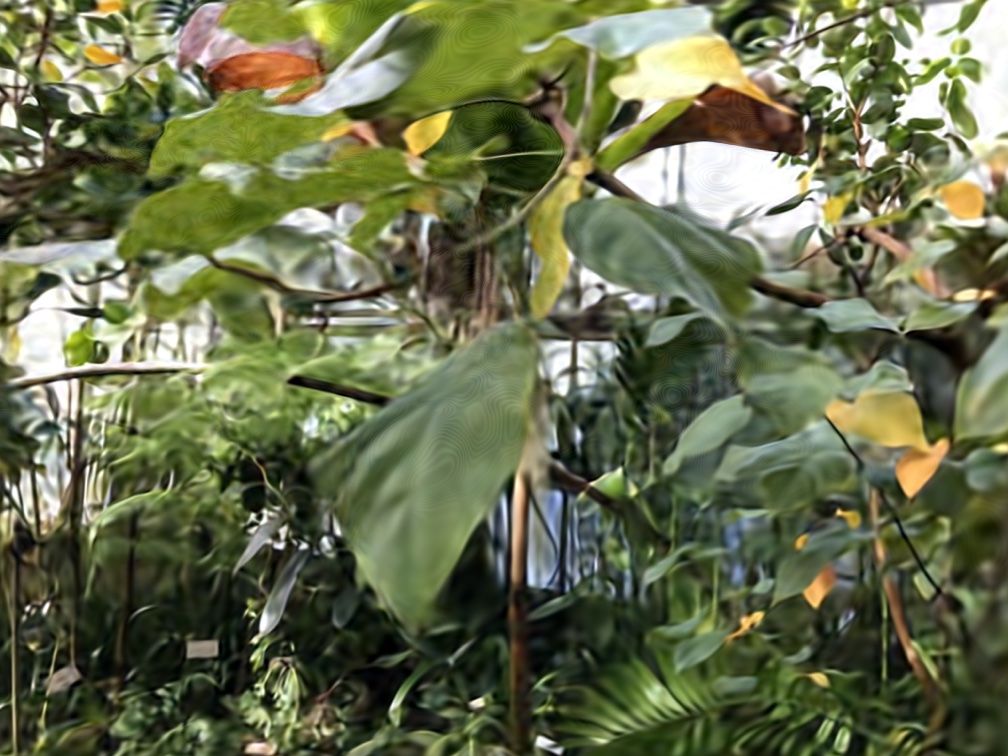}%

    \SuppLFSubfig{0 0 0 0}{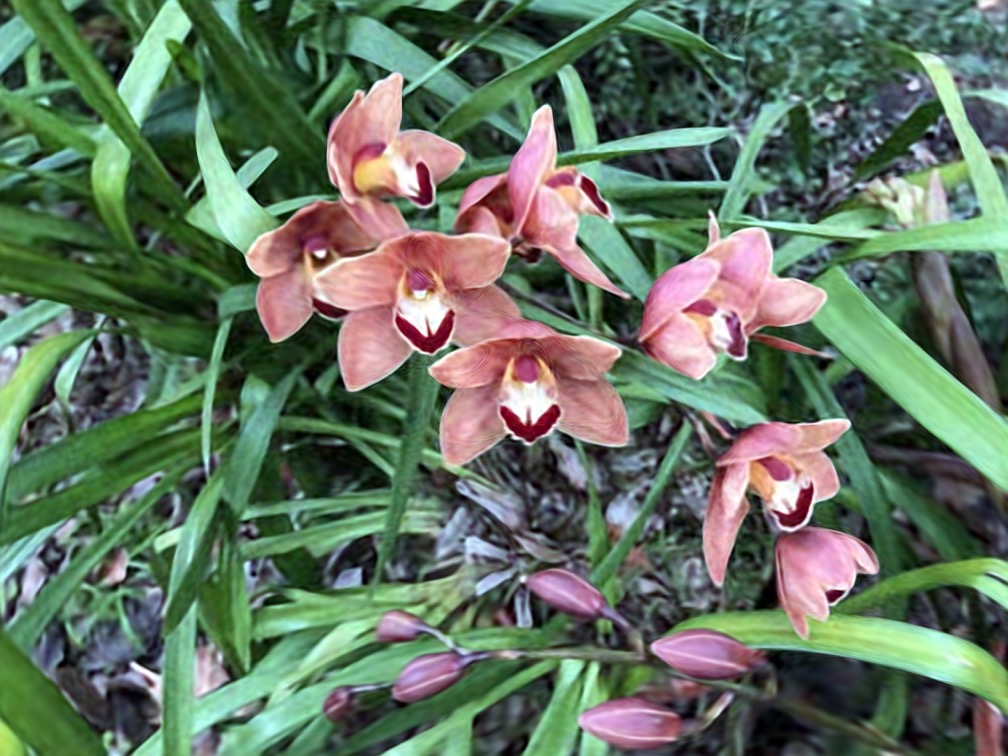}%
    ~
    \SuppLFSubfig{0 0 0 0}{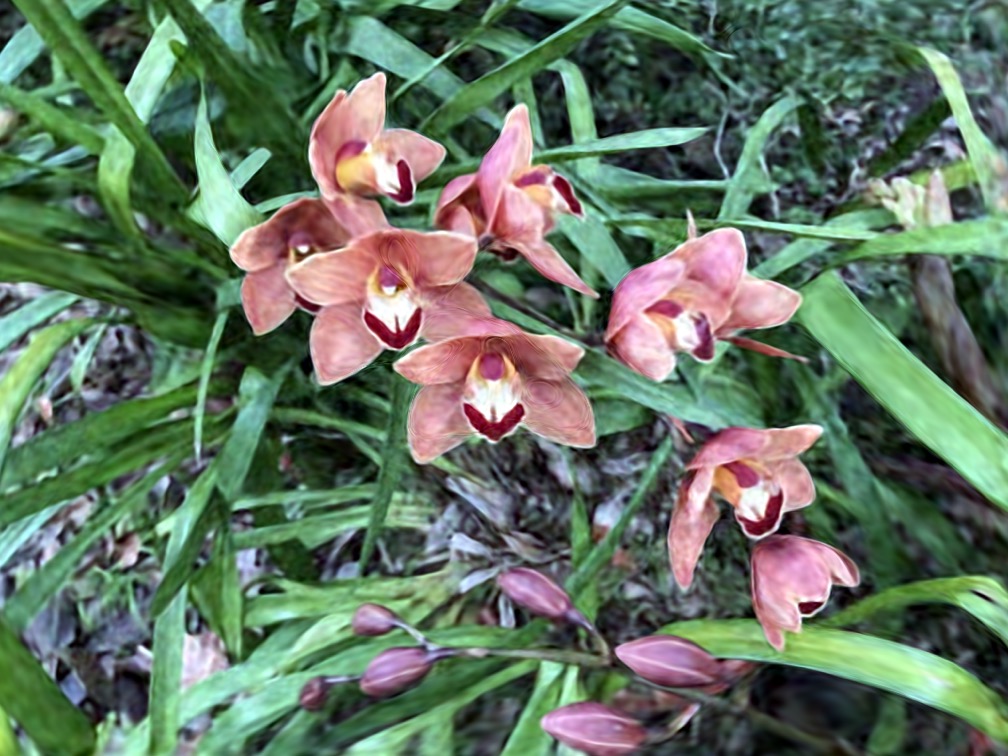}%
    ~
    \SuppLFSubfig{0 0 0 0}{./figures/LLFF/Orchids/14.jpeg}%
    ~
    \SuppLFSubfig{0 0 0 0}{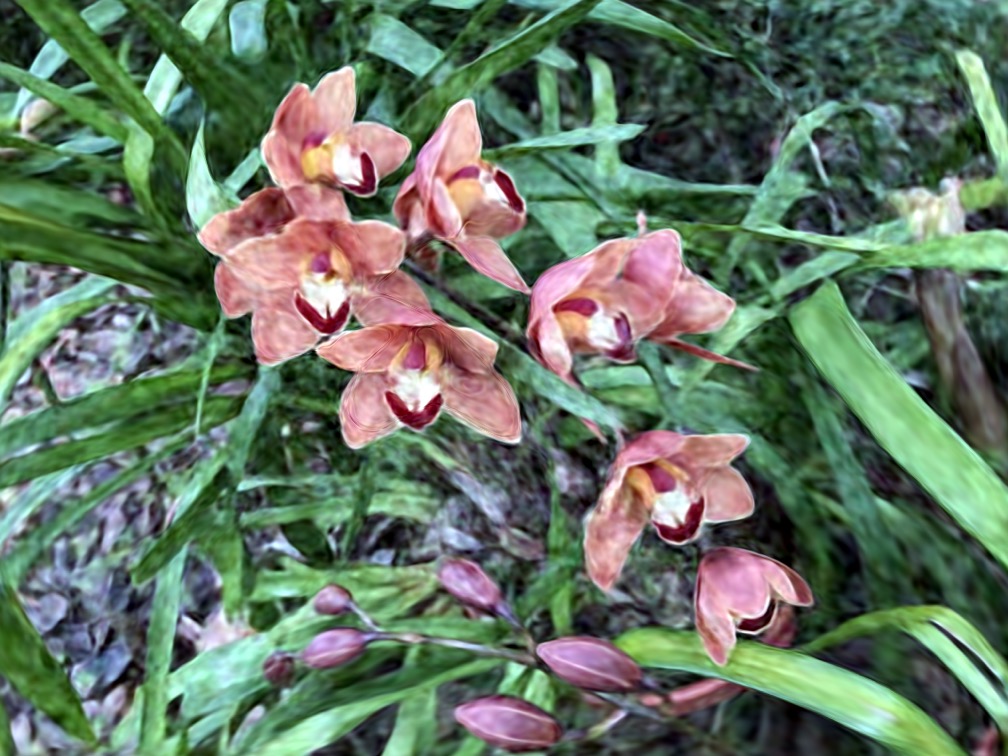}%
    ~
    \SuppLFSubfig{0 0 0 0}{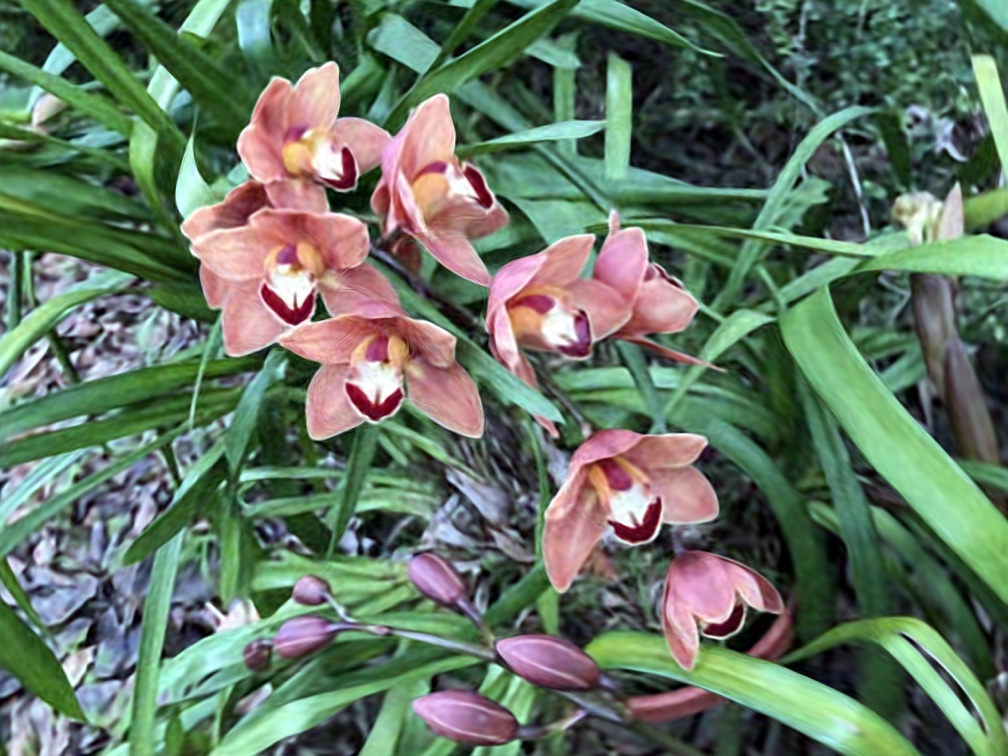}%

    \SuppLFSubfig{0 0 0 0}{./figures/LLFF/Room/0.jpeg}%
    ~
    \SuppLFSubfig{0 0 0 0}{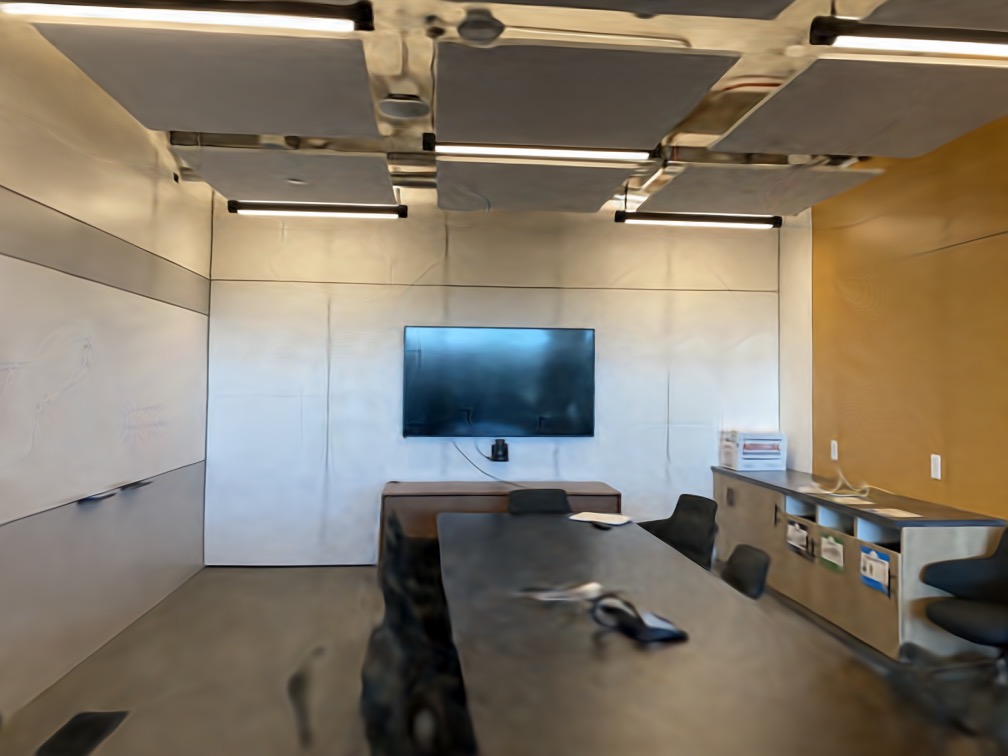}%
    ~
    \SuppLFSubfig{0 0 0 0}{./figures/LLFF/Room/14.jpeg}%
    ~
    \SuppLFSubfig{0 0 0 0}{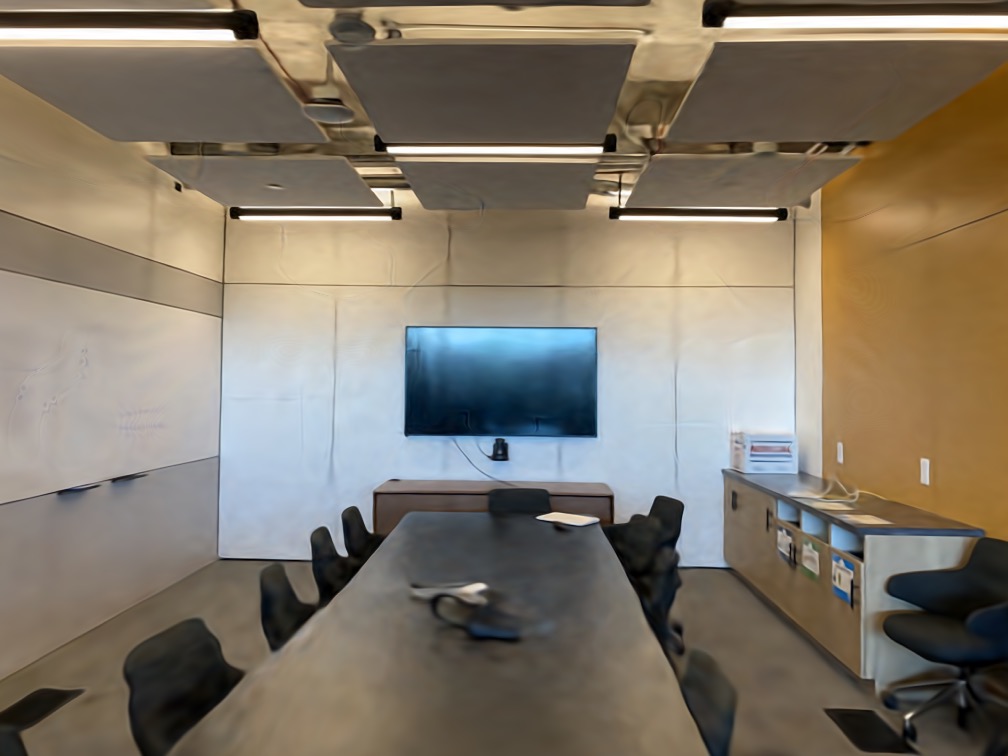}%
    ~
    \SuppLFSubfig{0 0 0 0}{./figures/LLFF/Room/29.jpeg}%
    
    \SuppLFSubfig{0 0 0 0}{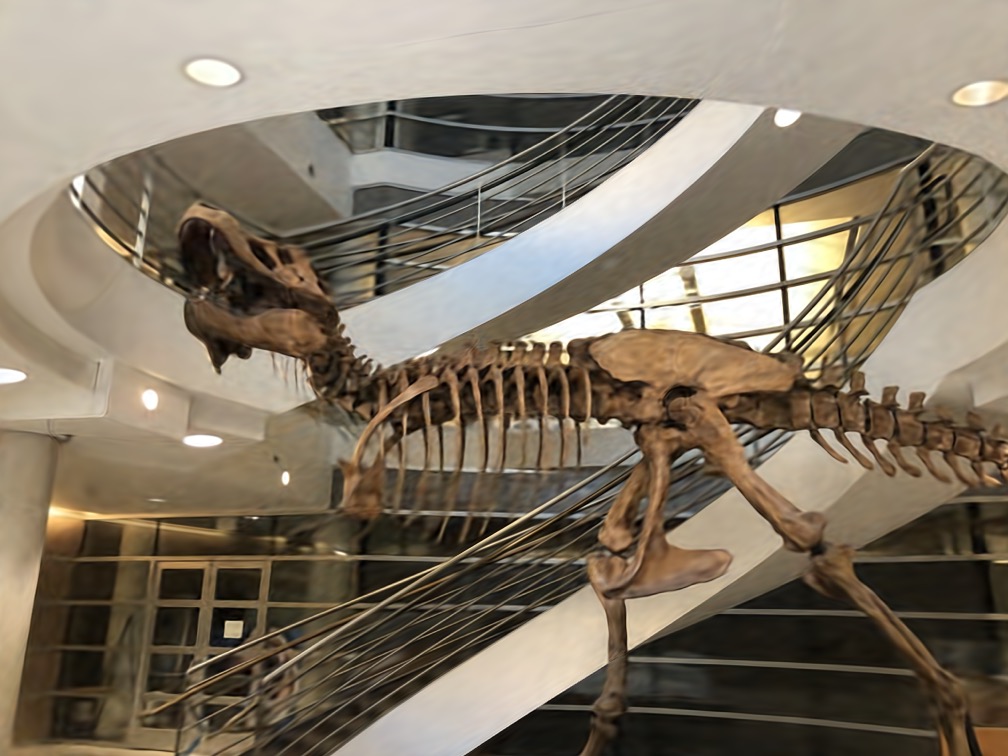}%
    ~
    \SuppLFSubfig{0 0 0 0}{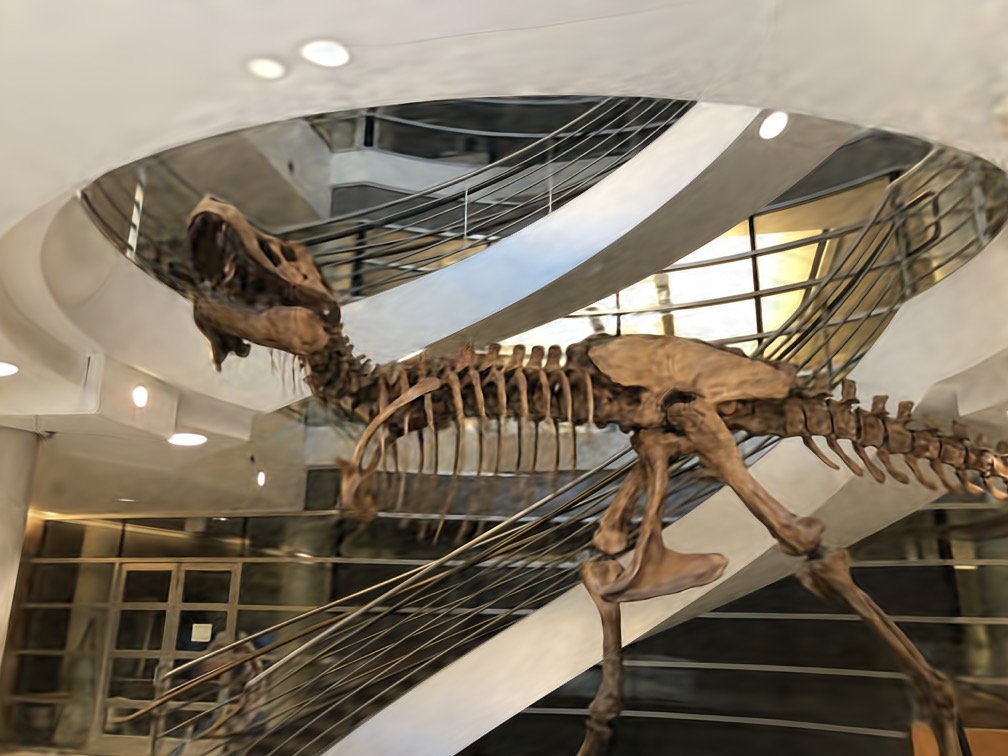}%
    ~
    \SuppLFSubfig{0 0 0 0}{./figures/LLFF/Trex/14.jpeg}%
    ~
    \SuppLFSubfig{0 0 0 0}{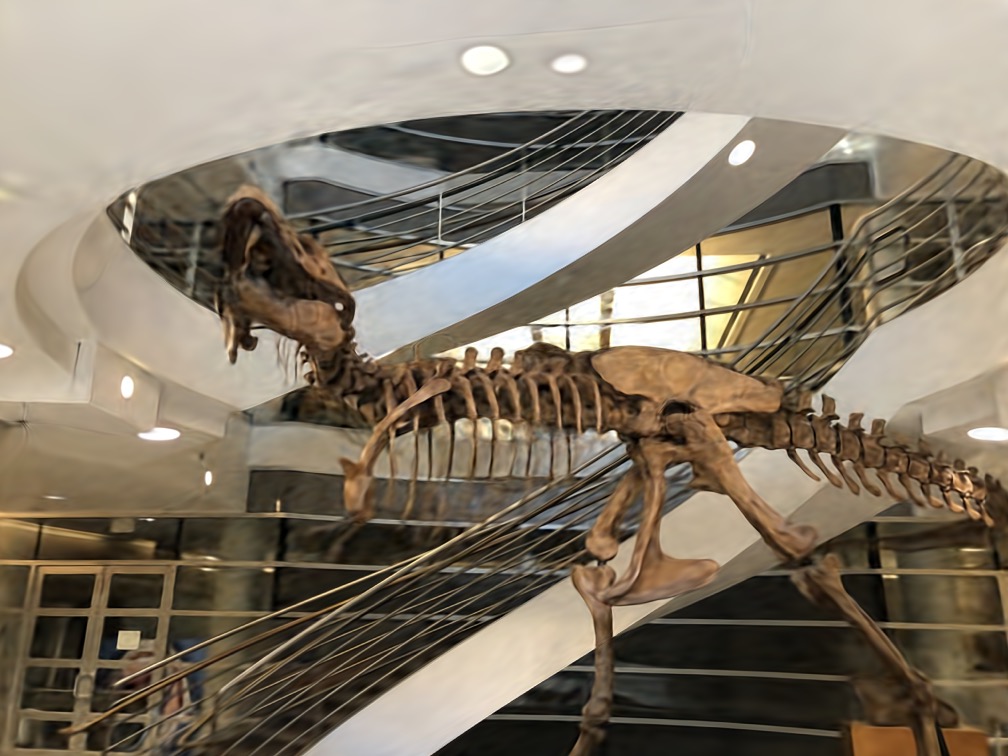}%
    ~
    \SuppLFSubfig{0 0 0 0}{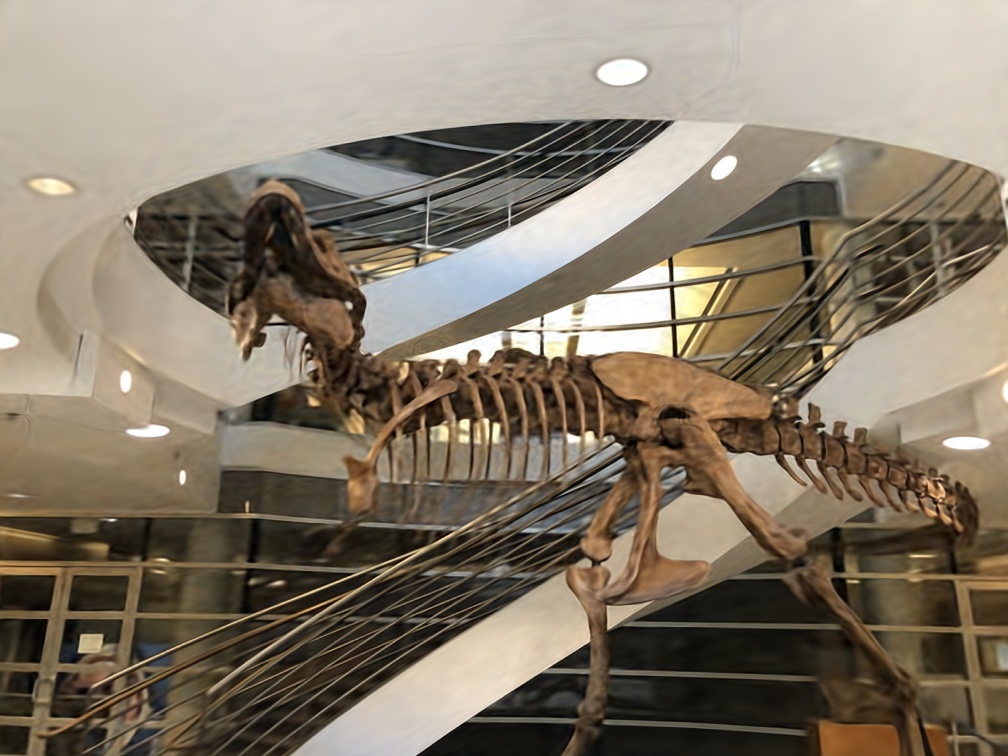}%

    \FigFiveSubfigCaption{$t=0$}%
    ~
    \FigFiveSubfigCaption{$t=0.25$}%
    ~
    \FigFiveSubfigCaption{$t=0.5$}%
    ~
    \FigFiveSubfigCaption{$t=0.75$}%
    ~
    \FigFiveSubfigCaption{$t=1$}%
    \vspace{-5pt}
	\caption{Unstructured Light Field Results. See supplementary video for better contrast.}
\end{figure*}

\clearpage
\newpage
\begin{table*}[!ht]
\begin{tabular}{c | ccc | ccc}
\toprule
Method & NeRF & LFN & Ours & NeRF & LFN & Ours\\
\midrule
Scene &  \multicolumn{3}{c}{SSIM} &  \multicolumn{3}{|c}{PSNR} \\
\midrule
Amethyst & 0.976 & 0.979 & 0.982 & 37.43 & 38.92 & 39.32\\
Beans & 0.984 & 0.996 & 0.997 & 37.51 & 46.61 & 47.85\\
Bracelet & 0.986 & 0.992 & 0.992 & 35.32 & 38.89 & 38.44\\
Bulldozer & 0.967 & 0.971 & 0.966 & 35.36 & 36.81 & 34.36\\
Bunny & 0.985 & 0.986 & 0.986 & 41.73 & 42.52 & 41.52\\
Chess & 0.987 & 0.988 & 0.987 & 39.71 & 41.39 & 39.17\\
Flowers & 0.962 & 0.977 & 0.970 & 33.80 & 38.28 & 35.24\\
Knights & 0.976 & 0.947 & 0.980 & 34.69 & 30.73 & 36.14\\
Tarot-L & 0.605 & 0.960 & 0.962 & 17.52 & 31.37 & 30.92\\
Tarot-S & 0.763 & 0.973 & 0.982 & 21.73 & 33.08 & 34.99\\
Treasure & 0.948 & 0.971 & 0.956 & 30.66 & 34.42 & 30.65\\
Truck & 0.977 & 0.980 & 0.981 & 38.00 & 39.05 & 38.63\\
\bottomrule
\end{tabular}
\caption{Detailed Known Views Results on Stanford Light Field Scenes.}
\label{table:PerSceneBegin}
\end{table*}

\begin{table*}[!hb]
\begin{tabular}{c | ccc | ccc}
\toprule
Method & NeRF & LFN & Ours & NeRF & LFN & Ours\\
\midrule
Scene &  \multicolumn{3}{c}{SSIM} &  \multicolumn{3}{|c}{PSNR} \\
\midrule
Amethyst & 0.977 & 0.960 & 0.979 & 37.56 & 32.67 & 38.14\\
Beans & 0.970 & 0.994  & 0.994 & 32.91 & 42.48 & 43.74\\
Bracelet & 0.989 & 0.962  & 0.992 & 36.46 & 29.22 & 37.81\\
Bulldozer & 0.969& 0.937  & 0.963 & 35.59 & 28.44 & 33.56\\
Bunny & 0.985& 0.975  & 0.984 & 41.77 & 35.61 & 40.83\\
Chess & 0.987 & 0.967 & 0.985 & 40.05 & 31.62 & 37.35\\
Flowers & 0.965 & 0.950  & 0.971 & 34.45 & 30.10 & 35.71\\
Knights & 0.976 & 0.782 & 0.976 & 34.11 & 19.26 & 34.54\\
Tarot-L & 0.521 & 0.933  & 0.937 & 15.68 & 27.00 & 25.90\\
Tarot-S & 0.738 & 0.971 & 0.980 & 21.09 & 32.33 & 34.04\\
Treasure & 0.954 & 0.931 & 0.962 & 31.28 & 26.72 & 31.62\\
Truck & 0.978 & 0.961 & 0.979 & 38.42 & 32.60 & 36.01\\
\bottomrule
\end{tabular}
\caption{Detailed Novel View Results on Stanford Light Field Scenes.}
\end{table*}

\begin{table*}[!hb]
\begin{tabular}{c | ccc | ccc}
\toprule
Method & NeRF & LFN & Ours & NeRF & LFN & Ours\\
\midrule
Scene &  \multicolumn{3}{c}{SSIM} &  \multicolumn{3}{|c}{PSNR} \\
\midrule
Fern & 0.860 & 0.838 & 0.814 & 25.61 & 25.02 & 24.53\\
Flower & 0.921 & 0.949 & 0.900 & 29.11 & 33.02 & 28.06\\
Fortress & 0.943 & 0.945 & 0.892 & 32.14 & 32.13 & 30.27\\
Horns & 0.898 & 0.893 & 0.847 & 28.15 & 29.27 & 26.51\\
Leaves & 0.790 & 0.833 & 0.674 & 21.65 & 23.84 & 20.01\\
Orchids & 0.794 & 0.832 & 0.816 & 22.55 & 24.67 & 23.79\\
Room & 0.975 & 0.962 & 0.949 & 34.24 & 30.87 & 30.28\\
Trex & 0.929 & 0.945 & 0.914 & 27.78 & 30.14 & 27.67\\
\midrule
M1 & 0.958 & 0.971 & 0.953 & 29.49 & 32.30 & 30.39\\
M2 & 0.987 & 0.988 & 0.989 & 35.75 & 36.48 & 35.78\\
W1 & 0.963 & 0.968 & 0.975 & 32.99 & 33.46 & 34.27\\
\bottomrule
\end{tabular}
\caption{Detailed Known Views Results on Unstructured Light Field Scenes.}
\end{table*}

\begin{table*}[!hb]
\begin{tabular}{c | ccc | ccc}
\toprule
Method & NeRF & LFN & Ours & NeRF & LFN & Ours\\
\midrule
Scene &  \multicolumn{3}{c}{SSIM} &  \multicolumn{3}{|c}{PSNR} \\
\midrule
Fern & 0.832 & 0.721 & 0.529 & 24.20 & 20.63 & 15.72\\
Flower & 0.905 & 0.827 & 0.529 & 28.21 & 23.72 & 15.929\\
Fortress & 0.948 & 0.810 & 0.734 & 32.39 & 25.09 & 23.90\\
Horns & 0.907 & 0.811 & 0.687 & 27.69 & 22.75 & 20.22\\
Leaves & 0.776 & 0.603 & 0.284 & 20.87 & 16.72 & 12.49\\
Orchids & 0.786 & 0.360 & 0.346 & 21.72 & 11.96 & 13.83\\
Room & 0.962 & 0.844 & 0.790 & 29.94 & 21.59 & 19.16\\
Trex & 0.940 & 0.875 & 0.816 & 28.34 & 23.78 & 22.35\\
\midrule
M1 & 0.946 & 0.894 & 0.844 & 25.43 & 18.13 & 13.57\\
M2 & 0.986 & 0.968 & 0.897 & 32.63 & 25.85 & 15.30\\
W1 & 0.964 & 0.958 & 0.854 & 27.23 & 24.66 & 12.37\\
\bottomrule
\end{tabular}
\caption{Detailed Novel View Results on Unstructured Light Field Scenes.}
\label{table:PerSceneEnd}
\end{table*}

\end{document}